\documentclass[a4paper,10pt,titlepage]{book}
\usepackage[dvipdfmx]{graphicx}
\usepackage{bmpsize}
\usepackage{geometry}
\usepackage{epsfig}
\usepackage{fancyhdr}
\usepackage{plain}
\usepackage{amsmath, amsthm, amssymb, amsfonts, bm}
\usepackage{mathtools}
\usepackage[sort&compress,square,semicolon,authoryear]{natbib}
\usepackage{multirow}
\usepackage{caption}
\usepackage{tabularx}
\usepackage{import}
\usepackage{booktabs, makecell}
\usepackage{hyperref}
\usepackage{algorithm}
\usepackage{algpseudocode}
\usepackage[skip=1ex]{caption}
\usepackage{subcaption}
\usepackage{todonotes}
\usepackage{tablefootnote}
\usepackage{pgfplots}
\usepgfplotslibrary{fillbetween}
\pgfplotsset{compat=1.18} 
\usepackage{pifont}
\usepackage{seqsplit}
\usetikzlibrary{shapes}
\usepackage{wasysym}
\usepackage{bbm}
\usepackage{bold-extra}
\usepackage{microtype}
\usepackage{placeins}
\hypersetup{
    colorlinks=true,
    linkcolor=blue,
    filecolor=blue,
    urlcolor=blue,
    citecolor=blue,
}
\graphicspath{{images/}}

\usepackage{sectsty}
\sectionfont{\LARGE}
\subsectionfont{\Large}
\subsubsectionfont{\large}
\paragraphfont{\large}

\newcommand{\clearemptydoublepage}{\newpage{\pagestyle{empty}\cleardoublepage}}

\newcommand{\specialcell}[2][c]{%
  \begin{tabular}[#1]{@{}c@{}}#2\end{tabular}}
  
\newcommand\bld[1]{{\fontseries{sb}\selectfont #1}}
\newcommand\bund[1]{\underline{\bld{#1}}}
\newcommand\und[1]{\underline{#1}}

\newcommand{\cmark}{\ding{51}}%
\newcommand{\xmark}{\ding{55}}%

\newcommand\symbolwithin[2]{%
  {\mathmakebox[\widthof{\ensuremath{{}#2{}}}][c]{{#1}}}}

\makeindex
\def\eqref#1{equation~\ref{#1}}

\def\1{\bm{1}}
\def\0{\bm{0}}

\def\vb{{\bm{\mathrm{b}}}}

\def\ve{{\bm{\mathrm{e}}}}

\def\vh{{\bm{\mathrm{h}}}}

\def\vk{{\bm{\mathrm{k}}}}

\def\vo{{\bm{\mathrm{o}}}}
\def\vp{{\bm{\mathrm{p}}}}
\def\vq{{\bm{\mathrm{q}}}}

\def\vv{{\bm{\mathrm{v}}}}

\def\vx{{\bm{\mathrm{x}}}}
\def\vy{{\bm{\mathrm{y}}}}

\def\mA{{\bm{A}}}

\def\mD{{\bm{D}}}
\def\mE{{\bm{E}}}
\def\mF{{\bm{F}}}

\def\mH{{\bm{H}}}

\def\mK{{\bm{K}}}
\def\mL{{\bm{L}}}

\def\mP{{\bm{P}}}
\def\mQ{{\bm{Q}}}

\def\mV{{\bm{V}}}
\def\mW{{\bm{W}}}

\DeclareMathAlphabet{\mathsfit}{\encodingdefault}{\sfdefault}{m}{sl}
\SetMathAlphabet{\mathsfit}{bold}{\encodingdefault}{\sfdefault}{bx}{n}

\def\sA{{\mathbb{A}}}

\def\sC{{\mathbb{C}}}
\def\sD{{\mathbb{D}}}
\def\sE{{\mathbb{E}}}

\def\sI{{\mathbb{I}}}

\def\sL{{\mathbb{L}}}

\def\sR{{\mathbb{R}}}

\def\sV{{\mathbb{V}}}

\def\sZ{{\mathbb{Z}}}

\DeclareMathOperator*{\argmax}{arg\,max}
\DeclareMathOperator*{\argmin}{arg\,min}

\begin{document}

\defcitealias{soldaini-moschitti-2020-cascade}{Soldaini et al., 2020}
\defcitealias{Falcon_PyTorch_Lightning_2019}{Falcon et al., 2019}
\defcitealias{tymoshenko-moschitti-2021-strong}{Tymoshenko et al., 2021}
\defcitealias{Lauriola2021}{Lauriola et al., 2021}

\pagestyle{plain}

\newpage
\clearemptydoublepage
\thispagestyle{empty}
\begin{center}

\begin{figure}[h!]
  \centerline{\psfig{file=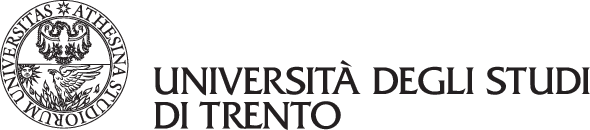,width=0.4\textwidth}}
\end{figure}

\hrulefill

DEPARTMENT OF INFORMATION ENGINEERING AND COMPUTER SCIENCE\\
\textbf{ICT International Doctoral School}\\

\vspace{0.6 cm} 
\Huge\textsc{Structural Self-Supervised Objectives for Transformers\\}
\vspace{0.3 cm} 
\Large\textsc{Exploiting Unlabeled Data and Structural Self-Supervised Objectives to align Pre-Training with Fine-Tuning}\\

\vspace{0.3 cm}

\begin{center}
\begin{tabular}{l}
\huge{Luca Di Liello}\\
\end{tabular}
\end{center}
\vspace{0.0 cm} 
\begin{flushleft}
\begin{tabular}{ll}
\multicolumn{2}{l}{\large Advisor}\\
 & \large Prof. Alessandro Moschitti \\
 & \large University of Trento \\
\end{tabular}
\end{flushleft}

\begin{flushleft}
\begin{tabular}{ll}
\multicolumn{2}{l}{\large Co-Advisor}\\
 & \large Prof. Olga Uryupina \\
 & \large University of Trento \\
\end{tabular}
\end{flushleft}

\begin{flushleft}
\begin{tabular}{ll}
\multicolumn{2}{l}{\large Mentor}\\
 & \large Siddhant Garg \\
 & \large Amazon Alexa AI \\
\end{tabular}
\end{flushleft}

\hrulefill

\normalsize
July $2023$
\end{center}

\newpage
\clearemptydoublepage
\thispagestyle{empty}
\large

{\bf \Huge Abstract}

\vspace{3cm}

\noindent
\emph{In this Thesis, we leverage unsupervised raw data to develop more efficient pre-training objectives and self-supervised tasks that align well with downstream applications. \newline\newline
In the first part, we present three alternative objectives to BERT's Masked Language Modeling (MLM), namely Random Token Substitution (RTS), \mbox{Cluster-based} Random Token Substitution \mbox{(C-RTS)}, and Swapped Language Modeling (SLM). Unlike MLM, all of these proposals involve token swapping rather than replacing tokens with BERT's [MASK]. RTS and C-RTS involve predicting the originality of tokens, while SLM tasks the model at predicting the original token values. Each objective is applied to several models, which are trained using the same computational budget and corpora. Evaluation results reveal RTS and \mbox{C-RTS} require up to 45\% less pre-training time while achieving performance on par with MLM. Notably, SLM outperforms MLM on several Answer Sentence Selection and GLUE tasks, despite utilizing the same computational budget for pre-training.
\newline\newline
In the second part of the Thesis, we propose self-supervised pre-training tasks that exhibit structural alignment with downstream applications, leading to improved performance and reduced reliance on labeled data to achieve comparable results. We exploit the weak supervision provided by large corpora like Wikipedia and CC-News, challenging the model to recognize whether spans of text originate from the same paragraph or document. To this end, we design (i) a pre-training objective that targets multi-sentence inference models by performing predictions over multiple spans of texts simultaneously, (ii) self-supervised objectives tailored to enhance performance in Answer Sentence Selection and its Contextual version, and (iii) a pre-training objective aimed at performance improvements in Summarization.
\newline\newline
Through continuous pre-training, starting from renowned checkpoints such as RoBERTa, ELECTRA, DeBERTa, BART, and T5, we demonstrate that our models achieve higher performance on Fact Verification, Answer Sentence Selection, and Summarization. We extensively evaluate our proposals on different benchmarks, revealing significant accuracy gains, particularly when annotation in the target dataset is limited. Notably, we achieve state-of-the-art results on the development set of the FEVER dataset and results close to state-of-the-art models using much more parameters on the test set. Furthermore, our objectives enable us to attain state-of-the-art results on ASNQ, WikiQA, and TREC-QA test sets, across all evaluation metrics (MAP, MRR, and P@1). For Summarization, our objective enhances summary quality, as measured by various metrics like ROUGE and BLEURT. We maintain that our proposals can be seamlessly combined with other techniques from recently proposed works, as they do not require alterations to the internal structure of Transformer models but only involve modifications to the training tasks.
}

\vspace{1cm}
\noindent
{\bf Keywords}

\noindent
self-supervised pre-training, structural objectives, transformer models, answer sentence selection, fact verification, summarization

\newpage
\thispagestyle{empty}

\pagenumbering{roman}
\tableofcontents
\cleardoublepage
\pagestyle{fancy}

\renewcommand{\chaptermark}[1]{\markboth{#1}{#1}}
\fancyhead[R]{}
\fancyhead[L]{\MakeUppercase{\chaptername\ \thechapter\ --\ \leftmark}}

\pagenumbering{arabic}

\chapter{Introduction}
\label{sec:intro}

In this Thesis, we focus on the development of alternative pre-training objectives for Language Models (LMs), to improve both efficiency and final accuracy on several downstream tasks. Language Models (LMs) are advanced Machine Learning (ML) systems designed for Natural Language Processing (NLP) and Natural Language Understanding (NLU). Nowadays, LMs are predominantly based on deep learning architectures, particularly the Transformer~\citep{vaswani2017attention}, as seen in models like GPT~\citep{Radford2018ImprovingLU} and BERT~\citep{devlin-etal-2019-bert}. These models have brought about a revolution in the NLP industry and achieved state-of-the-art performance across a wide range of language-related applications.

The main drawback of LMs regards efficiency. Large Language Models (LLMs) recently reached a remarkable size, with over 100 billion parameters in some cases~\citep{chowdhery2022palm, zhang2022opt, Scao2022BLOOMA1}. The computational requirements for both inference and pre-training of these models are substantial, demanding a significant amount of computational resources~\citep{Xu2021ASO, schwartz2020green, strubell-etal-2019-energy}. For example, the carbon footprint of BLOOM, a recently released model with 176B parameters, is estimated to be equivalent to 30 tonnes of CO$_2$~\citep{Luccioni2022EstimatingTC}. Other models like GPT-3~\citep{brown2020language}, Gopher~\citep{Rae2021ScalingLM} and LLaMa 2~\citep{touvron2023llama}, are estimated to require 500, 350 and 540 tonnes of CO$_2$ to be fully pre-trained, respectively.

The success of Language Models in NLP derives from two key factors: the Transformer architecture and the Transfer-Learning (TL) training technique.

\paragraph{The Transformer architecture}
The Transformer was initially proposed by \citet{vaswani2017attention} as a new architecture for Machine Translation (MT). The Transformer enables to capture long-range dependencies and contextual information in text. It consists of stacked neural layers, which allow the model to attend to different parts of the input text and learn meaningful representations and relations. This is performed by comparing multiple times the representation of each token with all the others in the input sequence.

\paragraph{Transfer-Learning}
Transfer-Learning (TL) is a widely used technique to train LMs by leveraging knowledge gained from one task to improve performance on another related task~\citep{west2007}. In traditional Machine Learning (ML) approaches, models are trained directly on specific tasks using large amounts of labeled data. However, TL is based on the assumption that most NLP tasks share common patterns and features, for example the same language and similar sentence/paragraph/document structures. TL involves two main steps: \emph{pre-training} and \emph{fine-tuning}~\citep{peters-etal-2018-deep}. During pre-training, a language model is trained from scratch on a vast amount of unlabeled data from sources like Wikipedia and CommonCrawl to capture general language understanding. In the fine-tuning step, the pre-trained model is specialized on a specific task or domain using smaller amounts of labeled data. This process allows the model to adapt its learned features to the target task, resulting in improved performance without requiring extensive labeled data for each individual application.

\

In the subsequent Sections, we present a comprehensive description of the contributions of this Thesis, highlighting both the accomplishments and limitations of the proposed methodologies. Additionally, we provide an in-depth examination of the structure of this work, including the experimental setup and the hardware employed. Lastly, we furnish an overview of the papers published during my Ph.D., relating them to the various chapters of this Thesis.

\section{Contributions}
\label{sec:thesis_contribution}

This Thesis examines the pre-training of Language Models from various angles. Initially, we demonstrate that the current state-of-the-art training tasks are not efficient when pre-training time is a constraint. We address this issue by providing alternative pre-training objectives which are both more efficient and lead to higher language model accuracies. Secondly, we design new pre-training tasks that align structurally with the downstream application. This allows us to fine-tune models over downstream datasets using fewer data while achieving better accuracies.

While today's research is shifting toward Auto-Regressive (AR) LLMs that can generate text for the user, we focus on Auto-Encoders (AE), which still represent an important technique for information embedding, retrieval and classification.

Out of all the downstream tasks examined in this research, special emphasis is placed on Answer Sentence Selection (AS2), a branch of Question Answering (QA) that constitutes the primary focus of my Ph.D. studies.

\subsection{Alternative Efficient Pre-Training Objectives}

In this research, we study the pre-training efficiency and time complexity of LMs Auto-Encoders. We develop alternative pre-training tasks that reach higher final accuracies while using less energy. Given the widespread utilization in recent works~\citep{liu2019roberta, he2020deberta, lan2020albert, sanh2020distilbert}, we adopt the Masked Language Modeling (MLM) pre-training task of BERT as the foundation for our experiments. MLM works by masking a small fraction of the input tokens and by tasking the model at predicting their original value. This procedure is expensive because the classification head of the model must span over the whole vocabulary, which usually contains thousands of tokens, to perform predictions. We design our objectives to reduce the classification head size and improve training time. Our proposals that target the inefficiency of MLM can be summarized as:

\begin{itemize}
\item \textbf{Random Token Substitution (RTS)}, in which input tokens are replaced with others rather than being masked. Subsequently, the model should predict only whether tokens are originals or replacements using a lightweight binary classification head;
\item \textbf{Cluster-base Random Token Substitution (C-RTS)}, which is similar to RTS but replacements are harder to be detected because we exploit a simple statistical approach to find the more challenging substitutions;
\item \textbf{Swapped Language Modeling (SLM)}, in which tokens are always replaced with others (and never masked) but the model should predict their original value, as in MLM.
\end{itemize}

We pre-training different models with our objectives over the same corpora and using a standard experimental setting. Then, we evaluate the effectiveness of our proposals on several downstream tasks. We show that RTS and C-RTS reduce pre-training time by up to 45\% when compared with MLM while maintaining a comparable level of accuracy. Finally, it becomes evident that MLM is a sub-optimal technique, as SLM achieves greater accuracy in numerous tasks while utilizing the same computational resources. More details are given in Chapter~\ref{cha:effective}.

\subsection{Specialized Task-oriented Structural Objectives}

The four works presented here introduce specialized pre-training techniques designed to adapt the model for specific downstream tasks during the pre-training phase. Our approach involves combining these objectives with the original training task in a continuous pre-training using various publicly available checkpoints. Continuous pre-training is a phase in which a model is trained further on large unlabeled datasets but with different objectives to align better with specific tasks. Remarkably, this continuous pre-training utilizes only 5\% to 10\% of the computational time and the same data as the original training, ensuring that any observed improvements are not attributed to the introduction of additional knowledge. Furthermore, we emphasize that our techniques are orthogonal to other optimization methods, such as pre-fine-tuning, which is training step performed before the real fine-tuning to further specialize the model using labeled data. We acknowledge that exploring the combination of our objectives with these alternative methods is a promising direction for future research.

\subsubsection{Self-supervised Objectives for Multi-Sentence Inference}

Most of the recently proposed LMs are pre-trained with token-level techniques, such as Masked Language Modeling (MLM) or Token Detection (TD), and sometimes also with some sentence-level tasks such as Next Sentence Prediction (NSP) or Sentence Order Prediction (SOP)~\citep{lan2020albert, devlin-etal-2019-bert, clark2020electra}. Token-level objectives task the model at reconstructing a corrupted input while sentence-level objectives task the model at classifying 1 or 2 input sentences, for example by predicting their order in the original document. This creates LM that perform well when the input is composed of at most 2 text spans but struggle at processing more than 3 inputs at a time. We call those models \emph{Pairwise} sentence classifiers.

In this work, we propose a novel architecture we denote as \emph{Jointwise}, that is specifically trained to reason over multiple input text spans at the same time. To train this multi-sentence model, we design a new pre-training task in which the model is provided $k+1$ sentences and should predict whether sentence $\{ s_i \}_{0 < i \le k}$ belongs to the same paragraph as $s_0$. We call this objective \textbf{Multi-Sentences in the Same Paragraph (MSPP)}. This task is self-supervised and does not require annotated data, because documents from large corpora such as Wikipedia, the BookCorpus or OpenWebText are already divided into several paragraphs by humans. Since every paragraph in a document describes the same general topic from different perspectives, our objective forces the model to reason at the semantic level over multiple text spans.

After the continuous pre-training, which is performed starting from a RoBERTa~\citep{liu2019roberta} checkpoint, we adapt our Jointwise architecture to perform Answer Sentence Selection and Fact Verification. In AS2, the LM is given a question and a set of possible answer candidate, and should re-rank them such that correct answers receive the higher score. Fact Verification is instead the task of determining whether a claim is supported or refuted by a set of evidence sentences. For AS2, we provide the model with the question and $k$ candidates, and it reasons about them jointly to find the best answer. For Fact Verification, we feed the model with the claim and a set of evidence sentences, and the model takes advantage of the latter to jointly predict whether the claim is supported or not.

By fine-tuning our models over different datasets, we show significant gains in performance over the baseline Pairwise classifier, also outperforming methods that exploit additional labeled data in some scenarios. This work is presented in Chapter~\ref{cha:jointwise}.

\subsubsection{Self-Supervised Objectives for AS2}

There are many cases in which the Jointwise model described before does not adapt well to the final task. For example, in Answer Sentence Selection there may be too few answer candidates to fully exploit the Jointwise architecture effectively. Moreover, the Jointwise inference speed might be slower than an equivalent Pairwise model because of the quadratic complexity growth in the input sequence length. This may be an issue in latency-constrained scenarios, like virtual assistants. For those reasons, here we present self-supervised objectives that target Pairwise model pre-training.

Pairwise models are usually pre-trained with weak supervision using sentence-level objectives such as Next Sentence Prediction or Sentence Order Prediction. We show that those objectives are solved with high accuracies by Transformer models, thus providing weak signals while training. Moreover, those objectives do not force the model at reasoning at the semantic level because are mostly designed only to understand the order of sentences in a given corpus.

We exploit the fact that large unlabeled corpora are divided into documents and paragraphs by humans as a weak form of supervision. We design 3 self-supervised tasks for pre-training that are harder to be solved than previous sentence-level objectives and that force the model to reason at the semantic level. In particular, we propose:

\begin{itemize}
\item \textbf{Sentences in the Same Paragraph (SSP)}: SSP challenges the model at predicting whether two spans of text are extracted from the same paragraph of a document;
\item \textbf{Sentences in Paragraph (SP)}: in SP, we extract a span of text from a paragraph and we provide it to the model along with the remaining text in the paragraph. Here, the model should predict if the span belongs to that paragraph in the original document;
\item \textbf{Paragraphs in the Same Document (PSD)}: PSD tasks the model with predicting whether two entire paragraphs belong to the same original document.
\end{itemize}

After performing continuous pre-training starting from several checkpoints of RoBERTa, ELECTRA and DeBERTa, we show that our pre-training objectives outperform the baselines on many Answer Sentence Selection datasets and we reach state-of-the-art performance on WikiQA and TREC-QA. Moreover, we test our objectives in the few-shot setting, underlining the optimal performance of our approaches when the data on the target task is scarce. All the details about these experiments are provided in Chapter~\ref{cha:pairwise}.

\subsubsection{Self-supervised objectives for Contextual AS2}

Recently, \citetalias{Lauriola2021} and \citet{han-etal-2021-modeling} showed that context can help re-ranking the best answer candidate in the top position. In Answer Sentence Selection happens that many answers are not well-formed or contain unresolved entities or relations. Additional context can help in solving this issue by adding details about them. In this work, our goal is to design pre-training objectives that help the model in understanding the role of the additional input context. We extend the original architectures of recently released Transformer models such as RoBERTa and DeBERTa to allow them to tokenize and process 3 input sequences: the question, the answer candidate and the context.

The objectives we propose are extensions of the SSP task described before. We define 3 different methods to gather additional context that will be fed into the model:

\begin{itemize}
\item \textbf{Static Document-level Context (SDC)}: in SDC, we select the first paragraph of a document as the context. The first paragraph is usually a summary of the document content and thus it can help in solving broken references;
\item \textbf{Dynamic Paragraph-level Context (DPC)}: DPC sets the context equal to the text that remains in the paragraph after the sampling of the two spans for the SSP objective;
\item \textbf{Paragraphs in the Same Document (PSD)}: PSD defines the context as the pair of sentences surrounding the second span that is fed to the SSP objective, which corresponds to the answer in the final task. Notice that this aligns well with the downstream AS2 application because the context is related to the answer and not to the question.
\end{itemize}

As before, we perform continuous pre-training starting from several state-of-the-art pre-trained models and then we evaluate our proposals on many Contextual Answer Sentence Selection datasets. We show that the accuracy of our models is superior to the baselines on all the benchmarks we consider, and we also reach the state-of-the-art on ASNQ when combining our objectives with larger models, such as DeBERTaV3\textsubscript{Large}. We describe in depth these experiments and results in Chapter~\ref{cha:context}.

\subsubsection{Self-supervised Objectives for Summarization}

Finally, we show that our methodologies can be adapted successfully also to generative tasks. In particular, we analyze specialized pre-training objectives designed for Summarization, in which a model is provided with a document and should generate a coherent and concise summary of the main document's points. We leverage the weak supervision of large corpora to design an objective that accustoms the model for the Summarization task already while pre-training.

More specifically, we propose \textbf{Static Document-level Summary (SDS)}, an objective in which the model receives as input all the paragraphs of a document except the first one, and it must learn to generate the latter. Notice that we chose the first paragraph of the document because it usually acts as a summary of the whole document's content.

We perform continuous pre-training starting from BART and T5 checkpoints for a few hundred thousand training steps. Then, we fine-tune our models and the baselines on several datasets for Summarization and we show how our methodologies can help to improve performance in Summarization. Furthermore, given the inherent subjectivity of evaluating summaries, we employ a Large Language Model with over 40 billion parameters to generate additional silver summaries. This approach allows for a more robust comparison between models and helps address the challenges associated with subjective evaluations. The full description of those techniques and the results are given in Chapter~\ref{cha:summarization}.

\section{Implementation \& Machines}
\label{sec:machines}

We provide the details of the hardware setup used all the experiments of this work:
\begin{itemize}
	\item CPU: 2 x Intel Xeon Platinum 8275CL @ 3.00GHz (48 cores/96 threads)
	\item RAM: 1152 GiB DDR4
	\item GPU: 8 x NVIDIA A100 GPU with 40GB HBM2 memory, connected with NVLink
	\item Cost: 32\$/h on AWS as of May 2023
\end{itemize}
Regarding the development of the code, we based our project on different libraries designed for Machine Learning:
\begin{itemize}
	\item \emph{PyTorch}~\citep{pytorch}, described by the authors as ``Tensors and Dynamic neural networks in Python with strong GPU acceleration'', is a deep learning library that offers high-performance tensors manipulation and auto-differentiation;
	\item \emph{HuggingFace Transformers \& Datasets}~\citep{wolf-etal-2020-transformers, lhoest-etal-2021-datasets}, a large collection of state-of-the-art architectures, models and datasets for NLP and CV;
	\item \emph{PyTorch Lightning}~\citepalias{Falcon_PyTorch_Lightning_2019}, a framework that allows researchers to abstract from the hardware and pleasantly train and test Neural Networks in different settings;
	\item \emph{TorchMetrics}~\citep{Detlefsen2022}, a collection of metrics to evaluate models performance.
\end{itemize}
The framework we developed that can be used to reproduce all the experiments described in this work and the pre-trained checkpoints will be released here: \url{https://github.com/lucadiliello/transformers-framework}. We train all experiments with \emph{fp16} (AMP) or \emph{bf16}, DeepSpeed and FuseAdam (which equivalent to the AdamW optimizer but more efficient)~\citep{deepspeed}, which decrease GPU memory consumption while improving training speed.

\paragraph{Evaluation and Statistical Analysis} In the tables displaying results on the downstream datasets, we emphasize our proposed models by highlighting them in bold. Moreover, we include the standard deviation for each fine-tuning on the target datasets. Unless specified otherwise, we perform 5 runs with different initialization seeds. Additionally, for each block in Tables, we underline results that are statistically superior to the corresponding baseline and emphasize the highest overall scores using bold formatting. To determine if there is a significant performance improvement, we conduct a statistical \textit{t}-test with a significance level $\alpha = 0.05$.

Finally, notice that we utilize various downstream benchmarks, such as ASNQ and WikiQA, across multiple Chapters of this work. As the experiments were conducted at different times during my Ph.D. pursuit, the baseline results improved over time due to advancements in training frameworks, hardware technologies, and bug fixes. We do not retrain every model with each new software version release, maintaining the original results, even if there are occasional misalignments among different Chapters.

\section{Publications}

In this Section, we provide a list of the publications at international conferences that led to the creation of this work:

\begin{itemize}

\nocite{di-liello-etal-2022-effective}
\item \textbf{\hyperlink{cite.di-liello-etal-2022-effective}{Effective Pretraining Objectives for Transformer-based Autoencoders}} \\
Luca Di Liello, Matteo Gabburo, Alessandro Moschitti. \\
\textit{Findings at The 2022 Conference on Empirical Methods in Natural Language Processing (EMNLP), 2022} \\

\nocite{di-liello-etal-2022-paragraph}
\item \textbf{\hyperlink{cite.di-liello-etal-2022-paragraph}{Paragraph-based Transformer Pre-training for Multi-Sentence Inference}} \\
Luca Di Liello, Siddhant Garg, Luca Soldaini, Alessandro Moschitti. \\
\textit{North American Chapter of the Association for Computational Linguistics (NAACL), 2022.} \\

\nocite{di-liello-etal-2022-pre}
\item \textbf{\hyperlink{cite.di-liello-etal-2022-pre}{Pre-training Transformer Models with Sentence-Level Objectives for Answer Sentence Selection}} \\
Luca Di Liello, Siddhant Garg, Luca Soldaini, Alessandro Moschitti. \\
\textit{The 2022 Conference on Empirical Methods in Natural Language Processing (EMNLP), 2022} \\

\nocite{di-liello-etal-2023-context}
\item \textbf{\hyperlink{cite.di-liello-etal-2023-context}{Context-Aware Transformer Pre-Training for Answer Sentence Selection}} \\
Luca Di Liello, Siddhant Garg, Alessandro Moschitti. \\
\textit{The 61st Annual Meeting of the Association for Computational Linguistics (ACL), 2023} \\

\end{itemize}

\section{Structure of this Thesis}
\label{sec:structure_of_this_thesis}

This Thesis begins with an introduction to Natural Language Processing and Transformer-based models, which is provided in Section~\ref{sec:background}. We describe in detail the branches of NLP, with a special focus on Question Answering and Answer Sentence Selection. Then, we give an outline of the Transformer architecture, from the tokenization algorithms and the embedding strategies to the structure of the Attention Mechanism. Follows an overview of various categories of Transformer models, such as Auto-Encoders and Auto-Regressive architectures, and a description of the most used training strategies and tasks.

Chapter~\ref{cha:related_works} describes the works related to the discoveries presented in this Thesis. We start by providing an overview of several recently released state-of-the-art models, which are our starting point for many experiments. Then, we describe the most relevant works related to the tasks we address in this Thesis, such as Answer Sentence Selection and Summarization. Finally, we list works related to the methodologies we developed in this publication, such as multi-sentence inference models and specialized continuous pre-training over weakly supervised data.

In Chapter~\ref{cha:datasets}, we describe every dataset used in this Thesis, providing details about the data collection process, the number of examples and the annotation procedure.

From Chapter~\ref{cha:effective} to~\ref{cha:summarization}, we elaborate on the main discoveries of this work. In particular, in Chapter~\ref{cha:effective} we analyze alternative pre-training objectives that reduce training time while increasing final accuracy on several downstream tasks. Then, in Chapter~\ref{cha:jointwise} we propose an innovative Transformer architecture that can process many input spans of text at a time while performing a different prediction for each of them. We also design a new training task that, when combined with the proposed architecture, provides benefits in Answer Sentence Selection and Fact Verification settings. In Chapter~\ref{cha:pairwise}, we study different pre-training objectives to improve models that reason over 2 spans of text by exploiting the weak supervision of large corpora, leading to improvements on various AS2 datasets. Follows Chapter~\ref{cha:context}, in which we extend the previous idea by adding context to the equation. We design custom pre-training objectives that teach the models at exploiting the context to better re-rank answer candidates while fine-tuning. Finally, in Chapter~\ref{cha:summarization}, we show that the weak supervision of unlabeled datasets can be exploited also for tasks different from Question Answering or Fact Verification. We describe self-supervised objectives to improve the performance of Auto-Regressive models in Summarization.

We present the conclusions of this Thesis in Chapter~\ref{cha:conclusions}, with an overview of the methods used, the best results achieved and a list of the possible future research directions.

\chapter{Background}
\label{sec:background}

In this Section, we start by providing an overview of the Natural Language Processing tasks that we address in this work, with a special focus on Question Answering, Fact Verification and Summarization. Then, we describe the most renowned architectures used for NLP. In particular, we elaborate on the Transformer, which is, at the time of writing, the state-of-the-art for almost all NLP tasks. Finally, we provide an overview of the most common Transformer models referenced in this work by describing their internal architecture, the data and the objectives used for pre-training. The mathematical notation and formulas in this Chapter follow the convention of indexing starting from 1.

\section{Natural Language Processing}
\label{sec:nlp}

Natural Language Processing is the study of algorithms that can process human-generated text. NLP is a mix of computer science and linguistics: while the first field provides the techniques to solve problems such as architectures design, data collection, data processing and models evaluation, linguistics studies the nature and the structure of languages, their syntax, semantics, morphology and their ambiguities~\citep{wiki:nlp}.

The first forms of symbolic NLP were developed in 1950. They exploited a series of rules to encode the structure of a language and allowed for automatic Machine Translation. However, systems were very limited and performance was far below expectations. Twenty years later, with the development of the first large vocabularies and ontologies, scientists designed the first chatbots that could extract information from knowledge bases. A few years later, in the 80s, many algorithms for automatic parsing, semantic~\citep{10.1145/318723.318728} and morphology extraction~\citep{koskenniemi-1984-general} were further developed. However, the main drawback of symbolic NLP was the need of writing complex rules by hand and for every language.

In the 90s, NLP researchers started to use statistical analysis to develop better systems, in particular, using Machine Learning algorithms. ML finally removed the need to manually write rules-based parsers for each language, which was complex and expensive. Great improvements in performance, especially in Machine Translation, were enabled by the growth of the World Wide Web, starting from 1995, which allowed large quantities of raw data to be publicly available. The possibility to use large-scale datasets of unlabeled text pushed on the development of new unsupervised, self-supervised and semi-supervised encoding algorithms, e.g. Word Embeddings~\citep{NIPS2002_d5e2fbef}. 

Recently, from the 2010s and onwards, algorithms based on Artificial Neural Networks (ANN) have become the standard in NLP. The practice is to encode a sentence by first splitting it into a sequence of tokens, which may be words, sub-words or even single characters. Then, tokens are encoded in a vector space and further processed together with various Neural Networks architectures. While before 2017 scientists exploited mainly Convolutional Neural Networks (CNNs) and Recurrent Neural Networks (RNNs)~\citep{attentive-2016-santos}, recently Transformers~\citep{vaswani2017attention} reached state-of-the-art performance in almost all NLP tasks~\citep{devlin-etal-2019-bert, liu2019roberta, Radford2018ImprovingLU}.

In the next Section, we provide an overview of the NLP task we address in this work. In particular, the focus is on Answer Sentence Selection, which is a branch of Question Answering that is rapidly gaining popularity, especially for virtual assistants such as Amazon Alexa or Google Home.

\subsection{Question Answering}
\label{sec:question_answering}

Question Answering (QA) is the task of automatically answer to questions posed in natural language. A QA pipeline is usually composed of multiple modules that deal with (i) the retrieval of data that contain relevant information and (ii) the extraction or generation of the correct answer span that is returned to the user.

The challenges in QA regard the organization of data for the retrieval phase, the parsing of unstructured text, the processing and disambiguation of the question and the elaboration of documents to extract the correct answer. For example, data for the retrieval phase can be organized in ontologies, graphs or simple lists of documents. While the first two methodologies allow fast access to key information, they are hard to create and maintain. On the other hand, a list of documents is easy to compose and update. However, information is stored in an unstructured way and it is not trivial to efficiently search for it. 

In this Thesis, we focus mostly on \textbf{Open Domain Question Answering}~\citep{voorhees-tice-2000-trec}, which is a branch of QA where questions are not restricted to a single domain but may potentially span over all human knowledge. In other more concrete words, ODQA is the task of replying to questions for which an answer may be found on the web. Open Domain QA defines new challenges because the system does not know the domain of the question a priori. Common techniques to solve the problem include powerful search mechanisms to retrieve relevant information and Machine Learning models with high generalization capabilities to extract or generate the final answer.

\begin{figure*}[t]
    \centering
    \includegraphics[width=\textwidth]{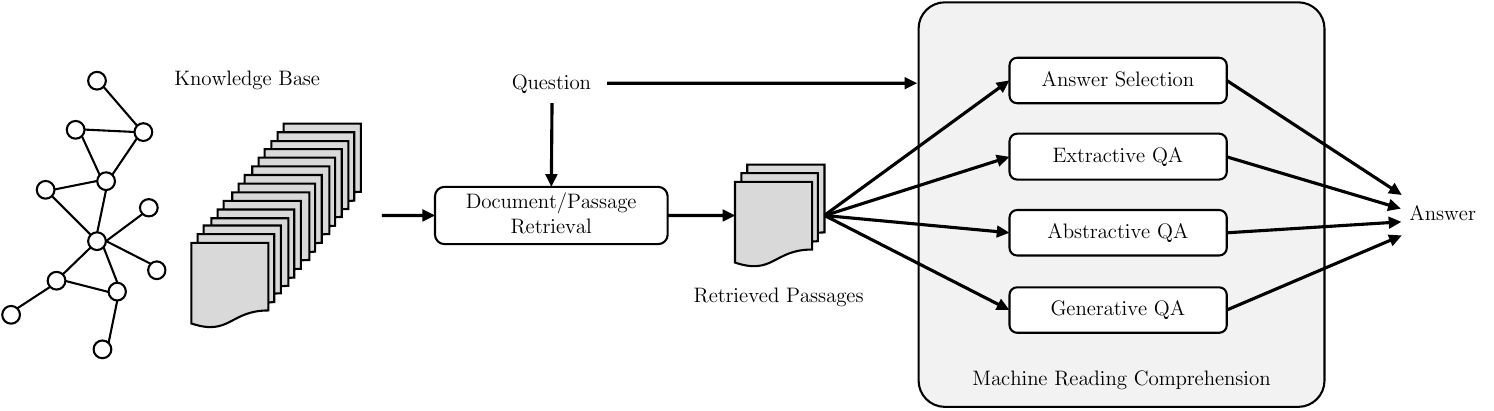}
    \vspace{-0.6em}
    \caption{\small Example of the common structure of an ODQA pipeline. Source documents are usually millions or billions of data points organized as graphs or lists. The Retrieval block finds a subset of relevant documents (a few hundred or thousands) related to the question and divides them into smaller entities for easier processing. Finally, Machine Reading Comprehension techniques are used to extract the correct answer span that is returned to the user.}
    \label{fig:qa}
\end{figure*}

The following Sections describe in depth the components of a QA pipeline. Figure~\ref{fig:qa} shows a high-level perspective of how data are retrieved and processed to find the best answer. In particular, after the gathering of relevant documents through some Information Retrieval system, 4 different Machine Reading Comprehension approaches could be exploited to obtain the final answer. MRC is an umbrella term for methods that, given a user query, process documents and try to extract or generate the correct answer. The main difference between Answer Sentence Selection and other techniques is that in AS2, the retrieved documents are split into smaller entities (e.g. sentences), which are processed separately, while Extractive, Abstractive and Generative QA usually process the whole input document to extract or generated the target answer. While the latter methods have potentially more context to find the correct answer span, Answer Selection techniques can work on much strict latency constraints because input textual sequences are shorter and can be processed in parallel~\citep{garg2019tanda}.

\subsubsection{Passage Retrieval}
\label{sec:passage_retrieval}

Passage Retrieval is a branch of Information Retrieval that studies how to organize, maintain, update and search over large bases of data points. In Question Answering, a passage retrieval system is given a query and should rank the available documents based on their relevance. Documents may contain structured or unstructured text, depending on the application. The comparison of the query with each document to compute a relevance score can be performed in different ways. Among the most common, we cite: (i) tokens match, in which the score is computed as the degree of overlap between the query and the document tokens; (ii) similarity score, in which both the query and the document are encoded in a vector space and some similarity metric (e.g. cosine similarity) is used to compute the relevance.

\paragraph{Token matching techniques}
Among the token matching techniques, TF-IDF~\citep{rajaraman_ullman_2011} and BM25~\citep{10.1561/1500000019} are two non-parametric approaches that recently gained popularity thanks to their effectiveness and efficiency.

TF-IDF stands for Term Frequency-Inverse Document Frequency, and it works by calculating the importance of a word in a document or corpus by multiplying the frequency of the word (TF) by the inverse frequency of the word in the corpus (IDF). Words that appear frequently in a document but not in the corpus will have a high TF-IDF score and are considered important to the document. In Passage Retrieval, the term frequency (TF) of the query term in the document is multiplied by the inverse document frequency (IDF) of the query term in the corpus. The scores are then summed across all query terms to give a final score for each document. In practice, a map between terms and documents is pre-computed to speed up ranking.

BM25 builds over TF-IDF to improve ranking quality by also taking into account the frequency of query terms in the document, the length of the document, and other factors such as the frequency of query terms in the whole collection of documents.

\paragraph{Representation similarity techniques}
Regarding the techniques based on representation similarity, we cite Dense Passage Retriever (DPR)~\citep{karpukhin-etal-2020-dense}, ColBERT~\citep{khattab-2020-colbert} and QUestion-Answer Database Retrieval (QUADRo)~\citep{Campese2023QUADRoDA}. DPR uses two BERT~\citep{devlin-etal-2019-bert} models to encode both query and each document to a real-valued vector space and applies cosine similarity to compute relevance. The BERT models are fine-tuned to minimize and maximize the encoded distance between positive and negative pairs respectively. In ColBERT, the authors encode the query and the document with a BERT model to obtain vectorial representations, similar to DPR. However, to compute the similarity score, they exploit a cheap interaction module that performs a more fine-grained comparison between the query and the document embeddings, leading to better re-ranking abilities. Finally, QUADRo operates on a database of previously answered question-answer pairs rather than a pool of documents. The authors encode the input query and millions of question-answer pairs, minimizing the distances between the embeddings when the query and the questions refer to the same concept, improving the retrieval and re-ranking accuracy of dense passage retrieval methods.

\

Common techniques to speed up lookup in Information Retrieval systems are indexes and inverted indexes, which use hash functions to map similar documents or parts of them to the same hash value. Examples of ready-to-be-used retrieval systems are ElasticSearch\footnote{\url{https://www.elastic.co}} and Facebook FAISS~\citep{johnson2019billion}. The first allows users to quickly get relevant documents by searching into sharded and distributed indexes by text matching, while the latter allows search by similarity thanks to the vectorial encodings of queries and documents.

For Question Answering, knowledge can be stored in many different modalities depending on the quantity of data and whether they are structured or not. More details about the different data organization methods are given in Appendix~\ref{app:structured_data}.

\subsubsection{Answer Sentence Selection}
\label{sec:answer_sentence_selection}

Answer Sentence Selection (AS2) is a ranking task in which a system is provided with a question and a set of answer candidates, and it should rank the latter based on how likely they answer the given question.

More formally, the system is provided with a question $q$ and a set of possible answer candidates $\sA = \{a_1, \dots, a_n \}$. The goal is to select the $a_k$ that best answers $q$. The common practice is to train a binary classifier $\mathcal{M}$ that is fed with pairs $(q, a_i), i \in \{1,{\dots},n\}$ and learns answers correctness $s_i = \mathcal{M}(q, a_i)$ by predicting whether $a_i$ answers $q$ or not. At inference time, the best answer to $q$ is chosen by selecting the candidates $a_k$ that scores the highest probability $s_k$ of being correct:
\begin{equation}
k = \argmax_i \mathcal{M}(q, a_i)
\end{equation}
In AS2, there are 2 main common practices used to compute scores $s_i$ for each answer $a_i$ given question $q$: Bi-Encoders and Cross-Encoders. Bi-Encoders are systems in which the question $q$ and the answer candidate $a_i$ are encoded separately, using two different networks $\mathcal{M}_1$ and $\mathcal{M}_2$. If the networks are the same and share the same internal parameters ($\mathcal{M}_1 = \mathcal{M}_2$), we are referring to a ``Siamese Networks'' system. Bi-Encoders create two embeddings $e_q = \mathcal{M}_1(q)$ and $e_{a_i} = \mathcal{M}_2(a_i)$ that are independent from each others because no information from $q$ is used to compute $e_{a_i}$ and vice-versa. The main advantage of Bi-Encoders is that every answer candidate embedding can be computed efficiently in advance, even before knowing the question. The final relevance score to determine whether $a_i$ is a correct answer to $q$ is typically computed with a similarity metric, such as cosine similarity:
\begin{equation}
s_k = \frac{e_q  \cdot e_{a_i}}{ ||e_q|| \cdot ||e_{a_i}|| }
\end{equation}
A common example of siamese networks used for Bi-Encoders is represented by Sentence-BERT~\citep{https://doi.org/10.48550/arxiv.1908.10084}.

On the other hand, in the Cross-Encoder setting, a model $\mathcal{M}$ computes a joint representation that depends both on the question $q$ and the answer candidate $a_i$:
\begin{equation}
s_k = \mathcal{M}(q, a_i)
\end{equation}
In this Thesis, we focus on Cross-Encoders because they provide the highest level of performance even though they have slightly higher computational requirements in practice. The higher computational cost arises from the encoding of a longer input sequence, which usually is the concatenation of the question and the answer candidate.

The evaluation of AS2 models requires metrics that measure the score quality compared to an ideal ranking of the answers. In an ideal ranking, every positive candidate has a higher score than all the negatives. Common metrics for AS2 include Mean Average Precision (MAP), Mean Reciprocal Rank (MRR) and Precision@K (P@K). MAP is a soft metric useful to evaluate the whole ranking quality because it measures the Precision at different positions. MRR and P@1 are hard metrics that measure the position of the first correct answer retrieved. In particular, Precision@1 is the fraction of times the model was able to score a correct answer at the top of the ranking.

\paragraph{Contextual Answer Sentence Selection}
Recent studies~\citep{Lauriola2021, han-etal-2021-modeling} showed that processing every answer candidate separately can lead to suboptimal results because local or global context helps in classifying answer candidates, especially when they are malformed or missing information or when they contain ambiguities or unresolved references to external entities. Contextual AS2 is an extension of AS2 in which the inputs of the language model are augmented with additional context to better disambiguate between answer candidates. More formally, in Contextual AS2 a model $\mathcal{M}$ receives in input a tuple $(q, a_i, c_i)$ where $q$ and $a_i$ are the original question and the answer candidate while $c_i$ is the additional context to be processed. As before, the model should predict a ranking score $s_i = \mathcal{M}(q, a_i, c_i)$. A typical approach to get the context is to extract the sentences before and after the answer candidate in the original document (local context) or to take the first paragraph of the same document (global context).

\subsubsection{Machine Reading Comprehension}
\label{sec:machine_reading}

Machine Reading Comprehension (MRC) is a task in NLP that involves building systems that can understand and reason about text in a similar way as humans do. This involves understanding the meaning of words and phrases, as well as the context in which they are used, and using this understanding to answer questions or make decisions based on the text.

There are several approaches to MRC, including rule-based, Information Retrieval-based, and Machine Learning-based systems. Machine Learning-based systems are the most common, and often use deep learning techniques such as neural networks to learn how to understand and reason about text. MRC has a wide range of applications, including Information Retrieval, Question Answering, Summarization, and dialogue systems and is an active area of research.

In the context of Question Answering, the term MRC typically refers to techniques in which the whole document is processed to find the target answer, such as Extractive or Generative Question Answering. While those techniques parse entire documents in a single step and have visibility over a lot of contexts, the elaboration of long sequences is expensive and results in higher latencies.

The three main approaches used to build a Machine Reading Comprehension pipeline are summarized as follows (more details are provided in Appendix~\ref{app:machine_reading_comprehension}):

\paragraph{Extractive Question Answering} Extractive QA involves the identification and selection of relevant information from a given text to directly answer a question. It relies on algorithms that analyze the text, identify key sentences or phrases, and choose the most appropriate portion as the answer. Extractive QA models do not generate new content but rather extract existing information.

\paragraph{Generative Question Answering} Generative QA entails the generation of a response to a question from scratch. It utilizes advanced language models, such as Recurrent Neural Networks or Transformer models, to understand the question and generate a coherent and contextually relevant answer.

\paragraph{Abstractive Question Answering} Abstractive QA goes beyond simple answer generation by summarizing relevant information and generating a concise response. It involves understanding the question and the source text, identifying key points, and generating a summary that captures the essence of the information.

\subsection{Fact Verification}
\label{sec:fact_verification}

Fact Verification is a task in NLP that requires verifying the accuracy of a statement or a claim~\citep{zhou-etal-2019-gear, liu-etal-2020-fine, zhong-etal-2020-reasoning}. This is accomplished by checking the sources of the information, comparing them, and evaluating their credibility. In particular, Fact Verification consists of 2 challenges: (i) the retrieval of relevant documents that may contain textual evidence for the claim and (ii) the classification of the claim as supported/\allowbreak refuted/\allowbreak neutral compared to the retrieved evidence.

The first part, which is an Information Retrieval task, is usually performed with reliable search engines such as ElasticSearch or FAISS, as described in Section~\ref{sec:passage_retrieval}. In this work, we address the second task, in which a model is provided with a claim and a set of evidence texts and should classify whether the claim is supported or not.

More formally, given a claim $c$ and the set of evidence text spans $E=\{e_1, \dots, e_n\}$ that were retrieved using some IR technique, the objective is to predict whether $c$ is supported/\allowbreak refuted/\allowbreak neutral using $E$. Specifically, to classify $c$, the model should find at least one evidence $c_i$ supporting/\allowbreak refuting $c$.

Fact verification is an important task in NLP because it helps to ensure that information is accurate and reliable, and it helps to prevent the spread of misinformation. It is also an area of active research and development, as new techniques and technologies are developed to improve the efficiency and effectiveness of Fact Verification processes.

\subsection{Summarization}
\label{sec:summarization}
In NLP, Summarization is the task that involves generating concise and coherent summaries that capture the key information from longer documents. The goal is to condense the content while retaining its essential meaning. There two main branches are Extractive Summarization and Abstractive Summarization~\citep{automatic_text_summarization}.

\paragraph{Extractive Summarization}
In Extractive Summarization, the task is to identify and select the most important sentences or phrases from the original text and arrange them to form a summary. The selected sentences are typically extracted verbatim from the source document without any modification. Extractive methods rely on techniques such as sentence ranking, keyword extraction, and clustering to determine the most salient information. This technique is often the easiest to implement and can preserve the factual accuracy of the original text.

\paragraph{Abstractive Summarization}
In this methodology, the model should generate a summary that may contain new phrases, rephrased sentences, or paraphrased content that captures the meaning of the original text. It involves understanding the source document and generating concise and coherent summaries in a more human-like manner. This task is usually performed by Generative Models, which are given the document in input and generate the summary one token at a time. Abstractive Summarization allows for more flexibility and creativity in generating summaries but can be more challenging due to the need for advanced language understanding and generation capabilities.

\section{Language Models}
\label{sec:language_models}

Language models in NLP are employed as statistical estimators to predict the likelihood of token sequences, assigning probabilities $P(w_1, w_2, \dots, w_k)$. These models find applications in tasks like Machine Translation, Text Classification, and Text Generation, enabling the classification or generation of human-like text.

Language Models can be constructed using various approaches, each offering its own set of advantages and drawbacks. One straightforward method involves using $n$-gram models~\citep{ngram1954}, which estimate the probability of a word given the preceding $n-1$ words in the sequence. For instance, a bigram model ($n=2$) estimates the probability of a word based solely on the preceding word, while a trigram model ($n=3$) takes into account the probabilities of the two preceding words. While $n$-gram models are relatively simple to implement and can perform well with small datasets, they may exhibit decreased accuracy compared to other approaches when dealing with large datasets or long sequences of text.

An alternative approach involves neural language models, which rely on Artificial Neural Networks (ANNs) and are trained on extensive text datasets. Neural LMs can learn statistical patterns in the data and generate more natural-sounding text compared to $n$-gram models. Additionally, they can handle longer sequences and accommodate a larger vocabulary size. Nonetheless, they may necessitate more computational resources for training and may require more data to achieve satisfactory results.

For a long time, Recurrent Neural Networks~\citep{rnn} and Convolutional Neural Networks~\citep{cnn} have been the state-of-the-art for a large variety of NLP tasks. Then, in 2017, \citet{vaswani2017attention} proposed the Transformer architecture that quickly became the standard de facto for most NLP tasks such as Machine Translation, Summarization, Question Answering and Natural Language Inference. For a detailed description of RNNs and CNNs, see Appendix~\ref{app:language_models_rnn_cnn}

This Section proceeds by describing extensively the Transformer, which is the building block of all the experiments reported in this Thesis. In Appendix~\ref{app:optimization}, we provide an overview of the loss function, the back-propagation and the various optimization algorithms commonly used to train Transformer models.

\subsection{Transformers}
\label{sec:transformers}

The Transformer~\citep{vaswani2017attention} is a type of Feed-Forward ANN based on the Attention Mechanism, which allows every token to \emph{attend} to every other token in the text. In particular, the input sequence is processed by a series of Self-Attention layers, which allow the model to directly incorporate information from the entire input sequence at each position. The Transformer also introduces the concept of Multi-Head Attention, which is the usage of multiple Self-Attention layers in parallel, each with its own set of parameters. This allows the model to attend to different parts of the input sequence simultaneously and to learn more complex relationships between the input tokens.

In more detail, the initial steps to process a sequence of words with a Transformer model involve tokenization and embedding. Tokenization is employed to segment the input text into a sequence of tokens defined in a vocabulary, such as words, sub-words, or characters. Embedding is instead a function that maps these tokens into fixed-size vector representations. Subsequently, the embedded sentence is fed into the Transformer model, which consists of a stack of one or more Transformer blocks. The output of each block, ranging from the first to the last, contains increasingly contextualized representations of each token. Lastly, the token representations outputted by the last layer are passed through a small neural network, typically referred to as a Classification Head, to obtain predictions over a set of predefined labels or to perform regression.

\subsubsection{Tokenization}
\label{sec:tokenization}
The tokenization is the process of splitting the input text into a sequence of tokens belonging to a fixed-size vocabulary $\sV$. Usually, a tokenization algorithm defines both the instructions to create the vocabulary $\sV$ from raw text and the procedure to map the text into a sequence of tokens belonging to $\sV$. For example, a simple tokenizer may divide the input text into individual words by splitting over whitespaces. It may happen that some input text substrings cannot be tokenized using only the tokens in $\sV$: in those cases, an OOV (out of vocabulary) token is assigned.

In Appendix~\ref{app:tokenization_methods}, we cover the four main tokenization algorithms used by recent LMs, which are Byte-Pair Encoding (BPE)~\citep{sennrich-etal-2016-neural}, WordPiece~\citep{Schuster2012JapaneseAK}, UnigramLM~\citep{kudo-2018-subword} and Sentence-Piece~\citep{kudo-richardson-2018-sentencepiece}.

\subsubsection{Word Embeddings}

Computers are inherently designed to process numerical data rather than strings. Therefore, every token generated by the tokenizer must be mapped to a vector space before being input to the LM. To achieve this, an embedding layer is utilized to convert categorical variables, such as tokens or integers, into continuous vector representations.

A trivial embedding layer is called \emph{one-hot} encoding, in which every input token is mapped to a binary vector. The binary vector corresponding to a token $t_i$ has the same length of the vocabulary $\sV$ and contains all 0s but from the position $i$ corresponding to the token $t_i$, which is set to 1. For example, given a vocabulary $\sV = \{ t_1, \dots, t_5 \}$, the one-hot encoding of $t_3$ is:
\begin{equation}
\text{one\_hot}(t_3) = [0, 0, 1, 0, 0]
\end{equation}
The primary issue with one-hot encoding lies in the data sparsity and the encoding size, which corresponds to the size of the vocabulary $|\sV|$, typically ranging from 30K to 120K and beyond in Transformer models~\citep{devlin-etal-2019-bert, he2021deberta}.

A more sophisticated embedding layer should satisfy the following properties: (i) create a dense representation of tokens and (ii) map related tokens to close positions. Embedding layers are typically implemented as a matrix of weights $\mE \in \sR^{|\sV| \times d}$, where each row $\ve_i$ corresponds to the representation of token $t_i$. While training, the embedding layer takes as input a token and returns the corresponding row from the matrix as the embedding vector.

While in some architecture the embedding matrix $\mE$ was learned before training, for example with Gensim~\citep{rehurek2011gensim} or word2vec~\citep{word2vec}, Transformer models' embedding layers are learned during the training process of the model, allowing to extract the most relevant information directly from the input data.
Embedding layers are useful for dealing with the large vocabularies of natural languages, allowing them to represent each word with a fixed-size vector, which can be easily processed by the Transformer. They also help to overcome the problem of sparsity in one-hot encoded representation and can effectively capture semantic and syntactic information of words.

\subsubsection{Positional Embeddings}

The Transformer architecture, which we will deepen in Section~\ref{sec:transformer_block}, is based on the Attention Mechanism and processes every input token representation at the same time, regardless of the position in the input text. This is different from recurrent architectures such as GRUs~\citep{cho-etal-2014-properties} and LSTMs~\citep{sepp1997lstm}, which intrinsically add positional information by processing sequentially one token at a time.

For this reason, Transformer models need an additional technique to inject positional information into the network. The original Transformer implementation used cosine positional embeddings~\citep{vaswani2017attention}, in which every positional embedding $\ve_i \in \sR^d$ to token $t_i$ is defined as:
\begin{equation}
\ve_{i,j} = \begin{cases} 
      \sin\frac{i}{10000^{2j/d}} & \text{if $j$ is even} \\
      \cos\frac{i}{10000^{2j/d}} & \text{if $j$ is odd} \\
   \end{cases}
\vspace{0.5em}
\end{equation}
However, recent models use either learnable absolute or relative positional embeddings.

\paragraph{Learnable Absolute Positional Embeddings}

Learnable absolute positional embeddings are similar to Word Embeddings, in which they are composed of a matrix $\mP \in \sR^{s \times d}$, where each row $\vp_i$ corresponds to the unique representation of a position $i \in \{1, \dots, L\}$ in the continuous vector space $\sR^d$. The main advantage is that learnable positional embeddings are easy to implement and work well. However, the sequence length $L$ must be fixed a priori. Finally, the embedding $\vh_i$ of each token $t_i$ is computed as the sum of the word and the positional embedding:
\begin{equation}
\vh_i = \text{embedding}(t_i) = \ve_i + \vp_i
\end{equation}
We indicate with $\mH \in \sR^{L \times d}$ the embedding matrix of size $d$ of a sequence of $L$ tokens that is provided in input to the Transformers architecture. Learnable absolute positional embeddings are used by many well-known models such as BERT~\citep{devlin-etal-2019-bert}, RoBERTa~\citep{liu2019roberta}, ALBERT~\citep{lan2020albert}, ELECTRA~\citep{clark2020electra}, DeBERTa~\citep{he2020deberta} and BART~\citep{lewis-etal-2020-bart}.

\paragraph{Relative Positional Embeddings}

Relative positional embeddings~\citep{shaw-etal-2018-self} are fixed values that are injected directly into the attention scores matrix to add positional information. The Attention Mechanism, described in Section~\ref{sec:transformer_block}, computes a score for every input tokens pair to estimate the relation between a token and all the others. This information is stored in a matrix $\mA \in \sR^{L \times L}$ where $L$ is the input sequence length. Relative embeddings are contained in a matrix $\mP \in \sR^{L \times L}$ that is added to $\mA$. Every position $i, j$ in $\mP$ contains a value that represents the distance between token $t_i$ and $t_j$. The actual computation of $\mP$ varies across models and architectures. Relative positional embeddings are used by models such as T5~\citep{2020t5}, Transformer-XL~\citep{dai-etal-2019-transformer} and XLNet~\citep{yang2020xlnet}.

\subsubsection{Transformer block}
\label{sec:transformer_block}

The original Transformer block is composed of 2 or 3 sub-layers, based on whether it is used as an encoder or as a decoder~\citep{vaswani2017attention}. At the core of the Transformer architecture, there is the Attention Mechanism, which is used to modify the representation of every token based on all the other tokens in the input sequence.

\paragraph{Attention Mechanism}

The Attention Mechanism is an architecture used in neural networks to allow a model to focus on a specific part of the input when processing it by comparing every input token representation with the others. In recent years, several Attention Mechanisms such as Additive Attention~\citep{bahdanau2016neural} and Multiplicative Attention~\citep{luong-etal-2015-effective} have been developed. The Transformer uses a modified version of the Multiplicative Attention, called Scaled Dot-Product Attention~\citep{vaswani2017attention}.

The Scaled Dot-Product Attention works as follows. First, each input hidden state $\vh_i$ containing the representation in $d$ dimensions of $t_i$ is projected from $\sR^d$ to $\sR^d$ using three different learnable matrices of weights $\mW_q$, $\mW_k, \mW_v \in \sR^{d \times d}$. More formally:
\begin{equation}
\begin{split}
\vq_i = & \mW_q \vh_i \\
\vk_i = & \mW_k \vh_i \\
\vv_i = & \mW_v \vh_i \\
\end{split}
\end{equation}
Then, projections are then packed into three matrices $\mQ$, $\mK$ and $\mV$, called respectively \emph{Queries}, \emph{Keys} and \emph{Values}, and the new hidden states $\mH'$ are computed with the Attention Mechanism as follows:
\begin{equation}
\mH' = \text{Attention}(\mQ, \mK, \mV) = \text{softmax} \left( \frac{\mQ \mK^T}{\sqrt{d}} \right) \mV
\label{eq:attention}
\vspace{0.2em}
\end{equation}
Notice that $\mA = \mQ \mK^T \in \sR^{L \times L}$ contains the attention scores and for every pair of tokens $t_i, t_j$, $\mA_{i,j}$ can be interpreted as the similarity between the two. The scaling factor $\sqrt{d}$ is used to smooth the \emph{softmax} function and stabilize gradient back-propagation by avoiding very large attention scores. Finally, the multiplication of the softmaxed attention scores with $\mV$ allows to influence every token representation with the attention scores, thus creating a more contextualized hidden state $\mH'$.

\paragraph{Multi-Head Attention}

In multi-head attention, the input hidden state is split into multiple heads or subsets, and attention is performed separately on each head. Each head performs a linear transformation on the input vectors to project them into a new space, and then applies the Attention Mechanism to the transformed vectors. The results of each head are then concatenated and projected back into the original input space. More specifically, given the number of attention heads $n$, the Queries, Keys and Values are computed with:
\begin{equation}
\begin{split}
\vq^{a}_i = & \mW^{a}_q \vh_i \\
\vk^{a}_i = & \mW^{a}_k \vh_i \\
\vv^{a}_i = & \mW^{a}_v \vh_i \\
\end{split}
\end{equation}
with $a \in \{1, \dots, n\}$ and $\mW^{a}_q$, $\mW^{a}_k$ and $\mW^{a}_v \in \sR^{d \times \frac{d}{n}}$. Then, attention is computed as before for each triple $(\mQ^{a}, \mK^{a}, \mV^{a})$, yielding to a set of new hidden states $\{ {\mH'}^{a} \}$. The final hidden state is then computed as a linear projection of the concatenation along dimension $d$ of $\{ {\mH'}^{a} \}$. More specifically, $\mH' = \text{concat}({\mH'}^{1}, \dots, {\mH'}^{n}) \mW_0$, with $\mW_0 \in \sR^{d \times d}$.

The advantage of multi-head attention is that it allows the model to capture different patterns and features from the input sequence, and to learn more complex relationships between the input and output. This can lead to improved performance on several tasks such as Machine Translation, Language Modeling, and Text Classification.

\paragraph{Feed-Forward Layer}

The Feed-Forward layer is a sequence of two affine transformations that are applied to each $\vh_i$ independently. The transformation matrices are $\mW_1, \mW_2 \in \sR^{d \times f}$ while the biases are $\vb_1 \in \sR^{f}$ and $\vb_2 \in \sR^{d}$. The computation is performed with:
\begin{equation}
\vh_i'' =  \mW_2  \max(0, \mW^T_1 \vh'_i + \vb_1 ) + \vb_2
\label{eq:ffn}
\end{equation}
In common Transformer models such as BERT~\citep{devlin-etal-2019-bert} or RoBERTa~\citep{liu2019roberta}, $f$ is called \emph{intermediate\_size}. The activation function is not always a \emph{relu} function as in Equation~\ref{eq:ffn}, but can also be \emph{gelu} or other custom activations depending on the model architecture.

\paragraph{Encoder}

The Transformer block used in encoder layers is composed of an \emph{Self-Attention} and a Feed-Forward layer. Self-Attention means that the Attention Mechanism uses the same hidden state $\mH$ to compute the projections $\mQ$, $\mK$ and $\mV$.

\paragraph{Decoder}

A decoder block is similar to the encoder block, with the addition of an extra attention layer that enables the incorporation of information from the encoder module. This additional block is called \emph{Cross-Attention}. The output of the encoder is used as $\mK$ and $\mV$ of every Cross-Attention layer in the decoder, such that the hidden state is heavily influenced by the information in output from the encoder.

\paragraph{Classification Head}

The Classification Head is a module used to predict a label $y$ from a set of possible labels $\sC$. This is usually accomplished by putting a linear layer on top of the output representation of the last Transformer Block~\citep{devlin-etal-2019-bert, lan2020albert}. More complex classification heads composed of multiple linear and activation functions are used in models such as RoBERTa~\citep{liu2019roberta} and BART~\citep{lewis-etal-2020-bart}.

\subsubsection{Transformer Models}
\label{sec:transformer_models}

\begin{figure*}[t]
	\centering
	\begin{subfigure}[t]{0.23\columnwidth}
		\centering
		\includegraphics[width=\linewidth]{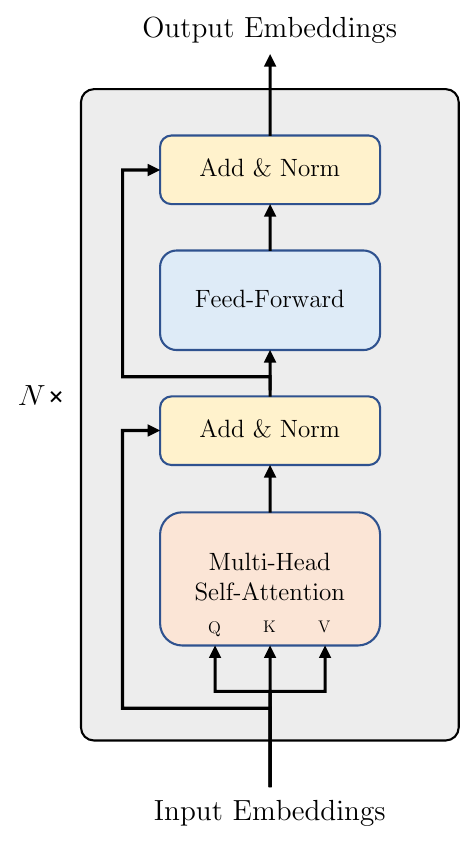}
		\caption{\small Encoder-only}
    		\label{fig:encoder}
	\end{subfigure}\hfill
	\begin{subfigure}[t]{0.48\columnwidth}
		\centering
		\includegraphics[width=\linewidth]{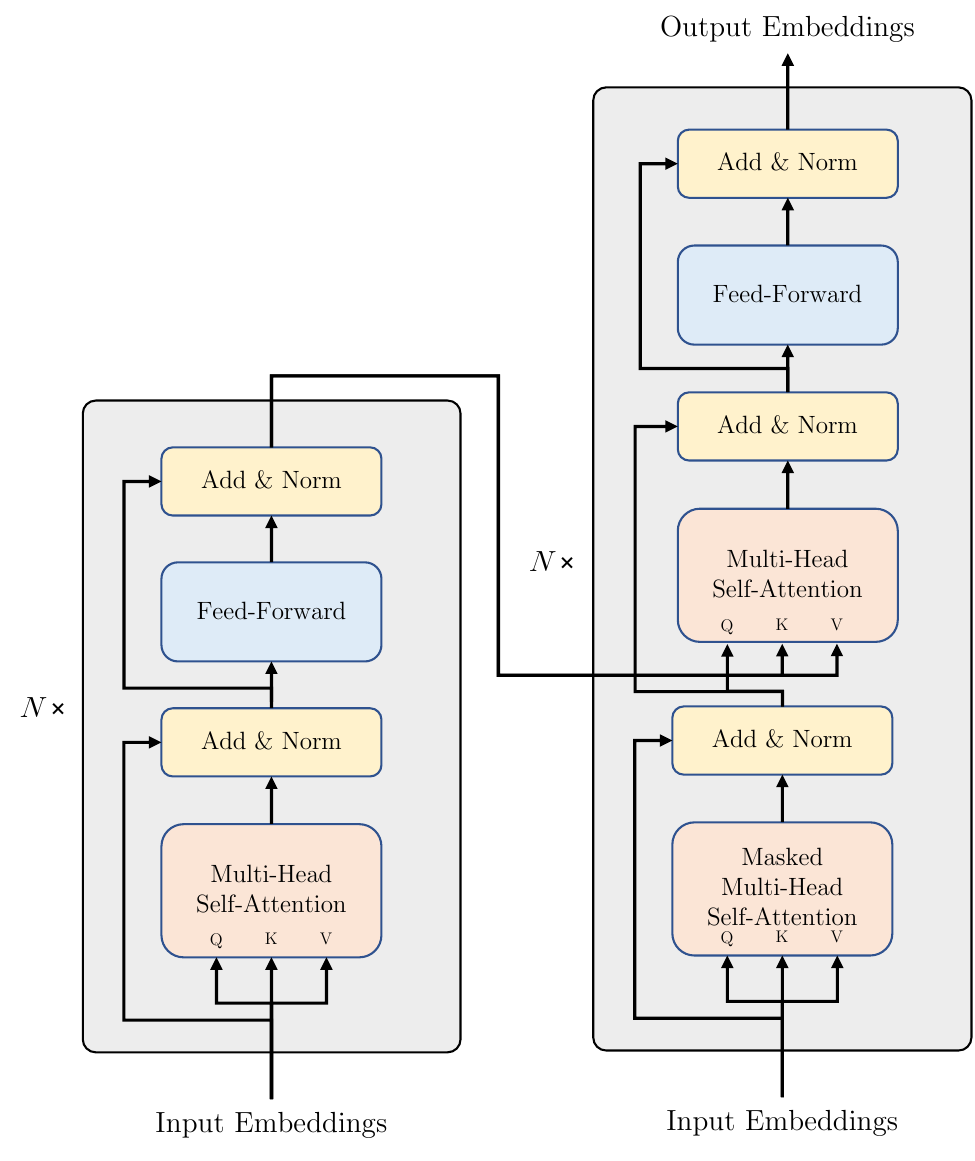}
		\caption{\small Encoder-Decoder}
    		\label{fig:encoder_decoder}
	\end{subfigure}\hfill
	\begin{subfigure}[t]{0.23\columnwidth}
		\centering
		\includegraphics[width=\linewidth]{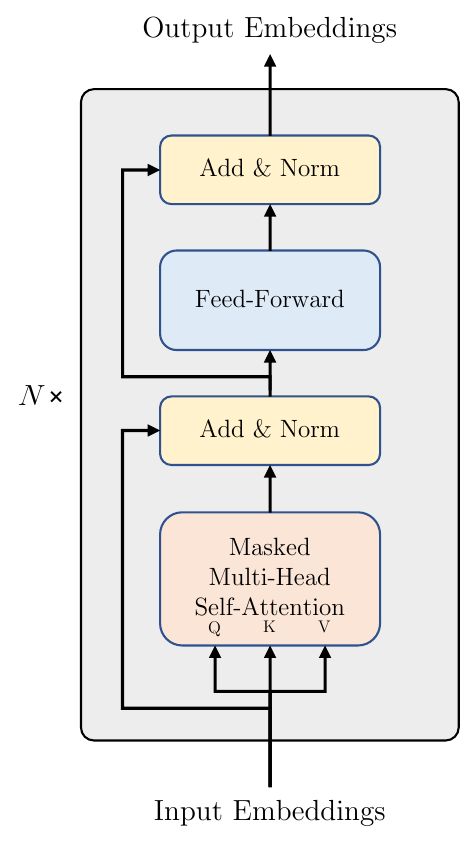}
		\caption{\small Decoder-only}
    		\label{fig:decoder}
	\end{subfigure}
	\caption{\small The three main Transformer configurations. $N$ is the number of stacked layers in the encoder or the decoder.}
\end{figure*}

Transformer-based can be categorized into three different classes based on their internal architectures:
\begin{itemize}
\item \textbf{Encoder-only}: the model is a stack of encoder blocks, such as ELECTRA~\citep{clark2020electra}, RoBERTa~\citep{liu2019roberta} or ALBERT~\citep{lan2020albert}. These models are usually trained with self-supervised objectives like BERT's Masked Language Modeling (MLM) or ELECTRA's Token Detection (TD). The structure of an Encoder-only model is shown in Figure~\ref{fig:encoder}.
\item \textbf{Encoder-Decoder}: the model comprises two sets of blocks: the encoder and the decoder. The encoder blocks take the input and generate a representation of it, while the decoder blocks generate a response for the user. For instance, if the input is a question, the encoder would process it, and the decoder would generate the answer. Similarly, in Summarization, if the input is a document, the encoder would create its representation, and the decoder would generate a summary. Examples of Encoder-Decoder models are BART~\citep{lewis-etal-2020-bart} and T5~\citep{2020t5}. Those models are usually pre-trained with objectives that challenge the model at reconstructing, from the decoder output, some corrupted input provided to the encoder. Figure~\ref{fig:encoder_decoder} shows the architecture of an Encoder-Decoder model.
\item \textbf{Decoder-only}: this class comprehends models composed only of decoder layers, such as the GPT family~\citep{Radford2018ImprovingLU, radford2019language, brown2020language}, OPT~\citep{zhang2022opt} and Bloom~\citep{Scao2022BLOOMA1}. Those models are usually pre-trained with GPT's Causal Language Modeling (CLM) objective, which consists in predicting the next token in a sequence. Figure~\ref{fig:decoder} shows an example of Decoder-only architecture.
\end{itemize}
Transformers can also be divided between Auto-Encoder and Auto-Regressive models based on the pre-training objective and the approach used for pre-training:
\begin{itemize}
\item \textbf{Auto-Encoders}: this class of Transformer models is well suited for tasks such as Sequence or Token Classification because each token in the sequence can attend to all the others. These models feature bi-directional attention and can create good vectorial text representations. For example, when trained at predicting some masked token $t_i$, the information from all the other tokens in the sequence $\{ t_j \}_{j \ne i}$ can be exploited. Auto-Encoders perform well on classification tasks in which the model should incorporate information from the whole input sequence to predict the target label.
\item \textbf{Auto-Regressive}: those models only use the tokens preceding the target token $t_i$ as a source of information. In other words, they are constrained to use only the tokens $\{ t_j\}_{j < i}$ for predicting $t_i$. Auto-Regressive models use masked attention layers to prevent the model from exploiting information from future tokens. Given the nature of Auto-Regressive Transformers, those models perform well in generative tasks, such as Summarization and Generative Question Answering, by predicting a token at a time.
\end{itemize}
Table~\ref{tab:transformers_architecture_table} shows a summary of the architecture and type of well-known Transformer models. Most Encoder-only models are Auto-Encoders while Decoder-only models are primarily Auto-Regressive. Mixed models such as T5 and BART use an Auto-Encoder to create a representation of some input text and an Auto-Regressive decoder to output a response for the user.

\begin{table*}
	\center
	\resizebox{0.75\textwidth}{!}{%
	\begin{tabular}{lccc}
		\toprule
		\textbf{Type} & \textbf{Encoder-only} & \textbf{Encoder-Decoder} & \textbf{Decoder-only} \\
		\toprule
		Auto-Encoder & \specialcell{BERT, RoBERTa, ALBERT, \\ DeBERTa, ELECTRA} & & \\
		\midrule
		Mixed & XLNet & \specialcell{T5, PEGASUS, \\ BART, BigBird } & \\
		\midrule
		Auto-Regressive & & & \specialcell{GPT, Falcon, Bloom, \\ OPT, Transformer XL} \\
		\bottomrule
	\end{tabular}%
	}
	\caption{\small Configuration of the most renowned Transformer models.}
	\label{tab:transformers_architecture_table}
\end{table*}

\subsection{Pre-Training with objectives}
\label{sec:pretraining_objectives}

The pre-training phase of Transformers is a crucial step for accustoming models to understand and generate text. It involves training on a large corpus of text data to learn the statistical patterns and relationships present in natural language. During pre-training, the model is exposed to a vast amount of raw text, such as books, articles, and web pages. The text is typically broken down into smaller units (see Section~\ref{sec:tokenization}), which allows the model to process and analyze the data more effectively.

Transformers utilize a self-supervised learning approach in pre-training, which means that the training data itself provides the supervision signal for learning. Based on the pre-training objective, such as Masked Language Modeling (MLM)~\citep{devlin-etal-2019-bert}, Causal Language Modeling (CLM)~\citep{Radford2018ImprovingLU} or Token Detection (TD)~\citep{clark2020electra}, the model is tasked at reconstructing the input which is corrupted in several ways. For example, in MLM some of the tokens are randomly replaced with a special [MASK] and the model's task is to predict the original value given the surrounding context. In CLM, the model should instead predict the next token in a sequence given all the preceding. By training the model to reconstruct corrupted text, it learns to capture the syntactic and semantic structures of the language.

The corpora used in pre-training consist of large amounts of raw data crawled from the web. Examples are Wikipedia, the BookCorpus and the CommonCrawl.

Pre-training continues for several epochs or iterations until the model achieves a satisfactory level of language understanding and predictions accuracy. Once the pre-training phase is complete, the model can proceed to the fine-tuning stage, where it is further adapted to specific downstream tasks. In the following Sections, we provide an overview of the most cited and used pre-training objective in the literature.

\subsubsection{Token-Level Pre-Training Objectives}
\label{sec:token_pretraining_objectives}

Token-level objectives challenge the model to predict missing, replaced or corrupted tokens by exploiting the knowledge of the surrounding context. The most effective token-level pre-training objectives in the literature are Masked Language Modeling, Token Detection, Causal Language Modeling and Denoising.

\paragraph{Masked Language Modeling}

Masked Language Modeling works by randomly replacing some of the input tokens, i.e. 15\%, with a special mask token, called [MASK]. The model is then tasked to reproduce the original tokens using the information from the surrounding context. This is usually also referred as the \emph{cloze} task. MLM for pre-training language models was firstly proposed by BERT~\citep{devlin-etal-2019-bert} and later was heavily utilized by many LMs such as RoBERTa~\citep{liu2019roberta}, ALBERT~\citep{lan2020albert} and XLM~\citep{lample2019crosslingual}. MLM requires bi-directional attention, thus it targets mainly Auto-Encoder models. The main drawbacks of MLM are that: (i) original tokens are predicted assuming they are independent of each other; (ii) there is a pre-training/fine-tuning discrepancy because the [MASK] token appears only while pre-training and not while fine-tuning; (iii) is not suited for text generation because tokens can attend to the right.

\paragraph{Token Detection}

Token Detection, which was initially proposed by ELECTRA~\citep{clark2020electra}, consists in challenging the model at predicting whether input tokens are originals or fakes created by another small Auto-Encoder network trained with MLM. First, a small Auto-Encoder model trained with MLM (the \emph{generator}) is used to replace some masked tokens with plausible replacements. Those new tokens are inserted in the original sequence in place of the [MASK] tokens. Then, a large discriminative model (the \emph{discriminator}) should classify every input token and tell whether it is original or a replacement created by the generator. After pre-training, the generator is discarded and the discriminator is fine-tuned on the downstream task. The advantages of Token Detection are that: (i) there is no discrepancy between pre-training and fine-tuning because the discriminator is never fed with the [MASK] token; (ii) the whole output of the discriminator is used to compute the loss, thus the training is faster because of the more effective error back-propagation. TD has been used to pre-train Auto-Encoder models such as ELECTRA, XLM-E~\citep{chi-etal-2022-xlm} and DeBERTaV3~\citep{he2021deberta}.

\paragraph{Causal Language Modeling}

Causal Language Modeling is a simple pre-training objective in which the task is to predict the next token in a sequence. CLM has been extensively used for pre-training Auto-Regressive models such as all the GPT architectures~\citep{Radford2018ImprovingLU, radford2019language, brown2020language}, OPT~\citep{zhang2022opt} and Bloom~\citep{Scao2022BLOOMA1}. The main drawbacks of CLM are that in inference, predicting one token at a time is expensive, even if the user applies caching techniques for previously computed $\mK$ and $\mV$ matrices.

\paragraph{Denoising}

Denoising is a class of objectives mostly used to train Encoder-Decoder models such as T5~\citep{2020t5} and BART~\citep{lewis-etal-2020-bart}. The task consists in corrupting the input text by removing a subset of tokens (and possibly replacing them with a [MASK] token) before feeding the encoder. Then, the decoder is tasked with generating the original sequence (BART) or just the missing spans (T5). This task helps the model in learning to handle noisy and incomplete input data and in generating coherent output sequences.

\subsubsection{Sentence-Level Pre-Training Objectives}

Sentence-level pre-training objectives teach the model to reason at the sentence level and not only on individual tokens. Follows a list of sentence-level objectives commonly used in pre-training of Transformers, usually along with some other token-level task.

\paragraph{Next Sentence Prediction}

NSP was initially proposed by the authors of BERT~\citep{devlin-etal-2019-bert} and consists in providing the model with two consecutive spans of text $s_1$ and $s_2$ and tasking the model at predicting whether $s_2$ appeared after $s_1$ in the original document. While this objective was shown to improve results on some Pairwise tasks such as Natural Language Inference, \citet{liu2019roberta} showed that in longer pre-trainings, the effect of the NSP objective is negligible. The main reason is that in long pre-trainings, NSP becomes a trivial task that can be solved by checking the main topic of $s_1$ and $s_2$, since negatives are created by sampling both spans from different documents. Thus, the signal that is back-propagated through the model is very weak.

\paragraph{Sentence Order Prediction}

SOP is a harder alternative to NSP that was proposed by \citet{lan2020albert}. In SOP, the model is fed with two spans of text $s_1$ and $s_2$ and should predict their order in the original document. Even if SOP is harder than NSP, large LMs achieve very high accuracy on this task, thus more challenging sentence-level objectives should be researched.

\paragraph{Sentence Structural Objective}

SSO was first proposed by \citet{wang2019structbert} and improves over NSP to create a more challenging objective. In SSO, the model is given two spans $s_1$ and $s_2$ and should perform a three-way classification to predict whether: (i) the spans were sampled from different documents; (ii) $s_1$ precedes $s_2$ in the original document, (iii) $s_1$ follows $s_2$ in the original document.

\paragraph{Sentences Shuffling}

Another sentence-level objective is Sentence Shuffling~\citep{lewis-etal-2020-bart}, which consists in shuffling the input sentences before feeding them to the encoder and tasking the decoder at reconstructing the original text.

\subsection{Fine-Tuning on downstream tasks}

Fine-tuning is the task of adapting a pre-trained model to perform well on some specific task, such as Natural Language Inference (NLI), Question Answering or Fact Verification.

For sentence-level classification tasks such as Natural Language Inference or Answer Sentence Selection, predictions are performed by plugging a small classification head on top of the Transformer model. For example, BERT~\citep{devlin-etal-2019-bert} uses a linear layer $\mL \in \sR^{h \times |\sC|}$ applied to the output embedding of the first token to predict a label $y \in \sC$, where $\sC$ is the set of possible label classes. RoBERTa~\citep{liu2019roberta} takes instead the average of all token output embeddings before performing the classification.

Regarding the token-level tasks such as POS Tagging, Named Entity Recognition or Extractive Question Answering, a small classification head similar to that described above is applied to all output token embeddings independently.

Finally, \emph{Teacher-Forcing} is used for the fine-tuning of generative architectures. In this case, models are not enhanced with additional parameters but are trained at generating the target sequence using the original language modeling head. In particular, Teacher-Forcing is a training technique where a generative model is trained by providing the correct output at each step. It helps the model learn faster and generate higher-quality sequences. During inference, the model generates output text based on past predictions.

Generally, fine-tuning can be performed in different ways based on the underlying model architecture and the task to be performed. We avoid describing how every model is adapted for every task because this would be very verbose and is out of the scope of this work.

\chapter{Related Work}
\label{cha:related_works}

This Chapter is divided into three main Sections. First, we describe all the language model architectures we exploit in this work. In the second Section, we report the works related to the tasks we address in this Thesis, such as Answer Sentence Selection. In the last Section instead, we discuss the related state-of-the-art techniques we compare with and from which we took inspiration.

\section{Language Models}
\label{sec:sota_lm}

\begin{table*}[t]
    \centering
    \resizebox{1.0\linewidth}{!}{%
    \begin{tabular}{lcccccccc}
        \toprule
        \textbf{Model}			& \textbf{Type}		& \textbf{\# Params}		& \textbf{Vocab} 		& \textbf{\# Layers}	& \textbf{Hidden size} 	& \textbf{\# Att. heads}	& \textbf{\# Interm. size}	& \textbf{\# Tokens} \\

        \toprule

    	BERT\textsubscript{Small}		& AE				& 13M				& 30K				& 12				& 256				& 4					& 1024				& 65B \\
    	BERT\textsubscript{Base}			& AE				& 109M				& 30K				& 12				& 768				& 12					& 3072				& 131B \\
    	BERT\textsubscript{Large}		& AE				& 335M				& 30K				& 24				& 1024				& 16					& 4096				& 131B \\

	\midrule

	RoBERTa\textsubscript{Base}		& AE				& 125M				& 50K				& 12				& 768				& 12					& 3072				& 2097B \\
	RoBERTa\textsubscript{Large}		& AE				& 355M				& 50K				& 24				& 1024				& 16					& 4096				& 2097B \\
	
	\midrule

    	ELECTRA\textsubscript{Small}		& AE				& 18M				& 30K				& 12				& 256				& 4					& 1024				& 65B \\
	ELECTRA\textsubscript{Base}		& AE				& 109M				& 30K				& 12				& 768				& 12					& 3072				& 524B$^\clubsuit$ \\
	ELECTRA\textsubscript{Large}		& AE				& 335M				& 30K				& 24				& 1024				& 16					& 4096				& 1835B$^\clubsuit$ \\

	\midrule

	ALBERT\textsubscript{Base}		& AE				& 12M				& 30K				& 12 (repeating)	& 768				& 12					& 3072				& 2097B$^\spadesuit$ \\
	ALBERT\textsubscript{Large}		& AE				& 17M				& 30K				& 24 (repeating)	& 1024				& 16					& 4096				& 2097B$^\spadesuit$ \\

	\midrule

	DeBERTa\textsubscript{Base}		& AE				& 139M				& 50K				& 12				& 768				& 12					& 3072				& 1048B \\
	DeBERTa\textsubscript{Large}		& AE				& 406M				& 50K				& 24				& 1024				& 16					& 4096				& 1048B \\
	DeBERTaV3\textsubscript{Base}	& AE				& 184M				& 128K				& 12				& 768				& 12					& 3072				& 2097B \\
	DeBERTaV3\textsubscript{Large}	& AE				& 434M				& 128K				& 24				& 1024				& 16					& 4096				& 2097B \\

	\midrule

	BART\textsubscript{Base}			& AR				& 139M				& 50K				& 6+6			& 768				& 12					& 3072				& 2097B \\
	BART\textsubscript{Large}		& AR				& 406M				& 50K				& 12+12			& 1024				& 16					& 4096				& 2097B \\

	\midrule

	T5\textsubscript{Base}			& AR				& 223M				& 32K				& 12+12			& 768				& 12					& 3072				& 1048B \\
	T5\textsubscript{Large}			& AR				& 738M				& 32K				& 24+24			& 1024				& 16					& 4096				& 1048B \\

        \bottomrule
    \end{tabular}
    }
    \caption{\small Comparison of several state-of-the-art Transformer models. ``Type'' indicates whether the model is an Auto-Encoder or Auto-Regressive model. ``Token'' is the total number of tokens seen while pre-training. $\clubsuit$: We report the number of tokens seen in pre-training of ELECTRA\textsubscript{Base}++ and ELECTRA\textsubscript{Large}-1.75M respectively, because those are the models that have been released by the authors. $\spadesuit$: We report the number of tokens seen by the models released in the corresponding repository.}
    \label{tab:pretrained_models}
\end{table*}

The statistics about the model's number of parameters, the data and the setting used to pre-train the architectures exploited in this work are reported in Table~\ref{tab:pretrained_models}.

\subsection{BERT}
\label{sec:sota_lm_bert}

BERT~\citep{devlin-etal-2019-bert} is an architecture released by Google Research that is built upon the Transformer architecture, which allows it to capture long-range dependencies in text. BERT is initially pre-trained on large amounts of unlabeled text data from sources like books and Wikipedia, using Masked Language Modeling and Next Sentence Prediction objectives. Unlike previous models that mainly relied on left-to-right or right-to-left context, BERT leverages bi-directional context to perform predictions while pre-training or fine-tuning. BERT was initially released in two versions based on the number of parameters, the number of layers and the hidden state size. However, recently the authors released also smaller, larger and multilingual variants.

\subsection{RoBERTa}
\label{sec:sota_lm_roberta}

RoBERTa's~\citep{liu2019roberta} architecture is the same as BERT. It was developed by Facebook AI Research in which the authors discovered that by training BERT for longer, on more data and with different hyper-parameters, the model could perform much better on a large number of downstream tasks. More specifically, they (i) removed the Next Sentence Prediction objective; (ii) applied dynamic masking in MLM; (iii) increased the batch size and the learning rate; (iv) used a training corpus almost 10 times larger and (v) changed the tokenizer from WordPiece with 30K tokens to BPE with 50K tokens. RoBERTa has been initially released in different variants based on the number of hidden layers, the hidden dimension and the total number of parameters.

\subsection{ELECTRA}
\label{sec:sota_lm_electra}

ELECTRA~\citep{clark2020electra} is yet another Transformers-based model released by Google Research which is trained as a discriminator rather than as a generator. ELECTRA is composed of a small generator network trained with MLM and a large discriminator model trained with Token Detection. The generator is provided with text in which 15\% of the tokens are masked and should find reasonable replacements. On the other hand, the discriminator should identify which tokens are original and which have been introduced by the generator. After the pre-training, the generator is discarded and the discriminator is fine-tuned for the desired task.

ELECTRA offers several advantages over BERT. One significant advantage is its use of the entire discriminator output to compute the loss value, providing a stronger signal for optimization. Additionally, the discriminator in ELECTRA is never exposed to input sequences containing the [MASK] token, as the generator replaces these tokens with challenging alternatives. This addresses a key issue in BERT, where the [MASK] token is present during pre-training but absent during fine-tuning, leading to a discrepancy between the two stages.

ELECTRA was shown to match BERT's performance using only half of the computational resources required by the latter.

\subsection{ALBERT}
\label{sec:albert}
ALBERT~\citep{lan2020albert} is a Transformer-based model that was developed by Google Research as a more efficient variant of BERT. ALBERT is designed to reduce the memory footprint and computational requirements of BERT while maintaining its effectiveness in various NLP tasks. ALBERT introduces three key differences with BERT. First, the model shares the parameters of all the layers, thus being much leaner in terms of memory footprint and faster in training/inference. Secondly, they refactor the embedding layer into two separate projection matrices that reduce the total number of parameters. Third, the substitute BERT's NSP objective with Sentence Order Prediction (SOP), that was shown to be more challenging and more effective when the model is fine-tuned on the downstream tasks. Finally, they pre-train the model for longer and over more data than BERT. Thanks to these optimizations, they achieve state-of-the-art results on a wide range of NLP tasks such as GLUE, SQuAD and RACE.

\subsection{DeBERTa}

\paragraph{DeBERTa} is a Transformer-based architecture developed by Microsoft Research that uses disentangled attention to improve the representation of tokens and positions~\citep{he2020deberta}. In other successful models such as BERT and RoBERTa, the position and the token embeddings are summed together before being fed to the Transformer layers. In DeBERTa, each token is represented with 2 distinct vectors: 1 for the relative positions and 1 for the token representation. In the Transformer layer, the Attention Mechanism is applied to tokens-to-tokens, tokens-to-positions and positions-to-tokens. Position-to-position attention scores are not computed because they are not informative. Moreover, DeBERTa is pre-trained with an enhanced version of MLM that uses absolute positions applied just before the language modeling head to help in predicting the right token in long sequences.

DeBERTa is shown to outperform RoBERTa on various tasks using less computational resources. The largest model released surpassed the human baseline on SuperGLUE~\citep{wang2019superglue} for the first time.

\paragraph{DeBERTaV2} further improves upon the original DeBERTa model\footnote{More information about DeBERTaV2 can be found here: \url{https://github.com/microsoft/DeBERTa}}. It introduces two key modifications: disentangled attention is now utilized in both Self-Attention and Cross-Attention (for decoding), and it utilizes an additional training objective called document-level language modeling. These enhancements are designed to improve the model's understanding of long-range dependencies and its generalization capabilities.

\paragraph{DeBERTaV3} is yet another improvement over DeBERTaV2 in which the model is trained in an ELECTRA-like configuration, i.e. the architecture is composed of a small generator network trained with MLM and a large discriminator trained with TD~\citep{he2021deberta}. At the time of writing, DeBERTaV3 is the most accurate language model among all the architectures with less than one billion parameters.

\subsection{BART}

BART (Bidirectional and Auto-Regressive Transformer) is a Sequence-To-Sequence (Encoder-Decoder) language model developed by Facebook AI Research~\citep{lewis-etal-2020-bart}. It is based on the Transformer architecture and is trained on large amounts of unlabeled text data, where it learns to reconstruct corrupted, shuffled or masked input sequences with Teacher-Forcing. This process helps the model to acquire a rich understanding of language patterns and semantics which resulted in state-of-the-art performance across many generative tasks such as Machine Translation, Summarization and Question Answering.

\subsection{T5}

T5 (Text-to-Text Transfer Transformer) is a versatile language model developed by Google AI~\citep{2020t5}. It is trained using a ``text-to-text'' framework, where a wide range of NLP tasks are cast as a text generation problem. T5 is trained to map a source text, such as a prompt or question, to a target text, which can be a summary, a translation, a sentiment analysis, or any other entity. The pre-training objectives of T5 include tasks like Translation, Summarization, Question Answering, Text Classification, and more. By training on several different tasks, T5 learns to generalize across different NLP domains and can be exploited for different applications in a zero-shot setting by only adapting the input prompt, without changes to the model architecture. T5 has achieved state-of-the-art results on various benchmark datasets across multiple tasks such as Machine Translation, outperforming previous models on standard translation benchmarks. T5 also excels in Summarization, Question Answering, and Text Classification, surpassing previous approaches and achieving competitive performance.

\section{Natural Language Processing Tasks}
\label{sec:related_tasks}

This Section provides an overview of the works related to the tasks we address in this Thesis.

\subsection{Answer Sentence Selection}
\label{sec:related_as2}

This Section relates to the experiments described in Chapters~\ref{cha:effective}, \ref{cha:jointwise}, \ref{cha:pairwise} and~\ref{cha:context}.
Answer Sentence Selection is a branch of Question Answering in which the task is to re-rank a set of possible answer candidates such that the correct answers to the given question are provided with the highest score. 

Previous approaches to AS2 involved using CNNs to learn and score question and answer representations. \citet{Severyn2015LearningTR} discusses the importance of learning a similarity function between pairs of text spans in ranking tasks such as Information Retrieval and AS2. The authors propose a CNN architecture that learns the optimal representation of text pairs and a similarity function in a supervised way. The model achieves state-of-the-art accuracy on the tasks mentioned before and shows comparable results in tweet re-ranking without the need for manual feature engineering or additional syntactic parsers.

Another relevant improvement in AS2 was achieved thanks to Alignment Networks~\citep{shen-etal-2017-inter}. Alignment Networks allow the model to measure the similarity between two sentences based on their interaction information. Specifically, the authors exploit a similarity matrix built over the word representation of both input sentences to discover alignments. Moreover, their system automatically computes a weight for every word in the sentences based on estimated importance. By applying this methodology to LSTM Networks, they reach state-of-the-art results in different AS2 benchmarks, such as TREC-QA and WikiQA.

Compare-and-Aggregate architectures for Question Answering have been extensively studied~\citep{Wang2017ACM,Bian2017ACM,Yoon2019ACM}. Compare-and-Aggregate differs from the previous techniques in the way the question and answer embeddings are used to perform the final prediction. Previous methods usually exploit a single vector representing the question and another vector for the answer to predict the target label. In Compare-and-Aggregate instead, the prediction is performed by comparing smaller units between the question and the answer, such as single-word embeddings. Then, results from all comparisons are aggregated and used for the final decision. It is worth mentioning that the authors of \citet{Yoon2019ACM} mixed Clustering techniques with Compare-and-Aggregate to improve final re-ranking even further.

In \citet{bonadiman-moschitti-2020-study}, the authors exploit separate encoders for the question and the candidate answer for a more efficient re-ranking. They show it is possible to achieve good accuracy with a training phase lasting only a few minutes. However, the exploitation of lighter architectures such as LSTMs leads to final lower results on various AS2 datasets when compared with methods based on large Transformer models.

Recently, \citet{garg2019tanda} achieved state-of-the-art results with their new methodology called Transfer-\textsc{and}-Adapt. In T\textsc{and}A, a large Transformers model is first fine-tuned on a large QA corpus (ASNQ) and then it is adapted to the downstream task, which can be a small dataset such as WikiQA and TREC-QA. The advantage is that the model can exploit previously learned relations between the question and the answer to perform better in the final task. This is especially true on small target datasets such as WikiQA and TREC-QA, on which T\textsc{and}A reaches state-of-the-art accuracy.

Another noteworthy contribution in the field of Answer Sentence Selection is presented by \citet{gabburo-etal-2022-knowledge}. In their work, the authors utilize Auto-Regressive models to generate answers based on the top$_k$ candidate sentences retrieved by a Pairwise re-ranker. Building upon GenQA~\citep{muller-etal-2022-cross}, they introduce a self-supervised technique that transfers knowledge from AS2 to Generative QA. This technique challenges the model to reconstruct the top answer retrieved by the AS2 system, leveraging only the next $k$ candidates in the ranking, thereby eliminating the need for labeled datasets.

\paragraph{Contextual Answer Sentence Selection}

The works described in this Section refer to experiments presented in Chapter~\ref{cha:context}. Contextual AS2 is an improvement over AS2 in which the model has access to additional knowledge for each answer candidate in order to perform a better re-ranking~\citep{barlacchi-etal-2022-focusqa}. The context is a very precious source of information because it can help in solving entities and relations in the answer candidate.

In \citet{shalini2016clstm}, the authors introduce CLSTM (Contextual LSTM), an extension of the LSTM network that incorporates contextual features into the model. Experimental results over three NLP tasks (Word Prediction, Next Sentence Selection, and Sentence Topic Prediction) show that using both words and topics as features improves CLSTM performance over corresponding baselines. The study demonstrates that enhancing models with additional context is essential to improve accuracy in tasks like Question Answering, Sentence Completion, Paraphrase Generation, and Next Utterance Prediction in dialog systems.

The authors of \citet{tran-etal-2018-context} enhance AS2 systems with context. To the best of our knowledge, this is the first work targeting specifically Contextual AS2. The authors propose an innovative architecture called Context-dependent Additive Recurrent Neural Network (CARNN) that incorporates strong external signals for a better learning process. The authors show state-of-the-art results on several datasets, including TREC-QA.

In \citet{Tan2018ContextAwareAS}, the authors exploit GRU Networks to model answer candidates and local context, improving performance on two AS2 datasets. Unlike previous approaches, that treated question-answer pairs independently, the authors emphasize the importance of contextual information within the passage. They propose a hierarchical GRU model that incorporates context information at both the word and sentence levels. The model outperforms state-of-the-art methods on the WikiQA and SQuAD datasets, demonstrating the effectiveness of their approach.

\citetalias{Lauriola2021} propose a Transformer encoder that uses both local and global document-level context to better disambiguate between answer candidates. Moreover, the authors perform experiments over the combination of Transfer-Learning (using the T\textsc{and}A methodology) and Contextual AS2, reaching state-of-the-art performance on several AS2 datasets, such as WikiQA.

Finally, \citet{han-etal-2021-modeling}, the authors exploit unsupervised retrieval techniques to incorporate contextual information into Answer Sentence Selection (AS2) models. Traditional AS2 models score question-answer pairs individually without considering the document context. The authors propose an approach that efficiently incorporates context by extracting relevant sentences using unsupervised similarity techniques and feeding them into the fine-tuned Transformer architecture. Their approach, leveraging a multi-way attention architecture, improves the state-of-the-art in AS2 while maintaining low system latency.

\subsection{Fact Verification}
\label{sec:related_fact_verification}

Fact Verification is the task of predicting whether a claim is supported, refuted or neutral compared to a set of evidence sentences. This Section is related to the discoveries of Chapter~\ref{cha:jointwise}. 

An initial significant approach to address the Fact Verification task is represented by GEAR~\citep{zhou-etal-2019-gear}. In GEAR, the authors address the issue that previous methods have not effectively integrated and reasoned over multiple pieces of evidence. The proposed solution, a graph-based evidence aggregating and reasoning framework, allows the transfer of information on a fully-connected evidence graph and the usage of different aggregators to collect evidence information. The authors exploit contextual embeddings or BERT to further enhance performance.

DOMLIN~\citep{stammbach-neumann-2019-team} is another technique that targets Fact Verification. In DOMLIN, the authors design a new system to select evidence sentences that does not only consider the claim but also the previously extracted evidence. They exploit BERT for both retrieval and claims classification and show promising performance gains on the FEVER dataset.

Another relevant work regarding Fact Verification is Kernel Graph Attention Network (KGAT)~\citep{liu-etal-2020-fine}. In this work, the authors build a graph containing the evidence sentences and apply two different kernels to it: (i) node kernels, which measure the relevance of the evidence compared to the given claim and (ii) edge kernels, which propagate evidence in the graph for a more sophisticated Fact Verification. The target is to create a network able to discover claims that are syntactical and semantically acceptable but not strongly supported by the retrieved evidence. KGAT reached state-of-the-art performance on the FEVER benchmark.

In Transformers-XH~\citep{zhao2020transformer-xh}, the authors address the problem that natural languages usually have a complex structure, which can be represented with trees or graphs. Thus, they propose to extend the Attention Mechanism with eXtra Hops, a technique that enables the model to attend also between text sequences than only single tokens. This innovation improves inference over tasks in which the model should reason over several text spans, such as Multi-Hop Question Answering and Fact Verification, reaching state-of-the-art performance on HotpotQA and FEVER.

\citet{tymoshenko-moschitti-2021-strong} propose two interesting baselines for Fact Verification. They show that when using Transformer-based models, simple transformations of the output embeddings allow them to reach very good accuracies. Specifically, they exploit the evidence sentence embeddings by applying a max-pooling layer or by computing the weighted sum. Results on FEVER show that max-pooling and weighted sum are effective techniques to achieve state-of-the-art performance.

Finally, we describe two related works that use very large language models to retrieve more evidence and to perform Fact Verification. In \citet{20.500.11850/453826}, the authors improve DOMLIN by utilizing a large model for evidence retrieval and by exploiting GPT-3~\citep{brown2020language} few-shot learning capabilities to summarize the retrieved evidence for each claim in the target dataset. Their new architecture, called DOMLIN++, achieves state-of-the-art results on the FEVER dataset. The other relevant work is DREAM~\citep{zhong-etal-2020-reasoning}, in which the authors propose a new methodology to reason about the semantic structure of evidence sentences. By exploiting pre-trained language models such as XLNet~\citep{yang2020xlnet}, they create a graph that describes the semantic similarity of words. Then, they propagate the information through the graph with Graph Convolutional Networks and Graph Attention Networks to contextualize the representation of words in nodes. Exploiting the described architecture, the authors achieve state-of-the-art results on FEVER on both evaluation metrics, namely Label Accuracy and FEVER score.

\subsection{Summarization}
\label{sec:related_summarization}

Abstractive Summarization comprehends a wide range of benchmarks, models and techniques. In this Section, we try to summarize the most relevant works to provide an overview of the actual state-of-the-art in this field.

Abstractive Text Summarization is a generative task and the actual state-of-the-art is based on Auto-Regressive Transformer models in the Encoder-Decoder or Decoder-only configuration. For this reason, most of the related works are based on architectures such as BART or T5. For details about those architectures, refer to Section~\ref{sec:sota_lm}, while for the differences between the Encoder-Decoder and Decoder-only configurations, read Section~\ref{sec:transformer_models}.

The first work we review is SimCLS~\citep{liu-liu-2021-simcls}, which is a technique that combines contrastive learning with generative models to create high-quality summaries. The authors break down the generation process into two separate stages. First, an Auto-Regressive model such as BART is trained and exploited to generate $n$ summaries with Beam Search. Secondly, they design a parametric function to evaluate each summary candidate independently. More specifically, they train a RoBERTa model with a contrastive loss to measure the similarity between the original document and each possible candidate. Finally, the authors show that applying SimCLS to BART or PEGASUS improves results on several benchmarks such as CNN/DailyMail and XSum.

PEGASUS~\citep{zhang2020pegasus} is another model specifically pre-trained for Abstractive Summarization, in which an algorithm selects a subset of sentences that are removed from the original document and used as pseudo-summaries. When released, PEGASUS was shown to outperform previous SOTA on 12 Abstractive Summarization benchmarks.

Another relevant work that addresses Abstractive Summarization is BRIO~\citep{liu-etal-2022-brio}. In this work, the authors address the mismatch between training and inference on generative models, called exposure bias. While training, the model predicts the next token using Teacher-Forcing based on the assumption that all the previous tokens are correct. In inference instead, the previous tokens have been generated by the model itself and there is no guarantee they are correct. Instead of training with maximum likelihood estimation, the authors assume a non-deterministic output distribution of token probabilities and propose an innovative training algorithm in which every summary generated by the model is assigned a probability mass based on the quality. They reached state-of-the-art performance on several Summarization datasets, such as CNN/DailyMail and XSum.

SummaReranker~\citep{ravaut-etal-2022-summareranker} is a Mixture-of-Experts model that addresses again the exposure bias problem by combining a model specialized for Summarization with a re-ranker that chooses the best summary among those generated by the former. By exploiting PEGASUS as the Summarization architecture, they reached state-of-the-art models on different datasets such as CNN/DailymMail, XSum and Reddit TIFU.

At the time of writing, the state-of-the-art model for Abstractive Summarization on the datasets we consider is MoCa~\citep{zhang2022momentum}. Here, the authors address the exposure bias issue in inference, where the model should choose the best summary among possible candidates that deviated from the gold summary. The authors propose an alternative generation algorithm in which the model generates several samples with beam search that are aligned to the gold summaries while training, utilizing the momentum moving average technique.

\section{Techniques}

\subsection{Efficient Objectives for Pre-Training}
\label{sec:related_efficient}

In this Section we analyze recent techniques that try to reduce training time and carbon footprint, to reach the same final accuracy as previous state-of-the-art models. These works are related to our research about efficient and effective objectives presented in Chapter~\ref{cha:effective}.

Recently, many self-supervised pre-training objectives have been proposed to train Transformer-based models. Those objectives are mostly self-supervised because the pre-training phase is performed over large corpora of text without explicit annotation.

An example of a work that targets reduced training time by proposing architectural modifications is ALBERT, described in Section~\ref{sec:albert}. In ALBERT the parameters of every Transformer layer are tied to save memory, enabling larger batch sizes. However, this approach reduces the expressive power of the models, requiring longer pre-trainings. Another approach used by \citet{sanh2020distilbert} and \citet{turc2019wellread} is distillation. Distillation exploits a large and already pre-trained teacher model to train a lighter architecture to comparable accuracy, thus compressing the knowledge of the teacher into the smaller model. Distillation reduces the final architecture size, but the pre-training phase remains expensive due to the large teacher model.

Pre-training is typically a time-consuming process, taking weeks and requiring costly machines~\citep{liu2019roberta, brown2020language}. Therefore, it is important to explore alternative training tasks for pre-training Transformers. In \citet{tay2020efficient}, the reader can find a comprehensive survey of recent advancements in Transformer efficiency.

Regarding the learning objectives, SpanBERT~\citep{joshi-etal-2020-spanbert} introduces two new training tasks: Span Masking (SM) and Span Boundary Objective (SBO). SM masks contiguous spans of text instead of individual tokens, while SBO predicts span content using only the output representation corresponding to the boundary tokens. Another research about Transformers efficiency is proposed by \citet{zhang-etal-2020-multi-stage}, in which the authors suggest to adapt the model to the final task during pre-training to enhance downstream performance, which is related to the methods we propose in Chapter~\ref{cha:pairwise},~\ref{cha:context} and~\ref{cha:summarization}.

In T5~\citep{2020t5}, the authors propose the use of Deshuffling~\citep{liu2019summae} to pre-train an Auto-Regressive model. This technique involves shuffling random spans of text and requiring the model to output tokens in the original order. It has shown promising results across various benchmarks.

Although we focus solely on Auto-Encoder architectures, it is worth mentioning the work of \citet{izsak-etal-2021-train}, which discusses the usage of multiple optimizations for faster pre-training in Transformers. They also suggest that using larger models with the same runtime can yield better results. However, inference speed and efficiency decrease as the number of parameters of the model increases.

\subsection{Multi-Sentence Inference}
\label{sec:related_multi_sentence}

Inference over multiple sentences has already been studied in the past, also for different problems than Answer Sentence Selection. In this work, we do not target new architectures for multi-sentence inference, but effective pre-training strategies to improve models' ability at collecting information from several spans of text.

Answer Sentence Selection is a challenging task because the model needs to effectively capture the semantic connections between questions and answers. In~\citep{compagg2017bian}, the authors propose two improvements over previous state-of-the-art methods, mostly based on Compare-Aggregate networks, which are described in Section~\ref{sec:related_as2}. First, they introduce Dynamic-Clip Attention, which focuses on reducing noise in the attention matrix to better find relations between the question and the answer tokens. Secondly, they feed their architecture with a question and a set of answer candidates and task the model with predicting the relative order of the latter. They reach state-of-the-art results on WikiQA and TREC-QA.

Another relevant work about inference over multiple text spans is \citet{qingyao2018}, which addresses the Information Retrieval task. In this work, the authors demonstrate that ranking documents independently for each query is suboptimal because they may follow a different distribution in the feature space. For this reason, they develop a Deep Listwise Context Model, which is applied to the most relevant documents retrieved for each query to improve their final ranking. The proposed architecture is designed to encode multiple documents and to model their interactions, and could be applied on top of other ranking systems. By training with an attention-based loss function, they achieve state-of-the-art results on several benchmarks for Information Retrieval.

The most relevant related works targeting Answer Sentence Selection with multiple answer candidates are \citet{bonadiman-moschitti-2020-study} and \citet{zhang-etal-2021-joint}. The first work employs cheap and fast neural networks to rank multiple sentences together. It leverages the original order of answer candidates when extracted from source documents for a better ranking. When tested over WikiQA or TREC-QA, the proposed architecture is hundreds of times faster than Transformer-based models such as BERT, and achieves better performance than previous approaches with a similar number of parameters. The second work exploits mutual information and relations between several answer candidates to improve the ranking quality. Specifically, their architecture performs a three-way classification between answer candidate pairs and should predict whether candidates support, refute or are neutral compared to the others, similar to a Fact Verification pipeline. They combine state-of-the-art AS2 systems with their multi-classifier through a joint layer, which outputs scores for the ranking. The proposed architecture obtains state-of-the-art results on several AS2 datasets, such as WikiQA and TREC-QA.

\subsection{Exploiting Document Structure}
\label{sec:related_document_structure}

The intrinsic structure of documents has been leveraged in various studies. For instance, RoBERTa~\citep{liu2019roberta} generates input examples for pre-training while avoiding crossing document boundaries. This approach yields a more accurate model across multiple benchmarks because the model focuses on a single topic at a time and can be fine-tuned more effectively for target tasks.

Another work that exploits document structure is REALM~\citep{guu2020realm}, in which the authors augment models pre-training with a neural retriever that collects additional knowledge from large sources of text, such as Wikipedia. By applying the described technique in both pre-training, fine-tuning and inference, they incorporate useful additional knowledge from millions of semi-structured documents, reaching state-of-the-art on several Open Domain Question Answering benchmarks. This work differs from our pre-training setting because it exploits external knowledge to improve the model's accuracy.

DeCLUTR~\citep{giorgi-etal-2021-declutr} is yet another approach that exploits the document structure to create better sentence-level embeddings. They design a self-supervised objective that challenges the model at predicting whether two sentences are extracted from the same paragraph of a document. However, as opposed to our discoveries, they do not exploit the Cross-Attention of Transformer models and work only at the individual sentence level. The continuous pre-training they apply starting from publicly available checkpoints reduces the performance gap with models trained over labeled datasets. They show improvements in accuracy over several tasks that require reasoning over two input sentences.

\chapter{Datasets}
\label{cha:datasets}

The architecture and the training objectives proposed in this Thesis have been evaluated on a wide range of benchmarks. In this Section, we describe the domain, the number of examples and the annotation strategy of various datasets. Moreover, we provide details about how data have been collected from the original sources.

\section{Pre-Training}
\label{sec:pretraining_datasets}

Large corpora of raw text are of fundamental importance for models pre-training. Statistics for all datasets are reported in Table~\ref{tab:pretraining_datasets}.

\begin{table*}[t]
    \centering
    \small
    \resizebox{0.9\linewidth}{!}{%
    \begin{tabular}{lccccccl}
        \toprule
        \textbf{Dataset}				& \textbf{\# Docs}		& \textbf{\# Words}		& \textbf{\# Words/Doc} 		& \textbf{\# Para/Doc}			& \textbf{\# Sent/Para}		& \textbf{Size} \\

        \toprule

        Wikipedia					& 4,184,712			& 1,610,571,882		& 384.8					& 7.9						& 2.3						& 9.9GB \\
	BookCorpus				& 17,868				& 1,149,401,064		& 64,327.3					& 2005.9					& 2.5						& 6.2GB \\
	Open-Web-Text				& 8,013,769			& 6,401,507,179		& 798.8					& 20.7					& 1.9						& 38.8GB \\
	CC-News					& 149,954,415			& 62,676,498,100		& 418.0					& 16.5					& 1.3						& 385.4GB \\

        \bottomrule
    \end{tabular}
    }
    \caption{\small Statistics about the pre-training corpora exploited in this work. Cleaning involves filtering HTML tags and JavaScript code from pages, images, tables and other entities which are not raw text.}
    \label{tab:pretraining_datasets}
\end{table*}

\paragraph{Wikipedia}

The Wikipedia dataset is a collection of text data derived from the Wikipedia online encyclopedia. It encompasses a diverse range of subjects, covering fields like science, history, geography, arts, and more. It contains a substantial amount of text, including introductory paragraphs, detailed explanations, citations, and references, providing a comprehensive and authoritative source of information for a wide array of topics\footnote{We publish our pre-processed version of Wikipedia here: \url{https://huggingface.co/datasets/lucadiliello/english_wikipedia}, which corresponds to the dump of November 1, 2021.}.

\paragraph{BookCorpus}

The BookCorpus dataset~\citep{zhu2015aligning} is a relatively small collection of text data created by Google Research as a means to provide a diverse and extensive source of textual information for Machine Learning purposes. We use the Open version of this dataset in our experiments, which consists of approximately 17,868 books from a wide range of genres and topics\footnote{We used the version of the BookCorpus released here: \url{https://huggingface.co/datasets/bookcorpusopen}}. The books included in the dataset were published between 2007 and 2020 and were released under and open-source license.

\paragraph{OpenWebText}

The OpenWebText dataset is a collection of text data derived from various sources on the internet, excluding the Wikipedia website. It aims to capture a broader range of language patterns, writing styles, and topics compared to datasets focused solely on Wikipedia. The dataset is typically compiled by scraping publicly available web pages, blogs, forums, news articles, and other online sources. The original scraping script was created by OpenAI but it was not publicly released. We use an open version of the OpenWebText, which is based on Reddit dumps that are filtered based on the Karma score ($>= 3$) of the threads\footnote{We used the version of OpenWebText released here: \url{https://github.com/jcpeterson/openwebtext}}.

\paragraph{CC-News}

CC-News is a dataset released by the Common Crawl Foundation that is exclusively centered around news articles gathered from the web. CC-News aims to offer a comprehensive and diverse compilation of news articles, encompassing various topics, languages, and sources. The dataset covers articles from a wide range of news websites, covering domains such as politics, sports, technology, entertainment, and more. We parse the data of the original CC-News dataset to keep a random subset of the English articles published between 2016 and 2020\footnote{We release our version of CC-News here: \url{https://huggingface.co/datasets/lucadiliello/cc_news}} and we filter away artifacts such as HTML code, headers, titles from the WARC files.

\section{Fine-Tuning}

\subsection{Answer Sentence Selection}
\label{sec:as2_datasets}

The main task addressed in this Thesis is Answer Sentence Selection. For this task, we evaluate our proposed models and objectives on several public and two industrial datasets. Moreover, we created 2 additional corpora derived from well-known Question-Answering datasets, similar to how the authors in \citet{garg2019tanda} created ASNQ. With a total of 8 datasets, we present an exhaustive evaluation setting that includes datasets of varying sizes, ranging from very few thousand examples to more than 10 million. Moreover, we include datasets where sentence candidates are extracted from both single and multiple source documents. We created \emph{clean} versions of each dataset, which means we removed questions without at least a positive and a negative answer candidate in the development and test set, which is the standard practice in AS2~\citep{garg2019tanda, attentive-2016-santos}. This allows us to better compare results because otherwise, there would have been upper bounds to the AS2 metrics dependent on the tested dataset.

\paragraph{ASNQ}

The Answer Sentence Natural Question (ASNQ) is a large dataset for AS2 derived from Google's Natural Questions corpora~\citep{kwiatkowski-etal-2019-natural}. The questions have been retrieved from Google's search engine and the system is required to reason over entire Wikipedia articles to find the correct answer. NQ is a challenging dataset because it contains thousands of questions asked by real users and answers extracted from long Wikipedia articles.
The annotations include both \emph{long} and \emph{short} answers. The short answers are very small text spans that contain the exact answer to the question. Long answers are instead paragraphs that contain the exact answer but also other sentences and text. Most of the short answers are contained in a long answer. There are very few rare cases in which the short answers are not contained in the long answers, probably due to annotation mistakes.
More specifically, NQ contains 307,373 train and 7,830 development user questions and, for each question, the most relevant Wikipedia article is provided. For the train set, a single long answer annotation is available. For the development set instead, the authors asked 5 different annotators to select a long answer. Regarding the short answers, on average there are 0.42 annotations in the train set and 2.10 in the development set.

ASNQ is derived from NQ by \citet{garg2019tanda} with the following rules. First, the authors split every retrieved Wikipedia document into a list of answer sentence candidates with the NLTK tokenizer~\citep{bird2009natural}. Then, they assign a positive label to answer candidates which are entirely contained in a long answer and that contain at least a short answer. Negative answer candidates comprehend instead: (i) sentences in a long answer but not containing a short answer; (ii) sentences containing a short answer but not contained in a long answer; (iii) sentences both not in a long answer and not containing a short answer. Notice that the formers are very hard negatives because they are very close to the short answer (in the same paragraph) and contained in a long answer. Moreover, the second case happens only because of a few mistakes in the annotation process. Finally, the authors discard questions not having at least a positive answer candidate.

The statistics for the resulting ASNQ dataset are reported in Table~\ref{tab:dataset_statistics}. For the experiments in Chapter~\ref{cha:effective}, we report results on the original ASNQ development set, because the dataset does not provide a test set. The results in Chapters~\ref{cha:jointwise}, ~\ref{cha:pairwise} and \ref{cha:context} are instead on the test set released by \citet{soldaini-moschitti-2020-cascade}, where the authors split the original development set for both validation and testing.

\paragraph{WikiQA}

\begin{table*}[t]
	\center
	\resizebox{0.9\textwidth}{!}{%
	\begin{tabular}{lcccccccccccc}
		\toprule
		\multirow{2}{*}{\textbf{Dataset}} & \multicolumn{4}{c}{\textbf{Train set}} & \multicolumn{4}{c}{\textbf{Development set}} & \multicolumn{4}{c}{\textbf{Test set}} \\
		\cmidrule(lr){2-5} \cmidrule(lr){6-9} \cmidrule(lr){10-13}
		& \textbf{Q} & \textbf{QA} & \textbf{A/Q} & \textbf{+/Q} & \textbf{Q} & \textbf{QA} & \textbf{A/Q} & \textbf{+/Q} & \textbf{Q} & \textbf{QA} & \textbf{A/Q} & \textbf{+/Q} \\
		\toprule
		ASNQ		& 57242	& 20377568	& 356.0	& 1.1		& 1336	& 463914	& 347.2	& 1.6		& 1336	& 466148		& 348.9 	& 1.6 \\
		WikiQA		& 2118	& 20360		& 9.6		& 0.5		& 122	& 1126	& 9.2		& 1.1		& 237	& 2341		& 9.9 	& 1.2 \\
		TREC-QA		& 1158	& 53310		& 46.0	& 5.4		& 69		& 1343	& 19.5	& 3.5		& 68		& 1442		& 21.2	& 3.5 \\
		\bottomrule
	\end{tabular}%
	}
	\caption{\small Statistics for ASNQ, WikiQA and TREC-QA. ``Q'' and ``QA'' stand for the number of unique question and question-answer pairs. ``A/Q'' is the average number of candidates per question while ``+/Q'' is the average number of positive answers per question.}
	\label{tab:dataset_statistics}
\end{table*}

WikiQA is a small and challenging dataset for Answer Sentence Selection released by Microsoft and based on queries asked to the Bing search engine~\citep{yang-etal-2015-wikiqa}. Based on research results, each question is associated with a Wikipedia page. Sentences from the summary (first paragraph) are then used as answer candidates and are annotated by crowd-workers. To ensure a high annotation quality, each sentence is independently analyzed by three different workers. Sentences with inconsistent labels were further parsed by a different set of workers and finally, labels were assigned based on majority vote. Statistics about the number of questions and answer candidates are given in Table~\ref{tab:dataset_statistics}.

\paragraph{TREC-QA}

TREC-QA is yet another popular dataset for Answer Sentence Selection created from the TREC-8 to TREC-13 tracks of Question Answering~\citep{wang-etal-2007-jeopardy}. The corpus is composed of factoid questions while answers are extracted from web documents by participating teams. In particular, the authors of \citet{wang-etal-2007-jeopardy} use the tracks TREC-8 to TREC-12 for training and the track TREC-13 for development and testing. Manual annotation of positive candidates is performed on both the development and test sets. In the training set, only the answers to 100 questions received gold labels while the others were automatically annotated, thus being more noisy. The split containing also the manually annotated candidate answers is called \emph{train-all}. We train over train-all, which is the standard practice~\citep{attentive-2016-santos}, because even if it is noisier, it contains much more question-answer pairs. Negative answer candidates are automatically extracted by selecting sentences with overlapping words with the question. Statistics for TREC-QA are given in Table~\ref{tab:dataset_statistics}.

\paragraph{IQAD}

IQAD (Internal Question Answering Dataset) is a large-scale internal dataset of the Alexa Search team used to evaluate QA systems. The dataset was built from Alexa traffic by de-identifying user questions asked to Alexa, which is a popular virtual assistant. The de-identification process removes all connections between questions and original Alexa customers. The dataset contains about 220K general topic questions and a large set of answer candidates extracted with ElasticSearch from a web index containing more than 1 billion pages. More detailed statistics are reported in Table~\ref{tab:dataset_statistics_2}. IQAD contains 2 different test sets that differ in the period in which data were gathered and in the annotation procedure. The first set contains 2.2K questions, with about 17 answer candidates per question, annotated manually by crowd-workers. The second test split contains 2K questions instead and about 16 answer candidates per question annotated by crowd-workers following strict fact verification guidelines. A manual analysis of a subsample of 100 questions for each dataset showed higher annotation quality in the second set. In the experimental Sections, we refer to the first test set as ``IQAD Bench 1'' while the second is named ``IQAD Bench 2''. In summary, IQAD is a very challenging dataset because it reflects real user questions, which may be vague, ambiguous or malformed.

\paragraph{WQA}

WQA (Web Question Answering) is yet another AS2 dataset created sampling questions asked to the Alexa virtual assistant in 2019~\citep{zhang-etal-2021-joint, hsu-etal-2021-answer}. Questions are non-representative and are anonymized to remove connections to the original users and geographical locations. Moreover, questions are not filtered by topic, so they span several different arguments. For each question, 500 pages were extracted from an index of more than 100M documents crawled from the web using ElasticSearch. Then, answer candidates has been ranked with a state-of-the-art AS2 system and the top 100 were annotated by crowd-workers. Statistics for this dataset are given in Table~\ref{tab:dataset_statistics_2}. Similarly to IQAD, WQA is a challenging corpus because questions mirror real Alexa customer queries, which usually are noisy, malformed or misleading.

\begin{table*}[t]
	\center
	\resizebox{0.9\textwidth}{!}{%
	\begin{tabular}{lcccccccccccc}
		\toprule
		\multirow{2}{*}{\textbf{Dataset}} & \multicolumn{4}{c}{\textbf{Train set}} & \multicolumn{4}{c}{\textbf{Development set}} & \multicolumn{4}{c}{\textbf{Test set}} \\
		\cmidrule(lr){2-5} \cmidrule(lr){6-9} \cmidrule(lr){10-13}
		& \textbf{Q} & \textbf{QA} & \textbf{A/Q} & \textbf{+/Q} & \textbf{Q} & \textbf{QA} & \textbf{A/Q} & \textbf{+/Q} & \textbf{Q} & \textbf{QA} & \textbf{A/Q} & \textbf{+/Q} \\
		\toprule
		\multirow{2}{*}{IQAD}& \multirow{2}{*}{221334}	& \multirow{2}{*}{3894129} & \multirow{2}{*}{17.6} 	& \multirow{2}{*}{-}	& \multirow{2}{*}{2434}	& \multirow{2}{*}{43369}	& \multirow{2}{*}{17.8} 	& \multirow{2}{*}{-}	& 2252 & 38587 & 17.1 & - \\
						& 					& 					 & 					& 				&					&					&			           	&				& 2088 & 33498 & 16.0 & - \\
		WQA			& 9984				& 149513				 & 15.0				& 0.23			& 5000				& 74805				& 15.0				& 0.21			& 5000 & 74712 & 14.9 & 0.21 \\
	\bottomrule
	\end{tabular}%
	}
	\caption{\small Statistics for IQAD and WQA.}
	\label{tab:dataset_statistics_2}
\end{table*}

\paragraph{NewsAS2}

NewsAS2\footnote{We release NewsAS2 here: \url{https://huggingface.co/datasets/lucadiliello/news_as2}} is a dataset derived from NewsQA~\citep{trischler-etal-2017-newsqa}, which was originally designed for Machine Reading.

NewsQA is a dataset composed of over 100K question-answer pairs generated by crowd-workers. First, the authors extracted 12.7K articles from the CNN/Daily Mail corpus with random selection. Those articles cover several topics, such as politics and business. Then, crowd-workers have been divided into three groups. The first group task was to generate questions by letting the workers see only the article's title and a short summary, thus promoting curiosity about the article's content. The second group was then asked to highlight answers in the documents, if present, or to reject questions when nonsensical or malformed. A positive label was assigned to answer on which at least two crowd-workers agreed. Finally, to improve the dataset quality, the authors asked the third group of workers to review the questions without an agreement on the answer and to make the final decision, or to reject them.

We downloaded NewsQA from the MRQA~\citep{fisch-etal-2019-mrqa} competition repository\footnote{\url{https://github.com/mrqa/MRQA-Shared-Task-2019}} and split the development set in both dev and test because the original test set in not labeled.
The conversion from MR to Answer Sentence Selection has been performed with the following algorithm. First, each sample $(q, \mD)$ from the original dataset was transformed in a tuple $(q, \{ s_1, s_2, \dots, s_n \})$ by splitting the document $\mD$ into multiple sentences with the NLTK tokenizer~\citep{bird2009natural}. Then, each sentence $s_i$ was labeled as positive if it entirely contained one of the original answer spans, as negative otherwise. NewsAS2 contains more than 70K questions and 1.8M question-answer pairs. The number of questions without at least a positive answer candidate is 1.5\% in the development set and 1.1\% for the test set. We remove those questions from the corresponding sets (clean setting). Statistics for NewsAS2 are reported in Table~\ref{tab:dataset_statistics_3}.
 
\paragraph{TriviaAS2}

TriviaAS2\footnote{We release TriviaAS2 here: \url{https://huggingface.co/datasets/lucadiliello/trivia_as2}} is a dataset for AS2 derived from TriviaQA \citep{joshi-etal-2017-triviaqa}, which was originally designed for Machine Reading.

TriviaQA is a challenging dataset for Reading Comprehension composed of questions and answers retrieved from 14 trivia and quiz-league websites. In particular, about 95K trivia questions authored by enthusiasts were extracted, and for each question, an average of 6 documents were crawled using the Bing search engine. Questions were filtered to have a minimum length of 4 tokens while documents were filtered when containing keywords like trivia, question and answer or when they contained ill-formatted text, such as PDFs. In addition, for each query, the authors extracted an additional set of about 2 documents from Wikipedia by searching the entities contained in the questions. Finally, documents were filtered when not containing the original answer to promote learning with distant supervision.

As for NewsQA, we download TriviaQA from MRQA and split the development into both development and test to have non-hidden annotations. The conversion from MR to AS2 is performed the same algorithm used for NewsAS2. Statistics for the resulting TriviaAS2 are given in Table~\ref{tab:dataset_statistics_3}.

\begin{table*}[t]
	\center
	\resizebox{0.9\textwidth}{!}{%
	\begin{tabular}{lcccccccccccc}
		\toprule
		\multirow{2}{*}{\textbf{Dataset}} & \multicolumn{4}{c}{\textbf{Train set}} & \multicolumn{4}{c}{\textbf{Development set}} & \multicolumn{4}{c}{\textbf{Test set}} \\
		\cmidrule(lr){2-5} \cmidrule(lr){6-9} \cmidrule(lr){10-13}
		& \textbf{Q} & \textbf{QA} & \textbf{A/Q} & \textbf{+/Q} & \textbf{Q} & \textbf{QA} & \textbf{A/Q} & \textbf{+/Q} & \textbf{Q} & \textbf{QA} & \textbf{A/Q} & \textbf{+/Q} \\
		\toprule
		NewsAS2		& 71561	& 1840533	& 25.7	& 1.7		& 2070	& 50895	& 24.6	& 1.7		& 2059	& 50842	& 24.7 	& 1.7 \\
		TriviaAS2		& 61688	& 1843349	& 29.9	& 1.7		& 1201	& 37470	& 31.2	& 5.3		& 1190	& 37691	& 31.7 	& 5.4 \\
		\bottomrule
	\end{tabular}%
	}
	\caption{\small Statistics for NewsAS2 and TriviaAS2.}
	\label{tab:dataset_statistics_3}
\end{table*}

\subsection{Fact Verification}
\label{sec:fact_verification_datasets}

Fact Verification is the task of predicting whether a claim is entailed, neutral or contradicted by a series of evidence sentences. Given the nature of this task, which requires the model to reason over multiple sentences, it is well suited for our Multi-Sentence Inference architecture, see Section~\ref{sec:self_supervised_for_multi_sentence_inference}.

\paragraph{FEVER}

FEVER~\citep{thorne-etal-2018-fever}, an acronym for ``Fact Extraction and VERification'', is a challenging dataset composed of about 185K claims and 920K evidence sentences extracted from Wikipedia.

The data has been generated with the following procedure. First, the initial section of 50K popular pages from Wikipedia has been extracted and cleaned. Then, a set of crowd-workers was tasked to generate claims containing a single statement related to the page topic. To avoid the creation of too complex claims, the crowd-workers were provided with a limited set of words collected from linked pages that could be used for the generation. Moreover, crowd-workers were allowed to modify existing claims, thus changing whether they were supported by the available evidence or not.

In the second phase, crowd-workers were provided with claims and were asked to find a set of evidence texts to support or refute the claim. If the claim could not be supported or contradicted, the workers should have provided evidence for their decision.

Since FEVER's original task involves both Information Retrieval of the evidence sentences and Recognizing Textual Entailment, we used the evidence collected by \citet{thorne-etal-2018-fever} using a BERT-based DocIR\footnote{We release our version of the FEVER dataset here: \url{https://huggingface.co/datasets/lucadiliello/fever}}. The final dataset statistics are reported in Table~\ref{tab:dataset_statistics_4}. On average, for each claim, a set of 5 evidence candidates is available to predict the target label. To evaluate our models, we submitted them to the FEVER shared task website\footnote{\url{https://competitions.codalab.org/competitions/18814}} because labels for the test set are hidden.

\begin{table*}[t]
	\center
	\resizebox{0.7\textwidth}{!}{%
	\begin{tabular}{lccccccccc}
		\toprule
		\multirow{2}{*}{\textbf{Dataset}} & \multicolumn{3}{c}{\textbf{Train set}} & \multicolumn{3}{c}{\textbf{Development set}} & \multicolumn{3}{c}{\textbf{Test set}} \\
		\cmidrule(lr){2-4} \cmidrule(lr){5-7} \cmidrule(lr){8-10}
		& \textbf{C} & \textbf{E} & \textbf{E/C} & \textbf{C} & \textbf{E} & \textbf{E/C} & \textbf{C} & \textbf{E} & \textbf{E/C}  \\

		\toprule
		FEVER		& 145405	& 722473	& 4.97		& 19998	& 98915	& 4.94	& 19998		& 98839	& 4.94 \\
		\bottomrule
	\end{tabular}%
	}
	\caption{\small Statistics for the FEVER dataset. ``C'' and ``E'' are the total number of claims and evidence sentences, while ``E/C'' is the average number of evidence sentences per claim.}
	\label{tab:dataset_statistics_4}
\end{table*}

\subsection{Summarization}
\label{sec:summarization_datasets}

The statistics for all datasets about Abstractive Summarization used as benchmarks in this Thesis are reported in Table~\ref{tab:dataset_statistics_5}.

\paragraph{CNN/DailyMail}

The CNN/DailyMail dataset is a widely used benchmark dataset for Abstractive Summarization which was developed by \citet{cnn-dm-2015}. The dataset is composed of news articles scraped from the CNN and The Daily Mail news websites between 2010 and 2015. Every article of the dataset is paired with summaries that were written by humans, providing a concise and informative representation of the article's main points. The dataset covers a wide range of topics, including politics, technology, science, sports, and entertainment. The articles are written in English and have a mix of short and long sentences, making the dataset challenging to summarize.

\paragraph{XSum}

XSum is a large-scale dataset for text Summarization that was created by \citet{narayan-etal-2018-dont}. The XSum dataset consists of approximately 227K news articles and their corresponding summaries. The articles were sourced from a wide range of news outlets, including The Guardian, The Independent, and The Daily Mail and cover a variety of topics, including politics, business and sports. The summaries in the XSum dataset are typically one or two sentences in length and provide a concise description of the main points of the article. The dataset was built with a two-stage process. First, a group of human annotators wrote a summary for each article. Then, a second group of human annotators reviewed the summaries and edited them to ensure they provided an accurate and comprehensive summary of the article, thus ensuring high-quality summaries.

\paragraph{Samsum}

The Samsum dataset is a corpus for conversations Summarization that was introduced by \citet{samsum}. The dataset consists of over 11,000 conversations between two participants with the scope of scheduling a meeting. Each conversation contains a set of messages exchanged between the participants, and the goal of the Summarization task is to produce a concise summary of the conversation that captures its most important aspects.

Conversational Summarization poses several challenges that are not present in traditional document Summarization. Conversations are often highly contextualized and contain a large amount of implicit information, such as the participants' intentions, preferences, and expectations. As a result, summarizing a conversation requires not only identifying important content but also understanding the underlying discourse and social interactions.

In the Samsum dataset, a set of small summary sentences were produced by human annotators. Each summary is intended to capture the key points of the conversation and provide a concise overview of its content.

\paragraph{Gigaword}

Gigaword is a large-scale text corpus for Abstractive Summarization created by the Linguistic Data Consortium (LDC)~\citep{graff2003english}. The dataset contains over one billion words and covers a wide range of topics and domains over more than two decades of events, from 1994 to 2015. The documents are news articles extracted from a variety of sources, such as the North American News, the AQUAINT corpus and the Agence France-Presse. In the Gigaword dataset, every article is paired with a very short summary that captures its key points. These summaries are typically one to three sentences long and were written by human editors.

\begin{table*}[t]
	\center
	\resizebox{0.9\textwidth}{!}{%
	\begin{tabular}{lcccccccccc}
		\toprule
		\multirow{2}{*}{\textbf{Dataset}} & \multirow{2}{*}{\textbf{Set}} & \multirow{2}{*}{\textbf{Examples}} & \multicolumn{4}{c}{\textbf{Document Length}} & \multicolumn{4}{c}{\textbf{Summary Length}} \\
		\cmidrule(lr){4-7} \cmidrule(lr){8-11}
		& & & \textbf{Mean} & \textbf{Std} & \textbf{Min} & \textbf{Max} & \textbf{Mean} & \textbf{Std} & \textbf{Min} & \textbf{Max} \\
		\toprule
		\multirow{3}{*}{CNN/DailyMail}	& Train	&  287113 	& 4033.6	& 1954.3	& 48		& 15925		& 294.7	& 120.1	& 14	& 7388 \\
								& Dev	&  13368		& 3923.8	& 1992.3	& 245	& 11412		& 328.0	& 146.2	& 52	& 8541 \\
								& Test	&  11490		& 3967.0	& 2024.7	& 293	& 11991		& 311.9	& 128.0	& 51	& 3410 \\
		\midrule
		\multirow{3}{*}{XSum	}		& Train	&  204017		& 2202.4	& 1795.5	& 15		& 174045		& 125.4	& 30.2	& 1	& 399 \\
								& Dev	&  11327		& 2174.7	& 1664.9	& 26		& 27345		& 125.5	& 30.4	& 5	& 500 \\
								& Test	&  11333		& 2214.5	& 1817.6	& 14		& 63937		& 125.3	& 30.5	& 17	& 410 \\
		\midrule
		\multirow{3}{*}{Samsum}		& Train	& 14732		& 511.1	& 402.6	& 0		& 5492		& 110.1	& 60.8	& 3	& 300 \\
								& Dev	& 819		& 499.3	& 403.3	& 57		& 2950		& 110.1	& 61.3	& 17	& 300 \\
								& Test	& 818		& 521.5	& 409.1	& 49		& 2793		& 108.7	& 58.4	& 17	& 300 \\
		\midrule
		\multirow{3}{*}{Gigaword}		& Train	& 3803957	& 181.5	& 47.3	& 35		& 636		& 51.0	& 14.7	& 6	& 265 \\
								& Dev	& 189651		& 181.5	& 47.3	& 40		& 520		& 51.8	& 14.4	& 7	& 158 \\
								& Test	& 1951		& 170.8	& 56.4	& 3		& 409		& 52.2	& 19.3	& 7	& 137 \\
	\bottomrule
	\end{tabular}%
	}
	\caption{\small Statistics for all Summarization datasets employed in this work. We report the number of examples for each split and statistics about the length of documents and summaries (in terms of characters, including whitespaces). As a rule of thumb, the number of characters per token is $\sim4$ for most Transformer models.}
	\label{tab:dataset_statistics_5}
\end{table*}

\section{Benchmark Suites}
\label{sec:benchmarks}

\begin{table*}[h]
	\center
	\resizebox{1.0\textwidth}{!}{%
	\begin{tabular}{lcccc@{\hskip 6pt}ccc@{\hskip 6pt}cc}
		\toprule
		\multirow{2}{*}{\textbf{Dataset}} & \multirow{2}{*}{\textbf{Task type}} & \multirow{2}{*}{\textbf{Classes}} & \multicolumn{2}{c}{\textbf{Train set}} & & \multicolumn{2}{c}{\textbf{Development set}} & & \textbf{Test set} \\
		\cmidrule{4-5} \cmidrule{7-8} \cmidrule{10-10}
		& & & \textbf{Examples} & \textbf{Examples/Class (\%)} & & \textbf{Examples} & \textbf{Examples/Class (\%)} & & \textbf{Examples} \\
		\toprule
		CoLA		& Classification	& 2		& 8551 & 29.5/70.5			& & 1043 & 30.9/69.1		& & 1063 \\
		MNLI-m		& Classification	& 3		& 392702 & 33.3/33.3/33.4	& & 9815 & 35.5/31.8/32.7	& & 9796 \\
		MRPC		& Classification	& 2		& 3668 & 32.6/67.4			& & 408 & 31.6/68.4			& & 1725 \\
		QNLI			& Classification	& 2		& 104743 & 50.0/50.0		& & 5463 & 49.5/50.5		& & 5463 \\
		QQP			& Classification	& 2		& 363846 & 63.1/36.9		& & 40430 & 63.2/36.8		& & 39095 \\
		RTE			& Classification	& 2		& 2490 & 50.2/49.8			& & 277 & 52.7/47.3			& & 3000 \\
		SST-2		& Classification	& 2		& 67349 & 44.2/55.8			& & 872 & 49.1/50.9			& & 1821 \\
		STS-B		& Regression	& -		& 5749 & 	-				& & 1500 & -				& & 1379 \\
		\bottomrule
	\end{tabular}
	}
	\caption{\small Statistics for all datasets inside the GLUE benchmark suite. ``\%Examples/Class'' shows the percentage of examples for each label class. The labels of the test set are hidden, thus we do not show statistics about their class distribution.}
	\label{tab:dataset_statistics_6}
\end{table*}

\paragraph{GLUE}

GLUE (General Language Understanding Evaluation)~\citep{wang-etal-2018-glue} is a collection of datasets for testing systems on a variety of tasks. It includes CoLA, a benchmark for Language Acceptability~\citep{warstadt-etal-2019-neural}; QNLI, MNLI and WNLI, three datasets for Natural Language Inference~\citep{wang-etal-2018-glue, williams-etal-2018-broad, 10.5555/3031843.3031909}; RTE, to test performance in Textual Entailment~\citep{giampiccolo-etal-2007-third}; QQP\footnote{\url{https://quoradata.quora.com/First-Quora-Dataset-Release-Question-Pairs}} and MRPC for paraphrasing~\citep{dolan-brockett-2005-automatically}; STS-B for sentence similarity~\citep{cer-etal-2017-semeval}; SST-2, to test models in Sentiment Analysis~\citep{socher-etal-2013-recursive}.

We do not include WNLI in our experiments because it is hard to beat even a trivial majority classifier~\citep{devlin-etal-2019-bert}. As for MNLI, we use the \emph{matched} version, usually indicated with MNLI-m. Statistics for each dataset are reported in Table~\ref{tab:dataset_statistics_6}. A more fine-grained analysis of each dataset contained in GLUE is given in Appendix~\ref{app:glue}. Notice that recently, a new version of this benchmark called SuperGLUE~\citep{wang2019superglue} has been released, and we encourage its usage for future research.

\chapter{Alternative Efficient Objectives}
\label{cha:effective}

The first part of this work focuses on Transformers pre-training, which is a core part of building Artificial General Intelligence. Transformers pre-training is a task that requires (i) costly and large computational infrastructures and a lot of energy, (ii) high-level skills to manage distributed training and peer-to-peer efficient communications, (iii) collection and cleaning/filtering of large amounts of raw data and (iv) well-tested code implementations to avoid nodes failure~\citep{strubell-etal-2019-energy, brown2020language}.

In this Chapter, we address the first issue mentioned before by proposing new pre-training tasks and objectives that require shorter pre-training to reach the same level of final performance. The other issues are generally addressed by large organizations which develop high-end frameworks to help researchers and developers. We mention: (i) PyTorch~\citep{pytorch} and Tensorflow~\citep{tensorflow2015}, which allow for fast ANN development and training through automatic back-propagation; (ii) DeepSpeed~\citep{deepspeed} and ColossalAI~\citep{bian2021colossal} that increment training speed and reduce inference time by sharding model's weights and through efficient communication protocols between nodes in distributed settings; (iii) CommonCrawl~\citep{commoncrawl} and The Pile~\citep{gao2020pile}, which are large sources of cleaned text freely available and (iv) PyTorch-Lightning~\citep{Falcon_PyTorch_Lightning_2019} and HuggingFace~\citep{wolf-etal-2020-transformers}, which help developers by proving frameworks to abstract from the underlying hardware and repositories with thousands of ready-to-be-used models.

Many recent works focused on improving pre-training efficiency through the implementation of innovative objectives or by improving the model architecture~\citep{lan2020albert,sanh2020distilbert,turc2019wellread}. For example, the authors of ELECTRA trained a BERT model as a discriminator instead of a generator~\citep{clark2020electra}. Instead of masking a random subset of input tokens, ELECTRA exploits a small Language Model trained with MLM to create challenging replacements. Then, the discriminator is tasked to predict which tokens are original and which are fakes introduced by the smaller generator network.

ELECTRA introduces many innovations, such as more efficient training and the missing discrepancy between pre-training and fine-tuning because no special [MASK] token is used in the discriminator training. However, it is not clear whether the improvements of ELECTRA over BERT derive more from the generator-discriminator architecture or the longer and more costly pre-training.

We provide an extensive description of the works related to this research in Section~\ref{sec:related_efficient}. Notice that they are orthogonal to our proposed methodology, which emphasizes the efficiency of the pre-training objective and the classification head size. Therefore, they can be combined with our alternative pre-training objectives. Additionally, works related to the models and the tasks we use in this Chapter can be found in Section~\ref{sec:sota_lm} and~\ref{sec:related_tasks}, respectively.

We study a batch of new pre-training strategies designed to reduce training costs while maintaining the same level of final performance. We provide a thoughtful study of both theoretical cost and experimental training time since the two do not always align well due to underlying hardware optimization techniques and different levels of memory usage. Our contributions can the summarized as the following three self-supervised objectives~\citep{di-liello-etal-2022-effective}:

\begin{itemize}
\item \textbf{Random Token Substitution (RTS)}: an efficient alternative to ELECTRA that drops the expensive generator network and works by randomly replacing some tokens with others and by tasking the model at recognizing which are originals and which are not;
\item \textbf{Cluster-based Random Token Substitution (C-RTS)}: a technique that improves over RTS using previous predictions to find more challenging replacements utilizing a simple and efficient statistical approach;
\item \textbf{Swapped Language Modeling (SLM)}: an alternative to BERT's MLM which randomly replaces tokens into others and asks the model to predict the original values.
\end{itemize}

The next Sections will cover a description of our proposed objectives and a comparison of the computational complexity with the SOTA pre-training tasks for Transformers. Then, we provide details about the experiment setting we used to evaluate our claims and results obtained over many different benchmarks. We also report negative findings in Appendix~\ref{app:effective_negative} to help future research.

\begin{figure*}[ht]
	\centering
	\begin{subfigure}{0.33\columnwidth}
		\centering
		\includegraphics[width=\linewidth,height=24em,keepaspectratio]{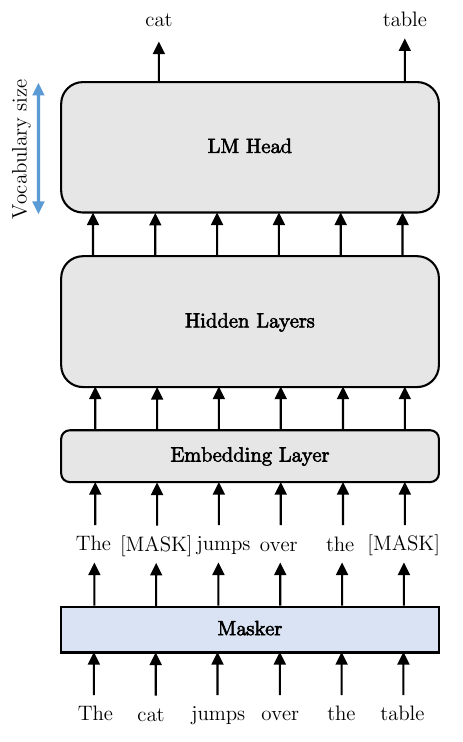}
		\caption{\small Masked Language Modeling}
    		\label{fig:bert}
	\end{subfigure}\hfill
	\begin{subfigure}{0.67\columnwidth}
		\centering
		\includegraphics[width=\linewidth,height=24em,keepaspectratio]{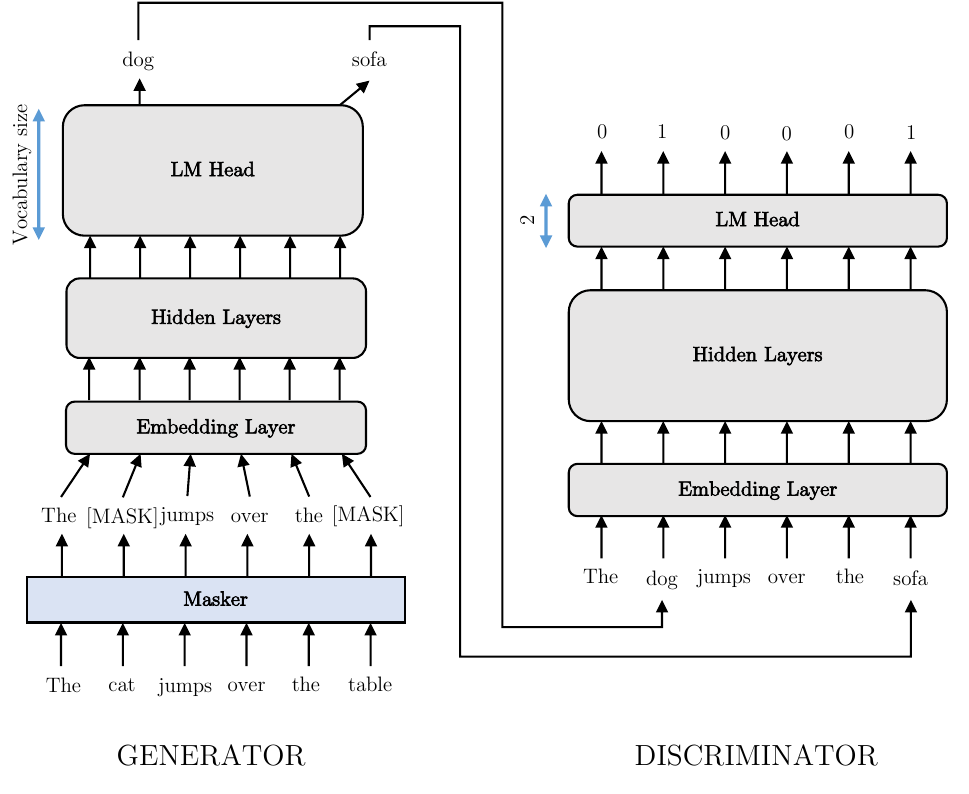}
		\caption{\small Token Detection}
    		\label{fig:electra}
	\end{subfigure}
	\caption{\small BERT's MLM and ELECTRA's TD objectives examples.}
\end{figure*}

\begin{figure*}[ht]
	\centering
	\begin{subfigure}{0.32\columnwidth}
		\centering
		\includegraphics[width=\linewidth,keepaspectratio]{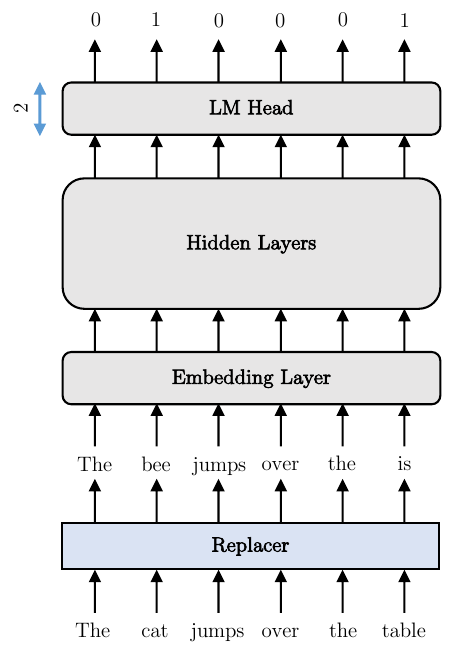}
		\caption{\small Random Token Substitution}
    		\label{fig:rts}
	\end{subfigure}\hfill
	\begin{subfigure}{0.35\columnwidth}
		\centering
		\includegraphics[width=\linewidth,keepaspectratio]{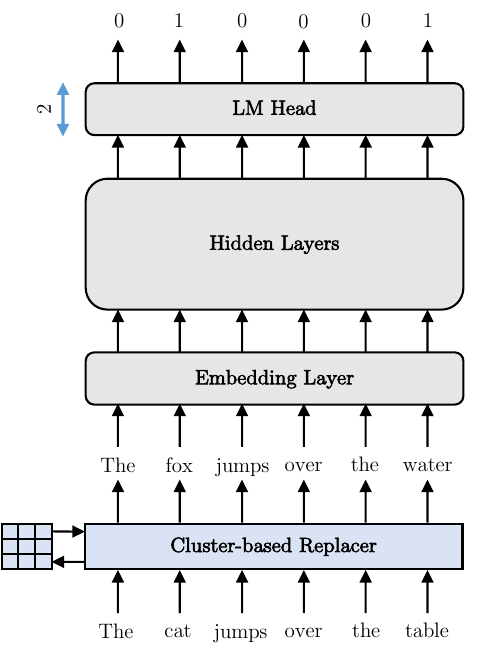}
		\caption{\small Cluster-based Random Token Substitution}
    		\label{fig:crts}
	\end{subfigure}\hfill
	\begin{subfigure}{0.32\columnwidth}
		\centering
		\includegraphics[width=\linewidth,keepaspectratio]{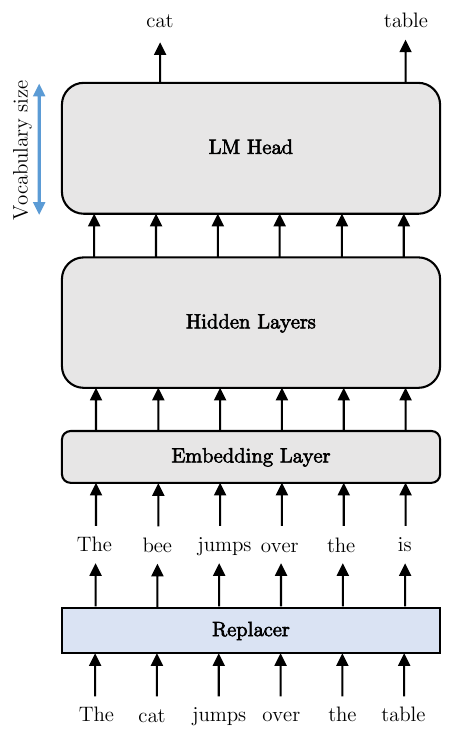}
		\caption{\small Swapped Language Modeling}
    		\label{fig:slm}
	\end{subfigure}\hfill
	\caption{\small Examples for RTS, C-RTS and SLM objectives.}
\end{figure*}

\section{Alternative Objectives}
\label{sec:alternative_objectives}

In this Section, we illustrate our proposed objectives, highlighting the advantages compared to state-of-the-art methods.

\subsection{Random Token Substitution}

RTS is a variant of the ELECTRA architecture in which the expensive generator network is substituted with a straightforward random sampling approach. Examples of how ELECTRA and RTS work are depicted in Figure~\ref{fig:electra} and~\ref{fig:rts}, respectively. RTS uses a very small classification head on top of the Transformer for binary classification, which reduces computational load and memory usage, allowing for larger batch sizes.

In RTS, we randomly select the 15\% of the input tokens and we change them into others sampled with uniform distribution from the vocabulary. Then, the model is tasked to recognize which tokens are original and which have been replaced with others.

\paragraph{Advantages}

RTS necessitates a minimal classification head, and predictions are not carried out across the entire vocabulary. Some might argue that removing the MLM head would not result in significant improvements since the embedding layer, which shares weights with the language modeling classification head, still acts as a bottleneck. However, this is not accurate, as the embedding layer functions by indexing the weights matrix to extract only the representation corresponding to each input token. In contrast, the language modeling head computes probabilities for all tokens in the vocabulary, making it computationally much more expensive by several orders of magnitude.

Another important advantage of RTS is that there is no pre-training/fine-tuning discrepancy because the [MASK] token is not used in both pre-training and fine-tuning~\citep{clark2020electra}.

\paragraph{Drawbacks}

RTS works well with Transformer models containing up to 100M parameters, however, we do not suggest its usage for larger models because it may lead to overfitting, as shown in the successive sections. When the model size increases, the RTS task can be solved with very high accuracies because most replaced tokens are easy to discover, leading to weak error signals and lower model quality.

\subsection{Cluster-based Random Token Substitution}

C-RTS is an improvement of RTS that targets the overfitting issue by exploiting the history of previous predictions to find more challenging replacements. The main idea of C-RTS is to model the miss-classification probability of the model for each token independently through a simple statical approach.

More formally, given a Transformer model $\mathcal{M}$ and sequence of tokens $[t_1, t_2, \dots, t_L]$ in input, $\mathcal{M}$ outputs a sequence of predictions $\vp$ for the labels $\vy$, where $y_i = 0$ means that $t_i$ is original and $y_i = 1$ indicates that $t_i$ was replaced with some other token $t_i'$ ($t_i \rightarrow t_i'$). Our objective is to maximize:
\begin{equation}
P(p_i = 0 \ | \ t_i \rightarrow t_i')
\label{eq:crts}
\end{equation}
since we want to find challenging replacements that are hard to be detected by $\mathcal{M}$.

Equation~\ref{eq:crts} could be estimated by counting the number of successes and failures of the model $\mathcal{M}$ at recognizing false input tokens in the previous training steps. An important difference between C-RTS and ELECTRA is that we do statistics over individual tokens, while ELECTRA's generator uses the whole input context to find challenging alternatives.

\paragraph{Clustering}

Since storing a matrix containing counts for every pair of tokens would lead to a very large memory footprint (vocabulary sizes range from  30K~\citep{devlin-etal-2019-bert} entries to over 120K~\citep{he2020deberta}), we split the vocabulary into $n$ clusters with similar size. We perform the clustering based on token similarity. However, since we do not have an already-trained embedding layer yet, we train a \emph{word2vec}~\citep{word2vec} model to obtain vectorial representations. We run word2vec on the same data that we use for pre-training, to avoid any performance improvement stemming from the usage of additional text.

In particular, we configure word2vec to use a context size of 2 words on both sides of the target token and we set the embedding size to 300. We use the PyTorch implementation of word2vec~\citep{pytorch} and the whole training takes less than 10 minutes on our machine. Given the negligible overhead of this step when compared to the pre-training time of Transformers, we omit those 10 minutes from the final training-time comparisons.

Once word embeddings are computed, we used the K-means algorithm~\citep{Lloyd82leastsquares} to group tokens into clusters. We implemented the script by exploiting the implementation of K-means in the \emph{scikit-learn}~\citep{scikit-learn} library. To choose the best number of clusters $n$, we first run the clustering algorithm with 20 random restarts for each value of $n \in \{30, 100, 300, 1000\}$. This experiment took about 20 minutes on the machine described in Section~\ref{sec:machines}. Then, we search for the best number of clusters by running a short pre-training of 5000 steps and by selecting the value that gives the smallest accuracy of C-RTS on the pre-training dev set.

\paragraph{Statistic-based replacement}

Once we partition the tokens of the vocabulary $\sV$ into clusters $\{ \sC_1, \dots, \sC_n \}$, we estimate the probability $P(p_i = 0 \ | \ t_i \rightarrow t_i')$ with the following methodology. First, we use a matrix $\mF \in \sZ^{n \times n}$ to count the number of successes and failures of the model. At each training step, if $t_i \in \sC_a \rightarrow t_i' \in \sC_b$, then $\mF_{a, b}$ is increased by 1 if $p_i = 0$ and decreased by 1 otherwise.

As in RTS, we select 15\% of the input tokens for replacement. If token $t_i \in \sC_a$ is chosen, we maximize the miss-classification probability by estimating $P(p_i = 0 \ | \ t_i \in \sC_a \rightarrow t_i' \in \sC_b)$ with $\mF$. Probability of replacing $t_i$ to $t_i'$ is computed with:
\begin{equation}
P(t_i \in \sC_a \rightarrow t_i' \in \sC_b) = P(t_i) \ P(t'_i \ | \ \sC_b) \ P(\sC_b \ | \ \sC_a)
\end{equation}
\newpage
\noindent
where:
\begin{itemize}
\item $P(t_i)$ is 15\% as in ELECTRA's TD or BERT's MLM;
\item $P(t'_i \ | \ \sC_b)$ is equal to $\frac{1}{|\sC_b|}$ because we pick a target token $t_i'$ with uniform probability from $\sC_b$;
\item $P(\sC_b \ | \ \sC_a)$ is computed taking advantage of history about previous predictions, which are stored in the matrix $\mF$.
\end{itemize}
In particular, the latter is computed by selecting the $a$-th row of $\mF$ and by considering it as a multinomial distribution over target clusters. To obtain values from $\mF_a$ interpretable as probabilities, we first apply the $min-max$ normalization:
\begin{equation}
\mF_a' = \frac{\mF_a - \min(\mF_a)}{\max(\mF_a) - \min(\mF_a)}
\end{equation} 
to restrict the values into the $[0, 1]$ interval. Then, we use \emph{$\gamma$-softmax} function to smooth the distribution, which allows us to control the temperature:
\begin{equation}
\mF_a'' = \frac{e^{\mF_{a,b}' / \gamma}}{\sum_{k=1}^n e^{\mF_{a,k}' / \gamma}}
\end{equation} 
The final vector $\mF_a''$ is then used as multinomial distribution to estimate $P(\sC_b \ | \ \sC_a)$. An example of C-RTS architecture is depicted in Figure~\ref{fig:crts}.

\subsection{Swapped Language Modeling}

The SLM objective is computationally equivalent to MLM, yet it diverges in its approach by omitting the use of a special [MASK] token, resulting in superior performance. In the MLM task, 15\% of the input tokens are selected and then treated in three different ways: (i) in 80\% of the cases, they are substituted with the special [MASK] token, (ii) 10\% of the time they are replaced by other tokens, and (iii) 10\% of the time they remain unchanged. SLM can be viewed as a variation of BERT's MLM, where tokens are consistently transformed into others. The remaining aspects of the model's architecture and loss functions remain unchanged. Figure~\ref{fig:slm} provides an illustrative example of SLM.

\section{Complexity Analysis}
\label{sec:complexity_analysis}

\begin{table*}[ht]
    \centering
    \resizebox{0.75\linewidth}{!}{%
    \begin{tabular}{lccccc}
        \toprule
        \textbf{Objective} & MLM & RTS & C-RTS & SLM & MLM + TD \\
        \toprule
        \textbf{LM head complexity} & $O(|\sV| \cdot d)$ & $O(d)$ & $O(d)$ & $O(|\sV| \cdot d)$  & $O(|\sV| \cdot d)$ + $O(d)$ \\
        \bottomrule
    \end{tabular}
    }
    \caption{\small Time complexity in number of FLOPS of the classification head employed by different objectives.}
    \label{tab:complexity}
\end{table*}

\noindent
In this Section, we analyze the space and time complexity of state-of-the-art objectives for Transformers used in the literature and of our proposals. We focus especially on the size of the language model head because the remaining part of the Transformer is shared across all architectures (ours comprehended). For more details about each objective, refer to Section~\ref{sec:token_pretraining_objectives}. Table~\ref{tab:complexity} summarizes the complexity of the language modeling head for MLM, TD and our proposed objectives.

\paragraph{Masked Language Modeling \& Swapper Language Modeling}

MLM and SLM predict the original value or masked or replaced tokens. Figure~\ref{fig:bert} shows an example of input data and targets for MLM while in Figure~\ref{fig:slm} we depict an example of the SLM objective. Both objectives require a classification head that spans the whole vocabulary $\sV$. Given the final hidden state $\vh_i \in \sR^{d}$ in output from the Transformer stack for each token $t_i$, the predictions for each class are performed by multiplying $\vh_i$ with a learnable matrix $\mW_{LM} \in \sR^{|\sV| \times d}$:
\begin{equation}
\vp_i = \text{softmax} \left( \mW_{LM} \vh_i \right)
\end{equation}
Thus, the computational complexity of the language modeling head for each token is $O(|\sV| \times d)$. We ignore the complexity of the softmax because it is linear in the hidden size $d$.

\paragraph{Token Detection \& Random Token Detection}

ELECTRA's architecture is composed of two blocks: a generator trained with MLM and a discriminator trained with TD. The complexity of MLM is studied in the previous Section. Token Detection is a binary prediction task, thus the language modeling head is composed of a small matrix $\mW_{LM} \in \sR^{2 \times d}$ followed by a softmax function. We can derive that the complexity of TD is $O(d)$. RTS and C-RTS have the same complexity as ELECTRA's discriminator since they are similar but the generator is substituted with our algorithms for selecting the token replacements.

\section{Experiments}

In this Section, we describe the testing environment we use to measure the accuracy of the reference and proposed objectives. First, we provide details about the pre-training setting, the datasets and the best hyper-parameters we found for each objective. Then, fine-tuning details such as datasets, training parameters and the hardware used to run the experiments are given. Finally, we present results as tables and plots, including comparisons against models released by the authors of BERT when available.

\subsection{Architectures}

Every objective has been tested by applying it to the same Transformer architecture, equivalent to that of BERT~\citep{devlin-etal-2019-bert} or RoBERTa~\citep{liu2019roberta}. For every model and architecture, we use an uncased vocabulary, which means WordPiece converts text to lowercase before tokenizing. We experiment with models of two different sizes: Base and Small.

\paragraph{Base} The Base architecture is equivalent to that of BERT\textsubscript{Base}, which has a vocabulary $\sV$ containing 30522 tokens based on WordPiece (see Paragraph~\ref{par:wordpiece}), a hidden size $d$ of 768, an intermediate size equal to 3072, 512 positional embeddings, 12 attention heads and 12 layers.

\paragraph{Small} The Small architecture we employ is similar to that of ELECTRA\textsubscript{Small}~\citep{clark2020electra}, which in prior experiments resulted to be better than BERT\textsubscript{Small}~\citep{turc2019wellread}. ELECTRA demonstrated that it is more effective to reduce the \emph{width} (the hidden size) of the Transformer instead of the \emph{height} (the number of layers). We use a vocabulary $\sV$ containing 30522 tokens based again on WordPiece, a hidden size $d$ of 256, an intermediate size of 1024, 512 positional embeddings 12 attention layers and 4 attention heads.

\begin{table*}[b]
    \centering
    \resizebox{0.85\linewidth}{!}{%
    \begin{tabular}{lcccccl}
        \toprule
        
        \multirow{2}{*}{\textbf{Models}} 	& \multirow{2}{*}{\textbf{FLOPS}} & \multirow{2}{*}{\textbf{Time}} & \multirow{2}{*}{\textbf{Memory}} & \multicolumn{2}{c}{\textbf{\# Parameters}} & \multirow{2}{*}{\textbf{Objective}} \\
 	\cmidrule{5-6} & & & & \textbf{Total} & \textbf{LM Head} & \\

        \toprule
 
        BERT\textsubscript{Base} + MLM			& $1.61\cdot10^{19}$		& 3d 14h					& 13.2GiB						& 109M		& 23M 	& MLM (1.0) \\
        BERT\textsubscript{Base} + RTS			& $1.54\cdot10^{19}$		& 2d 22h					& 9.9GiB						& 109M		& 1536 	& RTS (50.0) \\
        BERT\textsubscript{Base} + C-RTS			& $1.54\cdot10^{19}$		& 2d 22h					& 9.9GiB						& 109M		& 1536	& C-RTS (50.0) \\
        BERT\textsubscript{Base} + SLM			& $1.61\cdot10^{19}$		& 3d 14h					& 13.2GiB						& 109M		& 23M 	& SLM (1.0) \\
        ELECTRA\textsubscript{Base} + MLM/TD		& $1.98\cdot10^{19}$		& 3d 18h					& 18.0GiB						& 143M		& 23M + 1536 	& MLM (1.0) + TD (50.0) \\

        \bottomrule
    \end{tabular}
    }
    \caption{\small FLOPS, time, memory, number of parameters and objectives to pre-train Base models on two A100 GPUs of our machine. Notice that the number of parameters of MLM/SLM and RTS/C-RTS is the same because the weights of the embedding layer and language modeling head are shared. However, the memory usage of MLM/SLM is higher because gradients over the language modeling head are much larger in memory than gradients over the embedding layer. Notice also that FLOPS and training time of C-RTS differ slightly from those reported in \citet{di-liello-etal-2022-effective} because we offloaded computations over clusters to the CPU. Finally, we report the objectives along with their weight when computing the total loss.}
    \label{tab:flops_base}
\end{table*}

\subsection{Pre-Training}

In order to provide a meaningful and fair comparison between the objectives, we use the same pre-training dataset for all experiments. In more detail, we train all models on the same dump of the English Wikipedia and on the BookCorpus, which are cleaned before training from HTML tags and tables, from lists, indexes and everything not interpretable as raw text. Notice that the original BookCorpus used by \citet{devlin-etal-2019-bert} is not available anymore, and we do not know whether the version we use is of the same quality. For more details about the pre-training corpora, see Section~\ref{sec:pretraining_datasets}.

We copied the training set from \citet{devlin-etal-2019-bert}, in which the authors train the models for 900K steps with a maximum sequence length of 128 tokens and 100K additional steps with a sequence length of 512. This allows us to save a lot of pre-training time without a significant loss in performance. The increased efficiency derives from the fact that the attention module complexity is quadratic in the sequence length. Thus, for the first 900K steps, the model is theoretically 4 times faster\footnote{Actual speed-up is generally slightly lower because of the embedding layer and the language modeling head whose complexities are linear in the sequence length.} than when training with the full sequence length. However, larger and heavy-trained models such as RoBERTa~\citep{liu2019roberta} showed that attention weights and positional embeddings quality may suffer from pre-trainings with shorter sequence lengths due to the lack of computation of long-range dependencies between tokens. Nevertheless, we pre-train with the setting described by \citet{devlin-etal-2019-bert} because the comparison is still fair and because of our limited computational budget.

Regarding the training hyper-parameters, we apply a weight decay of 0.01 to every parameter apart from those in biases and LayerNorms modules and we set $\epsilon = 10^{-8}$, $\beta_1 = 0.9$ and $\beta_2 = 0.999$ in the optimizer. We use a triangular learning rate scheduler, with a peak of $1 \cdot 10^{-4}$ (Base) or $2 \cdot 10^{-4}$ (Small) and 10K warmup steps. The probability in drop-out layers is set to 0.1. Finally, we use a batch size of 256 examples for Base models and 1024 for the Small versions, for optimal utilization of the GPU memory.

We train also a Base and a Small ELECTRA architecture for comparison. Notice that the ELECTRA requires more FLOPS because of the presence of the additional generator network. FLOPS are a general and architecture-independent indicator of model requirements that measure the total number of floating point operations required for training. To address this issue, and train all models with the same compute budget, we reduce the total number of steps of ELECTRA from 900K@128~+~100K@512 to 689K@128~+~77K@512 for both Small and Base models. The generator size is set to $\frac{1}{3}$ and $\frac{1}{4}$ for Base and Small models as in the original work, respectively\footnote{Applies to the hidden size, the intermediate size and the number of attention heads.}. Inspired by \citet{clark2020electra}, we use FLOPS as a measure of compute requirements. However, actual training time is sometimes misaligned with FLOPS because of hardware acceleration such as NVIDIA Tensor Cores, mixed-precision training, low precision \emph{float32} matrix multiplication and other optimization techniques. Tables~\ref{tab:flops_base} and~\ref{tab:flops_small} summarize the training time and FLOPS required by the Base and Small models.

Each experiment is executed on two GPUs of our machine (which is described in Section~\ref{sec:machines}), using mixed-precision training.

\begin{table*}[t]
    \centering
    \resizebox{0.85\linewidth}{!}{%
    \begin{tabular}{lcccccl}
        \toprule
        
        \multirow{2}{*}{\textbf{Models}} 	& \multirow{2}{*}{\textbf{FLOPS}} & \multirow{2}{*}{\textbf{Time}} & \multirow{2}{*}{\textbf{Memory}} & \multicolumn{2}{c}{\textbf{\# Parameters}} & \multirow{2}{*}{\textbf{Objective}} \\
 	\cmidrule{5-6} & & & & \textbf{Total} & \textbf{LM Head} \\

        \toprule
 
        BERT\textsubscript{Small} + MLM			& $3.04\cdot10^{18}$		& 1d 7h				& 12.5GiB						& 13.5M		& 3.9M 		& MLM (1.0) \\
        BERT\textsubscript{Small} + RTS			& $2.73\cdot10^{18}$		& 17h				& 6.3GiB						& 13.5M		& 512		& RTS (20.0) \\
        BERT\textsubscript{Small} + C-RTS			& $2.73\cdot10^{18}$		& 17h				& 6.3GiB						& 13.5M		& 512		& C-RTS (20.0) \\
        BERT\textsubscript{Small} + SLM			& $3.04\cdot10^{18}$		& 1d 7h				& 12.5GiB						& 13.5M		& 3.9M		& SLM (1.0) \\
        ELECTRA\textsubscript{Small} + MLM/TD		& $4.87\cdot10^{18}$		& 1d 16h				& 17.4GiB						& 18.2M		& 3.9M + 512	& MLM (1.0) + TD (20.0) \\

        \bottomrule
    \end{tabular}
    }
    \caption{\small FLOPS, time, memory and number of parameters required to pre-train Small models on the same hardware. For more details, see Table~\ref{tab:flops_base}.}
    \label{tab:flops_small}
\end{table*}

\subsection{Fine-Tuning}

After pre-training every model on the same amount of data and with the same hyper-parameters, we evaluate them on different benchmarks.

\paragraph{GLUE} First, for a general performance evaluation, we use the GLUE benchmark suite, which contains datasets for many different tasks such as Questions Similarity and Natural Language Inference. For more details about the GLUE dataset, refer to Section~\ref{sec:benchmarks}. As in \citet{devlin-etal-2019-bert} and \citet{clark2020electra}, we omit WNLI from the GLUE datasets because even a trivial majority classifier is better than most Base or Small pre-trained models.

As a starting point, we copy the fine-tuning hyper-parameters of \citet{liu2019roberta} for every GLUE task. After that, we evaluate affine configurations to ensure good final accuracy of the models. The final hyper-parameters are described in Table~\ref{tab:glue_hparams}. Other parameters have been kept constant for all experiments: no weight decay, triangular learning rate with a warmup phase of 10\% of the total fine-tuning steps and a maximum sequence length of 128. Regarding the testing, we take the best model based on the evaluation metric of each GLUE dataset. Moreover, we apply early stopping with a patience of 5 and we validate at the end of each training epoch.

\begin{table*}[b]
    \centering
    \resizebox{0.9\linewidth}{!}{
        \begin{tabular}{lcccccccc}
            \toprule
            \textbf{Hyper-Parameter}	& \textbf{CoLA}	& \textbf{MNLI-m}	& \textbf{MRPC} 	& \textbf{QNLI}	& \textbf{QQP}	& \textbf{RTE}	& \textbf{SST-2}	& \textbf{STS-B} \\
            \midrule
            Batch Size			& 16		& 32		& 16		& 32		& 32		& 16		& 32		& 16 \\
            Learning Rate 		& $1\cdot 10^{-5}$ & $1\cdot 10^{-5}$ & $1\cdot 10^{-5}$ & $1\cdot 10^{-5}$ & $1\cdot 10^{-5}$ & $2\cdot 10^{-5}$ & $1\cdot 10^{-5}$ & $2\cdot 10^{-5}$  \\
            Max epochs			& 20		& 8		& 10		& 10		& 8		& 40		& 10		& 40 \\
            \bottomrule
        \end{tabular}
    }
    \caption{\small GLUE fine-tuning hyper-parameters.}
    \label{tab:glue_hparams}
\end{table*}

Regarding the evaluation metrics, we use those proposed by the original authors of GLUE~\citep{wang-etal-2018-glue} and similar works~\citep{devlin-etal-2019-bert, liu2019roberta, clark2020electra}. In particular, we use accuracy for all tasks apart from CoLA and STS-B, which are evaluated with Matthew and Spearman correlation coefficients, respectively.

Finally, we repeat every experiment with 5 random seeds (which influence the classification head initialization and the dataset shuffling), and we send the results of the best-performing checkpoints on the dev set to the GLUE Leaderboard for evaluation on the test set\footnote{https://gluebenchmark.com/leaderboard/}.

\paragraph{ASNQ}

We use ASNQ to evaluate models performance on large Answer Selection datasets. ASNQ contains more than 20M question-answer pairs in the training set, making it the largest dataset for AS2 to date. More details about ASNQ are given in Section~\ref{sec:as2_datasets}. We use a batch size of 2048 examples for each model and a learning rate equal to $1\cdot 10^{-5}$. The values are obtained with a greedy search: we try all the combinations of batch sizes in $\{ 128, 256, 512, 1024, 2048, 4096\}$ and learning rates in $\{ 2 \cdot 10^{-6}, 5 \cdot 10^{-6}, 1 \cdot 10^{-5}, 2 \cdot 10^{-5} \}$. Since most of the question-answer pairs in the dataset are shorter than 128 tokens, we set the maximum sequence length to this value. We allow models to train up to 10 epochs, and we do early stopping on the development set if MAP does not improve for 4 validations in a row. Other training parameters are the same of the fine-tuning on GLUE. Experiments are repeated 3 times with different initialization seeds to show average and standard deviation in the results.

We measure performance using Mean Average Precision (MAP) and Mean Reciprocal Rank (MRR). The first is a soft metric that measures the Average Precision at different levels of recall, while the second is a hard metric that measures only the reciprocal rank of the positive answer with the highest score. For more details about Answer Selection ranking, see Section~\ref{sec:answer_sentence_selection}. 

\paragraph{WikiQA, TREC-QA and Transfer-Learning}

As a last step, we evaluate the pre-trained models on two small datasets for Answer Sentence Selection: WikiQA and TREC-QA. We first fine-tune the pre-trained models directly on those two datasets. Then, as in T\textsc{and}A~\citep{garg2019tanda}, we measure performance in Transfer-Learning by taking the checkpoints of the models after a transfer step on ASNQ and fine-tuning them on WikiQA and TREC-QA. We indicate the two transfer steps and subsequent adaptations to the final task with ASNQ $\rightarrow$ WikiQA and ASNQ $\rightarrow$ TREC-QA.

Fine-tuning hyper-parameters are selected again with greedy search. For the batch size, we try values in $\{ 32, 64, 128\}$, while for the learning rate, we search in $\{ 2 \cdot 10^{-6}, 5 \cdot 10^{-6}, 1 \cdot 10^{-5}, 2 \cdot 10^{-5} \}$. The maximum sequence length is fixed to 128 tokens while the number of epochs is limited to 40. Again, we use early stopping to spare resources, with a patience of 5.

\section{Results}

\subsection{Pre-Training}

We start the comparison of the results with a discussion about pre-training times. For each Base model, Table~\ref{tab:flops_base} shows training time and the total number of FLOPS. RTS and C-RTS use 20\% less training time to complete the same number of pre-training steps. The difference in terms of FLOPS is instead smaller, around 5\%. Notice that RTS and C-RTS use 45\% GPU memory, thus training times could be reduced even further by doubling the batch size. However, we keep the same training parameters for all models to provide a fair comparison.

With Small models, the gains in training time are even larger. RTS and C-RTS allow to reduce pre-training time by 45\% and FLOPS by 10\%. Gaps are larger than Base models because the embedding layer and the language modeling head are bound to the vocabulary size $|\sV|$. We do not shrink the vocabulary size because recent works show significant drops in performance when reducing it~\citep{turc2019wellread} and because BERT's vocabulary size is already one of the smallest among recently released pre-trained models.
RTS and C-RTS have about $\frac{1}{3}$ of the memory footprint of MLM or SLM. As with Base models, we could have increased the batch size to speed up the training of RTS and C-RTS even further.

\subsection{Fine-Tuning}

In this Section, we present the comparison on different tasks of the models we pre-trained. The statistical significance comparison in the tables if performed compared to BERT\textsubscript{Base}~+~MLM and BERT\textsubscript{Small}~+~MLM. Results of the original BERT\textsubscript{Base}~+~MLM/NSP~\citep{devlin-etal-2019-bert} derive from our fine-tuning on the various tasks in the same setting as the other models, because the original authors do best model selection on the development set.

\subsubsection{Base Models}

\begin{table*}[t]
    \resizebox{1.0\linewidth}{!}{%
    \centering
    \begin{tabular}{l@{\hskip6pt}c@{\hskip7pt}c@{\hskip7pt}c@{\hskip7pt}c@{\hskip7pt}c@{\hskip7pt}c@{\hskip7pt}c@{\hskip7pt}c@{\hskip6pt}c}
        \toprule
        \textbf{Model} & \textbf{CoLA} & \textbf{MNLI} & \textbf{MRPC} & \textbf{QNLI} & \textbf{QQP} & \textbf{RTE} & \textbf{SST-2} & \textbf{STS-B} & \textbf{AVG} \\

        \toprule
        BERT\textsubscript{Base} + MLM/NSP~\citep{devlin-etal-2019-bert}	& 57.6 (1.8)              	& 84.3 (0.4) 		& 82.3 (1.3) 		& 91.0 (0.7) 		& 91.0 (0.2) 		& 68.9 (1.4) 		& 92.6 (0.1) 		& 89.1 (0.3) 		& 82.1 \\
        ELECTRA\textsubscript{Base} + MLM/TD               		& 63.4 (1.3) 		& 83.7 (0.2)		& 87.2 (0.8) 		& 90.4 (0.1) 		& 91.2 (0.1)  		& 74.6 (1.4) 		& 91.4 (0.4) 		& 88.5 (0.2) 		& 83.8 \\

        \midrule
        BERT\textsubscript{Base} + MLM              						& 58.1 (1.0)              	& \bld{83.4} (0.2)      	& 87.5 (0.5)            	& \bld{90.2} (0.3)  	& 90.9 (0.1)          	& 67.4 (1.2)            	& \bld{92.2} (0.3) 	& \bld{87.8} (0.3)     	& 82.2 \\
        \textbf{BERT\textsubscript{Base} + RTS}              						& 58.1 (1.1)              	& 82.7 (0.2) 		& \bld{87.6} (1.0)    	& 89.4 (0.3) 		& 90.9 (0.1)          	& 68.5 (1.4)            	& 91.5 (0.3) 		& 86.6 (0.4) 		& 81.9 \\
        \textbf{BERT\textsubscript{Base} + C-RTS}          						& 57.4 (0.7)              	& 82.0 (0.3) 		& 84.2 (0.4) 		& 89.6 (0.2)             	& 90.6 (0.1) 		& 66.6 (2.4)            	& 91.5 (0.2) 		& 87.0 (0.2)             	& 81.1 \\
        \textbf{BERT\textsubscript{Base} + SLM}               					& \bund{59.6} (1.0)	& \bld{83.4} (0.2)     	& 87.5 (0.4)            	& 89.9 (0.2)             	& \bld{91.0} (0.1) 	& \bund{69.2} (1.2) 	& 92.1 (0.1) 		& 87.6 (0.3)             	& \bld{82.5} \\
        \bottomrule
    \end{tabular}
    }
    \caption{\small Results on GLUE development set for Base models. We do not employ ensemble models. We measure performance with Accuracy on all tasks apart from CoLA and STS-B, in which we use Matthews correlation coefficient and Spearman's rank correlation coefficient, respectively. We use bold formatting for our proposals and the best results. We underline statistically significant improvements over BERT\textsubscript{Base} + MLM.}
    \label{tab:glue_base_dev}
\end{table*}
\begin{table*}[t]
    \small
    \centering
    \resizebox{\linewidth}{!}{%
    \begin{tabular}{lccccccccc}
        \toprule
        \textbf{Model} & \textbf{CoLA} & \textbf{MNLI} & \textbf{MRPC} & \textbf{QNLI} & \textbf{QQP} & \textbf{RTE} & \textbf{SST-2} & \textbf{STS-B} & \textbf{AVG} \\
        \toprule
        BERT\textsubscript{Base} + MLM/NSP~\citep{devlin-etal-2019-bert}	& 53.0 		& 82.7 		& 82.8 		& 89.6 		& 88.2 		& 62.8 		& 91.2 		& 80.6 		& 78.9 \\
        ELECTRA\textsubscript{Base} + MLM/TD    			& 60.6 		& 83.6 		& 84.1 		& 90.2 		& 89.0 		& 69.1 		& 92.4 		& 85.5 		& 81.8 \\
        \midrule
        BERT\textsubscript{Base} + MLM							& 53.7 		& \bld{83.6}	& 82.6 		& \bld{89.9}	& \bld{89.1}	& 63.1 		& \bld{92.3} 	& 83.6 		& 79.7 \\
        \textbf{BERT\textsubscript{Base} + RTS}     							& \bld{57.3} 	& 82.6 		& 81.3 		& 89.3 		& 88.9 		& \bld{66.2}	& 91.7 		& 82.2 		& 79.9 \\
        \textbf{BERT\textsubscript{Base} + C-RTS}       						& \bld{57.3}	& 82.8 		& 81.9 		& 89.4 		& 88.6 		& 60.9 		& 91.0 		& 82.2 		& 79.3 \\
        \textbf{BERT\textsubscript{Base} + SLM}          						& 57.0 		& 83.3 		& \bld{83.1}	& 89.7 		& 88.9 		& 65.0 		& \bld{92.3} 	& \bld{83.8}	& \bld{80.4} \\
        \bottomrule
    \end{tabular}
    }
    \caption{\small Results on the GLUE test set for Base models. For more details about the metrics, see Table~\ref{tab:glue_base_dev}. For the evaluation on the GLUE leaderboard, we select the best models on the development set.}
    \label{tab:glue_base_test}
\end{table*}

Results of Base models over the GLUE benchmarks are presented in Table~\ref{tab:glue_base_dev} for the development set (as in \citet{clark2020electra}) and in Table~\ref{tab:glue_base_test} for the test set. Results on the development set allow the comparison of the standard deviation in results. This is not possible on the test set because of the limited number of submissions allowed by the GLUE Leaderboard.

On GLUE, the models pre-trained with RTS and C-RTS objectives obtain comparable results with all the other techniques, except for ELECTRA-TD. However, while RTS and C-RTS require about 20\% less training time than a similar MLM-based model, ELECTRA-TD requires more time to complete the pre-training and consumes much more memory due to the presence of the additional generator network, thus reducing scalability. In particular, ELECTRA uses 36\% more memory than a similar MLM- or SLM-based model and almost doubles the memory required by RTS or C-RTS.
The SLM objective we designed outperforms MLM in most tasks while being trained with the same computational budget. In particular, SLM achieves an average GLUE score higher than our MLM-based BERT by 0.7 points and also higher than the original BERT, which was trained with MLM and NSP (+1.5). SLM shows also that the pre-training/fine-tuning discrepancy is a major issue of MLM-based models such as BERT or RoBERTa.

We compare also with the original BERT model released by \citet{devlin-etal-2019-bert}, which uses an additional sentence-level objective to improve performance. Notice that even if \citet{liu2019roberta} demonstrated that dropping the NSP objective does not affect final accuracy, we believe this claim is valid only with very long pre-trainings. RoBERTa~\citep{liu2019roberta} is trained for a full day on 1024 GPUs, which is about 53 more times the computational requirements of BERT~\citep{devlin-etal-2019-bert}. Thus, with models pre-trained with an academic budget and over tasks where the model has to work on two input sequences, such as AS2, our results show that NSP is still beneficial because it improves the representation of the CLS token, which is used for classification.

We do not compare against the ELECTRA models released by \citet{clark2020electra} because they were pre-trained on 20 times the total tokens seen by our models.

\begin{table*}[t]
    \centering
    \resizebox{0.9\linewidth}{!}{%
    \begin{tabular}{lcccccccc}
        \toprule
        \multirow{2}{*}{\textbf{Model}} 									& \multicolumn{2}{c}{\textbf{ASNQ}} & & \multicolumn{2}{c}{\textbf{WikiQA}} 	& & \multicolumn{2}{c}{\textbf{TREC-QA}} \\
	\cmidrule{2-3} \cmidrule{5-6} \cmidrule{8-9}

        															& \textbf{MAP} & \textbf{MRR}		& & \textbf{MAP} & \textbf{MRR}		& & \textbf{MAP} & \textbf{MRR} \\
        \toprule
        BERT\textsubscript{Base} + MLM/NSP~\citep{devlin-etal-2019-bert}  	& 66.8 (0.1) & 73.2 (0.2)			& & 82.8 (0.9) & 84.2 (1.0)				& & 87.2 (0.9) & 92.9 (1.0) \\
        ELECTRA\textsubscript{Base} + MLM/TD                  		& 64.9 (0.3) & 71.7 (0.4)			& & 81.8 (1.6) & 83.2 (1.6) 			& & 86.8 (1.4) & 92.2 (1.5)	 \\
        \midrule
        BERT\textsubscript{Base} + MLM                    						& 65.5 (0.2) & 72.2 (0.3)			& & 79.9 (1.1) & 81.2 (1.2)				& & 86.4 (0.7) & 91.5 (0.7)	 \\
        \textbf{BERT\textsubscript{Base} + RTS}                    						& 64.6 (0.1) & 71.1 (0.2)			& & 78.5 (2.5) & 80.1 (2.5)				& & 86.6 (1.5) & 91.8 (1.5)	 \\
        \textbf{BERT\textsubscript{Base} + C-RTS}                  						& 64.7 (0.1) & 71.5 (0.1) 			& & 79.0 (1.9) & 80.6 (1.6)				& & 87.0 (0.8) & 91.8 (1.1) \\
        \textbf{BERT\textsubscript{Base} + SLM}                    						& \bld{65.8} (0.3) & \bund{72.7} (0.3)	& & \bld{80.2} (1.5) & \bld{81.7} (1.6) 	& & \bld{87.3} (1.3) & \bld{92.1} (1.5) \\
        \bottomrule
    \end{tabular}
    }
    \caption{\small Results on ASNQ, WikiQA and TREC-QA for Base models. The results on ASNQ are on the dev.~set. We use bold formatting for our proposals and the best results. We underline statistically significant improvements over BERT\textsubscript{Base}~+~MLM.}
    \label{tab:efficient_as2_base_test_1}
\end{table*}

\begin{table*}[t]
    \centering
    \resizebox{0.75\linewidth}{!}{%
    \begin{tabular}{lccccc}
        \toprule
        \multirow{2}{*}{\textbf{Model}}			& \multicolumn{2}{c}{\textbf{ASNQ} $\xrightarrow{}$ \textbf{WikiQA}} & & \multicolumn{2}{c}{\textbf{ASNQ} $\xrightarrow{}$ \textbf{TREC-QA}} \\
	\cmidrule{2-3} \cmidrule{5-6}

        															& \textbf{MAP} & \textbf{MRR}		& & \textbf{MAP} & \textbf{MRR} \\
        \toprule
        BERT\textsubscript{Base} + MLM/NSP~\citep{devlin-etal-2019-bert}    	& 87.9 (0.4) & 89.3 (0.4)			& & 89.1 (0.4) & 93.4 (0.5) \\
        ELECTRA\textsubscript{Base} + MLM/TD                  		& 88.4 (0.4) & 89.8 (0.4)			& & 88.9 (0.3) & 92.0 (0.5)	 \\
        \midrule
        BERT\textsubscript{Base} + MLM                    						& \bld{88.9} (1.0) & \bld{90.2} (1.0)  	& & 87.6 (0.9) & 90.6 (1.1) \\
        \textbf{BERT\textsubscript{Base} + RTS}                    						& 87.9 (0.5) & 89.3 (0.5)   			& & 88.5 (0.5) & \bund{93.4} (0.4)	 \\
        \textbf{BERT\textsubscript{Base} + C-RTS}                  						& 87.1 (0.6) & 88.3 (0.6) 			& & \bund{88.7} (0.4) & \und{92.9} (0.7)	 \\
        \textbf{BERT\textsubscript{Base} + SLM}                    						& 87.7 (0.8) & 89.3 (0.7)          		& & 87.9 (0.6) & 91.0 (0.6)	 \\
        \bottomrule
    \end{tabular}
    }
    \caption{\small Results on WikiQA and TREC-QA for Base models after the transfer step on ASNQ.}
    \label{tab:efficient_as2_base_test_2}
\end{table*}

Regarding the results in Answer Sentence Selection, which are provided in Table~\ref{tab:efficient_as2_base_test_1} and~\ref{tab:efficient_as2_base_test_2}, RTS and C-RTS show comparable performance with MLM and SLM in most tasks. On WikiQA, results are slightly lower on average, while they are superior on TREC-QA. On ASNQ, RTS and C-RTS are below MLM and SLM by less than 1 point in MAP, while being pre-trained for 20\% less time. With the transfer step over ASNQ before the final fine-tuning on WikiQA and TREC-QA, relative differences between objectives do not change. Results are higher on average because the models are accustomed to performing Answer Selection on a large dataset before being fine-tuned. Notice also that the transfer step on ASNQ benefits more WikiQA than TREC-QA. We argue that the reason is that both ASNQ and WikiQA are created from documents sourced from Wikipedia, while TREC-QA is based on human-generated answers. 
On the other hand, SLM shows his superiority over MLM in 4 tasks out of 5. SLM outperforms a similar MLM-based model by 1.9 points in MAP on WikiQA, by 0.9 on TREC-QA, by 0.3 on ASNQ $\rightarrow$ WikiQA and by 0.3 on ASNQ, while requiring the same computational budget to pre-train. These results highlight again the advantage of removing the special [MASK] token from the pre-training.

\subsubsection{Small Models}

Here we present the results of applying our effective objectives to Small Transformer models. Table~\ref{tab:efficient_glue_small_dev} shows the results over the GLUE development set, while in Table~\ref{tab:efficient_glue_as2_small_test} we provide the results over 3 AS2 datasets and the average GLUE score on the test set. We do not show performance on Transfer-Learning from ASNQ to either WikiQA or TREC-QA because ASNQ is so large that after the transfer step, the differences in accuracy between the objectives are minimal and not statistically significant.

On the GLUE dev set, Small models trained with RTS and C-RTS outperform a similar MLM-based model on every task. The average GLUE score of RTS and C-RTS is respectively 2.0 and 2.1 points higher than MLM and even higher than SLM on some individual datasets. The only drawback of RTS and C-RTS is that the standard deviation is higher across most tasks, indicating a higher sensitivity to parameters' initialization. On the GLUE test set, RTS and C-RTS perform accordingly to the dev set, outperforming MLM by 1.3 and 1.6 points. We can conclude that RTS and C-RTS are well suited for Small models, since they require almost half of the pre-training time and deliver superior performance compared to MLM, SLM and ELECTRA's TD.

The AS2 dataset results exhibit a trend resembling those of GLUE: RTS and C-RTS consistently outperform MLM in all tasks. For instance, WikiQA showcases RTS's superiority over MLM by 2.1 points in Mean Average Precision, while ASNQ shows a 1.2-point difference in favor of RTS. While results between MLM and RTS on TREC-QA are quite similar, it is noteworthy that BERT with RTS surpasses ELECTRA\textsubscript{Small} trained with TD in all tasks, despite the latter requiring twice the pre-training time and three times the GPU memory.

\begin{table*}[t]
    \resizebox{\linewidth}{!}{%
    \centering
    \begin{tabular}{lc@{\hskip7pt}c@{\hskip7pt}c@{\hskip7pt}c@{\hskip7pt}c@{\hskip7pt}c@{\hskip7pt}c@{\hskip7pt}cc}
        \toprule
        \textbf{Model} 							& \textbf{CoLA} 		& \textbf{MNLI} 		& \textbf{MRPC} 	& \textbf{QNLI} 		& \textbf{QQP} 		& \textbf{RTE} 		& \textbf{SST-2} 	& \textbf{STS-B} 	& \textbf{AVG} \\
        \toprule
        ELECTRA\textsubscript{Small} + MLM/TD	& 51.2 (0.1) 	& 76.8 (0.2)          	& 85.5 (0.3) 		& 86.9 (0.0) 		& 88.4 (0.0) 		& 63.9 (0.3) 		& 87.0 (0.3) 		& 85.5 (0.1)   		& 78.1 \\
	\midrule
        BERT\textsubscript{Small} + MLM			& 42.2 (0.9)         		& 78.8 (0.2)              	& 80.8 (0.9)             	& 85.7 (0.6)            	& 88.6  (0.1)       	& 58.6 (1.3)         	& 89.4 (0.2)        	& 84.4 (0.2) 		& 76.1 \\
        \textbf{BERT\textsubscript{Small} + RTS}			& \und{51.2} (1.8) 		& \bund{79.9} (0.2) 	& 81.5 (0.7)            	& \bund{87.9} (0.1)	& \bund{89.4} (0.1) 	& \und{61.2} (0.6) 	& 88.4 (0.4) 		& \und{85.2} (0.2) 	& 78.1 \\
        \textbf{BERT\textsubscript{Small} + C-RTS}			& \bund{51.6} (0.9) 		& \und{79.8} (0.1)	& 81.3 (1.0)            	& \und{87.0} (0.4)	& \bund{89.4} (0.1)	& \bund{61.6} (1.4) 	& \bld{89.5} (0.2)	& \und{85.3} (0.3) 	& \bld{78.2} \\
        \textbf{BERT\textsubscript{Small} + SLM}			& \und{45.8} (0.5) 		& \und{79.1} (0.2)    	& \bund{83.3} (0.4)	& 86.3 (0.3)            	& \und{89.0} (0.1) 	& \und{60.8} (1.4) 	& 88.6 (0.4) 		& \bund{86.0} (0.2) 	& 77.4 \\
        \bottomrule
    \end{tabular}
    }
    \caption{\small Results on GLUE development set for Small models. For more details about the metrics, see Table~\ref{tab:glue_base_test}. We use bold formatting for our proposals and the best results. We underline statistically significant improvements over BERT\textsubscript{Small}~+~MLM.}
    \label{tab:efficient_glue_small_dev}
\end{table*}
\begin{table*}[t]
    \centering
    \small
    \resizebox{0.9\linewidth}{!}{%
    \begin{tabular}{lc@{\hskip7pt}c@{\hskip7pt}c c@{\hskip7pt}c@{\hskip7pt}c c@{\hskip7pt}c@{\hskip7pt}c c}
        \toprule
        \multirow{2}{*}{\textbf{Model}} & \multicolumn{2}{c}{\textbf{WikiQA}} & & \multicolumn{2}{c}{\textbf{TREC-QA}} & & \multicolumn{2}{c}{\textbf{ASNQ}} & & \textbf{GLUE} \\
        	\cmidrule{2-3} \cmidrule{5-6} \cmidrule{8-9} \cmidrule{11-11}

												& \textbf{MAP} & \textbf{MRR}			& & \textbf{MAP} & \textbf{MRR}		& & \textbf{MAP} & \textbf{MRR}			& & \textbf{AVG} \\
        \toprule
        ELECTRA\textsubscript{Small} + MLM/TD		& 75.8 (1.6) & 77.5 (1.5)          			& & 83.3 (1.0) & 88.2 (1.6)				& & 59.2 (0.0) & 66.3 (0.0)					& & 75.5 \\
	\midrule
        BERT\textsubscript{Small} + MLM					& 74.5 (1.9) & 75.8 (1.9)          			& & 85.3 (1.3) & 89.9 (1.2) 			& & 58.6 (0.3) & 66.0 (0.4)    				& & 74.1 \\
        \textbf{BERT\textsubscript{Small} + RTS}           				& \bund{76.6} (0.7) & \bund{78.4} (0.9)   	& & 85.2 (2.4) & 90.5 (2.8)				& & \bund{59.8} (0.1) & \und{66.9} (0.2)   		& & 75.4 \\
        \textbf{BERT\textsubscript{Small} + C-RTS}       				& 74.0 (1.1) & 75.8 (1.3)  				& & 86.1 (0.6) & \bund{92.5} (2.0) 		& & \und{59.6} (0.2) & \bund{67.1} (0.2)    		& & \bld{75.7} \\
        \textbf{BERT\textsubscript{Small} + SLM}           				& 73.6 (0.8) & 75.2 (0.9)   				& & \bld{86.3} (1.4) & 91.7 (1.8) 		& & \und{59.2} (0.2) & 66.5 (0.3)     			& & \bld{75.7} \\
        \bottomrule
    \end{tabular}
    }
    \caption{\small Results of our Small models pre-trained with RTS, C-RTS and SLM compared with MLM on AS2 datasets and the GLUE test set.}
    \label{tab:efficient_glue_as2_small_test}
\end{table*}

Regarding C-RTS, its performance, on average, is slightly lower than RTS in AS2 but slightly better on the GLUE test set, with differences in only a few decimal points. However, in the next Section, we will demonstrate that C-RTS exhibits clear advantages over RTS when trained for a longer duration, likely due to the slower convergence of the more challenging pre-training task.

Finally, SLM provides good performance on the GLUE test set, matching the average score of C-RTS and surpassing the expensive ELECTRA\textsubscript{Small} architecture and the MLM-based BERT. On the other Answer Sentence Selection tasks, SLM performance is better than those of BERT with MLM on 2 datasets out of three. Notice that BERT with SLM and MLM share the exactly same architecture, pre-training time and memory footprint.

\begin{table*}[t]
    \centering
    \resizebox{0.7\linewidth}{!}{%
    \begin{tabular}{lccccc}
        \toprule
        \textbf{Model} 						& \textbf{WikiQA}		& \textbf{TREC-QA}		& \textbf{ASNQ}		& \textbf{MRPC}		& \textbf{QNLI} \\
        \toprule
  
        \textbf{BERT\textsubscript{Small} + RTS}		& 76.6 (0.9)			& 85.5 (1.5)			& 59.9 (0.2)			& 81.5 (1.5)			& \bld{86.9} (0.3) \\
        \textbf{BERT\textsubscript{Small} + C-RTS}		& \bld{77.4} (1.0)		& \bld{86.4} (0.9)		& \bund{60.7} (0.1)		& \bund{84.2} (0.1)		& \bld{86.9} (0.1) \\

        \bottomrule
    \end{tabular}}
    \caption{\small Comparison between RTS and C-RTS. Results for WikiQA and TREC-QA are computed over the test set, while we report the best results over the dev.~set for the others. We compare results with MAP for WikiQA, TREC-QA and ASNQ, while we use Accuracy for the others. We underline statistically significant improvements over BERT\textsubscript{Small}~+~RTS.}
    \label{tab:rts_vs_crts}
\end{table*}

\subsection{RTS vs C-RTS}

The results of the previous Section show that C-RTS achieves accuracies similar to RTS in most benchmarks. To fully demonstrate the potential of a harder training task, we compared RTS and C-RTS when pre-trained for longer on the same data as before. In order to accomplish that, we use a new pre-training configuration in which the maximum sequence length is not reduced for the first steps (e.g.we use always 512 tokens). Then, we set the batch size to 1024 examples and train both models for 200K steps. After every 10K pre-training steps, we evaluate over 5 different benchmarks: ASNQ, WikiQA, TREC-QA, QNLI and MRPC. We selected those datasets because they cover 3 different domains and the sizes range from very few examples (WikiQA) to more than 20M (ASNQ). We exploit Small models for this experiment due to the cost of pre-training Base Transformers.

In Table~\ref{tab:rts_vs_crts} we report the results in the last pre-training step of the comparison described above. C-RTS obtains better results than RTS on 4 tasks out of 5 thanks to the harder training task. On the other hand, both objectives perform comparably from a statistical viewpoint on QNLI. This demonstrates empirically that selecting challenging replacements forces the model to find deeper relations between tokens. We provide the full plots of the comparison in Appendix~\ref{app:rts_vs_crts}

\begin{table*}[b]
    \centering
    \small
    \resizebox{\linewidth}{!}{%
    \begin{tabular}{lclc@{\hskip 30pt}|lcl}
        \toprule
        \textbf{Model} & \textbf{GLUE} & \textbf{FLOPS} & & \textbf{Model} & \textbf{GLUE} & \textbf{FLOPS} \\ 
        \toprule
        BERT\textsubscript{Small} + RTS                                & 75.4  & $\times$ 0.10 	& & BERT\textsubscript{Large} + MLM/NSP~\citep{devlin-etal-2019-bert}      & 83.3  & $\times$ 12 \\
        BERT\textsubscript{Small} + MLM                                & 74.1  & $\times$ 0.12 	& & ELECTRA\textsubscript{Base} + MLM/TD~\citep{clark2020electra}      & 85.7  & $\times$ 21 \\
        BERT\textsubscript{Small} + SLM                                & 75.7  & $\times$ 0.12 	& & RoBERTa\textsubscript{Base} + MLM~\citep{liu2019roberta}       & 86.3  & $\times$ 53 \\
        BERT\textsubscript{Base} + RTS                                & 79.9  & $\times$ 0.81 	& & ELECTRA\textsubscript{Large} + MLM/TD~\citep{clark2020electra}      & 88.6  & $\times$ 194 \\
        BERT\textsubscript{Base} + MLM                                & 79.7  & $\times$ 1 	& & RoBERTa\textsubscript{Large} + MLM~\citep{liu2019roberta}       & 88.8  & $\times$ 200 \\
        BERT\textsubscript{Base} + SLM                                & 80.4  & $\times$ 1 	& & ALBERT\textsubscript{Large} + MLM/SOP~\citep{lan2020albert}     & 90.0  & $\times$ 1937 \\
        \bottomrule
    \end{tabular}
    }
    \caption{\small Comparison of computational requirements for state-of-the-art large language models. We report the average GLUE score as well as an estimation of the FLOPS used for the pre-training, normalized on BERT\textsubscript{Base}~+~MLM.}
    \label{tab:large_models}
\end{table*}

\section{Pre-Training Computational Trends}

Our goal is to develop models that achieve comparable performance to MLM-based models while requiring less computational resources. Our findings indicate that the performance improvement follows a logarithmic pattern as the size of the pre-training dataset increases and models are trained for longer. Additionally, our computationally lighter models, such as RTS and C-RTS, demonstrate no statistically significant difference compared to more expensive models on most tasks. A comparison in Table~\ref{tab:large_models} reveals that the top-performing architectures only outperform MLM-based models when trained on significantly larger amounts of data and for much more time. For instance, ELECTRA\textsubscript{Base} utilizes 21 times more resources than BERT\textsubscript{Base}, and RoBERTa uses 53 times more. It is worth noting that achieving a score of 90.0 on the GLUE benchmark requires a model to be trained for 2000 times the duration of the original BERT model training, which is quite remarkable. For those reasons, we believe research in finding more efficient architectures, objectives or distillation techniques is extremely important to reduce the carbon footprint of models pre-training.

\chapter{Self-supervised Objectives for Multi-Sentence Inference}
\label{cha:jointwise}

In this Chapter, we propose pre-training objectives designed specifically to improve accuracy in multiple-sentence classification. Many tasks such as AS2, Fact Verification or NLI are solved by training a sentence pair classifier, for example by feeding the Transformer model with a question-answer or a claim-evidence pair. Recent studies~\citep{zhang-etal-2021-joint, tymoshenko-moschitti-2021-strong} showed that modeling multiple candidate sentences helps in learning long-range dependencies and allows the model to compare them, instead of scoring them individually.

A major problem is that off-the-shelf Transformer models struggle at modeling multiple related sentence candidates together. For this reason, we propose a new pre-training technique that exploits weak supervision of pre-training corpora to adapt the model at reasoning over multiple input sentences taken from the same paragraph. We call \emph{Jointwise} the class of models trained with objectives that target relations among multiple input sentences. At the same time, we call \emph{Pairwise} models the class of architectures designed to reason over 1 or 2 input sequences.

Regarding the related works, additional information about the RoBERTa architecture~\citep{liu2019roberta} can be found in Section~\ref{sec:sota_lm_roberta}. Additionally, Sections~\ref{sec:related_as2} and~\ref{sec:related_fact_verification} describe the actual state-of-the-art for Answer Sentence Selection and Fact Verification. In Section~\ref{sec:related_multi_sentence} instead, a description of works about multi-sentence inference is provided, while in Section~\ref{sec:related_document_structure} we analyze related research that exploits the weak structure of large corpora for supervision.

To demonstrate the superiority of our approach, we evaluate models pre-trained with our objectives on three datasets for Answer Sentence Selection (ASNQ, WikiQA and TREC-QA) and one dataset for Fact Verification (FEVER). Statistics and details about each dataset are given in Section~\ref{sec:as2_datasets}. We select RoBERTa\textsubscript{Base}~\citep{liu2019roberta} as the starting point because it provides good accuracy on a wide range of tasks and does not require expensive hardware to be fine-tuned. We compare against a baseline pre-trained only with common objectives (e.g. MLM) as well as models pre-trained over data formatted for our task but without explicit supervision.

\label{sec:self_supervised_for_multi_sentence_inference}
\section{Limits of Pairwise Models}

Examples of Pairwise models comprehend Transformers either (i) pre-trained with sentence-level objectives such as NSP~\citep{devlin-etal-2019-bert} or SOP~\citep{lan2020albert} or (ii) fine-tuned on Pairwise tasks such as Natural Language Inference, Answer Sentence Selection and Fact Verification.

When applied to downstream tasks, Pairwise models perform effectively in comparing two input sequences, thanks to the Cross-Attention of Transformers. In the AS2 task, they receive a question and a single answer candidate and produce a score indicating the degree of correctness. However, this approach has limitations as the model cannot efficiently compare and resolve references among multiple answer candidates to select the best answer.

Conversely, in Fact Verification, the current state-of-the-art approach involves providing the model with a claim and a set of evidence sentences. The model then determines whether the claim is supported, neutral, or refuted by the evidence sentences~\citep{liu-etal-2020-fine, tymoshenko-moschitti-2021-strong}. Nevertheless, Transformers are typically pre-trained with token-level and sentence-level objectives, focusing on at most two spans of text. Although models like RoBERTa are trained on very long sequences containing multiple sentences, they are not explicitly trained to identify connections and relations between these sentences. Consequently, they may struggle to learn meaningful dependencies between the candidates during fine-tuning, making them less adaptable to tasks requiring reasoning over multiple inputs.

\section{Multi-Sentence Transformer Model}
\label{sec:multi_sentence_transformer_model}

In this Section, we present the Transformer architecture and the pre-training objective we developed for the multi-sentence training and inference tasks.

\subsection{Joint Encoder Architecture}
\label{sec:joint_encoder_architecture}

We created an architecture specifically designed to process and reason over multiple input sequences. We indicate with $k+1$ the maximum number of text spans that can be provided to the model. The first input sentence is called \emph{pivot} and is of special importance because we study its relation with the other $k$ input sequences. For example, in AS2 we place the question in the first position and the model should predict its relation with the other $k$ answer candidates. Similarly, in Fact Verification we set the claim as the pivot and we predict whether the other $k$ evidence sentences support or reject the claim.

\begin{figure*}[t]
    \centering
    \includegraphics[width=\textwidth]{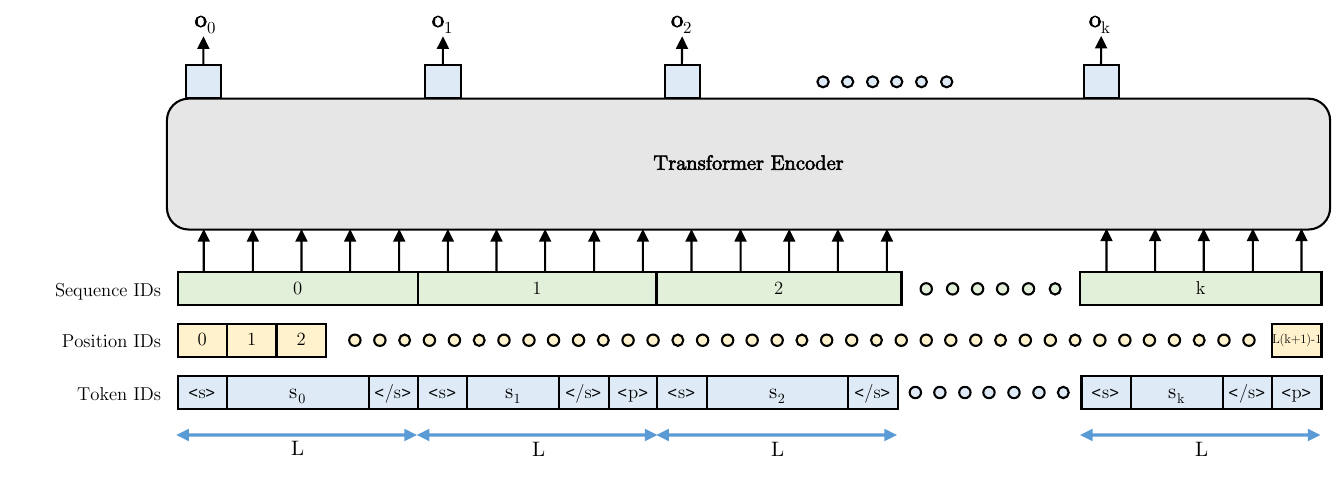}
    \caption{\small Example of Jointwise Fixed architecture.}
    \label{fig:joint_fixed}
\end{figure*}

We experimented with 2 different configurations: the first, referred to as \emph{Fixed} (Figure~\ref{fig:joint_fixed}), and the second, named \emph{Flexible} (Figure~\ref{fig:joint_flexible}), both consisting of a RoBERTa Encoder~\citep{liu2019roberta}. The main differences between the Fixed and Flexible variants are how inputs are formatted before being fed to the model and how multiple predictions are performed from the last hidden state.

\subsubsection{Fixed Model}

In the Fixed variant of our architecture, each of the $k+1$ input text spans has the same Fixed length $L$, measured in the number of tokens. We crop or pad on the right all the input sequences whose length is different than $L$. Notice that this may lead to pad tokens between different input sentences. This is not an issue because the attention mask will be aware of those special tokens and will disable attention scores in the corresponding positions. We add positional embeddings and sequence id embeddings to the word embeddings. Positional embeddings help the model in computing attention between pairs of tokens in different positions while sequence id embeddings are useful to let the model know to which sequence each token belongs. We are aware that having both positional ids and sequence ids in a Fixed architecture is redundant, however, we keep both embeddings because the overhead is negligible and we can use a single implementation for both Fixed and Flexible models. We also experimented using the same positional ids $0, \dots, L-1$ for every sentence, but we didn't measure significant gains. Notice that in this case, token type ids played an important role and couldn't be removed.

Jointwise Fixed models perform $k+1$ sentence-level predictions for each step, employing a classification head over output embeddings $[\vo_0, \vo_1, \dots, \vo_{k}]$, which correspond to the first token $<$s$>$ of each input sentence\footnote{$<$s$>$ is RoBERTa's ``begin of sentence'' token, equivalent to BERT's [CLS].}. In particular, given the last hidden state $\mH \in \sR^{(k+1)L \times d}$, the output embeddings $\vo_0, \vo_1, \dots, \vo_k \in \sR^d$ correspond to the rows $0, L, \dots, k L$ of $\mH$.

\subsubsection{Flexible Model}

\begin{figure*}[b]
    \centering
    \includegraphics[width=\textwidth]{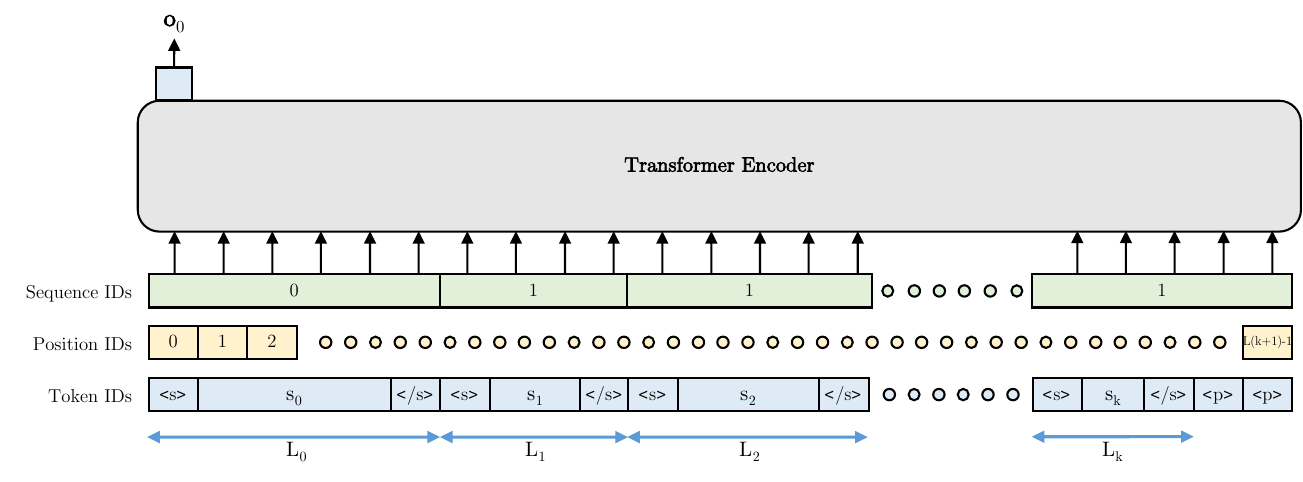}
    \caption{\small Example of Jointwise Flexible architecture.}
    \label{fig:joint_flexible}
\end{figure*}

The Flexible architecture is similar to the Fixed version but allows input sequences to have a variable length. In Figure~\ref{fig:joint_flexible}, the length of each input sequence is indicated with $L_0, \dots, L_k$. The only constraint in Flexible models is:
\begin{equation}
\sum_{i=0}^{k} L_i \le L
\end{equation}
If the input does not match the constraint, we iteratively remove one token at a time from the longest sequence until we do not match the requirement. We use again positional and sequence ids. In particular, preliminary experiments demonstrated that sequence ids are very important in this architecture because they help the model to better understand the context for each prediction. In Flexible models, we employ only 2 sequence ids instead of $k+1$. We assign a sequence id equal to 0 to the pivot and 1 to all the other $k$ sentence candidates. Predictions with Flexible models are simpler because we consider always only the output embedding $\vo_0$ of the first token to perform single or multi-label predictions.

\subsection{Multi-Sentence Inference Tasks}

Here, we describe how we adapt AS2 and Fact Verification for the multi-sentence inference task. We denote with $k$ the maximum number of predictions performed by our Jointwise model.

\paragraph{AS2}
We extend the definition of AS2 given in Section~\ref{sec:answer_sentence_selection} for multi-sentence inference. In particular, we partition the candidates $\sC$ in disjoint subsets $\sC_0, \dots, \sC_m$ of size $k$ with random selection. Then, we feed the model $\mathcal{M}$ with the question and one subset at a time, obtaining $k$ predictions over the candidates. If one subset $\sC_i$ is smaller than $k$, we add padding on the right until we match the sequence length of the other elements in the batch and we do not predict and back-propagate errors from the corresponding classification heads. Finally, we select the best answer among all subsets as the one having the highest overall score, as in the original task.

Our intuition is that by providing $\mathcal{M}$ with $k$ different candidates, it can model interrelated information to find the best answer $a$ to $q$~\citep{zhang-etal-2021-joint}.

\paragraph{Fact Verification} Let's consider a claim $c$ and a set of evidence sentences $\sE = \{ e_1, \dots, e_n \}$ retrieved using a BERT-based DocIR~\citep{liu-etal-2020-fine}. In the classic definition of the task, $c$ should be supported/refuted by at least one piece of evidence $e_i \in \sE$ to be labeled accordingly. More details about Fact Verification are given in Section~\ref{sec:fact_verification}.

As for Joint AS2, we provide the model $\mathcal{M}$ with the claim $c$ and the evidence $\sE$, such that different evidence sentences can be modeled jointly and complement each other~\citep{tymoshenko-moschitti-2021-strong}.

\subsection{Training and Inference using Joint Encoders}
\label{sec:inference_using_joint_encoder}

We describe how we exploit output embeddings $\vo_0, \dots, \vo_k$ of Fixed and $\vo_0$ of Flexible models to perform predictions. We differentiate between AS2, that requires to predict a label for each candidate, and Fact Verification, which requires to predict a single label for every input example.

\subsubsection{Predict a single label}

We experiment with two different strategies to predict a single label from the output embeddings of the model. 

\paragraph{IE$_1$} We place a linear layer $\mW \in \sR^{2 \times d}$ only on top of $\vo_0$, that corresponds to the embedding of the first input sentence (similar to BERT~\citep{devlin-etal-2019-bert}). This classification head can be used for both Fixed and Flexible models. We call this method Individual Evidence inference head and we obtain predictions $p$ with:
\begin{equation}
p = \mW \vo_0
\end{equation}
\paragraph{AE$_1$} Here, we apply a single linear layer $\mW \in \sR^{2 \times d}$ to the average of all the output embeddings $\vo_1, \dots, \vo_k$ to gather information from all input candidates. By definition, this classification head can be used only with Fixed models. We refer to this technique as Aggregated Evidence classification head:
\begin{equation}
p = \mW \left( \frac{1}{k} \sum_{i=1}^k \vo_i \right)
\end{equation}
\subsubsection{Predict $k$ labels}

To predict $k$ different labels, we experimented with 3 techniques (2 for Fixed and 1 for Flexible models).

\begin{figure*}[t]
    \centering
    \includegraphics[width=\textwidth]{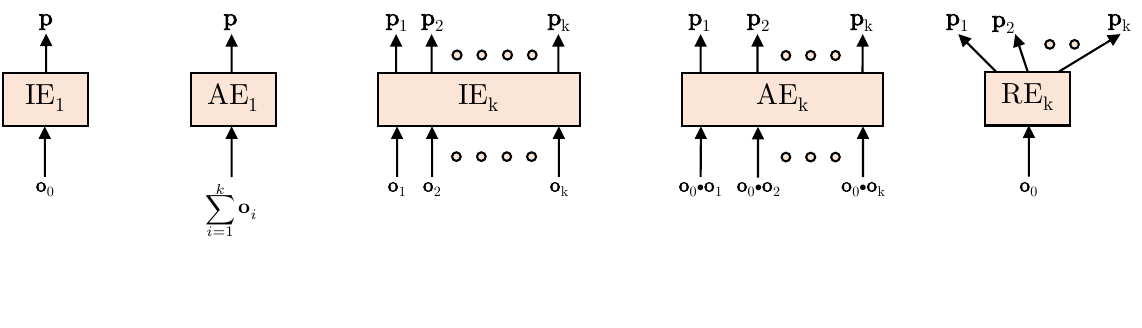}
    \vspace{-4em}
    \caption{\small Inference heads for Joint Transformer models.}
    \label{fig:joint_classification_heads}
\end{figure*}

\paragraph{IE$_k$} A linear layer with shared weights $\mW \in \sR^{2 \times d}$ is applied to each output embedding $\vo_i$ to obtain predictions. By definition, this head works only with Fixed models. We refer to this method as $k$-candidate Individual Evidence inference head:
\begin{equation}
p_i = \mW \vo_i \ \ \forall i \in 1, \dots, k
\end{equation}
\paragraph{AE$_k$} In this experiment, let the linear layer be a matrix of shared weights $\mW \in \sR^{2 \times 2d}$. We feed the linear layer with the concatenation of the first output embedding $\vo_0$ with each candidate embedding ($\vo_1, \dots, \vo_k$). This head is yet again specifically designed only for Fixed models. We call this technique $k$-candidate Aggregated Evidence (AE$_k$) classification head (``$\cdot$'' means vectors concatenation):
\begin{equation}
p_i = \mW ( \vo_0 \cdot \vo_i ) \ \ \forall i \in 1, \dots, k
\end{equation}
\paragraph{RE$_k$} This last classification head is designed only for Flexible models. Let $\{ \mW_i \in \sR^{2 \times d}, i \ \in \ 1, \dots, k \}$ be a set of $k$ linear layers. We apply all of them independently on the output embedding $\vo_0$ of the first input token to obtain $k$ predictions. We name this method $k$-candidate Reduced Evidence (RE$_k$). Predictions are computed as follows:
\begin{equation}
p_i = \mW_i \vo_0 \ \ \forall i \in 1, \dots, k
\end{equation}
In figure~\ref{fig:joint_classification_heads} we depict every classification head to better catch the differences between them. Furthermore, Table~\ref{tab:joint_classification_heads} provides a summary of the applicable classification heads for each model and task.

\begin{table*}[t]
    \centering
    \resizebox{0.80\linewidth}{!}{
    	\begin{tabular}{lccccccccccc}
    	\toprule

	\multirow{2}{*}{\textbf{Task}} & \multicolumn{5}{c}{\textbf{Fixed}} & & \multicolumn{5}{c}{\textbf{Flexible}} \\
	\cmidrule{2-6} \cmidrule{8-12}	& IE$_1$		& IE$_k$		& AE$_1$		& AE$_k$ 		& RE$_k$	& 	& IE$_1$		& IE$_k$		& AE$_1$		& AE$_k$ 		& RE$_k$ \\
	\midrule
	MSPP Pre-Training			&			& \cmark	& 			& \cmark 	& 			&& 			& 			& 			& 			& \cmark \\
	Answer Sentence Selection	& 			& \cmark	& 			& \cmark	& 			&&			& 			& 			& 			& \cmark \\
	Fact Verification			& \cmark	& 			& \cmark	& 			& 			&& \cmark	& 			&			& 			& \\
	\bottomrule
\end{tabular}}
    \caption{\small Compatibility matrix between models, classification heads and downstream tasks.}
    \label{tab:joint_classification_heads}
\end{table*}

\subsection{Continuous Pre-Training of Joint Encoders}
\label{sec:continuous_pretraining_of_joint_encoders}

\begin{figure*}[b!]
    \centering
    \includegraphics[width=0.90\textwidth]{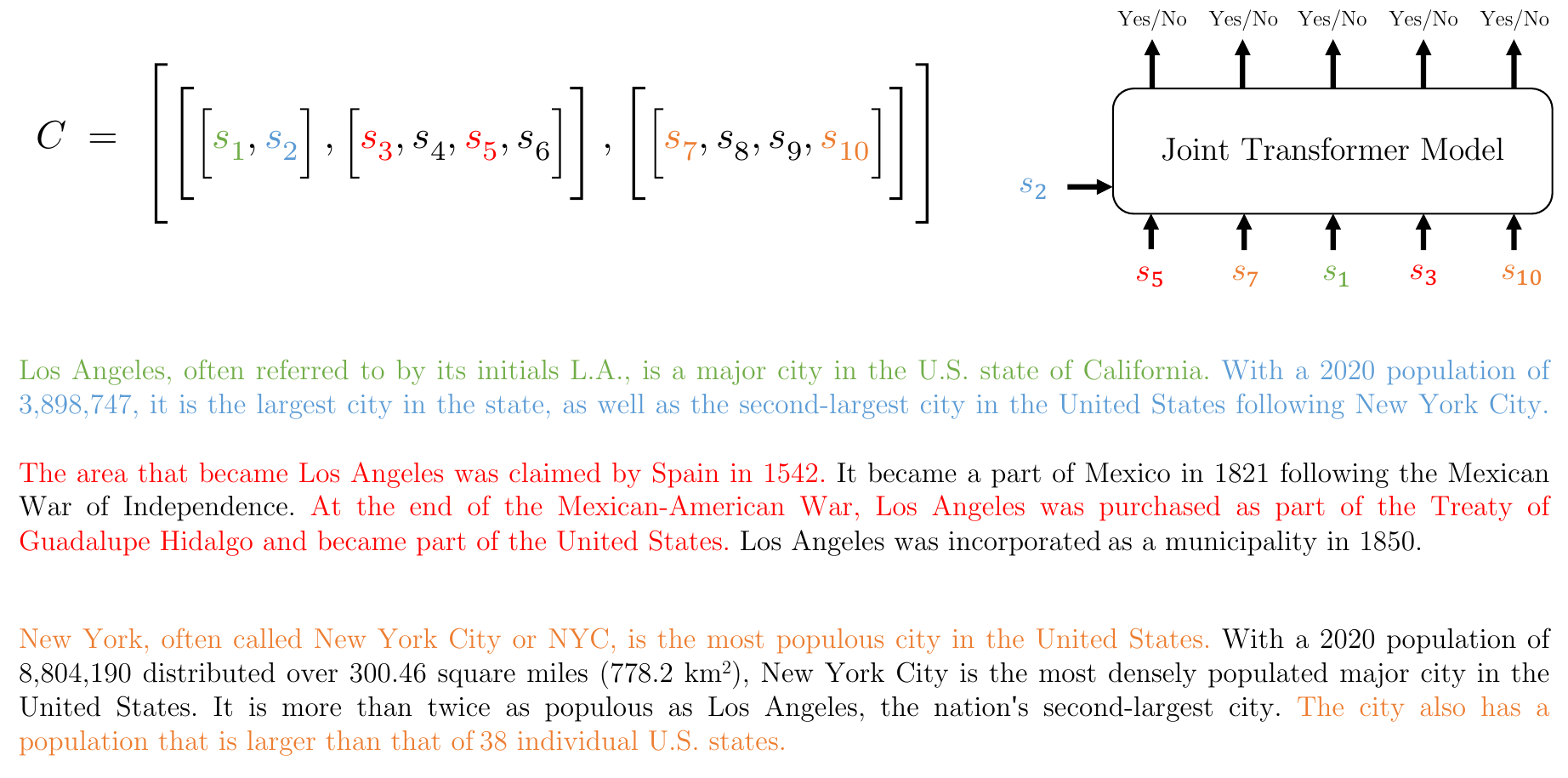}
    \caption{\small Example of MSPP objective.}
    \label{fig:mspp}
\end{figure*}

In long documents, paragraphs are used to explore the document's topic from various perspectives. Despite this rich source of information, most Transformer pre-training strategies fail to take advantage of it, leaving the unsupervised pre-training phase without any guidance. To improve the ability of Transformer models to understand the relationships between multiple sentences, we introduce a new pre-training task called Multi-Sentences in Paragraph Prediction (MSPP). In this task, the model is given $k+1$ sentences $\{s_0, \dots, s_k\}$ and is asked to predict whether sentences $\{s_1, \dots, s_k\}$ belong to the same paragraph $P$ as $s_0$, in the document $\mD$. We use the IE$_k$ and AE$_k$ prediction heads for Fixed models and RE$_k$ with Flexible models to perform the $k$ predictions. More formally, we create a label $y_i$ for each sentence $s_i$ with the following rule:
\begin{equation}
y_i = \begin{cases}
       1 \ \ \text{if} \ s_0, s_i \in P \ \text{in} \ \mD\\
       0 \ \ \text{otherwise} \\
     \end{cases} \forall \ i \in \{1,\dots,k\}
\end{equation}
During the pre-training phase, we randomly select a sentence from a paragraph as $s_0$. We then select $k_1$ other sentences from the same paragraph as positive examples, $k_2$ sentences from other paragraphs in the same document as hard negative examples, and $k_3$ sentences from documents other than $\mD$ as easy negatives. We ensure $k_1 + k_2 + k_3 = k$ by sampling more easy negatives ($k_3$) if paragraph $P$ is not large enough to sample $k_2$ hard negatives. This method tasks the model to learn the relationship between sentences in a paragraph, improving its ability to understand the underlying structure of long documents. An example of MSPP is given in Figure~\ref{fig:mspp}.

\section{Experiments}

To assess the performance of our joint Transformers, we conduct experiments on three AS2 datasets and one Fact Verification dataset. We do not use commonly used language model benchmarks for our evaluation, such as GLUE, because they only involve Pairwise classification tasks, which is not sufficient for our study.

\subsection{Continuous Pre-Training}

In our experiments, we want to avoid improvements stemming from the usage of additional data. For this reason, we use the same pre-training corpora of RoBERTa: English Wikipedia, BookCorpus, OpenWebText and CC-News. For more details about the datasets, see Section~\ref{sec:pretraining_datasets}. RoBERTa was also trained on the STORIES dataset, however, we omit it because it is not publicly available anymore\footnote{\url{https://github.com/tensorflow/models/tree/archive/research/lm\_commonsense\#1-download-data-files}}.

In all of our experiments, we set $k=5$ as in \citet{zhang-etal-2021-joint, tymoshenko-moschitti-2021-strong} because it aligns well with the number of candidates/evidence sentences in the datasets and because it does still allow an efficient training with a reasonable input sequence length. We set also $L=64$ in the experiments. Thus, Fixed models will have at most 6 input sequences of length 64 while Flexible models will also accept up to 6 input sequences whose total length is at most 384 tokens. However, since Flexible models do not apply internal padding, we found that setting the maximum sequence length $L$ to $256$ does not affect performance but increases training speed.

The pre-training dataset is created with the MSPP strategy by setting $k_1 = 1, k_2 = 2, k_3 = 2$. With these settings, we have a good balance of soft and hard negative examples. The resulting dataset contains about 180M examples. We do continuous pre-training starting from a RoBERTa\textsubscript{Base} checkpoint because training from scratch would require a very large computational budget\footnote{RoBERTa\textsubscript{Base} was trained on 1024 NVIDIA V100 GPUs for 1 day, which corresponds to an expense of more than 100,000\$ on Amazon Web Services or similar cloud computing providers.}. The original checkpoint is only augmented with sequence-level embeddings and a small classification head for binary predictions. While doing continuous pre-training, we found it beneficial to keep the original MLM objective along with MSPP. We train the model over the dataset described above for 100K steps with a batch size of 4096. We used a triangular learning rate with a peak value of $5 \cdot 10^{-5}$ and 10K warmup steps. We set the optimizer with $\beta_1 = 0.9$, $\beta_2 = 0.999$ and $\epsilon = 10^{-8}$ to update the weights of the model. We also apply a weight decay of $0.01$ and dropout with probability $0.1$ to regularize model training. Since we trained in mixed-precision (\emph{fp16}), we did not use gradient clipping because gradients are already scaled by the NVIDIA Apex algorithm automatically. For the pre-training, we experimented with 2 Fixed and 1 Flexible model: the first trained with IE$_k$, the second with AE$_k$ and the Flexible with RE$_k$. As in BERT and ALBERT~\citep{devlin-etal-2019-bert,lan2020albert}, we use an equal weight for the MLM and MSPP to compute the final loss value. The training takes about 2.2 days (Fixed) and 1.7 days (Flexible) on our machine, which is described in Section~\ref{sec:machines}, using mixed precision and the parameters described above. In terms of efficiency, our continuous pre-trainings account for about a 7\% (Fixed) and a 5\% (Flexible) increase in computation compared to the original RoBERTa pre-training.

\subsection{Fine-Tuning}

We evaluate the 3 continuously pre-trained checkpoints on ASNQ, WikiQA, TREC-QA and FEVER, using the correct classification heads, i.e. IE$_k$, AE$_k$ or RE$_k$ for AS2 and IE$_1$ or AE$_1$ for Fact Verification.

We compare against many baselines from other works. On AS2, we compare with (i) RoBERTa\textsubscript{Base}, used as a Pairwise Cross-Encoder, which is the standard practice~\citep{garg2019tanda} and (ii) RoBERTa\textsubscript{Base} with our classification heads but without MSPP pre-training. The latter allows us to evaluate how RoBERTa\textsubscript{Base} without the specialized pre-training objective adapts at processing $k+1$ input sequences.

On FEVER instead, we compare with GEAR~\citep{zhou-etal-2019-gear}, KGAT~\citep{liu-etal-2020-fine} and Transformer-XH~\citep{zhao2020transformer-xh}. We also include 3 models from \citet{tymoshenko-moschitti-2021-strong}: (i) a RoBERTa\textsubscript{Base} for multi-sentence classification with IE$_1$ prediction head, (ii) two Pairwise Cross-Encoder RoBERTa\textsubscript{Base} with heads that do max-pooling and weighted-sum to gather the context for the classification.

We optimize our models with a maximum sequence length of $384$ tokens for all experiments, which is $64$ multiplied by the number of sentences $k+1$. Regarding ASNQ, we train the model for a maximum of 10 epochs using a learning rate of $10^{-5}$ and a batch size of 512 examples, with the FuseAdam optimizer. We warmup for only 5000 steps, and we use early stopping based on the Mean Average Precision (MAP) on the development set. For WikiQA and TREC-QA, we train the model using a learning rate of $2 \cdot 10^{-6}$ and a batch size of 32, with 1000 warmup steps. We train for up to 40 epochs and use early stopping based on the MAP metric on the development set. Finally, for the FEVER dataset, we use a batches of 64 example, a learning rate of $10^{-5}$, and 1000 warmup steps. We use early stopping based on the Accuracy metric on the development set.

We evaluate models' performance on Answer Sentence Selection with Mean Average Precision (MAP), Mean Reciprocal Rank (MRR) and Precision@1 (P@1), as in similar works~\citep{garg2019tanda}. For the FEVER dataset, we measure performance with Label Accuracy, which is a standard metric in Fact Verification~\citep{liu-etal-2020-fine, tymoshenko-moschitti-2021-strong, zarharan-etal-2021-parsfever}. It evaluates the accuracy in predicting whether a claim is supported/refuted/neutral to the set of retrieved evidence sentences.

\section{Results}

In this Section, we report results obtained by fine-tuning the baselines and our MSPP pre-trained models on AS2 and Fact Verification benchmarks. We report results from original papers when available. On AS2, we analyze the results after a standard fine-tuning as well as the results of chaining our Jointwise models with a common Pairwise re-ranker. The statistical significance in AS2 is compared to RoBERTa\textsubscript{Base}~\citep{liu2019roberta}.

\subsection{Answer Sentence Selection}
\begin{table*}[t]
    \centering
    \resizebox{1.0\linewidth}{!}{
    \begin{tabular}{l c@{\hskip 6pt}c@{\hskip 6pt}c @{\hskip 3pt}c c@{\hskip 6pt}c@{\hskip 6pt}c @{\hskip 3pt}c c@{\hskip 6pt}c@{\hskip 6pt}c}
    \toprule
\multirow{2}{*}{\textbf{Model}} & \multicolumn{3}{c}{\textbf{ASNQ}} & &\multicolumn{3}{c}{\textbf{WikiQA}} & & \multicolumn{3}{c}{\textbf{TREC-QA}} \\

\cmidrule{2-4} \cmidrule{6-8} \cmidrule{10-12}					& \textbf{MAP} & \textbf{MRR} & \textbf{P@1}				& & \textbf{MAP} & \textbf{MRR} & \textbf{P@1}			& & \textbf{MAP} & \textbf{MRR} & \textbf{P@1} \\
    \toprule

RoBERTa\textsubscript{Base} Fixed (IE$_k$)				& 62.1 (0.8) & 65.6 (1.2) & 57.4 (1.3)  					& & 39.4 (1.6) & 40.3 (1.8) & 19.7 (1.9)           				& & 41.9 (2.4) & 50.8 (3.9) & 30.9 (5.4) \\
RoBERTa\textsubscript{Base} Fixed (AE$_k$)				& 62.3 (0.7) & 66.0 (1.0) & 57.7 (1.1) 					& & 39.0 (2.8) & 39.7 (2.9) & 18.7 (3.9)           				& & 42.3 (3.2) & 49.2 (5.0) & 29.7 (6.9) \\
RoBERTa\textsubscript{Base} Flexible (RE$_k$)		   		& 63.7 (0.3) & 70.2 (0.4) & 58.9 (0.6)						& & 77.8 (2.3) & 79.6 (2.4) & 69.5 (3.4)            				& & 84.8 (0.7) & 91.3 (1.7) & 84.8 (3.1) \\

	\midrule

	RoBERTa\textsubscript{Base}~\citep{liu2019roberta}    			& 66.9 (0.1) & 73.1 (0.1) & 61.8 (0.2)  					& & 85.3 (0.9)  & 86.5 (1.0) & 77.1 (2.1)            				& & 89.3 (0.9)   & 93.1 (1.0) & 87.9 (2.2) \\
	\textbf{RoBERTa\textsubscript{Base} Fixed (IE$_k$) + MSPP}  		& \und{67.2} (0.2) & \und{73.7} (0.2) & \und{63.0} (0.3) 		& & \und{88.5} (1.5) & \und{89.0} (1.5) & \und{82.7} (2.2)  		& & \bund{91.1} (0.5) & \bund{95.2} (1.3) & \bund{91.7} (2.2) \\
	\textbf{RoBERTa\textsubscript{Base} Fixed (AE$_k$) + MSPP}  		& \und{67.3} (0.2) & \und{73.7} (0.2) & \und{63.0} (0.3)		& & \und{87.9} (1.4) & \und{89.0} (1.5) & \und{81.9} (2.6)		& & 90.1 (1.0)  & 93.6 (0.6) & 88.7 (0.8) \\
\bottomrule
\end{tabular}}
    \caption{\small Results on ASNQ, WikiQA and TREC-QA. IE$_k$, AE$_k$ and RE$_k$ are the prediction heads we use both for MSPP pre-training and fine-tuning. We use bold formatting for our proposals and the best results. We underline statistically significant improvements over RoBERTa\textsubscript{Base}.}
    \label{tab:joint_table_as2}
\end{table*}

Results on all the AS2 datasets are reported in Table~\ref{tab:joint_table_as2}. We experiment using all multi-label heads defined in Section~\ref{sec:inference_using_joint_encoder}, i.e. IE$_k$, AE$_k$ and RE$_k$. Results are reported against a Pairwise baseline which uses the same Transformer encoder and has the same parameters as our Joint models (which in some cases have only a slightly larger classification head, with negligible effects on fine-tuning/inference time). The Table shows two main discoveries: (i) models fine-tuned directly on the downstream tasks in a multi-sentence prediction setting do not achieve high re-ranking scores; (ii) MSPP pre-training is fundamental to accustom the model to reason over multiple sentences.

Regarding the first point, the low performance of models fine-tuned directly on the multi-sentence tasks can be explained by considering that attention scores are not pre-trained to compare many sentences. Most pre-training objectives focus on token-level dependencies (e.g. MLM, TD, SLM) while sentence-level objectives such as NSP or SOP consider at most 2 input sequences. This is further confirmed by the results on ASNQ, which is a large dataset, and those on WikiQA and TREC-QA, which contain only a few thousand examples. Results on ASNQ are higher in the standard fine-tuning setting because the model can leverage the large quantity of annotated examples (about 4M, each containing 1 question and $k=5$ answer candidates) to learn multi-sentence relations. 

Regarding the Fixed models, both checkpoints pre-trained with MSPP and fine-tuned with either IE$_k$ and AE$_k$ improve over the Pairwise baseline by a large margin on WikiQA and TREC-QA. Specifically, the Joint Fixed model with IE$_k$ classification head improves by 5.6 points in Precision@1 on WikiQA and by 3.8 points on TREC-QA. Similarly, the Joint model with AE$_k$ head outperforms the baseline by 4.8 in Precision@1 on WikiQA and by 0.8 in TREC-QA. On ASNQ, improvements are modest because the dataset is larger and reduces the differences seen in pre-training, but are still consistent. In particular, the Joint model with IE$_k$ and AE$_k$ show a gain in MAP of 0.3 and 0.4 points respectively. When compared with Precision@1 instead, improvements are larger and equal to 1.2 points for both models.

The Flexible model we present reaches the best results on ASNQ and WikiQA, with a gap of 3.5 points in P@1 over the baseline on ASNQ and 6.4 points over the WikiQA baseline. On TREC-QA, MAP and P@1 are aligned with the other Joint models, with gains over the Pairwise baseline of 1.7 points in MAP and 3.3 in Precision@1. To the best of our knowledge, results on ASNQ are the actual state-of-the-art when compared with other models with a similar number of parameters\footnote{In Chapter~\ref{cha:pairwise} we will show that large models can achieve even higher scores.}. This shows that using a Flexible architecture is overall the best choice to design a multi-sentence inference model for Answer Sentence Selection.

Finally, we compare with T\textsc{and}A, which is the actual state-of-the-art in AS2 for WikiQA and TREC-QA. Our methods demonstrate similar performance on both datasets, with a maximum difference of 0.3 points when measured with MAP and MRR. Notice that T\textsc{and}A exploits a large labeled dataset for AS2 (ASNQ) to perform a pre-fine-tuning step and transfer knowledge into smaller datasets. We use instead only raw data from large corpora without manual annotation.

\subsubsection{Re-Rankers Chain}

\begin{table*}[t]
    \centering
    \resizebox{\linewidth}{!}{
    \begin{tabular}{l c@{\hskip 6pt}c@{\hskip 6pt}c @{\hskip 5pt}c c@{\hskip 6pt}c@{\hskip 6pt}c @{\hskip 5pt}c c@{\hskip 6pt}c@{\hskip 6pt}c}
    \toprule
\multirow{2}{*}{\textbf{Model}} & \multicolumn{3}{c}{\textbf{ASNQ}} & &\multicolumn{3}{c}{\textbf{WikiQA}} & & \multicolumn{3}{c}{\textbf{TREC-QA}} \\
\cmidrule{2-4} \cmidrule{6-8} \cmidrule{10-12} & 			\textbf{MAP} & \textbf{MRR} & \textbf{P@1}	& & \textbf{MAP} & \textbf{MRR} & \textbf{P@1}	& & \textbf{MAP} & \textbf{MRR} & \textbf{P@1} \\
    \toprule

RoBERTa\textsubscript{Base}~\citep{liu2019roberta}						& 66.9 (0.1) & 73.1 (0.1) & 61.8 (0.2)  				& & 85.3 (0.9)  & 86.5 (1.0) & 77.1 (2.1)            				& & 89.3 (0.9)   & 93.1 (1.0) & 87.9 (2.2) \\

\textbf{RoBERTa\textsubscript{Base} Fixed (IE$_k$) + MSPP} 				& \und{71.3} (0.4) & 73.1 (0.5) & \und{63.9} (0.8) 		& & \bund{88.5} (1.4) & \und{89.0} (1.5) & \und{82.7} (3.0)  	& & \bund{93.5} (1.2) & \bund{95.4} (1.4) & \bund{92.2} (1.8) \\
\textbf{RoBERTa\textsubscript{Base} Fixed (AE$_k$) + MSPP}				& \bund{71.5} (0.3) & \bld{73.4} (0.3) & \bund{64.3} (1.1)	& & \und{87.9} (0.8) & \und{88.7} (0.7) & \und{82.1} (1.1)		& & \bund{93.5} (1.1) & \und{94.9} (1.2) & \und{91.2} (1.4) \\
\textbf{RoBERTa\textsubscript{Base} Flexible (RE$_k$) + MSPP}			& \und{70.7} (0.5) & 72.5 (0.6) & \und{64.1} (1.0)		& & \und{88.1} (1.1) & \bund{89.1} (1.2) & \bund{82.8} (1.9)  	& & 90.7 (0.7) & 92.2 (0.8) & 89.7 (1.2) \\

\midrule

Superior limit												& \multicolumn{3}{c}{86.5}							& & \multicolumn{3}{c}{97.9}							& & \multicolumn{3}{c}{100.0} \\

\bottomrule

\end{tabular}}
    \caption{\small Results on ASNQ, WikiQA and TREC-QA for Joint Fixed and Flexible models when chained after the RoBERTa\textsubscript{Base} Pairwise re-ranker. In the last row, we show the maximum achievable performance based on the Hit-Rate at $k=5$ (HR@5) of the Pairwise re-ranker. We use bold formatting for our proposals and the best results. We underline statistically significant improvements over RoBERTa\textsubscript{Base}.}
    \label{tab:joint_table_top_k_as2}
\end{table*}

In this Section, we study whether Jointwise models can be chained with a Pairwise Cross-Encoder to improve the performance of the latter. More specifically, we use a RoBERTa\textsubscript{Base} Cross-Encoder to find the top $k$ more relevant answers for each question. After that, we apply the Jointwise model to better re-rank the top $k$ answers retrieved by the baseline Cross-Encoder.

Results are reported in Table~\ref{tab:joint_table_top_k_as2}. The last row of the Table shows the upper limit to Jointwise model performance. The limit is calculated as the Hit-Rate@k of the Pairwise re-ranker. This comparison is needed because if the Pairwise baseline is not able to rank at least one positive candidate in the top $k$, Jointwise encoders will process only negative examples and will receive a bad score by the evaluation metrics.

In this experiment, Fixed models provide better overall performance, especially on the TREC-QA dataset. On the other hand, the Flexible model provides similar performance to Fixed models on ASNQ and WikiQA, but falls short behind on TREC-QA, still providing a better re-ranking of the first $k$ candidates than the baseline Pairwise Cross-Encoder. On average, performance is lower than applying Jointwise models directly to the whole datasets: we suppose that our models work better when applied to all data because more candidates can be compared with each other. Moreover, the accuracy of Jointwise models when applied after a Pairwise re-ranker is limited by the latter's performance. This is further confirmed by checking the results on TREC-QA, in which most Jointwise models perform better than the previous setting because the Pairwise re-ranker was able to score always at least a positive candidate in the top $k$ (HR@k = 100.0\%).

\subsection{Fact Verification}

\begin{table*}[t]
    \centering
    \resizebox{0.70\linewidth}{!}{
	\begin{tabular}{lccc}
    	\toprule

	\multirow{2}{*}{\textbf{Model}} & & \multicolumn{2}{c}{\textbf{FEVER}} \\ 
	\cmidrule{3-4} & & \textbf{L.A. Dev} & \textbf{L.A. Test} \\

        \toprule
        GEAR~\citep{zhou-etal-2019-gear}															& & 70.69 				& 71.60 \\
        DOMLIN~\citep{stammbach-neumann-2019-team}												& & 71.54				& - \\
        KGAT with RoBERTa\textsubscript{Base}~\citep{liu-etal-2020-fine}									& & 78.29 				& 74.07 \\
        Transformer-XH~\citep{zhao2020transformer-xh} 												& & 78.05 				& 72.39 \\
        RoBERTa\textsubscript{Base} Pairwise + MaxPool~\citepalias{tymoshenko-moschitti-2021-strong} 			& & 79.82 				& - \\
        RoBERTa\textsubscript{Base} Pairwise + WgtSum~\citepalias{tymoshenko-moschitti-2021-strong} 			& & 80.01 				& - \\

	\midrule
	DOMLIN++~\citep{20.500.11850/453826}														& & 77.48 				& 76.60 \\
        DREAM~\citep{zhong-etal-2020-reasoning}													& & 79.16				& \bld{76.85} \\
        \midrule

        RoBERTa\textsubscript{Base} Fixed (IE$_1$) 												& & 78.92 (0.46)		& 73.01 \\
        RoBERTa\textsubscript{Base} Fixed (AE$_1$) 												& & 78.85 (0.36)		& 72.94 \\
        RoBERTa\textsubscript{Base} Flexible (IE$_1$)												& & 79.25 (0.27)		& 73.56 \\

        \midrule

        \textbf{RoBERTa\textsubscript{Base} Fixed (IE$_1$) + MSPP (IE$_k$)} 							& & 81.21 (0.24)    		& 74.39 \\
        \textbf{RoBERTa\textsubscript{Base} Fixed (AE$_1$) + MSPP (AE$_k$)}   							& & 81.21 (0.16)    		& 74.13 \\
        \textbf{RoBERTa\textsubscript{Base} Flexible (IE$_1$) + MSPP (RE$_k$)}   							& & \bld{82.10} (0.10)    	& 75.36 \\
        \bottomrule

	\end{tabular}}
    \caption{\small Results on FEVER dev and test sets. We use the prediction heads IE$_1$ and AE$_1$ for fine-tuning and either IE$_k$, AE$_k$ or RE$_k$ for the MSPP pre-training. `-' means results are not reported in the original paper. We use bold formatting for our proposals and the best results.}
    \label{tab:joint_table_fact_verification}
\end{table*}

We present results on the Fact Verification task by comparing our models against many baselines from the literature. Table~\ref{tab:joint_table_fact_verification} reports the Label Accuracy on the development and test sets of the FEVER dataset.

Flexible models without our ad-hoc pre-training task reach remarkable performance on both the development and test set, obtaining a Label Accuracy higher than some previous state-of-the-art systems. On the other hand, when our models are further trained with the MSPP objective, we obtain state-of-the-art results among models based on a Base architecture (about 120M of parameters). In particular, by testing models on the Fever Shared Task Leaderboard, we obtain a significant gain of 1.29 points when using our Jointwise Flexible model with MSPP. On the development set, all of our Joint models outperform the baselines by a margin of at least 1.20 points in Label Accuracy.

Finally, notice that our Joint Flexible model achieves a score slightly below the actual state-of-the-art on FEVER: DREAM and DOMLIN++. However, those models use a better retrieval system (we are limited by the performance of the BERT-based Doc-IR) and larger models for the Fact Verification task. In particular, DREAM Fact Verification systems is based on an XLNet\textsubscript{Large}~\citep{yang2020xlnet} model while DOMLIN++ uses a RoBERTa\textsubscript{Large}~\citep{liu2019roberta} architecture. Both models contain about 360M parameters, which is much more than the 124M parameters of our Jointwise RoBERTa\textsubscript{Base}.

\section{Latency Analysis}

Although our Joint model allows for longer input sequences in the Transformer, it also decreases the number of forward passes required by an equivalent Pairwise Cross-Encoder. To provide a simplified analysis of the latency in AS2, let $L$ be the length of the longest sentence. In the case of the Pairwise Cross-Encoder, $k$ forward passes (one for each candidate $c_i$) of the Transformer are needed with a sequence length of $2L$. However, our Joint model only requires a single forward pass over the Transformer with an input length of $L (k+1)$ (one pass with the query and $k$ answer candidates).

Since the Transformer's Self-Attention has a quadratic relationship with the input sequence length, we can expect the inference time of our Joint model to be approximately $O((k+1)^2 L^2)$, and $O(4 L^2 k)$ for the Pairwise Cross-Encoder. Thus, the ratio between the inference time of our Joint model and the Pairwise Cross-Encoder is:
\begin{equation}
r = \frac{O((k+1)^2 L^2)}{O(4 L^2 k)} = O(k)
\end{equation}
However, $k$ and $L$ are usually small numbers and the Transformer contains also many affine transformations and an embedding layer that scale linearly in the input sequence length. We observe empirically that when fine-tuning on the AS2 datasets with a batch size of 32, there is only a slight increase in the average inference latency equal to 14.1\% for Fixed and 9\% for Flexible models. Flexible models are faster because they allow to shrink batches dynamically when the input examples are all shorter than the allowed maximum sequence length.

\chapter{Self-Supervised Objectives for AS2}
\label{cha:pairwise}

In this Chapter, we study pre-training sentence-level objectives specifically designed for Answer Sentence Selection. Annotating large datasets is time-consuming and expensive (because of the large set of possible answer candidates retrieved for each question). For this reason, we aim at pre-training Transformer models that could be adapted to various AS2 datasets even if they contain a reduced number of training examples.

We propose three novel self-supervised sentence-level objectives that could be used in pre-training along with MLM to gather paragraph-level semantics from single or multiple documents into the model. The objectives we propose do not need additional expensive labeled datasets (such as in \citet{garg2019tanda}), but leverage semi-structured knowledge from large pre-training corpora such as Wikipedia or the CommonCrawl. In order to align with the sentence-pair structure of the Answer Sentence Selection task, our pre-training objectives are specifically designed to function with a pair of input text sequences.

The first objective can be summarized as predicting whether two random spans of text are extracted from the same paragraph of a document. The second objective instead provides the model with a paragraph and a sentence and should predict whether the sentence belongs to that paragraph in the original document. Finally, the third objective tasks the model to predict whether two paragraphs are extracted from the same original document.

We provide a detailed overview of the Transformer models used in this Chapter in Section~\ref{sec:sota_lm}. Moreover, works related to the task we address here can be found in Section~\ref{sec:related_as2}, while for similar techniques that exploit the weak supervision of raw documents, the reader should refer to Section~\ref{sec:related_document_structure}.

By doing continuous pre-training starting from different pre-trained checkpoints such as RoBERTa~\citep{liu2019roberta} and ELECTRA~\citep{clark2020electra}, we show that raw data can be exploited to increase final accuracy on different AS2 datasets. The next Section examines the reason why common pre-training tasks are not effective at structuring information from large pre-training corpora. Next, we describe our alternative objectives in detail. Then, we present the experimental setting and the results of our objectives when compared with common Transformer baselines. To help future research, we report a set of negative results in Appendix~\ref{app:pairwise_negative}. Finally, we show how we achieve new state-of-the-art results on some datasets by continuously pre-training larger models with our newly proposed tasks.

\section{The issue with common Objectives}
\label{sec:issue_common_pairwise_objectives}

Several Transformer models, including BERT, RoBERTa, ELECTRA, ALBERT, DeBERTa, and others, are trained using token-level objectives, sometimes supplemented with sentence-level tasks. BERT, RoBERTa, ALBERT, and DeBERTa primarily employ the Masked Language Modeling (MLM) objective as their main training strategy, while ELECTRA focuses on the Token Detection task. BERT also incorporates the Next Sentence Prediction (NSP) objective to enhance performance, particularly in classification tasks involving two sentences, such as Natural Language Inferencing, AS2, or Textual Entailment. However, \citet{liu2019roberta} demonstrated that by extending the training duration, NSP can be omitted without compromising final accuracy. Additionally, \citet{lan2020albert} indicated that NSP is a trivial task with easily attainable high accuracies, resulting in weak error signals for the model. Consequently, they proposed Sentence Order Prediction (SOP) as a more challenging alternative to NSP.

In this Chapter, we demonstrate that token-level objectives are good at capturing short-range dependencies in the input text, but are not optimal for reasoning over inputs composed of separated spans of text, e.g., a question and an answer. Most masked tokens in MLM or replaced tokens in TD can be correctly predicted by looking at the surrounding context~\citep{zaheer2020bigbird, beltagy2020longformer}. Moreover, we show that NSP and SOP are not challenging enough to train the model at reasoning over two input spans of text and do not align well with the final downstream AS2 task. 

\section{Alternative Objectives for AS2}

Large sources of data such as Wikipedia, the CommonCrawl or the Pile are collections of documents, and each document is divided into a list of paragraphs by humans. The paragraphs of a single document generally describe the same argument from different perspectives. The objectives we propose in this Chapter exploit this weak supervision to create 3 self-supervised tasks that do not need manually annotated data.

We define pre-training corpus $\sC$ in the following way:

\begin{itemize}
\item the corpus is a set containing millions of documents: $\sC = \{ \mD_1, \dots, \mD_k \}$;
\item each document is a sequence of paragraphs: $\mD_i = [ P_1, \dots, P_n ]$, where $n$ is the number of paragraphs in $\mD_i$;
\item paragraphs are lists of sentences: $P_j = [ s_1, \dots, s_m ]$, where $m$ is the number of sentences in $P_j$.
\end{itemize}

\noindent We use the notation $s_{ab}$ to indicate the subsequence of a paragraph $P_j$ that starts from sentence $a$ and ends with sentence $b$, $s_{ab}= s_a \cdot \ldots \cdot s_b$ (we use ``$\cdot$''to denote text concatenation). Then, we define our objectives as follows.

\subsection{Spans in the Same Paragraph (SSP)}
\label{sec:ssp}

\begin{figure*}[t]
    \centering
    \includegraphics[width=0.9\textwidth]{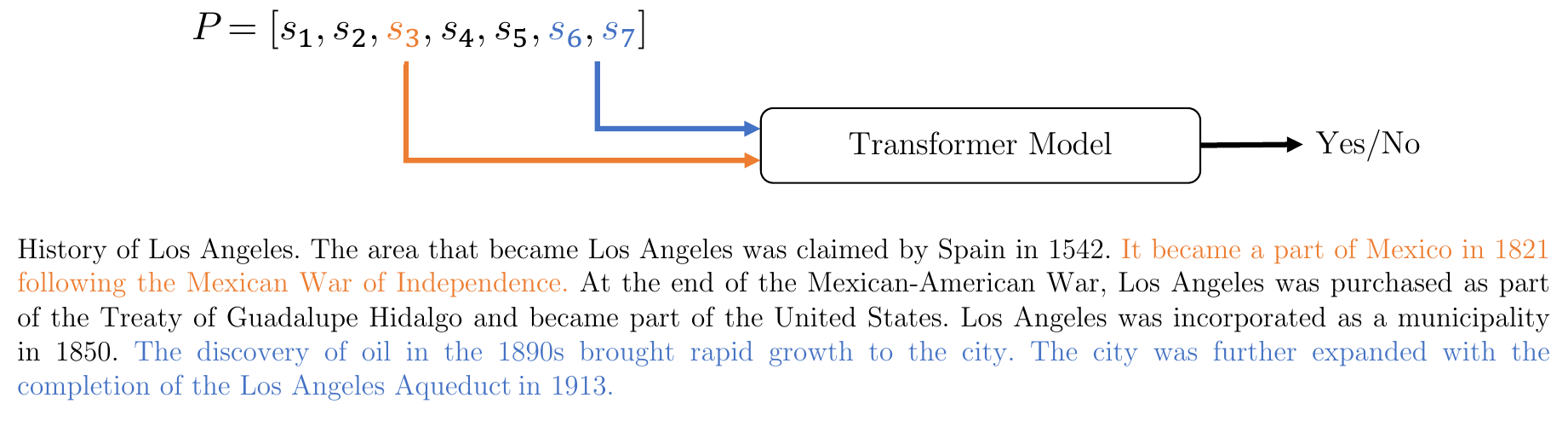}
    \caption{\small Example of SSP objective.}
    \label{fig:ssp}
\end{figure*}

The first task consists in detecting whether two spans of text $s_{ab}$ and $s_{cd}$ are extracted from the same paragraph $P_j \in \mD_i$. More formally, given two subsequences $s_{ab}, s_{cd}$ such that $s_{ab} \cap s_{cd} = \emptyset$, the task is to predict whether $s_{ab}, s_{cd} \in P_j$. We create positive example pairs by sampling two disjoint sequences $s_{ab}$ and $s_{cd}$ from the same paragraph $P_j$ of some document $\mD_i$. Then, we create $k_1$ hard negatives by sampling $s_{ab} \in P_j$ and $s_{cd} \in P_l$ from two different paragraphs in the same document, i.e $P_j, P_l \in \mD_i$ and $j \ne l$. This forces the model at recognizing that the general topic of the text spans is the same but described from different viewpoints. Finally, we create $k_2$ easy negatives by sampling $s_{ab} \in P_j$ and $s_{cd} \in P_l$ such that $P_j \in \mD_i, P_l \in \mD_f$ and $i \ne f$.

The task is designed to mimic the input of Transformer models for the Answer Sentence Selection task, which is composed of a question and
a candidate answer. The question corresponds to $s_{ab}$ while the answer is replaced by $s_{cd}$. Since $s_{ab}$ and $s_{cd}$ are two sequences of sentences in some paragraphs, we describe the sampling method we use to define their length. Questions in most cases are composed of a single sentence, with rare peaks of up to three sentences. Thus, we sample the length of $s_{ab}$ in the interval $[1, 3]$, with weighted probabilities of 70\%, 20\% and 10\%, respectively. Then, since most answers in the AS2 datasets are composed of one to five sentences, we randomly sample the length of $s_{cd}$ in $[1, 5]$ with probabilities of 14\% for 1 and 5, and 24\% for the other lengths.

Regarding the number of hard and easy negatives sampled for each positive pair, we try to obtain $k_1=2$ hard negatives from within the same paragraph and $k_2=2$ easy negatives from paragraphs in other documents. When there are not enough sentences in the paragraph to create $k_1$ hard negatives, we force $k_1 + k_2 = 4$ by sampling more easy negatives. An example of the SSP objective is depicted in Figure~\ref{fig:ssp}.

\subsection{Spans in Paragraph (SP)}

\begin{figure*}[b]
    \centering
    \includegraphics[width=0.9\textwidth]{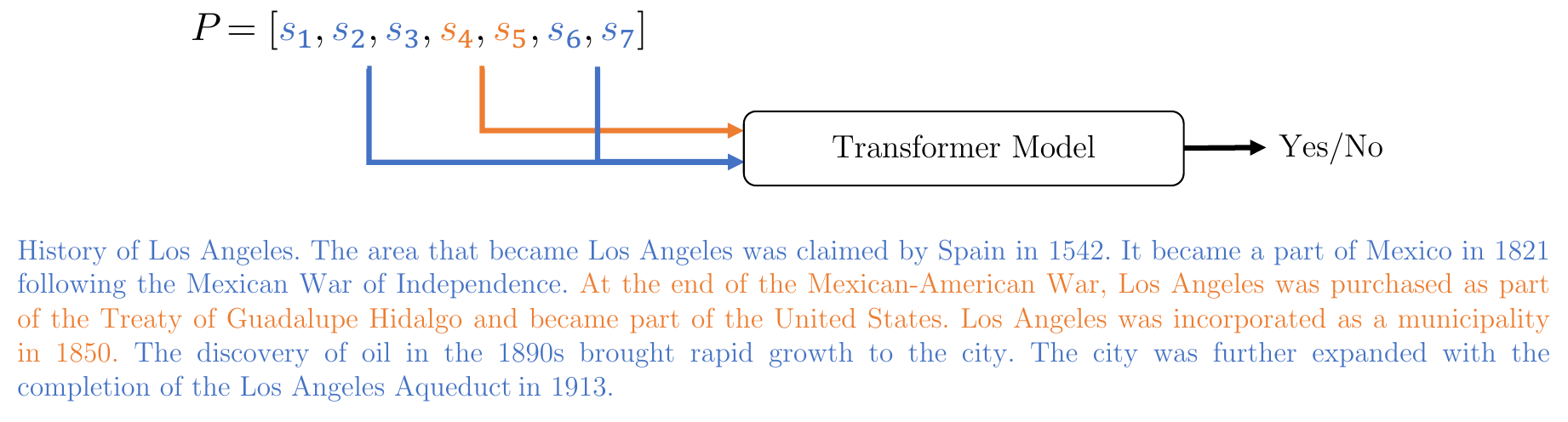}
    \caption{\small Example of SP objective.}
    \label{fig:sp}
\end{figure*}
\noindent
SP is a task that challenges the model at recognizing whether a span of text $s_{ab}$ is extracted from some paragraph $P_j \in \mD_i$. More specifically, we divide every paragraph $P_j$ in the corpus into two subsets $s_{ab}$ and $P_j \setminus s_{ab}$. Then, we create positive example pairs by sampling $s_{ab}$ and $P_j \setminus s_{ab}$, with $s_{ab} \in P_j$. For the hard negatives, we use the same $s_{ab} \in P_j$ but paired with other reduced paragraphs $P_l \setminus s_{cd}$ in the same document, such that $P_j, P_l \in \mD_i$ and $j \ne l$. This should force the model at recognizing that the topic is the same but addressed from different perspectives. Finally, we create easy negatives by pairing $s_{ab} \in P_j$ with reduced paragraphs $P_l \setminus s_{cd}$ sampled from other documents, i.e. $P_j \in \mD_i, P_l \in \mD_f$ such that $i \ne f$.

As in SSP, we define a sampling strategy for the length of $s_{ab}$. Since $s_{ab}$ represents the question when fine-tuning on the downstream task, we apply the same rule of SSP, by sampling the length of $s_{ab}$ in the interval $[1, 3]$, with weighted probabilities of 70\%, 20\% and 10\%. The second part is always the text that remains in paragraph $P_i$ after the split, and its length may vary considerably from one corpus to another. For example, Wikipedia has longer paragraphs than CC-News and OpenWebText on average, leading to longer pairs. We do not set any constraint on the length of the right part of the pair because the tokenizer will take care of truncating sequences that exceed the maximum sequence length allowed.

As in SSP, we sample up to $k_1=2$ hard negatives from within the same document and then we sample $k_2$ easy negatives until $k_1 + k_2 = 4$. An example of the SP objective is depicted in Figure~\ref{fig:sp}.

\subsection{Paragraphs in the Same Document (PSD)}

\noindent
The last new pre-training task we propose asks the model to classify whether two entire paragraphs are sampled from the same document. Formally, positive pairs are created by sampling two paragraphs $P_j$ and $P_l$ such that $P_j, P_l \in \mD_i$ and $j \ne l$. In PSD there is no distinction between easy and hard negatives because there is no way of creating negatives from within the same document. We create negative pairs by sampling $P_j$ and $P_l$ from different documents, i.e. $P_j \in \mD_i$ and $P_l \in \mD_f$ given $i \ne f$.

\begin{figure*}[hb]
    \centering
    \includegraphics[width=0.9\textwidth]{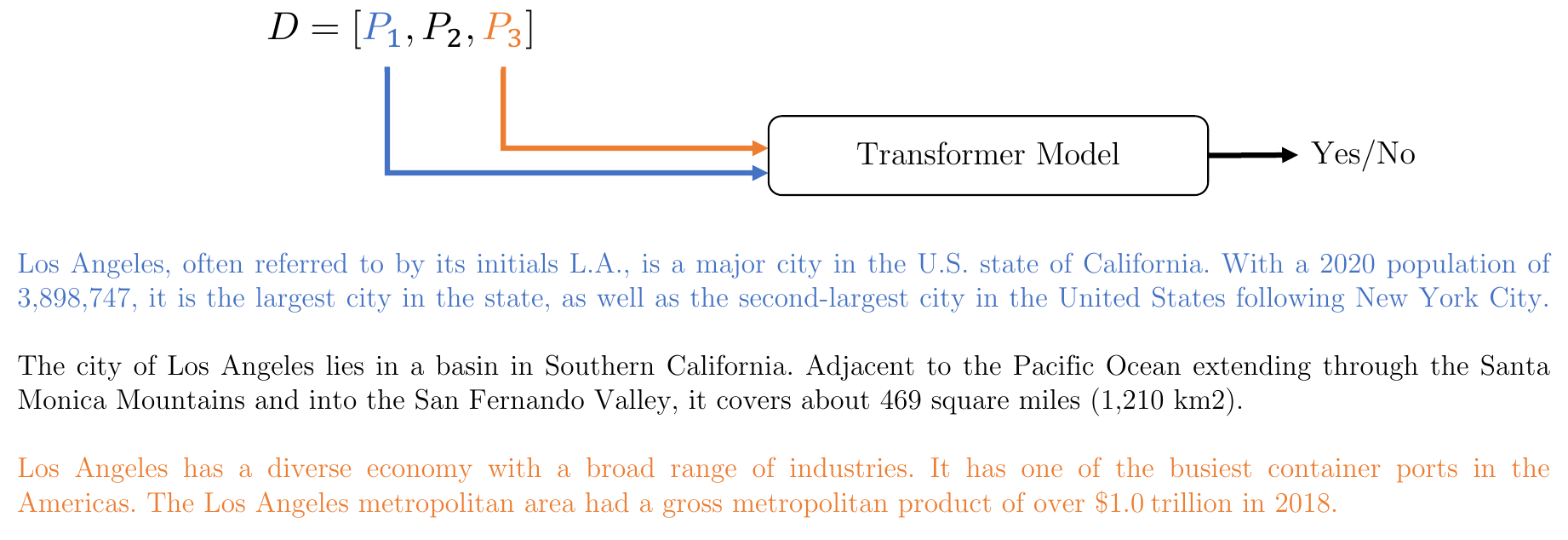}
    \caption{\small Example of PSD objective.}
    \label{fig:psd}
\end{figure*}

We do not crop paragraphs to reduce their length because this task will be performed by the tokenizer, by iteratively removing one token from the longest sequence in the pair until the total length is within the limit. For each positive pair, we create $k=4$ additional negative examples. An example of PSD is given in Figure~\ref{fig:psd}.

\section{Experiments}

In this Section, we describe the experimental setting used to evaluate the effectiveness of our pre-training objectives.

\subsection{Continuous Pre-Training}

\begin{table*}[t]
    \centering
    \resizebox{1.0\linewidth}{!}{%
    \begin{tabular}{lccccccl}
        \toprule
        \textbf{Model}				& \textbf{Steps}	& \textbf{LR}			& \textbf{BS}			& \textbf{MSL}		& \textbf{\# Tokens}		& \textbf{Effort}		& \textbf{Objectives}\\

        \toprule

    RoBERTa\textsubscript{Base} + SSP		& 400K		& $1 \cdot 10^{-4}$		& 4096				& 128				& 210B				& +2.5\%						& MLM (1.0) + SSP (1.0) \\
	RoBERTa\textsubscript{Base} + SP		& 200K		& $1 \cdot 10^{-4}$		& 4096				& 256				& 210B				& +5.0\%						& MLM (1.0) + SP (1.0) \\
	RoBERTa\textsubscript{Base} + PSD		& 200K		& $1 \cdot 10^{-4}$		& 4096				& 256				& 210B				& +5.0\%						& MLM (1.0) + PSD (1.0) \\
	RoBERTa\textsubscript{Base} + ALL		& 400K		& $1 \cdot 10^{-4}$		& 4096				& 128				& 210B				& +2.5\%						& MLM (1.0) + ALL (1.0) \\

	\midrule

        ELECTRA\textsubscript{Base} + SSP		& 400K		& $1 \cdot 10^{-4}$		& 4096				& 128				& 210B				& +5.0\% 						& MLM (1.0) + TD (50.0) + SSP (1.0) \\
	ELECTRA\textsubscript{Base} + SP		& 200K		& $1 \cdot 10^{-4}$		& 4096				& 256				& 210B				& +10.0\%						& MLM (1.0) + TD (50.0) + SP (1.0) \\
	ELECTRA\textsubscript{Base} + PSD		& 200K		& $1 \cdot 10^{-4}$		& 4096				& 256				& 210B 				& +10.0\%						& MLM (1.0) + TD (50.0) + PSD (1.0) \\
	ELECTRA\textsubscript{Base} + ALL		& 400K		& $1 \cdot 10^{-4}$		& 4096				& 128				& 210B				& +5.0\%						& MLM (1.0) + TD (50.0) + ALL (1.0) \\
	
	\midrule

	DeBERTa\textsubscript{Base} + ALL		& 100K		& $1 \cdot 10^{-4}$		& 4096				& 128				& 52B				& +1.25\%						& MLM (1.0) + SP (1.0) \\
	DeBERTa\textsubscript{Large} + ALL		& 100K		& $1 \cdot 10^{-4}$		& 4096				& 128				& 52B				& +1.25\%						& MLM (1.0) + ALL (1.0) \\
	
	\midrule

	DeBERTaV3\textsubscript{Base} + ALL		& 100K		& $1 \cdot 10^{-4}$		& 4096				& 128				& 52B				& +0.625\% 					& MLM (1.0) + TD (50.0) + ALL (1.0) \\
	DeBERTaV3\textsubscript{Large} + ALL	& 100K		& $1 \cdot 10^{-4}$		& 4096				& 128				& 52B				& +0.625\%					& MLM (1.0) + TD (50.0) + ALL (1.0) \\
        
        \bottomrule
    \end{tabular}
    }
    \caption{\small Hyper-parameters used for the continuous pre-training of several Transformer models with our specialized objectives. The ``Effort'' column indicates the amount of additional FLOPS required compared to the original pre-training. ``MSL'' is the maximum allowed input length. ``LR'' is the learning rate while ``BS'' refers to the training batch size. The last column describes the used training tasks along with their weight (in round brackets).}
    \label{tab:pairwise_pretraining_hparams}
\end{table*}

We do continuous pre-training starting from RoBERTa, ELECTRA, DeBERTa and DeBERTaV3 checkpoints. We apply all of our objectives to RoBERTa and ELECTRA, while for DeBERTa and DeBERTaV3, since they are more expensive to be continuously pre-trained, we apply only the objectives that show more consistent performance across all tasks, even though it may not always be the best. We also experiment by mixing the objectives and challenging the model to automatically identify the task and predict the correct label. We mix the datasets for the continuous pre-training at the example level, thus a batch can contain samples from both SSP, SP or PSD, with equal probability. The other pre-training parameters are the same as SSP. Those experiments are indicated with ``ALL'' in the following Sections.

To avoid improvements deriving from the usage of more data, we exploit the same corpora used for the original pre-training of the models. In particular, we do continuous pre-training with SSP, SP and PSD on English Wikipedia, BookCorpus, OpenWebText and CC-News. For ELECTRA, the BookCorpus is not included in the original pre-training, however, it accounts for less than 0.5\% of the training examples. After the pre-processing of the corpora with the techniques described above, we obtain about 420M examples for SP, 260M for SP and 200M for PSD.

In Table~\ref{tab:pairwise_pretraining_hparams}, we report the hyper-parameters used for each continuous pre-training experiment. For every model, we combine our objectives with the original tasks, i.e. MLM for RoBERTa and DeBERTa and TD for ELECTRA and DeBERTaV3. We found this beneficial since our long continuous pre-training may reduce models' generalization by only performing binary classification on the output embedding of the first token.

Regarding the length of the continuous pre-training, we also report the total number of tokens provided to each model. For reference, RoBERTa\textsubscript{Base} and DeBERTaV3\textsubscript{Base} were pre-trained over about 2000B tokens, ELECTRA\textsubscript{Base} over more than 500B tokens and DeBERTa\textsubscript{Base} over about 1000B tokens. DeBERTa\textsubscript{Large} and DeBERTaV3\textsubscript{Large} were trained over the same amount of tokens of the corresponding Base versions. For more details about the original pre-trained models, refer to Section~\ref{sec:sota_lm}. Notice that for DeBERTaV3 models, we couldn't exploit the original generators for Token Detection, as they were not released. We used DeBERTa\textsubscript{Small} and DeBERTaV3\textsubscript{Base} instead as the generators for DeBERTaV3\textsubscript{Base} and DeBERTaV3\textsubscript{Large}, respectively. We don't think this issue has noticeable effects on final accuracy. The column ``Effort'' of the table shows the effective amount of computational FLOPS required for our continuous pre-training compared to the original model training. This ratio is computed by taking into account the number of total training steps, the batch size and the maximum sequence length. Notice that the Attention Mechanism complexity is quadratic in the input length. Thus, by training with a limit of 128 or 256, we reduce the computational complexity, and thus training time, by $\frac{1}{16}$ and $\frac{1}{4}$ respectively (original models were all trained with a sequence length of 512). To further demonstrate that our achievements do not stem from this additional short pre-training, we compare models continuously pre-trained with and without our objectives in Section~\ref{sec:pairwise_only_mlm}.

Regarding the optimization and the hardware, we set $\beta_1 = 0.9$ and $\beta_2 = 0.99$, the weight decay to $0.01$ and we use a triangular learning rate with 10\% warmup steps. We trained all models on our machine (see Section~\ref{sec:machines} for more details), and each experiment required about 2.5 days of continuous pre-training.

\subsubsection{Comparison of objectives hardness}

Here we compare common sentence-level objectives with our proposals in terms of final accuracy. As explained before, we state that NSP~\citep{devlin-etal-2019-bert}, SOP~\citep{lan2020albert}, SSO~\citep{wang2019structbert} and other sentence-level tasks are trivial and do not provide meaningful signals to the model. In Table~\ref{tab:results_pairwise_pretraining} we report the accuracy on the development set after the continuous pre-training with our proposed objectives (SSP, SP and PSD) and the performance of other sentence-level tasks from the literature. Since we don't have access to the pre-training development sets of BERT, ALBERT and StructBERT, the performance of NSP, SOP and SSO are measured over a small random shard of OpenWebText and CC-News, which were not used in the pre-training of those two models. Thus, we expect slightly higher performance on the original data because here they are tested out of domain.

The results indicate that performing sentence-level predictions is more challenging for SSP and SP compared to NSP, SOP and SSO. Surprisingly, solving PSD proves to be even more difficult, despite having access to two large paragraphs for predictions, unlike SSP and SP, which only rely on two small text spans. Notably, we find it intriguing that the accuracy of SSP and SP is similar. This similarity can be attributed to the fact that the two tasks are not significantly different in practice, considering that the average paragraph length in the pre-training corpora is 1.4 sentences. Consequently, after filtering the paragraphs that are too small, it frequently occurs that the remaining portion of the paragraph in SP, after removing the text span, is merely another short text span, similar to the case in SSP.

\begin{table*}[t]
    \centering
    \resizebox{0.5\linewidth}{!}{
    \begin{tabular}{lccc}
    \toprule
    \textbf{Model} & \textbf{Accuracy} & \textbf{F1} \\
    \toprule
    RoBERTa\textsubscript{Base} + SSP      		& 91.8   & 83.1 \\
    ELECTRA\textsubscript{Base} + SSP      		& 90.4   & 79.9 \\
    \midrule
    RoBERTa\textsubscript{Base} + SP     		& 91.3   & 83.3 \\
    ELECTRA\textsubscript{Base} + SP     		& 89.9   & 80.1 \\
    \midrule
    RoBERTa\textsubscript{Base} + PSD     		& 83.5   & 61.4 \\
    ELECTRA\textsubscript{Base} + PSD     		& 82.3   & 57.1 \\
    \midrule
    BERT\textsubscript{Base}~\citep{devlin-etal-2019-bert} (NSP) 		& 96.9 	& 97.1 \\
    ALBERT\textsubscript{Base}~\citep{lan2020albert} (SOP) 		& 93.7 	& 94.7 \\
    StructBERT~\citep{wang2019structbert} (SSO) 	& 92.3 	& 93.4 \\
    \bottomrule
    \end{tabular}
    }
    \caption{\small Accuracy and F1-score of several objectives on the pre-training validation set.}
    \label{tab:results_pairwise_pretraining}
\end{table*}

\subsection{Fine-Tuning}

To demonstrate the superiority of our approach, we fine-tune all the models listed in Table~\ref{tab:pairwise_pretraining_hparams} on 6 different datasets for Answer Sentence Selection. Specifically, we use ASNQ, WikiQA, TREC-QA, WQA, NewsAS2 and TriviaAS2, which are all described in Section~\ref{cha:datasets}. We use the same hyper-parameters applied in the continuous pre-training apart from the learning rate, the batch size, the number of warmup steps of the triangular learning rate and the total number of epochs. In Table~\ref{tab:pairwise_as2_hparams} we provide a summary of the search space of all hyper-parameters divided by dataset. For every experiment, the number of warmup steps is computed to match the number of training steps in the first epoch. We optimize the hyper-parameters using the development set and using only the baselines, thus the best hyper-parameters for the models continuously pre-trained with our objectives may vary slightly. We do not conduct an extensive grid search because it is too expensive.

Performance of each model is measured with common metrics for AS2: Mean Average Precision, Mean Reciprocal Rank and Precision@1, as in related works~\citep{Lauriola2021, garg2019tanda}. In particular, in the description of the results we focus on Precision@1 because this is the rate at which a correct answer is returned to the user.

\begin{table*}[t]
    \centering
    \resizebox{1.0\linewidth}{!}{%
    \begin{tabular}{lcccccc}
        \toprule
        \textbf{Parameter}		& \textbf{ASNQ}					& \textbf{WikiQA \& TREC-QA}		& \textbf{WQA}				& \textbf{NewsAS2 \& TriviaAS2} \\

        \toprule

	Batch Size			& $\{ 1024, 2048 \} $					& $\{ 16, 32, 64, 128 \} $			& $\{ 64, 128, 256, 512 \} $	& $\{ 128, 256, 512 \} $ \\
	Learning Rate			& $\{ 1 \cdot 10^{-5}, 2 \cdot 10^{-5} \}$	& $\{ 5 \cdot 10^{-6}, 1 \cdot 10^{-5}, 2 \cdot 10^{-5} \}$	& $\{ 5 \cdot 10^{-6}, 1 \cdot 10^{-5}, 2 \cdot 10^{-5} \}$ & $\{ 1 \cdot 10^{-5}, 2 \cdot 10^{-5} \}$ \\
        Max epochs			& 6								& 40							& 20						& 10 \\
        \bottomrule
    \end{tabular}
    }
    \caption{\small Hyper-parameters search space for the fine-tuning over several AS2 datasets.}
    \label{tab:pairwise_as2_hparams}
\end{table*}
\FloatBarrier

\section{Results}

\begin{table*}[t]
    \centering
    \resizebox{\linewidth}{!}{
    \begin{tabular}{l c@{\hskip 6pt}c@{\hskip 6pt}c @{\hskip 3pt}c c@{\hskip 6pt}c@{\hskip 6pt}c @{\hskip 3pt}c c@{\hskip 6pt}c@{\hskip 6pt}c}
    \toprule
\multirow{2}{*}{\textbf{Model}} & \multicolumn{3}{c}{\textbf{ASNQ}} & &\multicolumn{3}{c}{\textbf{WikiQA}} & & \multicolumn{3}{c}{\textbf{TREC-QA}} \\

	\cmidrule{2-4} \cmidrule{6-8} \cmidrule{10-12}	& \textbf{MAP} & \textbf{MRR} & \textbf{P@1}				& & \textbf{MAP} & \textbf{MRR} & \textbf{P@1}			& & \textbf{MAP} & \textbf{MRR} & \textbf{P@1} \\

\toprule

T\textsc{and}A RoBERTa\textsubscript{Base} $^\clubsuit$					& - & - & -											& & 88.5 (0.8) & 89.9 (0.8) & 83.0 (1.3)					& & 90.1 (0.4) & 94.1 (0.6) & 89.7 (0.7) \\
CT RoBERTa\textsubscript{Base}~\citepalias{soldaini-moschitti-2020-cascade}	& 66.3 & 65.4 & 53.2  								& & 89.9 & 91.0 & -		            						& & 92.4 & 96.7 & - \\
T\textsc{and}A RoBERTa\textsubscript{Base}~\citep{garg2019tanda}			& - & - & -											& & 88.9 & 90.1 & -									& & 91.4 & 95.2 & - \\
RLAS-BIABC~\citep{rlas2022hamid}									& - & - & -  										& & 88.8 (3.6) & 89.1 (1.7) & -		            				& & 89.8 (1.5) & 90.6 (9.2) & - \\
\midrule

RoBERTa\textsubscript{Base}~\citep{liu2019roberta}						& 66.9 (0.1) & 73.1 (0.1) & 61.8 (0.1)  					& & 85.8 (1.3) & 87.2 (1.3) & 78.3 (2.8)            				& & 89.7 (0.7) & 94.4 (1.1) & 90.0 (1.9) \\

\textbf{RoBERTa\textsubscript{Base} + SSP}							& \und{68.1} (0.2) & \bund{74.5} (0.3) & \bund{64.1} (0.3)  	& & \bund{88.7} (0.3) & \bund{89.9} (0.4) & \bund{82.9} (0.7)	& & 89.3 (0.7) & 93.6 (0.6) & 88.5 (1.2) \\
\bottomrule
\end{tabular}}
    \caption{\small Results of our RoBERTa\textsubscript{Base} and ELECTRA\textsubscript{Base} models on ASNQ, WikiQA and TREC-QA. We report several baselines such as T\textsc{and}A, Cascade Transformer (CT) and RLAS-BIABC. $\clubsuit$: our versions of T\textsc{and}A. We use bold formatting for our proposals and the best results. We underline statistically significant improvements over either RoBERTa\textsubscript{Base} or ELECTRA\textsubscript{Base}.}
    \label{tab:pairwise_as2_base_1_1}
\end{table*}

\begin{table*}[t]
    \centering
    \resizebox{\linewidth}{!}{
    \begin{tabular}{l c@{\hskip 6pt}c@{\hskip 6pt}c @{\hskip 3pt}c c@{\hskip 6pt}c@{\hskip 6pt}c @{\hskip 3pt}c c@{\hskip 6pt}c@{\hskip 6pt}c}
    \toprule
\multirow{2}{*}{\textbf{Model}} & \multicolumn{3}{c}{\textbf{ASNQ}} & &\multicolumn{3}{c}{\textbf{WikiQA}} & & \multicolumn{3}{c}{\textbf{TREC-QA}} \\

	\cmidrule{2-4} \cmidrule{6-8} \cmidrule{10-12}	& \textbf{MAP} & \textbf{MRR} & \textbf{P@1}				& & \textbf{MAP} & \textbf{MRR} & \textbf{P@1}			& & \textbf{MAP} & \textbf{MRR} & \textbf{P@1} \\

\toprule

DeBERTa\textsubscript{Base}~\citep{he2020deberta}					& 68.2 (0.1) & 74.4 (0.1) & 64.4 (0.3)  					& & 88.0 (2.1) & \bld{89.3} (1.9) & 81.6 (3.1)            			& & 89.9 (1.2) & 94.1 (1.2) & 89.7 (2.5) \\
\textbf{DeBERTa\textsubscript{Base} + ALL}							& \bund{69.9} (0.2) & \bund{76.3} (0.2) & \bund{66.9} (0.3) 	& & \bld{88.2} (1.0) & \bld{89.3} (0.9) & \bld{82.4} (1.7)  		& & \bld{91.1} (0.9) & \bld{94.6} (1.1) & \bld{90.7} (2.2) \\

\toprule

DeBERTaV3\textsubscript{Base}~\citep{he2021deberta}					& 70.5 (0.1) & 77.1 (0.1) & 67.8 (0.2) 					& & 88.9 (2.3) & 90.1 (2.1) & 83.0 (3.6)					& & 90.0 (0.3) & 93.2 (0.6) & 89.1 (0.8) \\
\textbf{DeBERTaV3\textsubscript{Base} + ALL}							& \bund{71.0} (0.2) & \bld{77.2} (0.2) & \bund{68.2} (0.3) 		& & \bld{89.5} (1.4) & \bld{90.7} (1.5) & \bld{84.3} (2.2)           	& & \bund{91.3} (0.9) & \bld{94.1} (1.2) & \bund{91.5} (1.2) \\

\bottomrule
\end{tabular}}
    \caption{\small Results of our DeBERTa\textsubscript{Base} and DeBERTaV3\textsubscript{Base} models on ASNQ, WikiQA and TREC-QA. We underline statistically significant improvements over the corresponding baselines.}
    \label{tab:pairwise_as2_base_1_2}
\end{table*}
\begin{table*}[t]
    \centering
    \resizebox{\linewidth}{!}{
    \begin{tabular}{l c@{\hskip 6pt}c@{\hskip 6pt}c @{\hskip 3pt}c c@{\hskip 6pt}c@{\hskip 6pt}c @{\hskip 3pt}c c@{\hskip 6pt}c@{\hskip 6pt}c}
    \toprule
\multirow{2}{*}{\textbf{Model}} 					& \multicolumn{3}{c}{\textbf{WQA}} 						& &\multicolumn{3}{c}{\textbf{NewsAS2}} 					& & \multicolumn{3}{c}{\textbf{TriviaAS2}} \\

	\cmidrule{2-4} \cmidrule{6-8} \cmidrule{10-12}	& \textbf{MAP} & \textbf{MRR} & \textbf{P@1}				& & \textbf{MAP} & \textbf{MRR} & \textbf{P@1}			& & \textbf{MAP} & \textbf{MRR} & \textbf{P@1} \\

    \toprule

T\textsc{and}A RoBERTa\textsubscript{Base} $^\clubsuit$			& 69.1 (0.1) & 78.0 (0.2) & 66.2 (0.2)						& & 82.3 (0.2) & 85.1 (0.3) & 76.1 (0.5)					& & 77.3 (0.5) & 83.0 (0.6) & 75.5 (0.6) \\
CT RoBERTa\textsubscript{Base}~\citepalias{soldaini-moschitti-2020-cascade}	& 57.8 & 76.9 & 67.5  								& & - & - & -		            							& & - & - & - \\

\midrule

RoBERTa\textsubscript{Base}~\citep{liu2019roberta}				& 68.4 (0.1) & 77.3 (0.2) & 65.4 (0.2)  					& & 82.4 (0.2) & 85.2 (0.3) & 76.4 (0.6)            				& & 76.9 (0.6) & 82.2 (0.5) & 73.1 (0.5) \\

\textbf{RoBERTa\textsubscript{Base} + SSP}					& \und{69.2} (0.1) & \und{77.7} (0.1) & \und{65.7} (0.2)  	 	& & 82.3 (0.3) & 85.3 (0.3) & 76.5 (0.4)           				& & \bld{77.5} (0.3) & \bld{82.7} (0.4) & \bund{74.3} (0.6) \\
\textbf{RoBERTa\textsubscript{Base} + SP} 					& \und{69.4} (0.1) & \und{78.0} (0.2) & \und{66.0} (0.4) 		& & \bld{82.7} (0.4) & \bld{85.6} (0.4) & \bld{77.2} (0.7)      		& & 77.1 (0.2) & \bld{82.7} (0.3) & \und{74.0} (0.5) \\
\textbf{RoBERTa\textsubscript{Base} + PSD}					& \und{69.4} (0.1) & \und{78.0} (0.1) & \und{66.0} (0.2)		& & 82.4 (0.2) & 85.0 (0.2) & 76.3 (0.3)            				& & 76.0 (0.1) & 81.5 (0.1) & 72.2 (0.1) \\
\bottomrule
\end{tabular}}
    \caption{\small Results of our RoBERTa\textsubscript{Base} and ELECTRA\textsubscript{Base} models on WQA, NewsAS2 and TriviaAS2. $\clubsuit$: our versions of T\textsc{and}A. We use bold formatting for our proposals and the best results. We underline statistically significant improvements over either RoBERTa\textsubscript{Base} or ELECTRA\textsubscript{Base}.}
    \label{tab:pairwise_as2_base_2_1}
\end{table*}

\begin{table*}[t]
    \centering
    \resizebox{\linewidth}{!}{
    \begin{tabular}{l c@{\hskip 6pt}c@{\hskip 6pt}c @{\hskip 3pt}c c@{\hskip 6pt}c@{\hskip 6pt}c @{\hskip 3pt}c c@{\hskip 6pt}c@{\hskip 6pt}c}
    \toprule
\multirow{2}{*}{\textbf{Model}} 					& \multicolumn{3}{c}{\textbf{WQA}} 						& &\multicolumn{3}{c}{\textbf{NewsAS2}} 					& & \multicolumn{3}{c}{\textbf{TriviaAS2}} \\

	\cmidrule{2-4} \cmidrule{6-8} \cmidrule{10-12}	& \textbf{MAP} & \textbf{MRR} & \textbf{P@1}				& & \textbf{MAP} & \textbf{MRR} & \textbf{P@1}			& & \textbf{MAP} & \textbf{MRR} & \textbf{P@1} \\

    \toprule

DeBERTa\textsubscript{Base}~\citep{he2020deberta}		& 69.5 (0.0) & 78.5 (0.2) & 66.6 (0.3)  					& & \bld{82.9} (0.5) & 85.5 (0.5) & 77.0 (0.8)            		& & 77.8 (0.2) & 83.3 (0.4) & 75.1 (0.4) \\
\textbf{DeBERTa\textsubscript{Base} + ALL}				& \bund{69.7} (0.1) & \bld{78.6} (0.1) & \bund{68.0} (0.2)  		& & \bld{82.9} (0.3) & \bld{85.7} (0.4) & \bld{77.3} (0.6)      	& & \bund{79.5} (0.3) & \bund{84.2} (0.5) & \bund{76.5} (0.8) \\

\midrule

DeBERTaV3\textsubscript{Base}~\citep{he2021deberta}		& 70.7 (0.2) & 78.9 (0.1) & 67.8 (0.1)	 					& & 82.8 (0.4) & 85.6 (0.3) & 77.2 (0.5)				& & 80.4 (0.2) & 85.1 (0.4) & 77.2 (0.5) \\
\textbf{DeBERTaV3\textsubscript{Base} + ALL}				& \bund{71.1} (0.1) & \bund{79.4} (0.1) & \bund{68.4} (0.1)	& & \bld{83.1} (0.2) & \bld{85.8} (0.2) & \bld{77.8} (0.6)     	& & \bund{81.1} (0.3) & \bund{86.0} (0.6) & \bund{79.5} (0.8) \\

\bottomrule
\end{tabular}}
    \caption{\small Results of our DeBERTa\textsubscript{Base} and DeBERTaV3\textsubscript{Base} models on WQA, NewsAS2 and TriviaAS2. We underline statistically significant improvements over the corresponding baselines.}
    \label{tab:pairwise_as2_base_2_2}
\end{table*}

We divide results across the different models we used to improve readability. We begin by describing results on Base models and then we show the accuracy of Large architectures on the same tasks. Then, we demonstrate empirically that the improvements shown in this Chapter do not stem from the additional pre-training but rather from our harder tasks. Lastly, we offer insights into the reasons behind the varying performance of different objectives on different datasets. In the tables, the statistical significance is computed always with respect to the vanilla model of each block. We report results from the original papers when available, otherwise we fine-tuned the models on the different tasks.

\subsection{Base Models}

The results over Base models are reported in Tables~\ref{tab:pairwise_as2_base_1_1} and~\ref{tab:pairwise_as2_base_1_2} for ASNQ, WikiQA and TREC-QA, and in Tables~\ref{tab:pairwise_as2_base_2_1} and~\ref{tab:pairwise_as2_base_2_2} for WQA, NewsAS2 and TriviaAS2. Results are split across several tables for better readability. This Section proceeds by describing the results for each objective we propose.

\paragraph{SSP}

Our Spans in the Same Paragraph objective improves models' accuracy on all tasks compared to the baselines, apart from the combination with RoBERTa and ELECTRA on TREC-QA. The advantages are more marked on small datasets, which confirms that our SSP objective is particularly effective when fine-tuning data is scarce. More on this in Section~\ref{sec:results_pairwise_fewshot}. For example, SSP with either RoBERTa or ELECTRA improves the Precision@1 by 2.9 points on ASNQ and by 4.6 points on WikiQA. In the second table, SSP shows minor improvements in Precision@1 on WQA, NewsAS2 and TriviaAS2 of 0.6, 0.2 and 1.9 points respectively.

\paragraph{SP}

The Spans in Paragraph objective provides significant improvements across all 6 datasets and combined with every architecture. For example, when applied to RoBERTa or ELECTRA, it improves the Precision@1 by up to 3.6 points on ASNQ, 4.7 on WikiQA and 1.4 on TriviaAS2. As before, the general pattern shows discrete improvements on large datasets and very high gains on small datasets such as WikiQA. We believe this is the most flexible objective, and we empirically confirm this claim, because it is composed of short spans of text that mimic the question and paragraphs gathered from multiple documents thst mimic the answer.

\paragraph{PSD}

Models trained with Paragraph is the Same Document outperform the baselines on all datasets apart from the combination with ELECTRA on TREC-QA\footnote{The results of ELECTRA+PSD on TREC-QA are significantly lower than the average, we suspect there was an issue with the model initialization for this experiment.}. The improvements are consistent compared to the baselines, however, the average gains are smaller when compared with SSP or SP. We suspect that this is the consequence of PSD working at a higher level, by performing predictions over entire paragraphs, which may not align optimally with the AS2 task. Among the most significant results, we cite the 3.1 points of improvements in Precision@1 on ASNQ and the gains of 0.9 and 1.3 on WQA and TriviaAS2 respectively.

\paragraph{ALL}

In this experiment, we show the results of combining all the proposed objectives for a single continuous pre-training. When applied to RoBERTa, ELECTRA and DeBERTa, the results show improvements across all datasets and more agnostic results to the target tasks, which means that this objective should be the choice when testing on new datasets.
Regarding the two DeBERTa architectures, both models augmented with the combination of SSP, SP and PSD outperform the baselines by a significant margin on all datasets. In particular, we report the gains in Precision@1 of 2.5 points on ASNQ, 2.4 on TREC-QA and 2.3 points on TriviaAS2. \\

\noindent Finally, notice that we match the performance of T\textsc{and}A on several datasets such as WikiQA, WQA and NewsAS2, even outperforming it in some scenarios. T\textsc{and}A uses ASNQ to transfer knowledge about the Answer Sentence Selection task before fine-tuning on smaller datasets, leading to higher final accuracy. We match the performance of T\textsc{and}A in many experiments while using only unlabeled data that do not require expensive annotation processes. We compare our models also with the Cascade Transformer~\citep{soldaini-moschitti-2020-cascade}. Notice that this technique performs an additional training step over ASNQ before fine-tuning over the target datasets, as T\textsc{and}A does. We surpass the Cascade Transformer by a large margin on ASNQ with all of our continuously pre-trained models. Moreover, we surpass the performance of RLAS-BIABC with our DeBERTaV3\textsubscript{Base}~+~ALL model on both WikiQA and TREC-QA by a significant margin of 0.7 and 1.5 MAP points, respectively.

\FloatBarrier

\subsection{Large Models}
\label{sec:results_pairwise_large}

\begin{table*}[t]
    \centering
    \resizebox{\linewidth}{!}{
    \begin{tabular}{l c@{\hskip 6pt}c@{\hskip 6pt}c @{\hskip 3pt}c c@{\hskip 6pt}c@{\hskip 6pt}c @{\hskip 3pt}c c@{\hskip 6pt}c@{\hskip 6pt}c}
    \toprule
\multirow{2}{*}{\textbf{Model}} & \multicolumn{3}{c}{\textbf{ASNQ}} & &\multicolumn{3}{c}{\textbf{WikiQA}} & & \multicolumn{3}{c}{\textbf{TREC-QA}} \\

	\cmidrule{2-4} \cmidrule{6-8} \cmidrule{10-12}	& \textbf{MAP} & \textbf{MRR} & \textbf{P@1}				& & \textbf{MAP} & \textbf{MRR} & \textbf{P@1}			& & \textbf{MAP} & \textbf{MRR} & \textbf{P@1} \\

    \toprule

T\textsc{and}A BERT\textsubscript{Large}~\citep{garg2019tanda}			& - & - & -											& & 90.4 & 91.2 & -									& & 91.2 & 96.7 & - \\
T\textsc{and}A RoBERTa\textsubscript{Large}~\citep{garg2019tanda}		& - & - & -											& & 92.0 & 93.3 & -									& & 94.3 & 97.4 & - \\

\midrule

DeBERTa\textsubscript{Large}~\citep{he2020deberta}					& 72.6 (0.2) & 78.5 (0.2) & 68.8 (0.3)  					& & 89.5 (0.3) & 90.8 (0.4) & 84.0 (0.7)            				& & 91.9 (0.2) & 94.7 (1.1) & 90.7 (1.7) \\
\textbf{DeBERTa\textsubscript{Large} + ALL}							& \bund{73.4} (0.1) & \bund{78.9} (0.1) & \bund{69.5} (0.1)	& & \und{90.1} (0.1) & 91.2 (0.2) & \und{85.1} (0.5)          		& & 91.8 (1.1) & 94.6 (0.7) & 90.7 (1.7) \\

\textbf{T\textsc{and}A DeBERTa\textsubscript{Large} + ALL}				& - & - & -						 					& & \bund{91.4} (1.0) & \bund{92.4} (0.8) & \bund{86.8} (1.6)	& & \bld{92.3} (0.6) & \bld{94.9} (1.6) & \bld{93.1} (1.7) \\

\midrule

DeBERTaV3\textsubscript{Large}~\citep{he2021deberta}					& 73.8 (0.1) & 79.0 (0.1) & 69.5 (0.1) 					& & 90.9 (1.2) & 92.0 (1.1) & 86.2 (2.0)					& & 91.7 (0.7) & 94.7 (0.5) & 91.8 (0.8) \\
\bottomrule
\end{tabular}}
    \caption{\small Results of our Large models on ASNQ, WikiQA and TREC-QA. We use bold formatting for our proposals and the best results. We underline statistically significant improvements over either DeBERTa\textsubscript{Large} or DeBERTaV3\textsubscript{Large}.}
    \label{tab:pairwise_as2_large_1}
\end{table*}
\begin{table*}[t]
    \centering
    \resizebox{\linewidth}{!}{
    \begin{tabular}{l c@{\hskip 6pt}c@{\hskip 6pt}c @{\hskip 3pt}c c@{\hskip 6pt}c@{\hskip 6pt}c @{\hskip 3pt}c c@{\hskip 6pt}c@{\hskip 6pt}c}
    \toprule
\multirow{2}{*}{\textbf{Model}} 					& \multicolumn{3}{c}{\textbf{WQA}} 								& &\multicolumn{3}{c}{\textbf{NewsAS2}} 					& & \multicolumn{3}{c}{\textbf{TriviaAS2}} \\

	\cmidrule{2-4} \cmidrule{6-8} \cmidrule{10-12}	& \textbf{MAP} & \textbf{MRR} & \textbf{P@1}						& & \textbf{MAP} & \textbf{MRR} & \textbf{P@1}			& & \textbf{MAP} & \textbf{MRR} & \textbf{P@1} \\

    \toprule

DeBERTa\textsubscript{Large}~\citep{he2020deberta}		& 72.6 (0.0) & 80.8 (0.1) & 69.7 (0.1)  					& & \bld{84.1} (0.4) & \bld{86.7} (0.3) & 78.6 (0.5)            		& & 82.7 (0.3) & 86.8 (0.5) & 80.2 (0.9) \\
\textbf{DeBERTa\textsubscript{Large} + ALL}					& \bund{73.2} (0.0) & \bund{81.3} (0.1) & \bund{70.3} (0.1)  	& & \bld{84.1} (0.1) & \bld{86.7} (0.2) & \bld{78.9} (0.4)           	& & \bund{83.2} (0.3) & \bund{87.5} (0.3) & \bund{81.4} (0.5) \\

\midrule

DeBERTaV3\textsubscript{Large}~\citep{he2021deberta}		& 72.4 (0.2) & 80.3 (0.2) & 69.5 (0.3)	 					& & \bld{85.0} (0.1) & 87.5 (0.1) & 80.2 (0.3)				& & 82.5 (0.4) & 87.0 (0.7)  & 80.5 (1.1) \\
\textbf{DeBERTaV3\textsubscript{Large} + ALL}					& \bund{73.5} (0.2) & \bund{81.3} (0.3) & \bund{70.9} (0.5)	& & \bld{85.0} (0.3) & \bld{87.6} (0.3) & \bld{80.5} (0.4)   		& & \bund{84.3} (0.4)  & \bund{88.4} (0.3) & \bund{82.7} (0.6) \\

\bottomrule
\end{tabular}}
    \caption{\small Results of our Large models on WQA, NewsAS2 and TriviaAS2. We underline statistically significant improvements over the corresponding baselines.}
    \label{tab:pairwise_as2_large_2}
\end{table*}

In Tables~\ref{tab:pairwise_as2_large_1} and~\ref{tab:pairwise_as2_large_2} we report the results of combining larger models with our specialized objectives. On ASNQ, DeBERTa\textsubscript{Large} and DeBERTaV3\textsubscript{Large} provide an improvement of 0.7 and 1.7 points in Precision@1, respectively. To the best of our knowledge, DeBERTaV3\textsubscript{Large}~+~ALL is the actual state-of-the-art on ASNQ. On the two small datasets (WikiQA and TREC-QA), the improvements are smaller than with the Base models, because the baselines are already very capable of adapting to the AS2 task using less data, thanks to their larger size. Nevertheless, on TREC-QA and WikiQA we achieve the highest results to date among methods that do not exploit additional data, and the second place behind a T\textsc{and}A model based on RoBERTa\textsubscript{Large} otherwise. As a last experiment, we combined our continuously pre-trained checkpoints with T\textsc{and}A, by performing a transfer step over ASNQ with a batch size of $2048$ and learning rate of $10^{-5}$ (other hyper-parameters are the same of Table~\ref{tab:pairwise_as2_hparams}), followed by the fine-tuning over WikiQA and TREC-QA. Our T\textsc{and}A DeBERTaV3\textsubscript{Large}~+~ALL achieves state-of-the-art accuracy on both WikiQA and TREC-QA, with an impressive MAP of 92.7 and 95.4, respectively.

Regarding WQA, NewsAS2 and TriviaAS2, we show competitive performance on all benchmarks, outperforming the baselines in the best scenario by 1.4 points of P@1 on ASNQ, 0.3 on NewsAS2 and 2.2 points on TriviaAS2.

\subsection{Only MLM Continuous Pre-Training}
\label{sec:pairwise_only_mlm}

\input{tables/pairwise_as2_base_comparison}
\noindent
In this Section, we compare models continuously pre-trained using only the MLM objective and with the combination of MLM and our proposed tasks. We show that SSP, SP and PSD are of fundamental importance to reach the highest accuracy on Answer Selection datasets and that improvements should not be attributed to the additional data. Table~\ref{tab:pairwise_as2_base_comparison} summarizes the results on ASNQ, WikiQA, TREC-QA and WQA of our continuously pre-trained models compared with similar models trained over the same data but using only the MLM objective.

Our proposals consistently outperform models trained solely with the MLM objective across all tasks, even outperforming the baseline on smaller datasets. This can be attributed to the negative impact of resetting optimizer and scheduler states when training resumes. Additionally, adapting partitioned schedulers and optimizer states to new training configurations with a different number of training devices would have been challenging, even if the original complete checkpoints were available.

\FloatBarrier
\subsection{Few-Shot Setting}
\label{sec:results_pairwise_fewshot}

\input{tables/pairwise_as2_base_fewshot_1}
\input{tables/pairwise_as2_base_fewshot_2}

This Section analyzes the performance of our models in a few-shot scenario, demonstrating how our specialized continuous pre-training using unlabeled data can help when target data is scarce. 

In the Tables~\ref{tab:pairwise_as2_base_fewshot_1_1},~\ref{tab:pairwise_as2_base_fewshot_1_2},~\ref{tab:pairwise_as2_base_fewshot_2_1} and~\ref{tab:pairwise_as2_base_fewshot_2_2} we show the performance of our continuously pre-trained models in the few-shot setting. In this experiment, we reduce the size of the training sets by 1000 times for ASNQ and by 20 and 50 times for WikiQA and TREC-QA. Regarding WQA, NewsAS2 and TriviaAS2, the reduction factor is set to 150 for the former and 1000 for the last two. This leaves us with about 20K examples for ASNQ, 1000 for WikiQA, TREC-QA and WQA and 1800 for TriviaAS2 and NewsAS2. In order to ensure model convergence, we keep a higher number of examples for ASNQ. This adjustment is necessary due to the significantly small ratio of positive to negative examples, which is approximately 1 in 350. We leave the validation and test sets unchanged in this experiment to better compare with previous results. 

The results show that on average, our models outperform the baselines by a large margin featuring also a much lower standard deviation when the training data is scarce, thus confirming the benefits of our continuous pretraining. For example, on ASNQ and using Base models, our objectives increase the P@1 over the baselines from 6 to 24 points. On DeBERTa\textsubscript{Large} and DeBERTaV3\textsubscript{Large} the improvements are smaller but consistent, with gains of about 4 points in P@1.

WikiQA and TREC-QA have been reduced fewer times and thus improvements are more subtle. Nevertheless, every model we train with our specialized objective outperforms the corresponding baseline by a significant margin, with gains of up to 17 points in P@1. Notice that ELECTRA is the weaker model across all tasks. We believe that this is due to the shorted pre-training and thus it struggles at adapting to new tasks with less supervision. As a comparison, ELECTRA's number of pre-training tokens is half of those seen by DeBERTa and $\frac{1}{4}$ compared to RoBERTa and DeBERTaV3.

On WQA and NewsAS2, the results follow a similar trend. In the few-shot setting over WQA, which is the harder task, our continuously pre-trained models outperform the baselines from 2 to 10 points in Precision@1. For example, RoBERTa\textsubscript{Base}~+~PSD achieves 6.2 more points in P@1 when compared with the original RoBERTa\textsubscript{Base}. On NewsAS2, improvements range from 2 to 16 points, highlighting again how our specialized continuous pre-training is essential when target data is scarce.

Finally, we notice how our objectives combined with RoBERTa and ELECTRA struggle at improving performance on TriviaAS2. We suspect there is a misalignment between the structure of our task and the way data have been collected in the TriviaAS2 corpora. Since our objective show significant improvements when trained on the full dataset, we think that the reduced dataset is not large enough to correct the misalignment mentioned before.

\FloatBarrier

\subsection{General Considerations}
\label{sec:pairwise_considerations}

In this section, we provide an overview of the general performance of our models across various datasets and elucidate the factors contributing to the slight improvements observed in some scenarios.

\paragraph{Objective-dataset alignment}
Our objectives provide significant advantages on some datasets and more limited gains on others. First, consider that among the 6 datasets we use for the evaluation, 4 of them gather answer candidates from within the same document (ASNQ, WikiQA, NewsQAS2, TriviaAS2) while the other two contain answer candidates extracted from several documents across the web. We suspect this difference influences the performance of our models when trained with SSP, SP and PSD. The reason is that while SSP and SP work more at the paragraph/sub-paragraph level, PSD is trained only over paragraphs across different documents. For this reason, PSD aligns generally better than the other techniques on WQA and TREC-QA, while SSP and SP feature the greatest advantages on the other datasets. As a rule of thumb, we suggest exploiting PSD when the target task contains answer candidates gathered from multiple documents and when the question or the answer is longer than a pair of sentences. Otherwise, we suggest using SSP or SP. If the target dataset is very heterogeneous or not known in advance, we suggest choosing the models trained with the combination of our objectives (``ALL'') instead, as they provide more stable results across all datasets.

\paragraph{Limited improvements}
The gains on WQA and TREC-QA are limited when using our techniques or even state-of-the-art methods, such as T\textsc{and}A. For TREC-QA, this can be explained by noticing that we are close to a perfect re-ranking, and thus improving further is difficult. In \citet{soldaini-moschitti-2020-cascade}, the authors manually investigated the TREC-QA dataset and discovered that a few answer candidates are not annotated correctly. Considering also that the number of questions in the test set is only 68, this explains why it is hard to obtain significant gains on this dataset.

Regarding WQA, this dataset contains answer candidates extracted from billions of documents across the web. The answer candidates contain many typos, are sometimes malformed and do not always have correct annotations. For those reasons, our methods and other state-of-the-art methods struggle at improving the output ranking.

\chapter{Self-supervised objectives for Contextual AS2}
\label{cha:context}

In this research, we study the effects of providing Answer Selection architectures with additional context to improve accuracy in candidates' evaluation.

Previous studies~\citep{Lauriola2021, han-etal-2021-modeling} have exploited pre-trained Transformer encoders directly for Contextual AS2 tasks by fine-tuning them on inputs containing multiple sentences with different roles: the question, the answer candidate, and the surrounding context (previous and following sentences). However, this structured input poses challenges during fine-tuning as standard pre-training approaches do not align well with the downstream Contextual AS2 task. The language model lacks knowledge of the role of each sentence, requiring the learning of extended sentence-level embeddings during fine-tuning, which has shown empirical underperformance. This issue becomes more pronounced with limited fine-tuning data, indicating difficulties in leveraging context effectively.

To address these challenges, we propose three pre-training objectives that align structurally with the final Contextual AS2 task, which is described in Section~\ref{sec:answer_sentence_selection}. These objectives leverage the mild form of supervision contained in large datasets of raw text to pre-train the context slots in the Transformer text input. Specifically, we exploit the human subdivision of large documents in smaller paragraphs, which address the same general topic from different perspectives.

We report a description of the models used for the experiments in Section~\ref{sec:sota_lm}, while for works related to the Answer Sentence Selection and the Contextual AS2 tasks, refer to Section~\ref{sec:related_as2}. Additionally, we provide an overview of related research that exploits the document structure in Section~\ref{sec:related_document_structure}.

The effectiveness of our strategies is evaluated on 7 datasets using several popular pre-trained Transformers. The experimental results demonstrate that our approaches, which incorporate structural pre-training, effectively adapt Transformers to process contextualized input. Compared to the baselines, our methods achieve up to an 8\% improvement in accuracy on certain downstream datasets.

\section{Contextual Models}

Contextual models process 3 inputs at a time: the question, the answer candidate and the context. Contrary to the Jointwise models discussed in Chapter~\ref{cha:jointwise}, Contextual models focus on a single answer candidate and leverage the additional context to enhance the ranking accuracy.

\subsection{Injecting Context into Transformers}

\begin{figure*}[t]
    \centering
    \includegraphics[width=1.0\textwidth]{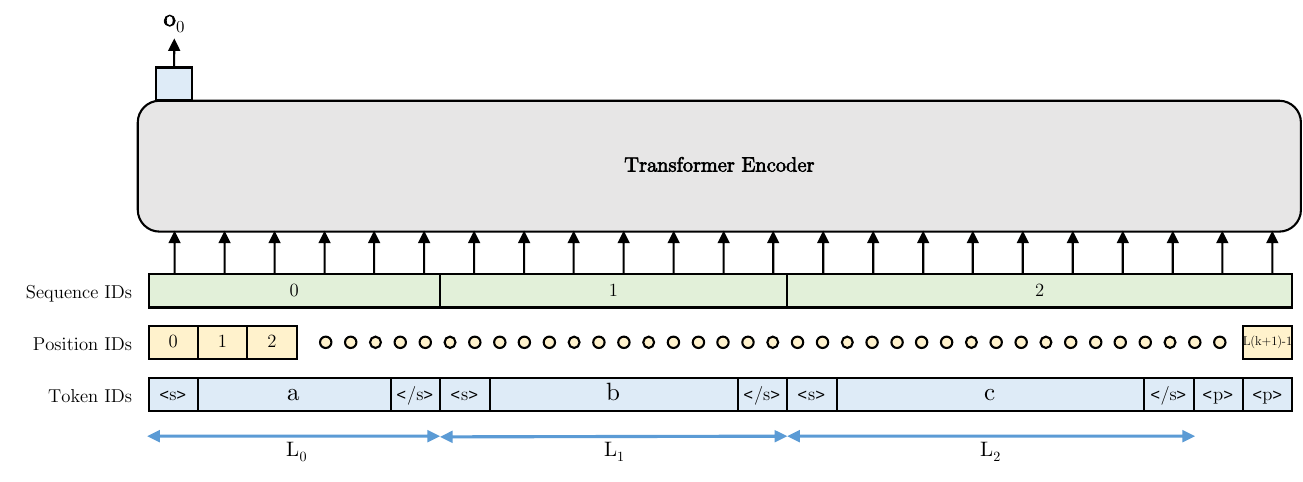}
    \caption{\small Example of Contextual model.}
    \label{fig:context_model}
\end{figure*}

Common pre-trained Transformers such as BERT~\citep{devlin-etal-2019-bert} and RoBERTa~\citep{liu2019roberta} struggle at reasoning over more than 2 spans of text. The reason is that they were trained either only with token-level objectives (e.g. MLM, TD) or with a combination of token-level and sentence-level objectives over two text spans, such as NSP or SOP. We invite the reader to refer to Section~\ref{sec:pretraining_objectives} and~\ref{sec:issue_common_pairwise_objectives} for further information about the common pre-training objectives and their issues, respectively.

We extend the original Transformer and tokenizer inputs by incorporating a third slot for context. An illustration of an extended model for Contextual training is shown in Figure~\ref{fig:context_model}. As depicted in the picture, there exists a token type id for each distinct input, and these embeddings require training either during the pre-training or directly in fine-tuning. Additionally, the pre-trained model checkpoints available from recent works do not encompass information about the contextual input. To address these concerns, in the next Section we propose several pre-training strategies for Contextual models pre-training.

\section{Context-aware Pre-training Objectives}

Contextual models are architectures designed to process more than 2 inputs efficiently. This poses practical challenges because common Transformer models are pre-trained on 1 or 2 spans of text at a time~\citep{devlin-etal-2019-bert, liu2019roberta, clark2020electra}. Thus, it is difficult to accustom those architectures to reason over many text spans only while fine-tuning. With the term ``span'', we mean a sequence of 1 or more sentences within the same document. Notice that even if most Transformer models are trained over long spans of text, there is no strong supervision over the relations between the different parts of the input. For example, the NSP and SOP objective of BERT and ALBERT tasks the model only to predict the order of 2 long text spans.

For those reasons, we propose 3 alternative pre-training objectives that are designed to adapt the model at exploiting additional context to better re-rank answer candidates in fine-tuning. The proposed objectives are extensions of SSP, which is described in Chapter~\ref{cha:pairwise}. We explain the intuition behind SSP with an example. 

Consider a passage from Wikipedia that consists of three sentences:

\begin{itemize}
\item $s_1$: Harry lives in a cupboard under the stairs in the house of the Dursleys, his aunt, uncle, and cousin, Dudley.
\item $s_2$: At the age of 11, Harry starts attending Hogwarts School of Witchcraft and Wizardry, where he takes classes in Charms, Transfiguration, and others.
\item $s_3$: He also participates in Potions lessons, where he encounters Professor Severus Snape, who immediately displays a dislike for him.
\end{itemize}

When presented with a question like ``Which classes did Harry take in Hogwarts?'', a conventional language model can easily identify answers that follow the format ``$X$ takes classes in $Y$''. It achieves this by matching the subject of the question with the object of the answer, focusing on the shared predicate of taking classes. However, the same language model would struggle to select answers of the form ``$X$ participates in $Y$'', as it requires understanding the entire predicate's argument structure of taking vs participating. By pre-training a language model using the SSP task, it can learn the implicit connections between concepts mentioned in $s_3$, such as ``participating in Potions lessons'', and the concept of ``taking classes in Charms, Transfiguration, and others'' mentioned in $s_2$. These sentences belong to the same paragraph, enabling the model to reason about concepts and establish relationships between entities.

The learned semantics, which involve connecting sentences within the same paragraph, prove to be valuable in downstream tasks. The model can leverage previously acquired relations between entities and concepts and apply them when matching questions with potential answers. By utilizing relations from one sentence, it can generate questions that can be answered by information retrieved from another sentence. This tendency is particularly common among sentences within the same paragraph since every paragraph provides different perspectives on the same general topic.

We study three different methods to augment SSP with suitable contextual information, denoted as $c$. Below, we provide a detailed explanation of how we sample the context $c$ from the documents for the contextual pre-training of the models. Regarding $a$ and $b$, they are sampled in the same way described in Section~\ref{sec:ssp}. For each positive pair, we sample up to $k_1=2$ hard negatives from within the same document and $k_2$ easy negatives from other documents until $k_1 + k_2 = 4$, as in SSP.

\subsection{Static Document-level Context}

\begin{figure*}[t]
    \centering
    \includegraphics[width=0.9\textwidth]{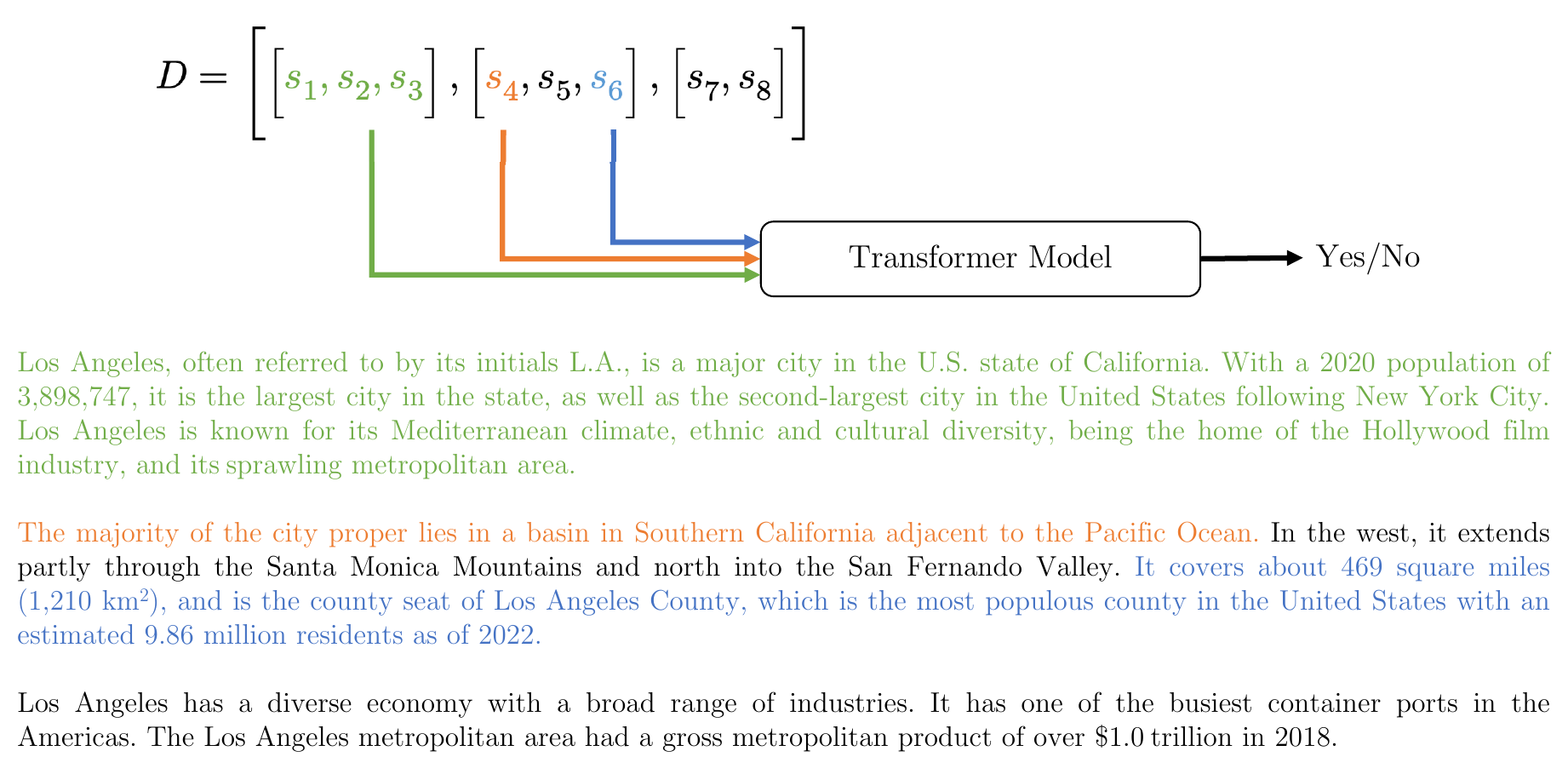}
    \caption{\small Example of SDC objective. $a$ is in orange, $b$ in blue and the context $c$ in green.}
    \label{fig:sdc}
\end{figure*}

In this initial experiment, we established the context $c$ to be the first paragraph $P_1$ of a document $\mD = [ P_1, \dots, P_n ]$ from which the answer $b$ is selected. The rationale behind this choice is that the first paragraph of most documents provides a general overview of the document's topic~\citep{Chang2020PretrainingTF}. By doing this, the context $c$ assists in predicting whether $a$ is extracted from the same paragraph as $b$. An example of the SDC objective is provided in Figure~\ref{fig:sdc}.

We refer to this as static document-level context because the contextual information $c$ remains constant for any answer $b$ obtained from the same document $\mD$. Specifically, we generate positive examples by randomly sampling $a$ and $b$ from a single paragraph $P_i \in \mD$, where $i > 1$. For the selected $a$, we create challenging negative examples by randomly selecting a sentence $b$ from different paragraphs $P_j \in \mD$, where $j \ne i \wedge j > 1$. In these negative examples, we still set $c = P_1$ since $b$ belongs to the same document $\mD$. We generate easy negatives for a given $a$ by sampling $b$ from a random paragraph $P'_i$ in another document $\mD' \ne \mD$. In this scenario, we choose $c$ as the first paragraph $P'_1$ of $\mD'$ because the context in the downstream AS2 task relates to the answer candidate rather than the question.

\subsection{Dynamic Paragraph-level Context}

In Dynamic Paragraph-level Context (DPC), we select the context $c$ as the remaining text of the paragraph $P_i$ from which $b$ is extracted. Positive examples are created by sampling both $a$ and $b$ from the same paragraph $P_i \in \mD$. For the positives, the context $c$ is equal to $c = P_i \setminus \{a, b\}$. Notice that removing $a$ and $b$ from $P_i$ is of fundamental importance not to make the task trivial, otherwise, the model could exploit $P_i$ to decide whether $a$ and $b$ originated from the same paragraph. Hard negatives are created by randomly selecting $b$ from another paragraph $P_j \in \mD, j \ne i$. In this case, the context $c$ is set to whatever remains in $P_j$ after the removal of $b$: $c = P_j \setminus \{ b \}$. Finally, we create easy negatives by sampling $b$ from a paragraph $P'_i$ in another document $\mD' \ne \mD$. Here, we set the context to $c = P'_i \setminus \{ b\}$.

\begin{figure*}[h]
    \centering
    \includegraphics[width=0.9\textwidth]{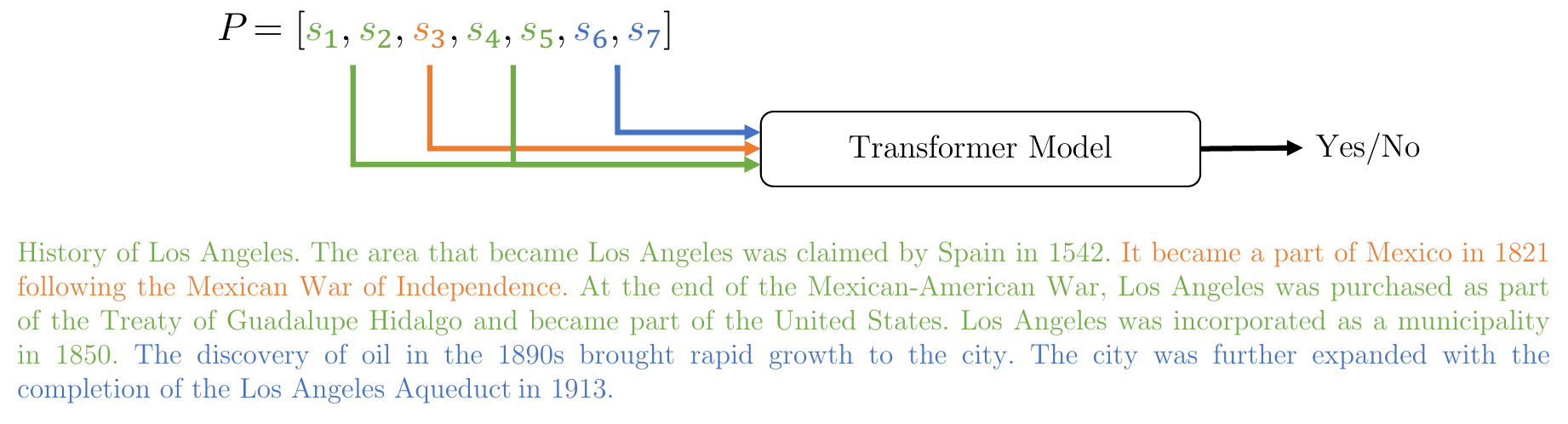}
    \caption{\small Example of DPC objective. $a$ is in orange, $b$ in blue and the context $c$ in green.}
    \label{fig:dpc}
\end{figure*}

Notice that the context is always extracted from the same document or paragraph of $b$. The reason is that in Question Answering, the context is tied to the answer, and should provide more details on it rather than on the question. An example of DPC objective is depicted in Figure~\ref{fig:dpc}.

\subsection{Dynamic Sentence-level Local Context}

In this approach, we designate the context $c$ as the surrounding text encompassing the sentence $b$, which includes the sentences preceding and succeeding $b$ within the paragraph $P_i$. To address exceptional cases, we require that either the previous or the next sentence of $b$ must exist (e.g., if $b$ is the final sentence in the paragraph $P$, the next sentence may not be present). We refer to this as Dynamic Sentence-level Local Context (DSLC), as the contextual information $c$ is defined at the sentence level and varies accordingly for each sentence $b$ extracted from the paragraphs.

Similar to the SDC and DPC, we create positive pairs by randomly sampling $a$ and $b$ from the same paragraph $P_i \in \mD$, with $c$ representing the local context surrounding $b$ within $P_i$ (ensuring $a$ is not part of $c$). We automatically exclude paragraphs that are insufficient in length to guarantee the formation of a positive example. For hard negatives, we sample $b$ from another paragraph $P_j \in \mD$, where $j \ne i$. As for easy negatives, we sample $b$ from a paragraph $P'_i \in \mD'$, where $\mD' \ne D$ (in both cases, $c$ is defined as the sentences around $b$).

\begin{figure*}[h]
    \centering
    \includegraphics[width=0.9\textwidth]{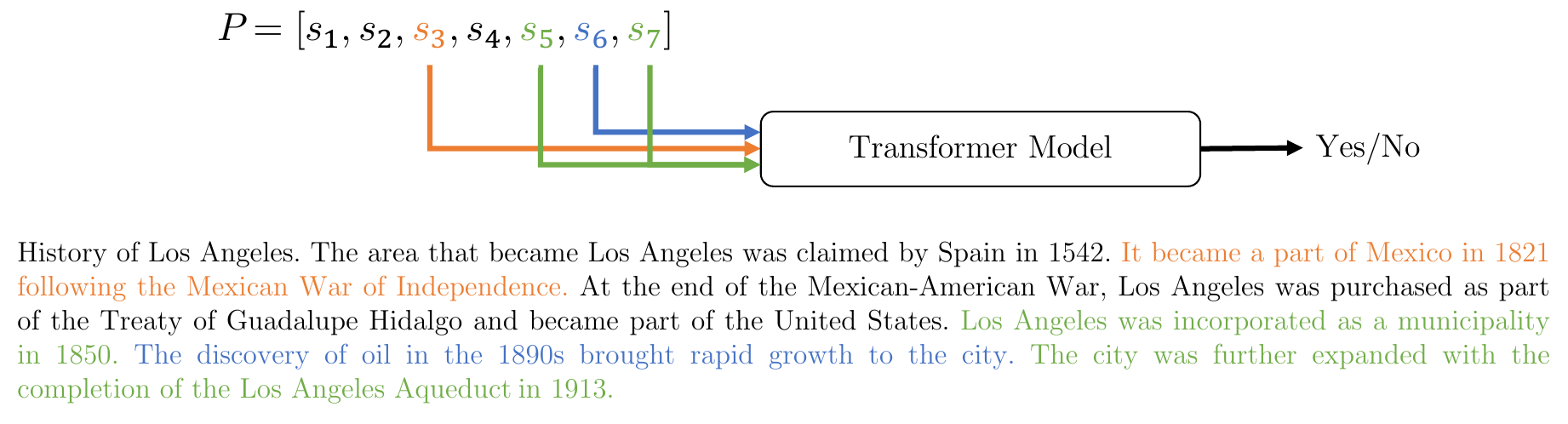}
    \caption{\small Example of DSLC objective. $a$ is in orange, $b$ in blue and the context $c$ in green.}
    \label{fig:dslc}
\end{figure*}

An example of DSLC objective is given in Figure~\ref{fig:dslc}.

\FloatBarrier
\section{Experiments}

In this Section, we describe the setting in which we pre-trained and fine-tuned the models. We highlight the datasets we use, the hyper-parameters we use in training and the amount of compute effort for each execution.

\subsection{Continuous Pre-Training}

\input{tables/context_pretraining}
We perform continuous pre-training starting from checkpoints of several state-of-the-art models. In particular, we test all of our objectives over RoBERTa and ELECTRA and then, based on which objective provides the better performance overall in preliminary experiments, we train also two DeBERTaV3 models. We also experiment by mixing all objectives, thus forcing the model to gather information from three different types of context to better understand entities and relations in the answers. We indicate those experiments with ``ALL'', as in the previous Chapter.

To avoid improvements derived from the usage of more data, we exploit the training corpora used to pre-train the original checkpoints of RoBERTa, ELECTRA and DeBERTaV3. We do continuous pre-training over the English Wikipedia, the BookCorpus, OpenWebText and CC-News by using a combination of our objectives (SDC, DPC and DSLC) with the original pre-training tasks, i.e. MLM for RoBERTa and TD for ELECTRA and DeBERTaV3. As before, we found that keeping the original training tasks allows us to reach higher levels of accuracy. Moreover, notice that for DeBERTaV3\textsubscript{Large} models, we can not exploit the original generator for Token Detection, as it was not released. Thus, we use a DeBERTaV3\textsubscript{Base} instead as the generator for DeBERTaV3\textsubscript{Large}. We pre-process the datasets described above according to the definition of our objectives, obtaining about 330M of samples for SDC and 200M for DPC and DSLC.

In Table~\ref{tab:context_pretraining} we show the hyper-parameters we use in each experiment, highlighting also the additional computational effort required over the original pre-training. One of the advantages of our approaches is that once continuously pre-trained, our checkpoints can be applied to different AS2 settings with minimal effort. The amount of tokens seen in pre-training by each initial checkpoint is described in Section~\ref{sec:sota_lm}. Notice that as in SSP, we keep a constant maximum sequence length of 128 tokens because in preliminary experiments we didn't find particular advantages with higher values. This reduces the computational complexity by a factor of 16 compared to the original sequence length of 512 tokens.

Regarding the optimization hyper-parameters and the hardware used, we set $\beta_1 = 0.9$ and $\beta_2 = 0.99$ in the FuseAdam optimizer, a weight decay of $0.01$ and a triangular learning rate with 10\% warmup steps. All models are trained on our machine (see Section~\ref{sec:machines}) and require about 3.0 days of training each with \emph{fp16} and DeepSpeed~\citep{deepspeed}. There may be slight variations compared with the results published in \citet{di-liello-etal-2023-context} because we fixed an important bug regarding the initialization of the token types embedding layer.

\subsection{Fine-Tuning}

We evaluate the models described in the previous Section on several datasets for Contextual Answer Sentence Selection. More specifically, we fine-tune over ASNQ, WikiQA, TREC-QA, IQAD, NewsAS2 and TriviaAS2. We do not test on WQA as in the previous Chapter because it provides no way of retrieving the original documents, but we use IQAD instead, which follows a similar data distribution and provides context. In all datasets, the context is always composed of the sentences before and after the candidate answers in the original document. The statistics for the datasets can be retrieved in Section~\ref{sec:as2_datasets}. For TREC-QA, we do not have context because answers are not extracted from documents (or the documents have not been publicly released). In this case, we exploit a large generative language model to generate additional context for each answer candidate. Specifically, we utilize Falcon~\citep{falcon40b}, a recently released LLM with 40B parameters. We use the version fine-tuned on various datasets from Baize~\citep{xu2023baize}, which contains instruction-based conversations between real users about several topics\footnote{The model we use can be found here: \url{https://huggingface.co/tiiuae/falcon-40b-instruct}}. We provide the model with the following prompt: ``Provide additional context to the text `\{answer\}' in two sentences: '', by dynamically substituting \{answer\} with the answer candidate in the dataset. For the generation, we use a temperature of 0.7,  a top$_k$ of 8 and sampling. Moreover, we force the output length to be between 16 and 64 tokens.

\input{tables/context_as2_hparams}

For the fine-tuning, we use the same hyper-parameters used in the pre-training but for the batch size, the learning rate, the number of warmup steps and the number of training epochs. Summaries of the hyper-parameters search spaces for each downstream dataset are reported in Table~\ref{tab:context_as2_hparams}. The number of warmup steps in the scheduler is set to match the number of steps in the first epoch. We also increase the maximum sequence length to 256 tokens in every experiment over ASNQ, NewsAS2, TriviaAS2 and IQAD, since the concatenation of the question, the answer candidate and the context of those datasets often exceeds 128 tokens. As in previous Chapters, we use the development set to find the best batch size and learning rate for each combination of baseline models and target datasets. We do not conduct a grid search over our continuously pre-trained models because it is too expensive and they adapt well to the best hyper-parameters chosen for the baselines.

We measure performance again with Mean Average Precision (MAP), Mean Reciprocal Rank (MRR) and Precision@1 (P@1) as in related works~\citep{Lauriola2021, garg2019tanda}.

\section{Results}

In this section, we present the outcomes obtained by refining our pre-trained models over various AS2 datasets. To facilitate comprehension of the enhancements, we divide the results into base and large models. We report results from the original papers when available, otherwise, we fine-tune the models on the different tasks. Statistical significance is computed compared to models from \citetalias{Lauriola2021}, which are fine-tuned by us for a fair comparisons because the authors perform an additional transfer step on ASNQ and because they report results on the development set.

\subsection{Base Models}

\input{tables/context_as2_base_1}
\input{tables/context_as2_base_2}

Results for this experiment are reported in Table~\ref{tab:context_as2_base_1_1} and~\ref{tab:context_as2_base_1_2} for ASNQ, WikiQA and TREC-QA, in Table~\ref{tab:context_as2_base_2} for both IQAD test splits and in Table~\ref{tab:context_as2_base_3_1} and~\ref{tab:context_as2_base_3_2} for NewsAS2 and TriviaAS2. This Section continues by describing the results of our Spans in the Same Paragraph (SSP) objective augmented with the different context extraction techniques we developed. We compare against the original pre-trained checkpoints of each model as well as against models fine-tuned with context, as in \citetalias{Lauriola2021}.

\paragraph{SDC}

Our continuously pre-trained models with SSP and Static Document-level Context improve over the baselines fine-tuned with either only question-answer pairs or question-answer-context tuples. On ASNQ, our RoBERTa\textsubscript{Base}~+~SSP~(SDC) increases the P@1 by 5.2 and 1.1 points compared to the two baselines, respectively. Similarly, ELECTRA\textsubscript{Base}~+~SSP~(SDC) improves the P@1 by 4.6 and 1.5 points over the corresponding baselines.

On smaller datasets such as WikiQA, improvements are larger because the baselines struggle at reasoning over 3 input text spans since fine-tuning data is scarce and pre-training was performed on at most 2 input text spans. For these reasons, our models augmented with SDC improve the P@1 by more than 4 points on WikiQA for both RoBERTa and ELECTRA. On TREC-QA, SDC struggles to outperform both baselines, probably because the first paragraph of a document used in SDC does not align well with the type of context generated by the large LLM we employ.

In the industrial test splits (IQAD Bench 1\&2), our models feature significant gains over both baselines. For example, on IQAD Bench 1, RoBERTa\textsubscript{Base}~+~SSP~(SDC) improves the P@1 by 2.8\%. Results are shown relative to the corresponding baseline because those datasets are subjected to Amazon's internal data license. Notice that the baselines augmented with context struggle to outperform the original Pairwise baseline because IQAD does not contain enough data to accustom the model to effectively exploit the contextual information.

Finally, we analyze the improvements we achieve on NewsAS2 and TriviaAS2. On these datasets, RoBERTa and ELECTRA combined with SSP (SDC) outperform the baselines augmented with context by 1 about point in P@1 for every combination. A particular case is represented by ELECTRA over TriviaAS2, in which our model outperforms the corresponding baseline by 3.5 points in P@1.

\paragraph{DPC}

In DPC, the context is created by retrieving the remainder of the paragraph after two spans of text are extracted for SSP. On ASNQ, the checkpoints of RoBERTa and ELECTRA continuously pre-trained with SSP~(DPC) improve the P@1 by up to 1.6 points over the baselines augmented with context, demonstrating that specialized objectives in pre-training are very important to accustom the model at the final task.

On the two small datasets (WikiQA and TREC-QA), both RoBERTa and ELECTRA with SSP (DPC) outperform the two baselines by a relevant margin. For example, ELECTRA\textsubscript{Base}~+~SSP~(DPC) improves the P@1 by 2.3 and 5.3 points on TREC-QA compared to the Pairwise and the contextual baseline, respectively.

On the two IQAD internal benchmarks, the results of exploiting SSP (DPC) are mixed. Our objective combined with RoBERTa shows consistent gains on both the test splits, with improvements up to 1.3\% in P@1. With ELECTRA, our objective provides consistent improvements over the contextual baseline but struggles at competing with the Pairwise model. This further demonstrates that context is a precious source of information that is difficult to be exploited properly, especially on IQAD, which contains real user questions that are noisy, contain typos and may not be semantically correct.

Finally, on NewsAS2 and TriviaAS2, our ELECTRA\textsubscript{Base} augmented with SSP (DPC) provides improvements over both datasets, with gains in the range of 1-3 P@1 points over the baselines. At the same time, RoBERTa\textsubscript{Base}~+~SSP~(DPC) outperforms the Pairwise baseline but falls short behind the contextual baseline.

\paragraph{DSLC}

The last context type we analyze is DSLC, in which we exploit the sentences around the second span of SSP as the context. On ASNQ, RoBERTa and ELECTRA augmented with SSP (DSLC) outperform the stronger baseline by 1.4 and 1.9 points in P@1 respectively. Similarly, on WikiQA and TREC-QA, every combination of our models with SSP (DSLC) shows improvements over the baselines, with gains between 0.4 and 4.4 points of P@1.

On Amazon's internal datasets, RoBERTa with SSP (DSLC) outperforms both the Pairwise and the contextual baseline by a significant margin. For example, on IQAD Bench 1, our model improves the MRR and P@1 by 0.5 and 1.9 points respectively. When combining SSP (DSLC) with ELECTRA and testing over IQAD Bench 1, our model features significant gains of 1.0 MAP and 0.6 P@1 points. On IQAD Bench 2 instead results of our models are significantly better when compared with the contextual baseline and also slightly better than the Pairwise RoBERTa\textsubscript{Base}.

On NewsAS2 and TriviaAS2, the gains of our models are similar to those on ASNQ. In particular, RoBERTa\textsubscript{Base}~+~SSP~(DSLC) outperforms the Pairwise and contextual baselines on NewsAS2 by a margin of 1.9 and 0.8 P@1 points respectively. When tested on TriviaAS2 instead, the results are in line with the contextual baselines. Regarding ELECTRA combined with SSP (DSLC), it outperforms the strongest baseline by 0.9 and 2.8 P@1 points on NewsAS2 and TriviaAS2 respectively.

\paragraph{ALL}

In this experiment, we combine all the proposed context-gathering methodologies for a single continuous pre-training, indicated with SSP (ALL). When applied to RoBERTa, ELECTRA and DeBERTa, the trend is clear: our pre-trained models outperform the baseline on almost all benchmarks. The only exception is represented by DeBERTa\textsubscript{Base}~+~SSP~(ALL) when tested over WikiQA, in which the additional context is enough to hurt performance, while SSP (ALL) leads to even lower accuracies. We suspect there is an interaction between the DeBERTa\textsubscript{Base} pre-training setting and the context of the WikiQA dataset. Since the results of SSP (ALL) are more agnostic compared to the target dataset, we suggest the usage of this objective when testing on new or unknown data.

Among the best results obtained among Base models, we cite DeBERTaV3\textsubscript{Base}~+~SSP~(ALL), which reaches a P@1 of 71.3 on ASNQ. Moreover, DeBERTa\textsubscript{Base}~+~SSP~(ALL) evaluated on NewsAS2 achieves the highest P@1 of 79.0 among all the models in this work built over a Base architecture.

\FloatBarrier
\subsection{Large Models}
\label{sec:results_context_large}

\input{tables/context_as2_large_1}
\input{tables/context_as2_large_2}

In Tables~\ref{tab:context_as2_large_1} and~\ref{tab:context_as2_large_2} we report the results of combining larger models with our SSP objective with context gathered in three different ways (ALL).

On ASNQ, our DeBERTa\textsubscript{Large} and DeBERTaV3\textsubscript{Large}~+~SSP~(ALL) reach remarkable performance on all metrics (second and first place respectively in the global ranking). When combining DeBERTa\textsubscript{Large} with SSP (ALL) and testing over WikiQA, there is a degradation of the performance, as with DeBERTa\textsubscript{Base} (more details about this issue are given in the previous Section). On the other hand, DeBERTaV3\textsubscript{Large}~+~SSP~(ALL) on WikiQA shows improvements of up to 3.9 P@1 points. On TREC-QA, DeBERTa\textsubscript{Large} and DeBERTaV3\textsubscript{Large} combined with our specialized continuous pre-training lead to improvements of 1.5 and 0.6 points in P@1, respectively. We experiment also the combination of our techniques with T\textsc{and}A, which results in our models obtaining the third and fourth place in the global ranking of WikiQA. Notice that the transfer step over ASNQ is able to override the misalignment between DeBERTa\textsubscript{Large}~+~SSP~(ALL) and WikiQA. On TREC-QA, we obtain the third place on the global ranking with T\textsc{and}A DeBERTaV3\textsubscript{Large}~+~SSP~(ALL), just a few decimal MAP points behind our T\textsc{and}A DeBERTaV3\textsubscript{Large}~+~ALL from Chapter~\ref{cha:pairwise} and T\textsc{and}A RoBERTa\textsubscript{Large} from~\citet{garg2019tanda}.

On NewsAS2 and TriviaAS2, the combination of Large DeBERTa models with our SSP objective provides increases in accuracy on both datasets, with gains up to 1.6 and 1.7 respectively, when measuring performance with P@1.

\subsection{General Considerations}

This Section analyzes the general pattern of performance gains and provides suggestions to choose between the objectives for different use cases.

\paragraph{Objective-dataset alignment}

As in Section~\ref{sec:pairwise_considerations}, we describe the alignment between the different objectives and the target datasets. First, notice that our datasets can be divided into two categories based on how answer candidates are extracted from relevant documents. In ASNQ, WikiQA, NewsAS2 and TriviaAS2, all the answer candidates for a given question are extracted from a single document (Wikipedia for ASNQ and WikiQA, CNN/DailyMail for NewsAS2 and quiz-league websites for TriviaAS2). On the other hand, IQAD and TREC-QA contain answer candidates extracted from multiple documents sourced from the web. This results in a more homogeneous context for the former datasets because all answer candidates belong to the same document and a more heterogeneous context for the latter.

Our DPC (Document Paragraph-level Context) and DSLC (Document Sentence-level Local Context) pre-training methods share similarities in terms of the context used to assist SSP. While DPC employs the remaining of paragraph $P$ as context after excluding $a$ and $b$, DSLC utilizes the preceding and subsequent sentences to $b$ within $P$. Through empirical observation, we find that often the contexts employed by DPC and DSLC partially overlap because of the limited number of sentences per paragraph in the original corpora. This explains why models trained using both approaches yield comparable performance.

Regarding the IQAD dataset, we observe that the SDC (Static Document-level Context) approach surpasses the performance of DPC and DSLC. In SDC, the context $c$ can significantly differ from $a$ and $b$, representing the first paragraph of the document. This distinction proves beneficial for leveraging information and effectively ranking answer candidates extracted from multiple documents, potentially from diverse domains, as required in IQAD. Consequently, we recommend employing DPC and DSLC when answer candidates originate from the same document, while SDC is preferable when candidates are sourced from multiple documents.

Finally, we expected TREC-QA to benefit more from SDC than DPC and DSLC, given the nature of the dataset, in which answer candidates are extracted from multiple documents. However, we suspect that the additional context generated by the Large Language Model we employ relates strictly to every answer candidate, thus leveraging better the continuous pre-training with DPC and DSLC.

\paragraph{Limited improvements}
Similar to the results over WQA in Chapter~\ref{cha:pairwise}, the improvements in IQAD are comparatively smaller than those observed in the other datasets. There are several reasons for this. Firstly, the questions in IQAD are genuine queries from users, and they are frequently ambiguous and inadequately structured. Additionally, IQAD incorporates answer candidates obtained from a vast collection of web documents extracted with ElasticSearch. These documents have not been reviewed by humans and may contain errors in syntax, poorly constructed sentences, and inaccurate information. Moreover, the annotation process relies on user feedback, which is not always reliable in terms of accuracy.

\chapter{Self-supervised objectives for Summarization}
\label{cha:summarization}

In this last Chapter, we show that specialized pre-training objectives can improve accuracy in tasks other than Answer Sentence Selection or Fact Verification. For this experiment, we set the goal of developing ad-hoc objectives for Summarization using Auto-Regressive models.

Summarization involves condensing a given long text or document into a shorter version while preserving the key information and main points. It aims to provide a concise and coherent summary that captures the essential meaning and context of the original text. In this experiment, we focus on Abstractive Summarization, which differs from Extractive Summarization in many ways. Abstractive Summarization involves generating new sentences that capture the essence of the source text, while Extractive Summarization involves selecting and combining existing sentences. More information about the Summarization tasks is given in Section~\ref{sec:summarization}. For a detailed description of the models exploited in this Chapter, see Section~\ref{sec:sota_lm}. Moreover, we provide works related to Summarization in Section~\ref{sec:related_summarization}, while regarding the exploitation of large corpora to create self-supervised tasks, refer to Section~\ref{sec:related_document_structure}.

Abstractive Summarization requires models able to process long documents and re-write them in a more concise form. Thus, in this experiment, we employ two state-of-the-art Generative Transformer-based models: BART~\citep{lewis-etal-2020-bart} and T5~\citep{2020t5}. Both models are built over an Encoder-Decoder architecture, in which the encoder is fed with the original document and the decoder generates the summary, one token at a time.

\section{Summarization-oriented Pre-training Objectives}

High-quality Summarizations datasets are scarce for several reasons:
\begin{itemize}
\item Human Effort: Creating high-quality Summarization datasets requires significant human effort. Skilled annotators or domain experts need to read and understand the source texts and then generate appropriate summaries. This process can be time-consuming and costly, especially for large-scale datasets;
\item Subjectivity and Complexity: Summarization is a subjective task, and different annotators may generate different summaries for the same source text. This subjectivity makes it challenging to achieve a consensus on what constitutes a ``good'' summary. Additionally, summarizing complex or technical content often requires domain expertise, further complicating the annotation process;
\item Lengthy Source Texts: Summarization datasets often consist of long documents or articles that need to be condensed into shorter summaries. This adds to the complexity and time required for annotation;
\end{itemize}
For example, all 4 datasets we use for the evaluation of our approaches have important weaknesses:
\begin{itemize}
\item CNN DailyMail~\citep{cnn-dm-2015}: Summaries in the CNN DailyMail dataset are often longer than human-written summaries, which can hinder the production of concise and coherent summaries by models trained on this dataset;
\item Samsum~\citep{samsum}: The small size of the Samsum dataset restricts topic diversity and generalizability for models trained on this dataset;
\item XSum~\citep{narayan-etal-2018-dont}: The XSum dataset consists of extremely short summaries, typically a single sentence. While this simplifies the Summarization task, it limits the expressiveness and depth of the summaries, as they may not capture the full context of the source articles;
\item Gigaword~\citep{graff2003english}: The Gigaword dataset is generated using a rule-based approach, resulting in extractive summaries that may not fully capture the key ideas of the source articles. This limitation restricts the dataset's utility for training models that aim to perform abstractive Summarization.
\end{itemize}
For the reasons mentioned above, we develop a specialized training objective that adapts the models to perform Summarization already while pre-training, such that afterward, the model is more independent of the fine-tuning dataset quality.

\subsection{Static Document-level Summary}

\begin{figure*}[b]
    \centering
    \includegraphics[width=0.9\textwidth]{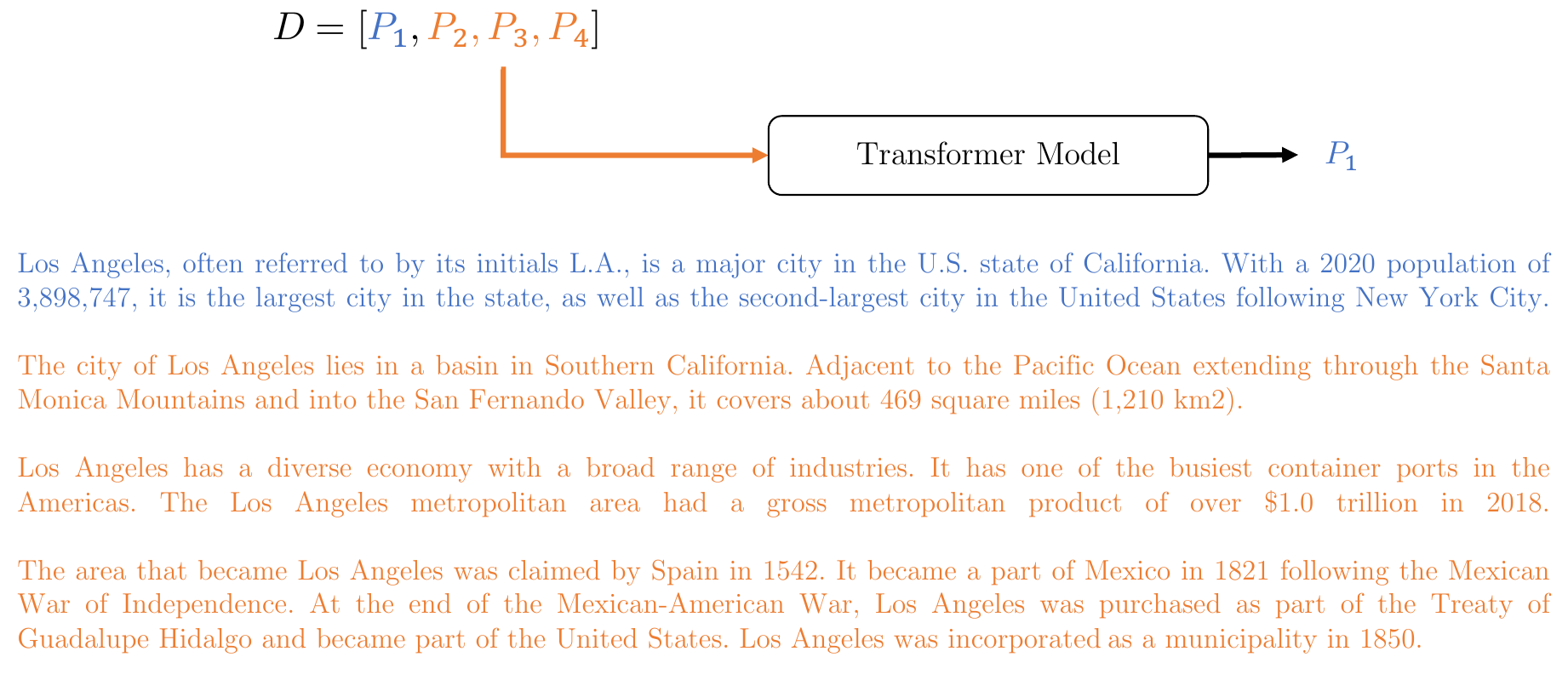}
    \caption{\small Example of SDS objective. The input is in orange and the gold labels in blue.}
    \label{fig:sds}
\end{figure*}

Static Document-Level Summary (SDS) is an objective built over the ideas developed in the previous Chapters. We exploit the raw subdivision of large documents in paragraphs as a light form of supervision. We task the model at predicting the first paragraph given all the successive. The reason is that the first paragraph acts often as an introduction or summary of the whole document's content. More formally, given a document $\mD = [P_1, \dots, P_m]$, the model is fed with $[P_2, \dots, P_m]$ and should predict $P_1$. The training is performed with Teacher-Forcing, and we utilize several generative algorithms, such as Beam Search~\citep{freitag-al-onaizan-2017-beam}, while evaluating. An example of the objective is depicted in Figure~\ref{fig:sds}.

\section{Experiments}

We describe the experimental setting for the pre-training and the evaluation of SDS.

\subsection{Continuous Pre-Training}

We do continuous pre-training starting from BART\textsubscript{Base} and T5\textsubscript{Base} publicly available checkpoints. Training those models from scratch would have required a huge amount of resources, which are out of the scope of this Thesis. Moreover, by showing that a small amount of continuous pre-training is enough to improve performance in Summarization, we can state that our techniques can be easily adapted to other architectures and corresponding pre-trained models, such as GPT~\citep{Radford2018ImprovingLU}.

We do continuous pre-training on the same datasets used to pre-train the original checkpoints of BART\textsubscript{Base} and T5\textsubscript{Base}. More specifically, we exploit Wikipedia, the BookCorpus, CC-News and OpenWebText for BART and the clean split of the C4 dataset for T5. The datasets have been processed for the SDS task described above and we filter documents with first paragraphs shorter than 2 sentences or with less than 50 characters. Finally, we reduced the resulting datasets to 100M training examples with random sampling.

In Table~\ref{tab:summarization_pretraining} we report the main hyper-parameters used while doing continuous pre-training. We fix the maximum input sequence length of the encoder to 512 tokens, while the limit over the decoder input, which is the gold summary, is set to 256 tokens. We use the Teacher-Forcing (TF) objective for the training and generative algorithms such as Beam Search for fine-tuning. Teacher-Forcing is an effective technique for training generative models, wherein the decoder receives the correct gold summaries as input, and gold labels are generated by shifting the summary tokens to the left by one position.

Regarding the optimization, we employ FuseAdam with $\beta_1 = 0.9$, $\beta_2 = 0.99$ and a weight decay of $0.01$ as in the other experiments, and we use a triangular learning rate scheduler which warms up to the first 10\% of the training steps. We run the two pre-trainings on our machine (see Section~\ref{sec:machines} for more details) with \emph{fp16} and DeepSpeed~\citep{deepspeed}. BART\textsubscript{Base} requires about 3.5 days before completing the 100K training steps while T5 needs 4 days even if it is fed with half of the tokens of the former. This can be explained by the architectures: BART\textsubscript{Base} is a Transformer model with 6 layers in the encoder and 6 in the decoder. T5\textsubscript{Base} instead uses 12 layers in both the encoder and the decoder, thus being a much larger model.

\begin{table*}[b]
    \centering
    \resizebox{0.75\linewidth}{!}{%
    \begin{tabular}{lccccccl}
        \toprule
        \textbf{Model}				& \textbf{Steps}	& \textbf{LR}			& \textbf{BS}			& \textbf{MSL}		& \textbf{\# Tokens}		& \textbf{Effort}		& \textbf{Objectives} \\

        \toprule

        BART\textsubscript{Base} + SDS		& 100K		& $1 \cdot 10^{-4}$		& 2048				& 512/256				& 104B/52B				& +5.0\%						& TF (1.0) \\

	\midrule

        T5\textsubscript{Base} + SDS			& 100K		& $1 \cdot 10^{-4}$		& 1024				& 512/256				& 52B/26B				& +5.0\% 						& TF (1.0) \\

        \bottomrule
    \end{tabular}
    }
    \caption{\small Hyper-parameters used for the continuous pre-training with the SDS objective. ``MSL'' is the maximum allowed input length on both the encoder and the decoder. ``\# Tokens'' shows the tokens seen by the encoder and the decoder respectively. ``Effort'' is the amount of FLOPS used in this experiment compared to the original pre-training of the models. ``TF'' is the Teacher-Forcing objective used for the training.}
    \label{tab:summarization_pretraining}
\end{table*}

\subsection{Fine-Tuning}

We extensively evaluate our approaches and the baselines on 4 datasets for Summarization: CNN-DailyMail, XSum, Samsum and Gigaword. They cover a wide range of Summarization tasks branches: (i) CNN-DailyMail consists of news articles accompanied by verbose human-generated summaries; (ii) XSum contains short news articles and single-sentence summaries; (iii) Samsum is composed of dialogues between fictional characters along with multi-sentence summaries and (iv) Gigaword is a large-scale dataset consisting of news articles and their machine-extracted headline summaries. More information about each Summarization dataset is given in Section~\ref{sec:summarization_datasets}.

We report the summary of the hyper-parameters search space in Table~\ref{tab:summarization_hparams}. We adapt the document and summary maximum length based on the datasets' statistics to reduce training time and energy consumption. For the generation, we report the main hyper-parameters in Table~\ref{tab:summarization_generation_hparams}. Notice that results may vary slightly from those reported in the original works because (i) we do not perform best model selection and (ii) we use a different decoding setting. Regarding the optimization, we exploit the same optimizer we use for the continuous pre-training, and we dynamically set the learning scheduler warmup to match the number of steps in the first epoch.

\input{tables/summarization_hparams}
\input{tables/summarization_generation_hparams}

\section{Evaluation}

Evaluating generative tasks is not trivial because natural language is ambiguous and there are many ways of expressing the same concept. Thus, to measure the performance of our models and perform an accurate comparison, we employ several different metrics: ROUGE-\{1, 2, L, LSum\}~\citep{lin-2004-rouge}, Bleu~\citep{papineni-etal-2002-bleu} and BLEURT~\citep{sellam-etal-2020-BLEURT}.

BLEU calculates the overlap of $n$-grams between the generated summary and the gold summaries. It focuses on precision, measuring how well the generated text matches the references. ROUGE, on the other hand, is a comprehensive evaluation package with multiple metrics. One of its metrics, ROUGE-$n$ (with $n \in \{1, 2\}$ in our test setting), is similar to BLEU as it counts common $n$-grams. However, ROUGE-$n$ is recall-oriented and emphasizes the ability to retrieve important information rather than exact matching. ROUGE-L evaluates the Longest Common Subsequence (LCS) between the generated summary and the references. It captures similarities beyond simple word matches. Finally, ROUGE-S measures similarity using skip-bigram co-occurrence statistics between the generated summary and the set of target references.

Among the parametric metrics, we utilize BLEURT, which stands out from BLEU or ROUGE by focusing specifically on comparing the encodings of the generated text and the target references. The advantage of using encodings is that concepts expressed differently but with similar meanings should have comparable embedding vectors. BLEURT employs a supervised learning approach, using human judgments to create a training dataset. The dataset comprises pairs of system-generated responses and reference responses, along with quality scores indicating the text's quality. These pairs train a regression model that maps the generated response's features to a quality score. BLEURT then outputs a similarity score, reflecting the degree of similarity between the generated and reference summaries.

We do not measure performance with BERT-Score~\citep{zhang-2019-bertscore} because preliminary evaluations showed a similarity between generated and target summaries of $\sim 96\%$ for every experiment, even using a RoBERTa\textsubscript{Large} model as sentences encoder. Moreover, BERT-Score is not fair and includes Social Biases~\citep{sun-etal-2022-bertscore}.

A key limitation of these metrics is that the evaluation becomes more accurate as the number of target summaries increases. Since every dataset we consider has only a single gold summary for each input document, in Section~\ref{sec:summarization_additional} we show a possible solution to this issue.

\section{Results}

\input{tables/summarization_base_1}
\vspace{-2em}
\input{tables/summarization_base_2}

In this Section, we analyze the effects of the continuous pre-training on the final Summarization tasks. In Table~\ref{tab:summarization_base_1} and~\ref{tab:summarization_base_2} we report the results achieved on CNN-DailyMail, Xsum, Samsum and Gigaword. 

The general pattern shows that models augmented with our SDS objective outperform the baselines on all tasks. For example, T5\textsubscript{Base}~+~SDS improves ROUGE-1 and ROUGE-2 F1-Scores by 0.9 and 0.8 points on CNN/DailyMail. BART\textsubscript{Base}~+~SDS also features gains over the baselines of 0.3 points of ROUGE-1 and 0.4 points when summaries are compared with BLEURT. Regarding Xsum, both T5\textsubscript{Base} and BART\textsubscript{Base} outperform the corresponding baselines by 0.7 and 0.4 ROUGE-1 points, respectively. On BLUE and BLEURT, the gap increases up to 0.8 points. Notice that almost all gains over the baselines are statistically significant thanks to the very small standard deviation of the evaluations.

We state again that a perfect model does not score 100.0 in metrics such as ROUGE and BLEU because the same concepts can be expressed in different ways that are equally correct. We consider a ROUGE-1 of about 50.0 as a very good result, given that the actual state-of-the-art in Abstractive Summarization, MoCa~\citep{zhang2022momentum}, achieves 48.9 on CNN/DailyMail. We do not compare with this technique because (i) the experimental setting is different, (ii) they use a different architecture and (iii) the total number of parameters of the models is different.

On the Samsum benchmark, the improvements are even larger. BART\textsubscript{Base}~+~SDS outperform its baseline by 1.3 points when performance is measured with ROUGE-2 F1-Score while T5\textsubscript{Base} augmented with SDS provides gains of 1.6 ROUGE-1 and 1.3 BLEURT points.

Finally, on Gigaword our methodologies provide mixed improvements based on the model's architecture. On BART\textsubscript{Base}, our model is on par with the baseline while T5\textsubscript{Base}~+~SDS outperforms the original model by a small but statistically significant gap. This demonstrates empirically that benchmarks such as Gigaword that do not align well with our objectives are not significantly penalized by the additional continuous pre-training. The reason Gigaword does not align well is that its summaries are very short and automatically generated, while we use the entire first paragraph of documents from various sources as a form of self-supervision. We leave the study of exploiting the titles of large collections of documents as summaries for future work.

\FloatBarrier

\subsection{Evaluation with Silver Summaries}
\label{sec:summarization_additional}

\input{tables/summarization_additional_1}
\vspace{-2em}
\input{tables/summarization_additional_2}

As we mentioned before, the evaluation of generated summaries is optimal when there is a large number of references for each document. We try to solve this issue by generating $n$ additional silver summaries for each document exploiting a very large language model: Falcon~\citep{falcon40b}. Specifically, we use the version fine-tuned on instruction-based conversations between users, called ``falcon-40B-instruct''. We provide the model with the following prompt: ``Document \{document\}\textbackslash n\textbackslash nSummary:'', which is what is suggested by the authors. We generate 8 additional summaries for each document in the datasets test sets and we exploit the best checkpoints of the previous Section as a starting point. Since the additional summaries have not been reviewed by humans, the results may vary based not only on the performance of our models and the baselines but also on the capacity of Falcon in creating high-quality and diverse summaries. Results are shown in Table~\ref{tab:summarization_additional_1} and~\ref{tab:summarization_additional_2}.

Generally, the results are higher than the previous Section because we report the highest similarity between the generated summaries and all the references. On CNN/DailyMail, our methods outperform the baselines on ROUGE and BLEU metrics, with increments of up to 0.9 points. On XSum, BART and T5 combined with our SDS objective improve the BLUERT by 0.6 and 0.4 points respectively.

The performance on the Samsum dataset, which contains summaries of dialogues, is improved by BART\textsubscript{Base}~+~SDS by 1.2 ROUGE-1/2 points and by more than 1.3 on ROUGE-L and ROUGE-LSum, compared to the baseline. At the same time, T5\textsubscript{Base}~+~SDS features accuracies similar to the baseline, probably because the additionally generated summaries align better with the baseline than our model.

Finally, we observe notable and statistically significant enhancements in our models when employing SDS on Gigaword, as compared to the baselines. It is worth noting that improvements in this dataset typically fall within the range of decimal points to a few integer points~\citep{zhang2022momentum}. This demonstrates again that specialized pre-training can help models at adapting to the final task already while pre-training.

\FloatBarrier

\subsection{General Considerations}

In this Section, we provided an overview of the improvements achievable by performing a continuous pre-training on Auto-Regressive models such as BART\textsubscript{Base} and T5\textsubscript{Base}, by exploiting only unlabeled documents that can be crawled freely from the web. Notice that our methodologies are orthogonal to related works that apply custom decoding strategies, specialized fine-tuning and other techniques. For example, in MoCa~\citep{zhang2022momentum} or BRIO~\citep{liu-etal-2022-brio}, our models could easily replace BART and T5 as starting checkpoints before the application of the ad-hoc fine-tuning for Summarization, which is described in the referenced works. We leave this research direction as a future work.

\chapter{Conclusions}
\label{cha:conclusions}

In this Chapter, we summarize the discoveries and the best results achieved in this Thesis. We study several innovative objectives for the efficient pre-training and the specialized continuous pre-training of Transformer-based models. We perform experiments on 4 different pre-training datasets, namely Wikipedia, BookCorpus, OpenWebText and CC-News and we evaluate our models on 21 benchmarks, covering a wide range of tasks: Fact Verification, Question Answering (Answer Sentence Selection), Summarization, Linguistic Acceptability, Sentiment Analysis, Paraphrasing, Natural Language Inference and Textual Entailment. We also release 2 new datasets for Answer Sentence Selection, which we call NewsAS2 and TriviaAS2 (derived from NewsQA and TriviaQA).

\section{Alternative Efficient Objectives}

Regarding the efficient pre-training, we develop 3 alternative objectives to BERT's MLM and ELECTRA's TD, which we name RTS, C-RTS and SLM. Those results are presented in Chapter~\ref{cha:effective}.

We discover through empirical evaluation that RTS and C-RTS can pre-train a language model to the level of performance of MLM while using fewer computational resources. Specifically, RTS and C-RTS save 20\% (base models) and 45\% (small models) training time to match or even outperform the performance of MLM on several tasks. For example, BERT\textsubscript{Base}~+~RTS outperforms BERT\textsubscript{Base}~+~MLM by 0.2 points on the GLUE benchmark and by 0.2 MAP points on TREC-QA. C-RTS instead outperforms a similar BERT\textsubscript{Small} model trained with MLM by 1.6 points on GLUE and by 0.8 and 1.0 MAP points on TREC-QA and ASNQ, respectively.

We also propose SLM, which focuses instead on providing a higher level of performance while requiring the same computational budget of MLM for the pre-training. SLM applied to BERT\textsubscript{Base} outperforms MLM by 0.7 points on the GLUE benchmark and has superior performance on 4 out of 5 Answer Sentence Selection datasets.

Future works in this research direction include the application of our objectives to more efficient models such as ALBERT~\citep{lan2020albert} or DistilBERT~\citep{sanh2020distilbert}. We noticed that with our objectives, the smaller the number of parameters, the larger the improvements. Moreover, we plan to train a larger language model with SLM, which was shown to outperform MLM in most tasks while requiring the same computational budget for pre-training.

\section{Specialized Objectives}

We design several training functions that align structurally with the final downstream task. The main advantage of our specialized training tasks is that they do not require manually annotated data, because they exploit the weak structure of large corpora like Wikipedia as a form of self-supervision. Moreover, they are orthogonal to related works, because our checkpoints can be easily exploited as a starting point for more advanced techniques since they share the same architectures of the original models.

\subsection{Multi-Sentence Inference}

In this research, we develop a self-supervised pre-training objective we call MSPP, that accustoms a Transformer model to efficiently perform inference over many input text spans. Multi-Sentence Inference can provide better accuracies than common Pairwise classifiers in different tasks. For example, in Answer Sentence Selection it can compare different sentence candidates to find the best answer to a question. In Fact Verification instead, it can leverage several retrieved evidence sentences to predict whether the claim is supported.

We perform continuous pre-training on raw data starting from a RoBERTa\textsubscript{Base} checkpoint and we evaluate our models over several datasets for AS2 and for Fact Verification. We show that checkpoints continuously pre-trained with MSPP can outperform Pairwise baselines by a large margin. For example, we improved the P@1 on ASNQ by 3.5 points, on WikiQA by 8.2 points and on TREC-QA by 3.8 points. When tested in Fact Verification on FEVER, we reach state-of-the-art Label Accuracy on the dev set and results close to other methods using larger models on the test set. Results for these experiments are described in Chapter~\ref{cha:jointwise}.

In future work, we plan to apply our models to evidence retrieved with a more powerful IR pipeline, since we are currently limited by the performance of a BERT-based DocIR. Moreover, we would like to exploit larger language models to achieve state-of-the-art accuracy on the FEVER test set and even better re-rankings on the AS2 datasets.

\subsection{Answer Sentence Selection}

In this research direction, we develop ad-hoc pre-training objectives specifically for the Answer Sentence Selection task. We exploit sources of raw text such as CC-News to create 3 self-supervised objectives over two text spans that align structurally with AS2, called SSP, SP and PSD. The experiments are provided in Chapter~\ref{cha:pairwise}.

After a continuous pre-training step over large unlabeled corpora, we show that our objectives can help Transformers at predicting whether an answer is correct for a given question. For example, our RoBERTa\textsubscript{Base}~+~SSP outperforms the corresponding baseline by 2.3, 4.6 and 1.2 P@1 points on ASNQ, WikiQA and TriviaAS2, respectively. We also reach very high scores on ASNQ when combining our objectives with our DeBERTaV3\textsubscript{Large}, surpassing the 70.0 points in P@1 for the first time. For NewsAS2 and TriviaAS2, since they are freshly released within this work, we set very strong baselines for future research. By performing a transfer step over ASNQ as in T\textsc{and}A before fine-tuning and starting from our DeBERTaV3\textsubscript{Large}~+~ALL checkpoint, we reach also the new state-of-the-art on WikiQA and TREC-QA, with MAPs of 92.7 and 95.4 points, respectively.

Finally, we show that our objectives can help especially when data is scarce. We study the consequences of reducing all the training datasets to contain only a few thousand examples. The results show improvements in P@1 up to 16 points depending on the objective and the target dataset.

As a future research direction, we plan to combine our pre-trained checkpoints with more refined techniques, such as RLAS-BIABC or T\textsc{and}A. Moreover, if the computational resources will allow it, we plan to apply our specialized objectives to larger models, such as DeBERTaV2-XXLarge~\citep{he2020deberta}.

\subsection{Contextual Answer Sentence Selection}

In Chapter~\ref{cha:context}, we continue the development of pre-training tasks by addressing Contextual AS2. In this downstream task, the system is provided with additional text that could be exploited for resolving entities in answer candidates, and thus performing a better re-ranking. We extended the SSP objective described before with additional context extracted from source documents with 3 different techniques: SDC, DPC and DSLC.

After a continuous pre-training over several millions of documents using only self-supervision, we show that our objectives outperform similar baselines augmented with context by a large margin. The reason is that even though the baselines receive the context as input, they don't know how to effectively take advantage of it for a better re-ranking. On the contrary, after continuous pre-training our models have learned to fully exploit this additional information.

We achieve the state-of-the-art on ASNQ with DeBERTaV3\textsubscript{Large}~+~SSP~(ALL), with a MAP of 78.8 and a P@1 of 75.1 points, and we also set very strong baselines for contextual NewsAS2 and TriviaAS2. Other examples of performance improvements are shown on WikiQA and TREC-QA, where our specialized training improves performance by up to 8.0 points in P@1.

In the future, we plan to exploit Large Language Models (LLMs) to generate better context, with architectures like Falcon~\citep{falcon40b}, as we did for TREC-QA. Moreover, we want to perform a better mixed continuous pre-training using all types of context (SDC, DPC, DSLC) together by challenging the model at predicting both the binary label for SSP and the type of context provided.

\subsection{Summarization}

The last task we address in this work is Summarization, in Chapter~\ref{cha:summarization}. We develop a special training objective (SDS) that tasks the model to predict the first paragraph of a model given all the others. We perform a continuous pre-training with our objective and Auto-Regressive models such as BART and T5.

By extensively evaluating our continuously pre-trained models on 4 datasets for Summarization, we show statistically significant gains compared to vanilla models. For example, our T5\textsubscript{Base}~+~SDC improves ROUGE-1 by 0.9 and 0.4 points on CNN/DailyMail and XSum, respectively. Additionally, our BART\textsubscript{Base}~+~SDS outperforms its corresponding baseline by 1.6 and 0.6 ROUGE-1 points on Samsum and Gigaword, which is remarkable considering that typically the improvements in Summarization range between a few decimal points and a few units.

Notice that the performance on Summarization is difficult to evaluate, because each document may have several correct summaries. To address this issue, we designed an experiment in which additional silver summaries for the test sets are generated with a Large LM, e.g. Falcon. By testing on the augmented test sets, we show significant improvements in all the metrics used for the evaluation. For example, our BART\textsubscript{Base}~+~SDS model improves the BLEURT score by 0.8 and 0.6 points on Samsum and XSum, respectively. Similarly, T5\textsubscript{Base}~+~SDS increases the BLUE score by 0.8 points on CNN/DailyMail and by 2.0 points on XSum.

Future research directions include the application of our techniques to large models, such as BART\textsubscript{Large} and T5\textsubscript{Large}, as well as exploiting LLMs like BLOOMZ~\citep{Scao2022BLOOMA1} to generate additional high-quality summaries both for training and evaluation.

\clearemptydoublepage


\thispagestyle{empty}
\makeatletter
\addcontentsline{toc}{chapter}{Bibliography}
\bibliographystyle{plainnat}

\bibliography{anthology, PhD-Thesis}

\clearemptydoublepage

\appendix

\chapter{Background \& Related Work}

\section{Data Organization}
\label{app:structured_data}

\paragraph{Structured data}
Data are structured when they are defined as a set of entities/concepts and relations between them. Some of the oldest and most common techniques to store structured data are tables and relational databases, such as MySQL~\citep{10.5555/560480} or MariaDB\footnote{\url{https://mariadb.org}}. Those databases are usually queried in a specific language (e.g. SQL) to extract interesting entities and relations.

Other relevant techniques to store structured data are formal ontologies and knowledge graphs. Ontologies are representations of a domain through the usage of categories, entities and relations. The goal of an ontology is to represent a domain with all its characteristics and properties. Those structures are usually represented with graphs or databases.

On the other hand, knowledge graphs are similar to ontologies but more data-driven, and the representation of data may change based on the scope. When building a knowledge graph, the objective is to allow users to find target concepts and related entities as quickly as possible. Those structures are usually backed by a graph database (GDB).

\paragraph{Semi-structured data}

Data are semi-structured when they are organized similarly to structured data but attributes are variable and there is no general schema that defines concepts and entities. Moreover, entities and relations may contain attributes that relate to unstructured data. An example of a database to store semi-structured data is MongoDB\footnote{\url{https://www.mongodb.com}}, in which the user is allowed to store custom dictionaries of key-value objects. Another example of semi-structured data is Wikipedia: the entities are the different arguments described by every page and relations are the links between different pages.

The growth of the World Wide Web is increasing exponentially the amount of semi-structured data available for Information Retrieval. Most of the data on the WWW is semi-structured because each blog, newspaper or website uses a different organization schema, or doesn't use one at all. Today, most of the algorithms based on neural networks for NLP are first pre-trained on huge amounts of semi-structured data before being specialized on downstream applications. For example, in our experiments, we create self-supervised training functions exploiting the weak structure of large corpora, which only define a paragraph-level subdivision.

\paragraph{Unstructured data}

Unstructured data in NLP are represented by raw text, without any kind of supervision. Unstructured text can be seen as a sequence of words, numbers, dates and other forms of free text. Recent advances in NLP allow the conversion from unstructured to semi-structured data through methods that search for structures inside the text, such as entity extraction algorithms (NER) and pattern recognition systems.

\section{Language Models}
\label{app:language_models_rnn_cnn}

\subsection{Recurrent Neural Networks}
\label{sec:rnn}

Recurrent Neural Networks (RNNs) are a specialized type of neural network suited for sequential data processing, including natural language, time series, and audio. While RNNs share their origin with Feed-Forward neural networks, they belong to a distinct category due to the potential presence of cycles in the information flow graph.

When applied to text, RNNs create a sentence-level representation by traversing the input text from left to right. RNNs consist of recurrent units, where each unit's output becomes the input for the same unit at the next time step. This design enables the network to maintain an internal state (memory) that can capture dependencies between elements in the input sequence. In processing a sequence of elements, an RNN typically passes the input elements through the network one at a time, using each output as the input for the next time step. The network's internal state is updated based on both the current input and the previous internal state.

Training RNNs involves employing backpropagation through time, a variant of the backpropagation algorithm used for standard Feed-Forward neural networks. The network is unrolled over time to treat it as a standard Feed-Forward neural network, allowing for gradient computation and weight updates through gradient descent.

However, RNNs suffer from certain drawbacks, including inefficiency due to their inability to process tokens in parallel and the vanishing information problem for early tokens in long sequences~\citep{5264952}.

\subsection{Convolutional Neural Networks}
\label{sec:cnn}

Convolutional Neural Networks (CNNs) represent a type of Artificial Neural Networks (ANNs) initially designed for Computer Vision tasks, including image classification and object recognition, but they have also been successfully adapted for NLP tasks.

In CNNs, convolution operates by extracting features from input data using a set of learnable filters. These filters slide over the input data, performing a dot product between the filter entries and the input data, generating a feature map.

An important characteristic of CNNs is weight sharing, where the same filters are utilized to scan the entire input. This feature reduces the number of parameters the network needs to learn and improves efficiency. CNNs also incorporate pooling layers and activation functions, which respectively downsample the feature maps and introduce non-linearity into the model.

For NLP tasks, CNNs treat the input text as a 2D matrix, with each word represented as a row and each word embedding represented as a column. Similar to image recognition tasks, the filters in CNNs for NLP slide over the input text, performing dot products between the filter entries and the input text, resulting in feature maps.

A notable distinction between CNNs for NLP and image recognition lies in the use of pooling. While pooling is employed in image recognition to downsample feature maps, it is typically avoided in NLP tasks to prevent the loss of vital information. Instead, max pooling is often employed to extract the most salient features from the feature map.

One significant advantage of CNNs over RNNs is their efficiency in processing input text in parallel, enabling them to take advantage of hardware acceleration technologies for faster multiplication of large matrices.

\section{Tokenization methodologies}
\label{app:tokenization_methods}

Here we provide an in-depth analysis of the most common tokenization techniques.

\subsection{Byte-Pair Encoding} BPE is a tokenization algorithm that follows the philosophy of representing frequent words with single tokens while splitting uncommon words in multiple tokens. It is derived from the compression algorithm~\citep{Gage1994ANA} that iteratively replaces the most frequent pair of bytes with a new, unseen, byte. In the BPE tokenizer, the vocabulary $\sV$ is built by iteratively merging the token pair that appears more frequently in the training data and adding it to $\sV$. Algorithm~\ref{alg:bpe} provides more details about the vocabulary creation procedure. This approach is bottom-up because it starts from a small vocabulary $\sV$ containing only single characters and builds up new tokens through the concatenation of frequent pairs.
\begin{algorithm*}
	\begin{algorithmic}[1]
		\Require Training documents $\sD$, vocabulary size $k$
		\Function{BPE}{$\sD$, $k$}
  			\State $\sV \gets \text{characters}(\sD)$  \Comment{Initialize $\sV$ with all the single characters in the training data}
 			\While{$|\sV| < k$}
				\State $\{ ((t_i, t_{i+1}), c_i) \}_{i \in \sI} \gets \text{get\_bigrams\_count}(\sD, \sV)$  	\Comment{Get bigrams counts splitting over $\sV$}
				\State $j \gets \argmax_{i \in \sI} c_i$										\Comment{Select most frequent token pair}
				\State $t \gets t_j \cdot t_{j+1}$ 											\Comment{``$\cdot$'' is the concatenation operator}
				\State $\sV \gets \sV \cup \{ t \}$
				\State $\sD \gets \text{replace\_bigram\_with\_unigram}(\sD, [ t_i, t_{i+1} ], t)$
			\EndWhile
			\State \Return $\sV$
		\EndFunction
	\end{algorithmic}
	\caption{\small Example of vocabulary creation with BPE tokenizer.}
	\label{alg:bpe}
\end{algorithm*}
BPE's encoding and decoding are performed with a greedy algorithm. The encoding procedure searches iteratively for the longest token $t \in \sV$ that matches the start of the given input sequence. Usually, before the vocabulary creation input sentences are split over whitespaces and punctuation to ensure no token crosses word boundaries. Moreover, special characters are be prepended or appended to every input word to indicate the start or the end. Regarding the decoding, it is a simple concatenation of every token. The division in words by inserting whitespaces is ensured to be correct by taking advantage of the additional special characters inserted before the tokenization. The following is an example of how the BPE tokenizer of RoBERTa~\citep{liu2019roberta}, which is a very prominent language model, splits text into tokens:

\vspace{-1.5em}
\begin{gather*}
  \text{``Harry Potter and the Philosopher's Stone''} \\
   \symbolwithin{\Downarrow}{=} \\
  [\text{``Harry''}, \text{``ĠPotter''}, \text{``Ġand''}, \text{``Ġthe''}, \text{``ĠPhilos''}, \text{``opher''}, \text{``'s''}, \text{``ĠStone''}]
\end{gather*}
\vspace{-1.0em}

\noindent
Notice how the word ``Philosopher's'' will be decoded correctly thanks to the special character ``Ġ'' use by RoBERTa to indicate the start of a new word in the original text.

Among the advantages of BPE there is the ability to tokenize words that were not seen at training time by only using the available tokens. In the worst case, the new word will be split in single characters, which were part of the initial vocabulary. BPE is used by many Transformer models such as GPT-2~\citep{radford2019language}, GPT-3~\citep{brown2020language} and RoBERTa~\citep{liu2019roberta}.

\subsection{WordPiece}
\label{par:wordpiece}

WordPiece is a tokenization algorithm derived from BPE in which the selection of the token pair to merge is performed with a more sophisticated technique. 
In particular, WordPiece does not select the symbol pairs that are most commonly used, but rather the ones that will increase the likelihood of the training data when added to the vocabulary. 

This means that it calculates the likelihood of each symbol pair by dividing the joint probability of the pair by the probability of the individual symbols. The symbol pair with the highest ratio is then selected. This approach is different from BPE, as WordPiece takes into account the potential loss of information when merging two symbols to determine if it is a worthwhile decision.

The algorithm to create the vocabulary $\sV$ is the same as BPE (Algorithm~\ref{alg:bpe}), after changing line 5 into:
\begin{equation}
j \leftarrow \argmax_{i \in \sI} \frac{P(t_i, t_{i+1})}{P(t_i) P(t_{i+1})}
\end{equation}
One of the benefits of using WordPiece is that it has been found to be less prone to over-segmenting words, and it has been shown to be effective when used in neural Machine Translation models. WordPiece is used in well known Transformer models such as BERT~\citep{devlin-etal-2019-bert}, DistilBERT~\citep{sanh2020distilbert} and ELECTRA~\citep{clark2020electra}.

\subsection{UnigramLM}

UnigramLM is a top-down tokenization approach which iteratively shrinks the vocabulary size by removing tokens. UnigramLM usually initializes the vocabulary with all the words and their substrings in training data. Then, given some loss function such as the log-likelihood, at each iteration the algorithm trains a simple unigram language model over the input data $\sD$ and selects a subset of tokens that, if removed, would increase the loss over the data the lowest. In other words, at each iteration the subset of tokens that influence the loss the less are removed. Common implementations remove the about the 10\% of the tokens at each iteration until the vocabulary reaches the desired size. It is also a good practice to include in the vocabulary all the single characters of a language to ensure every word can be tokenized and without recurring to the OOV token. Algorithm~\ref{alg:unilm} provides an example of UnigramLM vocabulary creation.
\begin{algorithm*}
	\begin{algorithmic}[1]
		\Require Training documents $\sD$, vocabulary size $k$, fraction $\alpha$ of tokens to remove at each iteration
		\Function{UnigramLM}{$\sD$, $k$}
  			\State $\sV \gets \text{words\_and\_subwords}(\sD)$								\Comment{Initialize $\sV$ with words and subwords in the training data}
 			\While{$|\sV| > k$}
				\State $M \gets \text{train\_unigram\_language\_model}(\sD, \sV)$  				\Comment{Train Unigram LM over $\sD$}
				\State $\sL = \{ \}$													\Comment{Will contain loss decrease for each token}
				\For{$t \in \sV$}	
					\State $M' \gets \text{train\_unigram\_language\_model}(\sD, \sV \setminus \{ t \})$
					\State $\sL_t \gets P_M(\sD) - P_{M'}(\sD)$							\Comment{Loss difference by removing token $t$ from $\sV$}
				\EndFor
				\For{$(|\sV| \cdot \alpha)$ times}										\Comment{Remove the $\alpha$ fraction of less influent tokens}
					\State $t = \argmin_i \sL_i$
					\State $\sV \gets \sV \setminus \{ t \}$								\Comment{Trim vocabulary}
					\State $\sL \gets \sL \setminus \{ \sL_t \}$								\Comment{Avoid selecting again $\sL_t$ from $\sL$}
				\EndFor
			\EndWhile
			\State $M \gets \text{train\_unigram\_language\_model}(\sD, \sV)$  					\Comment{Train final Unigram LM over $\sD$}
			\State \Return $\sV, M$													\label{alg:unilm:return}
		\EndFunction
	\end{algorithmic}
	\caption{\small Example of vocabulary creation with UnigramLM tokenizer.}
	\label{alg:unilm}
\end{algorithm*}
Since UnigramLM is based on vocabulary reduction and not on merging rules like BPE and WordPiece, there may se several ways of tokenizing the same string. This may hurt language models performance because different tokenization of the same string may be provided to the model. For this reason, an UnigramLM tokenizer saves also the unigram language model over the final vocabulary, see line~\ref{alg:unilm:return} of Algorithm~\ref{alg:unilm}. The final language model $M$ is then used to compute the likelihood of every possible tokenization. In practice the most likely tokenization is chosen to have a deterministic split of text in subwords.

UnigramLM alone is not used by any relevant language model. However, it's vocabulary creation technique is often implemented along with Sentence-Piece tokenization.

\subsection{Sentence-Piece} 

Sentence-Piece is not a new tokenization technique by itself, but rather a variant of the previously discussed algorithms in which the input text is not split over whitespaces to separate words. By treating the whitespaces between words as all the other characters, Sentence-Piece can be trained on streams of raw text in several languages without the need of ad-hoc adaptations. For example, it can successfully tokenize Chinese of Japanese text, which does not contain whitespaces, as well as English or German.

The vocabulary creation of a Sentence-Piece tokenizer can be based on either Byte-Pair Encoding or UnigramLM. Thus, it can tokenize almost any text without producing OOV tokens. Sentence-Piece is used in many start-of-the-art language models such as ALBERT~\citep{lan2020albert}, XLNet~\citep{yang2020xlnet}, DeBERTaV2~\citep{he2020deberta}, DeBERTaV3~\citep{he2021deberta} and Google T5~\citep{2020t5}.

\subsection{Loss functions, Back-Propagation and Optimization}
\label{app:optimization}

During the pre-training phase of language models, the primary objective is typically framed as a classification task. With token-level objectives, the model is trained to predict the original input by reconstructing it from various forms of corruption applied during training. To achieve this, a classification layer, such as a linear projection, is employed to map the output embeddings of the model to different categorical labels, enabling it to effectively classify and reconstruct the original input. In sentence-level tasks and fine-tuning, the output embeddings of the model are exploited to predict a categorical label that represents a feature of the whole input, such as the opinion of a user in Sentiment Analysis, or the relation between question and answer in Question Answering.

\paragraph{Loss function} The loss function is a metric that measures the accuracy of the predictions of a model compared to the gold labels. Transformer models commonly use the Cross-Entropy Loss function for both pre-training and fine-tuning stages. The Cross-Entropy Loss measures the average number of bits required to identify an event drawn from the distribution of the true labels $\mathcal{L}$ when using a scheme optimized for the predicted probabilities distribution $\mathcal{P}$. It provides a measure of the dissimilarity between these two distributions, indicating how well the predicted probabilities align with the gold labels. In the context of classification, it calculates the loss by comparing the predicted class probabilities with the categorical labels. Given a label $y \in \sC$, where the latter is the set of possible classes, and a vector of predictions over the classes $[ p_1, \dots, p_{|\sC|} ]$, the Cross-Entropy loss can be computed as follows:
\begin{equation}
L_{CE}(\vp, y) = - \sum_{i=1}^{|\sC|} \mathbbm{1}_{\{y = i\}} \log \frac{e^{p_i}}{\sum_{i=0}^{|\sC|} e^{p_i}}
\end{equation}
Moreover, the Cross-Entropy definition can be simplified when the task is binary classification and the predictions are already softmaxed:
\begin{equation}
L_{CE}(p, y) = - \big( y \log p + (1 - y) \log (1 - p) \big)
\end{equation}
Once the loss over the predictions and the labels is computed, the back-propagation algorithm is used to compute the gradients for the parameters of the model.

\paragraph{Back-propagation}

Given the error in the predictions, the back-propagation algorithm computes the gradients of all the parameters, enabling the model to update its weights and improve its performance through gradient descent. 

Mathematically, let's consider a model $\mathcal{M}$ that is composed of a sequence of parametric functions $f_1, \dots, f_n$ which are applied in order to an input $\vx$:
\begin{equation}
\vp = f_n \circ f_{n-1} \circ \dots \circ f_1 (\vx)
\end{equation}
We call $\vh_i$ the output of the layer $f_i$. Given some loss function, such as the Cross-Entropy defined before, and a target label $y$, we can compute the error $L$ with some loss function, such as Cross-Entropy:
\begin{equation}
L = L_{CE} (\vp, y)
\end{equation}
Then, by applying the Chain Rule~\citep{Goodfellow-et-al-2016}, we can compute the partial derivatives for each layer $f_i$ in the model $\mathcal{M}$. This rule allows us to propagate gradients through multiple layers of a neural network, by unrolling the gradients computed over the whole model to the gradients of each layer. This is performed as follows:
\begin{equation}
\nabla \mathcal{M} = \frac{\partial L}{\partial \vx} = \frac{\partial L}{\partial f_n} \cdot \frac{\partial f_n}{\partial f_{n-1}} \cdot \ \cdots \ \cdot \frac{\partial f_2}{\partial f_1} \cdot \frac{\partial f_1}{\partial \vx}
\end{equation}
Then, for each layer $f_i$, we can exploit the output of the previous layer $\vh_{i-1}$, the output of the current layer $\vh_i$ and the derivatives computed for the next layer $\frac{\partial f_{i+1}}{\partial f_i}$ to compute the gradients of the current layer parameters.

\paragraph{Optimization}

By efficiently applying the Chain Rule, we can propagate the gradients backward through the network, enabling the calculation of parameter updates and optimization through gradient descent. This process allows the network to learn and improve its performance by adjusting the weights based on the computed gradients. Suppose you have a layer $i$ which is a parametric function $f_i$ with weights $\mW_i$ and gradients $\nabla \mW_i$, previously computed with back-propagation. The common approach to update $\mW_i$ is gradient descend:
\begin{equation}
\mW_i' = \mW_i - \eta \ \nabla \mW_i
\end{equation}
where $\eta$ is called the learning rate and determines the speed at which the model converges. The original Gradient Descent algorithm is no longer used due to its sensitivity and instability while searching for the local minima. Common optimizer algorithms such as Adam~\citep{kingma2017adam}, RMSProp~\citep{tieleman2012lecture} and Adagrad~\citep{adagrad2019} improve over Gradient Descend by storing additional data for each weight in the model. They allow to maintain a more general direction toward the local minima while optimizing, thus increasing the training stability and the model generalization capabilities.

\section{Machine Reading Comprehension Branches}
\label{app:machine_reading_comprehension}

\subsection{Extractive Question Answering}

Extractive Question Answering involves finding specific answers to questions within a given text or set of texts. This is different from Generative (or Abstractive) QA, which involves generating new text as an answer to a question, rather than extracting an answer from existing text.

In Extractive QA, the system is given a question and a set of documents, and is expected to identify the specific piece of text within the documents that provides the answer to the question. This typically involves identifying the relevant parts of the text, such as named entities or specific phrases, and extracting them as the answer.

More formally, in Extractive QA a model $\mathbf{C}$ is given a question $q$ and a set of documents $\sD$, which may contain answer spans $a_i$ in some $D_i \in \sD$. The goal of the model is to extract the answer spans $a_i$ from $\sD$ given the question $q$. Typically, the performance are measured by checking the degree of overlap between the gold answer spans and those predicted by the model. The prediction of the answer spans by the model $\mathbf{C}$ can be performed with different techniques. For example, the model may output the start and end positions of the correct answer or a binary classifier over each token of the documents in $\sD$ could be used to predict the belonging to the answer.

\subsection{Generative Question Answering}

Generative Question Answering regards the generation of new text as an answer to a question, rather than extracting an answer from existing documents. This involves using Machine Learning techniques to learn how to generate coherent and appropriate responses to questions, based on a large dataset of examples. The goal is to generate correct answers that should be indistinguishable from human-generated text. The input set of retrieved documents could be used as a source of knowledge to help in the answer generation. However, some recent models have such a large number of internal parameters that are able to memorize very large quantities of information already while pre-training, and thus can answer most questions without external knowledge~\citep{brown2020language, 2020t5}.

In general, measuring the performance of generative models is a hard task, and evaluating generated answers is no exception. The main problem is that the model may output text with the same meaning of the gold answer but with a different formulation and using different terms. Thus, the best evaluation of generative model's outputs is manual annotation by well-trained crowd-workers. However, this method it is very expensive and the quality of the annotation is subject to variations based on annotator's salary, educational background and geographical location.

As an alternative, automatic evaluation techniques can be applied to evaluate generated answers. The most common metrics for automatic evaluation are BLEU~\citep{papineni-etal-2002-bleu} and ROUGE~\citep{lin-2004-rouge}, which were developed for Machine Translation and Summarization respectively. BLEU works by counting the ratio of $n$-grams that overlap between the generated answer and the target answer(s). On the other hand, ROUGE is an evaluation package containing different metrics. ROUGE-N for example, is similar to BLEU and works by counting common $n$-grams, however it is recall oriented and not precision-based as BLEU. ROUGE-L searches instead for the Longest Common Subsequences (LCS) between the generated answer and the gold references. Finally, we mention ROUGE-S, which uses skip-bigram co-occurrence statistics to measure the similarity between the generated string and the set of target references. Since generated answers may contain the same meaning of the target reference but expressed in a different form, the larger the number of gold references for each question, the better the evaluation.

\subsection{Abstractive Question Answering}

Abstractive Question Answering, is a type of generative QA that involves generating answers by summarizing or paraphrasing the information in a given document. This requires the system to understand and reason about the text, and to identify the most important or relevant information to include in the answer. Abstractive QA is strictly related to applications such as Summarization, where the goal is to generate a concise and coherent summary of a document or set of documents.

Since Abstractive QA is a generative task such as Generative QA, the metrics used for the evaluation are mostly the same: human evaluation when possible or automatic evaluation with metrics like BLEU, ROUGE and BERT-Score otherwise.

\chapter{Datasets}

\section{GLUE}
\label{app:glue}

The GLUE (General Language Understanding Evaluation) benchmark is a collection of diverse datasets designed to evaluate the performance of Natural Language Understanding (NLU) models. The benchmark consists of nine individual datasets, each targeting a different aspect of language understanding. We omit the WNLI dataset in our evaluation as similar works~\citep{devlin-etal-2019-bert, clark2020electra} because even a trivial majority classifier performs better than most of the models considered in this work. Here are detailed descriptions of each dataset:

\begin{itemize}
\item \textbf{CoLA (Corpus of Linguistic Acceptability)}: This dataset focuses on grammatical correctness. It contains sentences from a variety of sources, including books, articles, and websites. The task is to determine whether a given sentence is grammatically correct or not.
\item \textbf{SST-2 (Stanford Sentiment Treebank)}: The SST-2 dataset is derived from movie reviews, where each sentence is labeled with its sentiment (positive or negative). The task is to classify the sentiment of each sentence accurately.
\item \textbf{MRPC (Microsoft Research Paraphrase Corpus)}: This dataset comprises sentence pairs extracted from various sources, such as online forums and news. The goal is to determine whether the two sentences in each pair are paraphrases or not.
\item \textbf{QQP (Quora Question Pairs)}: The QQP dataset is created from pairs of questions asked on Quora, a question-and-answer platform. The task is to predict whether two questions are semantically equivalent or not.
\item \textbf{STS-B (Semantic Textual Similarity Benchmark)}: STS-B contains sentence pairs along with their similarity scores. The dataset covers diverse domains such as news, captions, and forum threads. The objective is to predict the degree of semantic similarity between the two sentences.
\item \textbf{MNLI (Multi-Genre Natural Language Inference)}: MNLI consists of sentence pairs from various genres, such as fiction, government reports, and telephone conversations. The goal is to determine the relationship between the two sentences: entailment, contradiction, or neutral.
\item \textbf{QNLI (Question-answering Natural Language Inference)}: QNLI is derived from the Stanford Question Answering Dataset (SQuAD). It involves sentence pairs where one sentence is a question and the other is a snippet from a Wikipedia article. The task is to determine whether the given question can be answered correctly using the provided snippet.
\item \textbf{RTE (Recognizing Textual Entailment)}: The RTE dataset consists of sentence pairs collected from news articles. The objective is to determine if the premise sentence entails, contradicts, or remains neutral compared to the hypothesis sentence.
\item \textbf{WNLI (Winograd Natural Language Inference)}: WNLI focuses on resolving pronoun references using contextual understanding. The dataset contains sentence pairs with ambiguous pronouns, and the task is to determine if the pronoun in the second sentence refers to the same entity as in the first sentence.
\end{itemize}
These datasets cover a wide range of Natural Language Understanding tasks, including syntactic and semantic analysis, sentiment classification, paraphrase identification, textual entailment, and pronoun resolution. They serve as a standardized benchmark to evaluate the performance of different language models and techniques in a variety of linguistic tasks.

\chapter{Alternative Efficient Objectives}

\section{RTS vs C-RTS}
\label{app:rts_vs_crts}

In this Section, we provide more details about the comparison between RTS and C-RTS on a longer pre-training, which should better highlight the differences between the two algorithms.

\begin{figure*}[p]
\centering

\begin{tikzpicture}
    \begin{axis}[%
    	xlabel=Pre-Training Steps,
    	ylabel=\textbf{WikiQA} MAP,
    	grid=both,
    	xmin=10000, xmax=200000, xstep=10000,
        ymin=68, ymax=80,
    	minor grid style={gray!25},
    	major grid style={gray!25},
    	width=0.8\columnwidth,
    	height=0.32\linewidth,
    	legend pos=south east]

    \addplot[line width=1pt,color=red] %
    	table[x=steps,y=map_avg,col sep=comma]{csvs/wikiqa_rts_topdown.csv};
        \addlegendentry{RTS};
    \addplot[name path=rts_top,color=red!50] %
    	table[x=steps,y=map_std_top,col sep=comma]{csvs/wikiqa_rts_topdown.csv};
    \addplot[name path=rts_down,color=red!50] %
    	table[x=steps,y=map_std_down,col sep=comma]{csvs/wikiqa_rts_topdown.csv};
    \addplot[red!30,fill opacity=0.5] fill between[of=rts_top and rts_down];

    \addplot[line width=1pt,color=blue] %
    	table[x=steps,y=map_avg,col sep=comma]{csvs/wikiqa_crts_topdown.csv};
        \addlegendentry{C-RTS};
    \addplot[name path=us_top,color=blue!50] %
    	table[x=steps,y=map_std_top,col sep=comma]{csvs/wikiqa_crts_topdown.csv};
    \addplot[name path=us_down,color=blue!50] %
    	table[x=steps,y=map_std_down,col sep=comma]{csvs/wikiqa_crts_topdown.csv};
    \addplot[blue!30,fill opacity=0.5] fill between[of=us_top and us_down];

    \legend{RTS,,,,C-RTS,,,,};
   \end{axis};
\end{tikzpicture}

\begin{tikzpicture}
    \begin{axis}[%
    	xlabel=Pre-Training Steps,
    	ylabel=\textbf{TREC-QA} MAP,
    	grid=both,
    	xmin=10000, xmax=200000, xstep=10000,
        ymin=78, ymax=90,
    	minor grid style={gray!25},
    	major grid style={gray!25},
    	width=0.8\columnwidth,
    	height=0.32\linewidth,
    	legend pos=south east]

    \addplot[line width=1pt,color=red] %
    	table[x=steps,y=map_avg,col sep=comma]{csvs/trecqa_rts_topdown.csv};
    \addplot[name path=rts_top,color=red!50] %
    	table[x=steps,y=map_std_top,col sep=comma]{csvs/trecqa_rts_topdown.csv};
    \addplot[name path=rts_down,color=red!50] %
    	table[x=steps,y=map_std_down,col sep=comma]{csvs/trecqa_rts_topdown.csv};
    \addplot[red!30,fill opacity=0.5] fill between[of=rts_top and rts_down];

    \addplot[line width=1pt,color=blue] %
    	table[x=steps,y=map_avg,col sep=comma]{csvs/trecqa_crts_topdown.csv};
    \addplot[name path=crts_top,color=blue!50] %
    	table[x=steps,y=map_std_top,col sep=comma]{csvs/trecqa_crts_topdown.csv};
    \addplot[name path=crts_down,color=blue!50] %
    	table[x=steps,y=map_std_down,col sep=comma]{csvs/trecqa_crts_topdown.csv};
    \addplot[blue!30,fill opacity=0.5] fill between[of=crts_top and crts_down];

    \legend{RTS,,,,C-RTS,,,,};
   \end{axis};
\end{tikzpicture}

\begin{tikzpicture}
    \begin{axis}[%
    	xlabel=Pre-Training Steps,
    	ylabel=\textbf{MRPC} Accuracy,
    	grid=both,
    	xmin=10000, xmax=200000, xstep=10000,
        ymin=75, ymax=86,
    	minor grid style={gray!25},
    	major grid style={gray!25},
    	width=0.8\columnwidth,
    	height=0.32\linewidth,
    	legend pos=south east]

    \addplot[line width=1pt,color=red] %
    	table[x=steps,y=acc_avg,col sep=comma]{csvs/mrpc_rts.csv};
    \addplot[name path=rts_top,color=red!50] %
    	table[x=steps,y=acc_std_top,col sep=comma]{csvs/mrpc_rts.csv};
    \addplot[name path=rts_down,color=red!50] %
    	table[x=steps,y=acc_std_down,col sep=comma]{csvs/mrpc_rts.csv};
    \addplot[red!30,fill opacity=0.5] fill between[of=rts_top and rts_down];

    \addplot[line width=1pt,color=blue] %
    	table[x=steps,y=acc_avg,col sep=comma]{csvs/mrpc_crts.csv};
    \addplot[name path=crts_top,color=blue!50] %
    	table[x=steps,y=acc_std_top,col sep=comma]{csvs/mrpc_crts.csv};
    \addplot[name path=crts_down,color=blue!50] %
    	table[x=steps,y=acc_std_down,col sep=comma]{csvs/mrpc_crts.csv};
    \addplot[blue!30,fill opacity=0.5] fill between[of=crts_top and crts_down];

    \legend{RTS,,,,C-RTS,,,,};
   \end{axis};
\end{tikzpicture}

\begin{tikzpicture}
    \begin{axis}[%
    	xlabel=Pre-Training Steps,
    	ylabel=\textbf{QNLI} Accuracy,
    	grid=both,
    	xmin=10000, xmax=200000, xstep=10000,
        ymin=83, ymax=88,
    	minor grid style={gray!25},
    	major grid style={gray!25},
    	width=0.8\columnwidth,
    	height=0.32\linewidth,
    	legend pos=south east]

    \addplot[line width=1pt,color=red] %
    	table[x=steps,y=acc_avg,col sep=comma]{csvs/qnli_rts.csv};
    \addplot[name path=rts_top,color=red!50] %
    	table[x=steps,y=acc_std_top,col sep=comma]{csvs/qnli_rts.csv};
    \addplot[name path=rts_down,color=red!50] %
    	table[x=steps,y=acc_std_down,col sep=comma]{csvs/qnli_rts.csv};
    \addplot[red!30,fill opacity=0.5] fill between[of=rts_top and rts_down];

    \addplot[line width=1pt,color=blue] %
    	table[x=steps,y=acc_avg,col sep=comma]{csvs/qnli_crts.csv};
    \addplot[name path=crts_top,color=blue!50] %
    	table[x=steps,y=acc_std_top,col sep=comma]{csvs/qnli_crts.csv};
    \addplot[name path=crts_down,color=blue!50] %
    	table[x=steps,y=acc_std_down,col sep=comma]{csvs/qnli_crts.csv};
    \addplot[blue!30,fill opacity=0.5] fill between[of=crts_top and crts_down];

    \legend{RTS,,,,C-RTS,,,,};
   \end{axis};
\end{tikzpicture}

\caption{\small Performance comparison of RTS and C-RTS on WikiQA, TREC-QA, MRPC and QNLI. We show both average and standard deviation after different fine-tunings with 5 different random seeds.}
\label{fig:rts_vs_crts}

\end{figure*}
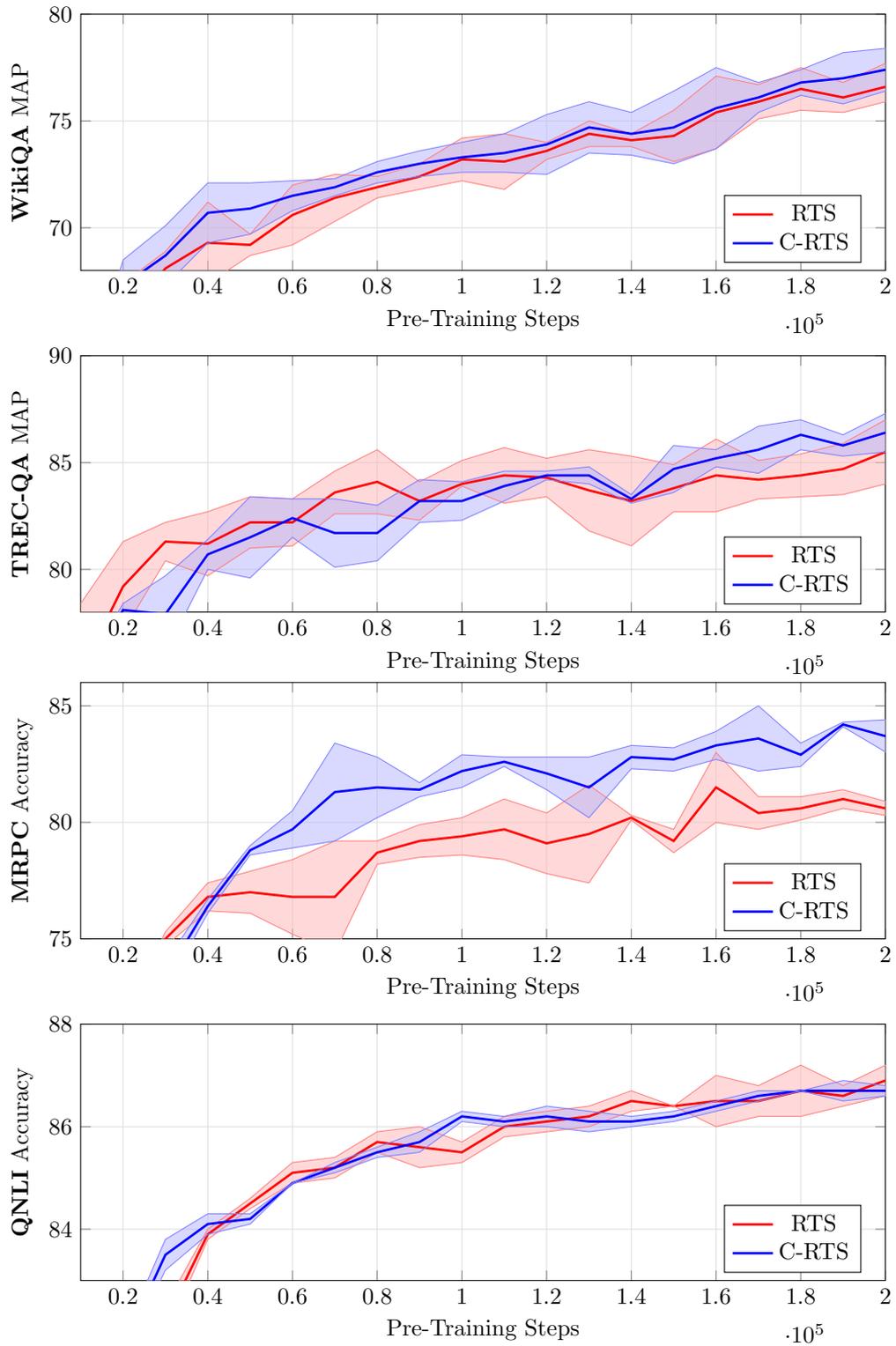

In Figure~\ref{fig:rts_vs_crts} we show the results of the comparison between RTS and C-RTS over the 4 datasets. We do not show the plot for ASNQ because fine-tuning the models on this dataset every 10K steps is very expensive. We repeated every fine-tuning on the downstream tasks 5 times with different initialization seeds. On 3 benchmarks out of 4, C-RTS is superior thanks to the harder training task. Choosing more challenging replacements instead of random ones forces the model to find subtler relations between tokens. On QNLI, both objectives perform equally from a statistical viewpoint. We argue that Small models are more influenceable than Base counterparts because they have less parameters. Thus, larger datasets with thousands of examples such as QNLI reduce the differences seen while pre-training.

\section{Negative Results}
\label{app:effective_negative}

In scientific research, negative results hold valuable lessons for future investigations. This Section presents experiments and observations that diverged from initial hypotheses, offering important insights to guide future researchers.

\subsection{C-RTS sampling always from the same Cluster}
We conduct an experiment where we simplify C-RTS by replacing tokens only with tokens within the same cluster. However, we discover that it is essential to allow sampling from other clusters as well because the model demonstrates the ability to learn how tokens are clustered after a significant number of training steps.

\subsection{Position-based techniques}
In this experiment, we explore altering the positions of tokens by masking certain positional encodings, similar to the MLM approach. The objective is to determine the original position of masked tokens in the original text. Although the classification head for this technique is smaller than MLM (slightly larger than RTS due to BERT's 512 possible positions), the results on the GLUE benchmark are about 3.0 points worse.

We also define another objective where we (i) shuffle the positions of some input tokens and (ii) predict their original positions. We randomly select 15\% of the tokens and permute them. Although this approach performs 1.5 points below MLM on the GLUE average, it is almost as fast as RTS. Combining position-based objectives with token-level objectives could be a potential future research direction.

\subsection{LM head-on ELECTRA's discriminator}
Based on the successful performance of SLM, we have implemented and tested a version of SLM for ELECTRA. However, applying SLM directly to the ELECTRA discriminator is not feasible because predictions are only made for altered tokens, which undermines the task of detecting fake tokens. To address this, we propose an alternative objective called SLM-All. In SLM-All, instead of predicting only on tampered tokens, the model is tasked with predicting the original values of all tokens while also reproducing the unchanged inputs in the output.

We evaluated this approach on GLUE, WikiQA, ASNQ, and TREC-QA. Although it required more time to be pre-trained (1.42 times the time required by a similarly sized MLM-based model), the results were generally inferior compared to other approaches.

The significant efficiency gap between SLM-All and other models is due to the fact that the latter predicts MLM-like tokens for every output embedding of each token, not just the 15\% as in SLM or MLM. Even with a 25\% reduction in the number of training steps for the ELECTRA model (to balance the presence of the additional generator with size 1/3), it still utilizes slightly more FLOPS than BERT-MLM. Specifically, it achieved a GLUE average score of 80.02 on the test leaderboard and MAP scores of 78.7, 86.7, 64.9 in WikiQA, TREC-QA, and ASNQ, respectively.

\chapter{Self-Supervised Objectives for AS2}

\section{Negative Results}
\label{app:pairwise_negative}

In this Section, we briefly describe the experiments that didn't lead to accuracy improvements of our language models. We believe negative results are important to help other authors at selecting future research directions.

\subsection{Additional positives from Large Language Model}

\input{tables/pairwise_as2_base_additional}

Most of the datasets for AS2 have a limited number of positive question-answer pairs. For example, the training set of ASNQ has about 1.1 positive answers every 356 candidates. The other datasets have instead a slightly higher ratio between positive answer and the total number of candidates sentences for each query: $\frac{0.5}{9.6}$ in WikiQA, $\frac{5.4}{46}$ in TREC-QA, $\frac{0.23}{15}$ in WQA and about $\frac{1.7}{28}$ in NewsAS2 and TriviaAS2. However, the gap between positives and negatives examples is still very large.

For those reasons, we try to increment the number of positive answer candidate by exploiting Falcon-40B-instruct~\citep{falcon40b}, a Large Language Model with 40 billion parameters fine-tuned for instruction-based conversations. Specifically, we generate a single additional positive answer for each query by providing Falcon with the question and possibly by adding a question mark at the end if it was missing.

Notice that we do not add positive answer candidates to the development and test sets to perform a fair comparison with the results provided in the previous Sections. We test the vanilla RoBERTa\textsubscript{Base} and ELECTRA\textsubscript{Base} checkpoints as well as the models further pre-trained with our specialized objectives on ASNQ and WikiQA. The latter should provide a meaningful test suite being a very large dataset with very few positives and a small dataset with a reasonable number of positive answers, respectively.

We report the results of the experiment in Table~\ref{tab:pairwise_as2_base_additional}. On ASNQ, there are rare combinations of models and objectives that benefits from the additional positives. For example, the two baselines RoBERTa\textsubscript{Base} and ELECTRA\textsubscript{Base} show an improvements of the P@1 by 0.6 points and 1.2 points respectively. Our continuously pre-trained models generally perform lower by 1.0 P@1 points when fine-tuned with the additional positives, apart from RoBERTa\textsubscript{Base}~+~SSP, which gains 0.2 points in P@1.

On WikiQA, no model benefits from the additional positives. In particular, there is a decrease between 0.3 and 4.0 points in P@1, depending on the combination between models architectures and specialized our pre-training objectives.

\end{document}